\documentclass[10pt,twocolumn,letterpaper]{article}

\usepackage[pagenumbers]{cvpr} %

\usepackage[dvipsnames]{xcolor}

\newcommand{\bc}{\mathbf{c}}

\newcommand{\bI}{\mathbf{I}}
\newcommand{\bJ}{\mathbf{J}}

\newcommand{\bO}{\mathbf{O}}
\newcommand{\bp}{\mathbf{p}}

\newcommand{\bR}{\mathbf{R}}
\newcommand{\bs}{\mathbf{s}}
\newcommand{\bt}{\mathbf{t}}

\newcommand{\bx}{\mathbf{x}}

\newcommand{\bSigma}{\boldsymbol{\Sigma}}

\newcommand{\cG}{\mathcal{G}}

\newcommand{\figref}[1]{Figure~\ref{#1}}
\newcommand{\secref}[1]{Section~\ref{#1}}

\newcommand{\eqnref}[1]{Eq.~\ref{#1}}
\newcommand{\tabref}[1]{Table~\ref{#1}}

\makeatletter
\DeclareRobustCommand\onedot{\futurelet\@let@token\@onedot}
\def\@onedot{\ifx\@let@token.\else.\null\fi\xspace}
\def\eg{e.g\onedot} 
\def\ie{i.e\onedot} 
\def\cf{cf\onedot} 

\def\vs{vs\onedot}

\def\etal{et~al\onedot}

\makeatother

\definecolor{yellow}{rgb}{1, 1, 0.7}
\definecolor{orange}{rgb}{1, 0.85, 0.7}
\definecolor{tablered}{rgb}{1, 0.7, 0.7}
\definecolor{red}{rgb}{1, 0, 0}

\definecolor{wincolor}{rgb}{0.85, 0.0, 0.0}

\definecolor{darkyellow}{rgb}{0.8, 0.8, 0.5}
\definecolor{darkred}{rgb}{0.7, 0.3, 0.3}
\definecolor{darkgreen}{rgb}{0.3, 0.7, 0.3}
\definecolor{blue}{rgb}{0.251, 0.498, 0.824}
\definecolor{green}{rgb}{0, 1.0, 0}
\definecolor{pink}{rgb}{1, 0.4, 0.7}

\definecolor{realred}{rgb}{0.95, 0.1, 0.0}
\newcommand{\red}[1]{{\color{realred}#1}}
\newcommand{\blue}[1]{{\color{blue}#1}}

\newcommand{\boldparagraph}[1]{\vspace{0.1cm}\noindent{\bf #1:}}

\newcommand{\scenename}[1]{\textit{#1}}

\definecolor{cvprblue}{rgb}{0.21,0.49,0.74}
\usepackage[pagebackref,breaklinks,colorlinks,citecolor=cvprblue]{hyperref}
\usepackage{pifont}
\usepackage{color}
\usepackage{xcolor}
\usepackage{nicefrac}       %
\usepackage{multirow}
\usepackage{multicol}
\usepackage{dsfont}

\usepackage{colortbl}
\usepackage[percent]{overpic}

\newtheorem{theorem}{Condition}

\def\paperID{5103} %
\def\confName{CVPR}
\def\confYear{2024}

\title{Mip-Splatting: Alias-free 3D Gaussian Splatting}

\author{
	Zehao Yu$^{1,2}$\quad Anpei Chen$^{1,2}$\quad Binbin Huang$^{3}$\quad Torsten Sattler$^{4}$\quad Andreas Geiger$^{1,2}$\vspace{8pt}\\
	$^{1}$University of T\"ubingen\qquad $^{2}$T\"ubingen AI Center\qquad $^{3}$ShanghaiTech University
	\\
	$^{4}$Czech Technical University in Prague\vspace{8pt}\\
	\url{https://niujinshuchong.github.io/mip-splatting}
}

\begin{document}

\twocolumn[{%
	\renewcommand\twocolumn[1][]{#1}%
	\maketitle
	\vspace{-2em}
	\includegraphics[width=1.\linewidth]{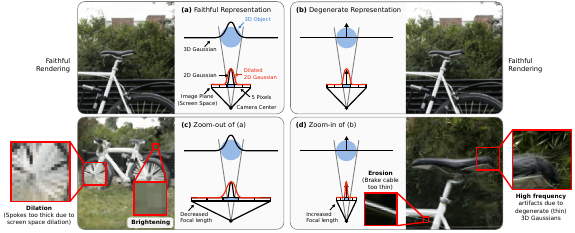}
	
    \captionof{figure}{\textbf{3D Gaussian Splatting} \cite{kerbl3Dgaussians} renders images by representing \blue{\textbf{3D Objects}} as \textbf{3D Gaussians} which are projected onto the image plane followed by \red{\textbf{2D Dilation}} in screen space as shown in (a). The method's intrinsic shrinkage bias leads to degenerate 3D Gaussians exceed sampling limit as illustrated by the $\delta$ function in (b) while rendering similarly to 2D due to the dilation operation. However, when changing the sampling rate (via the focal length or camera distance), we observe strong dilation effects (c) and high frequency artifacts (d).}
	\label{fig:teaser}
	\vspace{1.3em}
}]

\begin{abstract}
\vspace{-0.3cm}
Recently, 3D Gaussian Splatting has demonstrated impressive novel view synthesis results, reaching high fidelity and efficiency. However, strong artifacts can be observed when changing the sampling rate, \eg, by changing focal length or camera distance. We find that the source for this phenomenon can be attributed to the lack of 3D frequency constraints and the usage of a 2D dilation filter. To address this problem, we introduce a 3D smoothing filter which constrains the size of the 3D Gaussian primitives based on the maximal sampling frequency induced by the input views, eliminating high-frequency artifacts when zooming in. Moreover, replacing 2D dilation with a 2D Mip filter, which simulates a 2D box filter, effectively mitigates aliasing and dilation issues.
Our evaluation, including scenarios such a training on single-scale images and testing on multiple scales, validates the effectiveness of our approach. %
\end{abstract}

\section{Introduction}
Novel View Synthesis (NVS) plays a critical role in computer graphics and computer vision, with various applications including virtual reality, cinematography, robotics, and more. 
A particularly significant advancement in this field is the Neural Radiance Field (NeRF)~\cite{mildenhall2020nerf}, introduced by Mildenhall \etal in 2020. NeRF utilizes a multilayer perceptron (MLP) to represent geometry and view-dependent appearance effectively, demonstrating remarkable novel view rendering quality.
Recently, 3D Gaussian Splatting (3DGS)~\cite{kerbl3Dgaussians} has gained attention as an appealing alternative to both MLP~\cite{mildenhall2020nerf} and feature grid-based representations~\cite{yu2022plenoxels,muller2022instant,Chen2022ECCV,Sun2022CVPR,liu2020neural}. 3DGS stands out for its impressive novel view synthesis results, while achieving real-time rendering at high resolutions. This effectiveness and efficiency, coupled with the potential integration into the standard rasterization pipeline of GPUs
represents a significant step towards practical usage of NVS methods.

Specifically, 3DGS represents complex scenes as a set of 3D Gaussians, which are rendered to screen space through splatting-based rasterization. The attributes of each 3D Gaussian, \ie, position, size, orientation, opacity, and color, are optimized through a multi-view photometric loss. Thereafter, a 2D dilation operation is applied in screen space for low-pass filtering. Although 3DGS has demonstrated impressive NVS results, it produces artifacts when camera views diverge from those seen during training, such as zoom in and zoom out, as illustrated in ~\figref{fig:teaser}.
We find that the source for this phenomenon can be attributed to the lack of 3D frequency constraints and the usage of a 2D dilation filter.
Specifically, zooming out leads to a reduced size of the projected 2D Gaussians in screen space, while applying the same amount of dilation results in dilation artifacts.
Conversely, zooming in causes erosion artifacts since the projected 2D Gaussians expand, yet dilation remains constant, causing erosion and resulting in incorrect gaps between Gaussians in the 2D projection.

To resolve these issues, we propose to regularize the 3D representation in 3D space. Our key insight is that the highest frequency that can be reconstructed of a 3D scene is inherently constrained by the sampling rates of the input images. We first derive the multi-view frequency bounds of each Gaussian primitive based on the training views according to the Nyquist-Shannon Sampling Theorem~\cite{Nyquist1928IEEE,Shannon1949IEEE}. By applying a low-pass filter to the 3D Gaussian primitives in 3D space during the optimization, we effectively restrict the maximal frequency of the 3D representation to meet the Nyquist limit. Post-training, this filter becomes an intrinsic part of the scene representation, remaining constant regardless of viewpoint changes. Consequently, our method eliminates the artifacts presents in 3DGS~\cite{kerbl3Dgaussians} when zooming in, as shown in the $8\times$ higher resolution image in~\figref{fig:teaser_artefacts}.

Nonetheless, rendering the reconstructed scene at lower sampling rates (\eg, zooming out) results in aliasing. Previous work~\cite{Barron2021ICCV,barron2022mipnerf360,Barron2023ICCV,Hu2023ICCV} address aliasing by employing cone tracing and applying pre-filtering to the input positional or feature encoding, which is not applicable to 3DGS. Thus, we introduce a 2D Mip filter (à la “mipmap”) specifically designed to ensure alias-free reconstruction and rendering across different scales.
Our 2D Mip filter mimics the 2D box filter inherent to the actual physical imaging process~\cite{Shirley2023RTW1,szeliski2022computer,mildenhall2022rawnerf}.by approximating it with a 2D Gaussian low pass filter.
In contrast to previous work~\cite{Barron2021ICCV,barron2022mipnerf360,Barron2023ICCV,Hu2023ICCV} that rely on the MLP's ability to interpolate multi-scale signals during training with multi-scale images, our closed-form modification to the 3D Gaussian representation results in excellent out-of-distribution generalization: Training at a single sampling rate enables faithful rendering at various sampling rates different from those used during training as demonstrated by the $\nicefrac{1}{4}\times$ down-sampled image in~\figref{fig:teaser_artefacts}.

\noindent In summary, we make the following contributions:
\begin{itemize}
\item We introduce a \textbf{3D smoothing filter} for 3DGS to effectively regularize the maximum frequency of 3D Gaussian primitives, resolving the artifacts observed in out-of-distribution renderings of prior methods~\cite{kerbl3Dgaussians,zwicker2001ewa}.
\item We replace the 2D dilation filter with a \textbf{2D Mip filter} to address aliasing and dilation artifacts.
\item Experiments on challenging benchmark datasets~\cite{mildenhall2020nerf,barron2022mipnerf360} demonstrate the effectiveness of Mip-Splatting when modifying the sampling rate.
\item Our modifications to 3DGS are principled and simple, requiring only few changes to the original 3DGS code.
\end{itemize}

\newcommand{\aliasimage}[2]{
	\begin{overpic}[width=0.78in]{figs/teaser_artefacts/#1}
	\put (48,2) {\scriptsize #2}
    \end{overpic}
}

\begin{figure}[t]
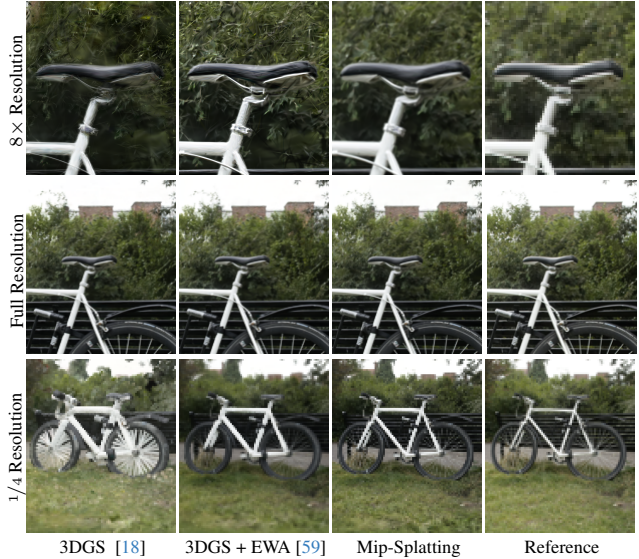

    \centering
    \scriptsize
    \begin{tabular}{@{}c@{}c@{}c@{}c@{}@{}c@{}}
    \multirow{1}{*}[15ex]{\rotatebox{90}{$8\times$ Resolution}} &
    \aliasimage{3dgs_up.png}{} & 
    \aliasimage{ewa_up.png}{} & 
    \aliasimage{ours_up.png}{} & 
    \aliasimage{gt_up_from_down.png}{} \\
    
    \multirow{1}{*}[15ex]{\rotatebox{90}{Full Resolution}} & 
    \aliasimage{3dgs_mid.png}{} & 
    \aliasimage{ewa_mid.png}{} & 
    \aliasimage{ours_mid.png}{} & 
    \aliasimage{gt_mid.png}{}\\
    
    \multirow{1}{*}[15ex]{\rotatebox{90}{$\nicefrac{1}{4}$ Resolution}} & 
    
    \aliasimage{3dgs_down.png}{} & 
    \aliasimage{ewa_down.png}{}  & 
    \aliasimage{ours_down.png}{} & 
    \aliasimage{gt_down.png}{} \\

     &3DGS ~\cite{kerbl3Dgaussians}& 3DGS + EWA~\cite{zwicker2001ewa} & Mip-Splatting & Reference 
    \end{tabular}
    \vspace{-0.1in}
    \caption{
    We trained all the models on single-scale (full resolution here) images and rendered images with different resolutions by changing focal length. While all methods show similar performance at training scale, we observe strong artifacts in previous work~\cite{kerbl3Dgaussians,zwicker2001ewa} when changing the sampling rate. By contrast, our Mip-Splatting renders faithful images across different scales.%
    }
    \label{fig:teaser_artefacts}
    \vspace{-0.1in}
\end{figure}
\section{Related Work}
\label{sec:related_work}
\boldparagraph{Novel View Synthesis} 
NVS is the process of generating new images from viewpoints different from those of the original captures~\cite{gortler2023lumigraph,levoy2023light}. NeRF~\cite{mildenhall2020nerf}, which leverages volume rendering~\cite{drebin1988volume,levoy1990efficient,max2005local,max1995optical}, has become a standard technique in the field. NeRF utilizes MLPs~\cite{Mescheder2019CVPR,Park2019CVPR,Chen2019CVPR} to model scenes as continuous functions, which, despite their compact representation, impede rendering speed due to the expensive MLP evaluation that is required for each ray point. Subsequent methods~\cite{Reiser2021ICCV,Hedman2021ICCV,yu2021plenoctrees,Reiser2023SIGGRAPH,yariv2023bakedsdf} distill a pretrained NeRF into a sparse representation, enabling real-time rendering of NeRFs. Further advancements have been made to improve the training and rendering of NeRF with advanced scene representations~\cite{liu2020neural,Sun2022CVPR,yu2022plenoxels,kulhanek2023tetra,Chen2022ECCV,muller2022instant,xu2022point,chen2023neurbf,kerbl3Dgaussians}. In particular, 3D Gaussians Splatting (3DGS)~\cite{kerbl3Dgaussians} demonstrated impressive novel view synthesis results, while achieving real-time rendering at high-definition resolutions. Importantly, 3DGS represents the scene explicitly as a collection of 3D Gaussians and uses rasterization instead of ray tracing.
Nevertheless, 3DGS focuses on in-distribution evaluation where training and testing are conducted at similar sampling rates (focal length/scene distance). In this paper, we study the out-of-distribution generalization of 3DGS, training models at a single scale and evaluating it across multiple scales. 

\boldparagraph{Primitive-based Differentiable Rendering} 
Primitive-based rendering techniques, which rasterize geometric primitives onto the image plane, have been explored extensively due to their efficiency~\cite{grossman1998point,gross2011point,sainz2004point,pfister2000surfels,zwicker2001ewa,zwicker2001surface}. Differentiable point-based rendering methods~\cite{Wang2019DSS,wiles2020synsin,Peng2021SAP,ruckert2021adop,lassner2021pulsar,Zheng2023pointavatar,Prokudin_2023_ICCV} offer great flexibility in representing intricate structures and are thus well-suited for novel view synthesis. Notably, Pulsar~\cite{lassner2021pulsar} stands out for its efficient sphere rasterization. The more recent 3D Gaussian Splatting (3DGS) work~\cite{kerbl3Dgaussians} utilizes anisotropic Gaussians~\cite{zwicker2001ewa} and introduces a tile-based sorting for rendering, achieving remarkable frame rates. Despite its impressive results, 3DGS exhibits strong artifacts when rendering at a different sampling rate. We address this issue by introducing a 3D smoothing filter to constrain the maximal frequencies of the 3D Gaussian primitive representation, and a 2D Mip filter that approximates the box filter of the physical imaging process for alias-free rendering.

\boldparagraph{Anti-aliasing in Rendering} %
There are two principal strategies to combat aliasing: \textit{super-sampling}, which increases the number of samples~\cite{supersampling}, and \textit{prefiltering}, which applies low-pass filtering to the signal to meet the Nyquist limit~\cite{summed_area_texture_mapping,pyramidal_parametrics,heckbert1989fundamentals,swan1997anti,mueller1998splatting,zwicker2001ewa}. For example, EWA splatting~\cite{zwicker2001ewa} applies a Gaussian low pass filter to the projected 2D Gaussian in screen space to produce a band limited output respecting the Nyquist frequency of the image. While we also apply a band-limited filter to the Gaussian primitives, our band-limited filter is applied in 3D space and the filter size is fully determined by the training images not the images to be rendered. While our 2D Mip filter is also a Gaussian low pass filter in screen space, it approximates the box filter of the physical imaging process, approximating a single pixel. Conversely, the EWA filter limits the frequency signal's bandwidth to the rendered image, %
and the size of the filter is chosen empirically. 
A critical difference to \cite{zwicker2001ewa} is that we tackle the reconstruction problem, optimizing the 3D Gaussian representation via inverse rendering while EWA splatting only considers the rendering problem. 

Recent neural rendering methods integrate pre-filtering to mitigate aliasing~\cite{Barron2021ICCV,barron2022mipnerf360,Barron2023ICCV,Hu2023ICCV,zhuang2023anti}. Mip-NeRF~\cite{Barron2021ICCV}, for instance, introduced an integrated position encoding (IPE) to attenuate high-frequency details. A similar idea is adapted for feature grid-based representations~\cite{Barron2023ICCV,Hu2023ICCV,zhuang2023anti}. However, these approaches require multi-scale images for supervision. In contrast, our approach is based on 3DGS~\cite{kerbl3Dgaussians} and determines the necessary low-pass filter size based on pixel size, allowing for alias-free rendering at scales unobserved during training.

\newcommand{\RRR}{\mathbb{R}}

\section{Preliminaries}
In this section, we first review the sampling theorem in~\secref{sec:sampling_theorem}, laying the foundation for understanding the aliasing problem. Subsequently, we introduce 3D Gaussian Splatting (3DGS)~\cite{kerbl3Dgaussians} and its rendering process in~\secref{sec:3d_gs_splatting}.

\subsection{Sampling Theorem}
\label{sec:sampling_theorem}

The Sampling Theorem, also known as the Nyquist-Shannon Sampling Theorem~\cite{Nyquist1928IEEE,Shannon1949IEEE}, is a fundamental concept in signal processing and digital communication that describes the conditions under which a continuous signal can be accurately represented or reconstructed from its discrete samples. To accurately reconstruct a continuous signal from its discrete samples without loss of information, the following conditions must be met:
\begin{theorem}
The continuous signal must be band-limited and may not contain any frequency components above a certain maximum frequency $\nu$.
\end{theorem}

\begin{theorem}
The sampling rate $\hat{\nu}$ must be at least twice the highest frequency present in the continuous signal: $\hat{\nu} \geq 2\nu$. 
\end{theorem}

In practice, to satisfy the constraints when reconstructing a signal from discrete samples, a low-pass or anti-aliasing filter is applied to the signal before sampling. The filter eliminates any frequency components above $\frac{\hat{\nu}}{2}$ and attenuates high-frequency content that could lead to aliasing.

\subsection{3D Gaussian Splatting}
\label{sec:3d_gs_splatting}

Prior works~\cite{zwicker2001ewa,kerbl3Dgaussians} propose to represent a 3D scene as a set of scaled 3D Gaussian primitives $\{\cG_k | k=1, \cdots, K\}$ and render an image using volume splatting.
The geometry of each scaled 3D Gaussian $\cG_k$ is parameterized by an opacity (scale) $\alpha_k\in[0,1]$, center $\bp_k \in \RRR^{3 \times 1}$ and covariance matrix $\bSigma_k \in \RRR^{3\times3}$ defined in world space:
\begin{equation}
\cG_k(\bx) = e^{-\frac{1}{2} (\bx-\bp_k)^T \bSigma_k^{-1}(\bx-\bp_k)}
\end{equation}
To constrain $\bSigma_k$ to the space of valid covariance matrices, a semi-definite parameterization $\bSigma_k = \bO_k \bs_k\bs_k^T \bO_k^T$ is used. Here,  $\bs \in \RRR^{3}$ is a scaling vector and $\bO \in \RRR^{3 \times 3}$ is a rotation matrix, parameterized by a quaternion~\cite{kerbl3Dgaussians}.

To render an image for a given view point defined by rotation $\bR \in \RRR^{3 \times 3}$ and translation $\bt \in \RRR^{3}$, the 3D Gaussians $\{\cG_k\}$ are first transformed into camera coordinates:
\begin{equation}
\bp'_k = \bR \, \bp_k + \bt, \quad \bSigma'_k = \bR \,\bSigma_k \, \bR^T
\end{equation}
Afterwards, they are projected to ray space via a local affine transformation %
\begin{equation}
\bSigma''_k = \bJ_k\, \bSigma'_k \, \bJ_k^T
\end{equation}
where the Jacobian matrix $\bJ_k$ is an affine approximation to the projective transformation defined by the center of the 3D Gaussian $\bp'_k$. %
By skipping the third row and column of $\bSigma''_k$, we obtain a 2D covariance matrix $\bSigma^{2D}_k$ in ray space, and we use $\cG_k^{2D}$ to refer to the corresponding scaled 2D Gaussian, see \cite{kerbl3Dgaussians} for details.

Finally, 3DGS \cite{kerbl3Dgaussians} utilizes spherical harmonics to model view-dependent color $\bc_k$ and renders image via alpha blending according to the primitive's depth order $1,\dots,K$: %
\begin{equation}
\bc(\bx) = \sum^K_{k=1} \bc_k\,\alpha_k\,\cG^{2D}_k(\bx) \prod_{j=1}^{k-1} (1 - \alpha_j\,\cG^{2D}_j(\bx))
\end{equation}

\boldparagraph{Dilation} To avoid degenerate cases where the projected 2D Gaussians are too small in screen space, \ie, smaller than a pixel, the projected 2D Gaussians are dilated as follows: 
\begin{equation}
\cG^{2D}_k(\bx) = e^{-\frac{1}{2} (\bx-\bp_k)^T \, {(\bSigma^{2D}_k + \, s \, \bI)}^{-1} \, (\bx-\bp_k)}
\label{eqn:dilation}
\end{equation}
where $\bI$ is a 2D identity matrix and $s$ is a scalar dilation hyperparameter. Note that this operator adjusts the scale of the 2D Gaussian while leaving its maximum unchanged. As this effect is similar to that of dilation operators in morphology, we called it a 2D screen space dilation operation\footnote{The dilation operation is not mentioned in original paper.}.

\boldparagraph{Reconstruction}
As the rendering process is fast and differentiable, the 3D Gaussian parameters can be efficiently optimized using a 
multi-view loss.
During optimization, 3D Gaussians are adaptively added and deleted to better represent the scene.
We refer the reader to~\cite{kerbl3Dgaussians} for details.

\section{Sensitivity to Sampling Rate}
\label{sec:challenges}
In traditional forward splatting, the centers $\bp_k$ and colors $\bc_k$ of Gaussian primitives are predetermined, whereas the 3D Gaussian covariance $\bSigma_k$ are chosen empirically~\cite{zwicker2001ewa,ren2002object}. In contrast, 3DGS~\cite{kerbl3Dgaussians}, optimizes all parameters jointly through an inverse rendering framework by backpropagating a multi-view photometric loss. 

We observe that this optimization suffers from ambiguities as illustrated in \figref{fig:teaser} which shows a simple example involving one object and an image sensor with 5 pixels. Consider the 3D object in (a), its approximation by a 3D Gaussian and its projection into screen space (blue pixel).
Due to screen space dilation (\eqnref{eqn:dilation}) with a Gaussian kernel (size $\approx$ 1 pixel), the degenerate 3D Gaussian represented by a Dirac $\delta$ function in (b) leads to a similar image. This illustrates that the scale of the 3D Gaussian is not properly constrained. In practice, due to its implicit shrinkage bias, 3DGS indeed systematically underestimates the scale parameter of 3D Gaussians during optimization.

While this does not affect rendering at similar sampling rates (\cf \figref{fig:teaser} (a) \vs (b)), it leads to erosion effects when zooming in or moving the camera closer. This is because the dilated 2D Gaussians become smaller in screen space. In this case, the rendered image exhibits high-frequency artifacts, rendering object structures thinner than they actually appear as illustrated in \figref{fig:teaser} (d).

Conversely, screen space dilation also negatively affects rendering when decreasing the sampling rate as illustrated in \figref{fig:teaser} (c) which shows a zoomed-out version of (a). In this case, dilation spreads radiance in a physically incorrect way across pixels. Note that in (c), the area covered by the projection of the 3D object is smaller than a pixel, yet the dilated Gaussian is not attenuated, accumulating more light than what physically reaches the pixel. This leads to increased brightness and dilation artifacts which strongly degrade the appearance of the bicycle wheels' spokes.

The aforementioned scale ambiguity becomes particularly problematic in representations involving millions of Gaussians.
However, simply discarding screen space dilation results in optimization challenges for complex scenes, such as those present in the Mip-NeRF 360 dataset~\cite{barron2022mipnerf360}, where a large number of small Gaussian are created by the density control mechanism~\cite{kerbl3Dgaussians}, exceeding GPU capacity. Moreover, even if a model can be successfully trained without dilation, decreasing the sampling rate results in aliasing effects due to the lack of anti-aliasing~\cite{zwicker2001ewa}.

\section{Mip Gaussian Splatting}

To overcome these challenges, we make two modfications to the original 3DGS model.
In particular, we introduce a 3D smoothing filter that limits the frequency of the 3D representation to below half the maximum sampling rate determined by the training images, eliminating high frequency artifacts when zooming in. Moreover, we demonstrate that replacing 2D screen space dilation with a 2D Mip filter which approximates the box filter inherent to the physical imaging process and effectively mitigates aliasing and dilation issues.
In combination, Mip-Splatting enables alias-free renderings\footnote{Note that we use alias to refer to multiple artifacts discussed in the paper, including dilation, erosion, oversmoothing, high-frequency artifacts and aliasing itself.} across various sampling rates. We now discuss the the 3D smoothing and the 2D Mip filters in detail.

\subsection{3D Smoothing Filter}
\label{sec:reg_filter}
3D radiance field reconstruction from multi-view observations is a well-known ill-posed problem as multiple distinctly different reconstructions can result in the same 2D projections~\cite{kaizhang2020,barron2022mipnerf360,Yu2022MonoSDF}. Our key insight is that the highest frequency of a reconstructed 3D scene is limited by the sampling rate defined by the training views.
Following Nyquist's theorem~\ref{sec:sampling_theorem}, we aim to constrain the maximum frequency of the 3D representation during optimization. 

\boldparagraph{Multiview Frequency Bounds}
\begin{figure}%
    \includegraphics[width=0.95\linewidth]{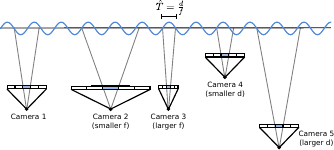}
    \caption{\textbf{Sampling limits.} A pixel corresponds to sampling interval $\hat{T}$. We band-limit the 3D Gaussians by the maximal sampling rate (\ie, minimal sampling interval) among all observations. This example shows 5 cameras at different depths $d$ and with different focal lengths $f$. Here, camera 3 determines the minimal $\hat{T}$ and hence the maximal sampling rate $\hat{\nu}$.}
    \label{fig:sampling_limit}
    \vspace{-0.1in}
\end{figure}
Multi-view images are 2D projections of a continuous 3D scene. The discrete image grid determines where we sample points from the continuous 3D signal. This sampling rate is intrinsically related to the image resolution, camera focal length, and the scene's distance from the camera. For an image with focal length $f$ in pixel units, the sampling interval in screen space is $1$. When this pixel interval is back-projected to the 3D world space, it results in a world space sampling interval $\hat{T}$ at a given depth $d$, with sampling frequency $\hat{\nu}$ as its inverse:
\begin{equation}
\hat{T} = \frac{1}{\hat{\nu}} = \frac{d}{f}
\label{eqn:sampling_rate}
\end{equation}
As posited by Nyquist's theorem~\secref{sec:sampling_theorem}, given samples drawn at frequency $\hat{\nu}$, reconstruction algorithms are able to reconstruct components of the signal with frequencies up to $\frac{\hat{\nu}}{2}$, or $\frac{f}{2d}$. Consequently, a primitive smaller than $2\hat{T}$ may result in aliasing artifacts during the splatting process, since its size is below twice the sampling interval. 

To simplify, we approximate depth $d$ using the center of the primitive $\bp_k$, and disregard the impact of occlusion for sampling interval estimation. Since the sampling rate of a primitive is depth-dependent and differs across cameras, we determine the maximal sampling rate for primitive $k$ as
\begin{equation}
\hat{\nu}_k = \text{max}\left(\left\{ \mathds{1}_n(\bp_k) \cdot \frac{f_n}{d_n}\right\}^{N}_{n=1}\right)
\end{equation}
where $N$ is the total number of images, $\mathds{1}_n(\bp)$ is an indicator function that assesses the visibility of a primitive. It is true if the Gaussian center $\bp_k$ falls within the view frustum of the $n$-th camera. 
Intuitively, we choose the sampling rate such that there exists at least one camera that is able to reconstruct the respective primitive.
This process is illustrated in~\figref{fig:sampling_limit} for $N=5$. In our implementation, we recompute the maximal sampling rate of each Gaussian primitive every $m$ iterations as we found the 3D Gaussians centers remain relatively stable throughout the training.

\boldparagraph{3D Smoothing}
Given the maximal sampling rate $\hat{\nu}_k$ for a primitive, we aim to constrain the maximal frequency of the 3D representation. This is achieved by applying a Gaussian low-pass filter $\cG_{\text{low}}$ to each 3D Gaussian primitive $\cG_k$ before projecting it onto screen space: 
\begin{align}
\cG_k(\bx)_{\text{reg}} = (\cG_k \otimes \cG_{\text{low}})(\bx)
\label{eqn:filter_3d_simple}
\end{align}
This operation is efficient as convolving two Gaussians with covariance matrices $\bSigma_1$ and $\bSigma_2$ results in another Gaussian with variance $\bSigma_1 + \bSigma_2$. Hence,
\begin{equation}
\cG_k(\bx)_\text{reg} = \sqrt{\frac{|\bSigma_k|}{|\bSigma_k + \frac{s}{\hat{\nu}_k} \cdot \bI|}} \, \, e^{-\frac{1}{2} (\bx-\bp_k)^T \, (\bSigma_k + \frac{s}{\hat{\nu}_k} \cdot \bI)^{-1} \, (\bx-\bp_k)}
\label{eqn:filter_3d_full}
\end{equation}
Here, $s$ is a scalar hyperparameter to control the size of the filter. %
Note that the scale $\frac{s}{\hat{\nu}_k}$ of the 3D filters for each primitive are different as they depend on the training views in which they are visible. By employing 3D Gaussian smoothing, we ensure that the highest frequency component of any Gaussian does not exceed half of its maximal sampling rate for at least one camera. 
Note that $\cG_{\text{low}}$ becomes an intrinsic part of the 3D representation, remaining constant post-training.

\subsection{2D Mip Filter}
\label{sec:mip_filter}

\begin{table*}[]
    \renewcommand{\tabcolsep}{1pt}
    \centering
    \resizebox{0.95\linewidth}{!}{
    \begin{tabular}{@{}l@{\,\,}|ccccc|ccccc|ccccc}
    &   \multicolumn{5}{c|}{PSNR $\uparrow$} & \multicolumn{5}{c|}{SSIM $\uparrow$} & \multicolumn{5}{c}{LPIPS $\downarrow$}  \\
    & Full Res. & $\nicefrac{1}{2}$ Res. & $\nicefrac{1}{4}$ Res. & $\nicefrac{1}{8}$ Res. & Avg. & Full Res. & $\nicefrac{1}{2}$ Res. & $\nicefrac{1}{4}$ Res. & $\nicefrac{1}{8}$ Res. & Avg. & Full Res. & $\nicefrac{1}{2}$ Res. & $\nicefrac{1}{4}$ Res. & $\nicefrac{1}{8}$ Res & Avg.  \\ \hline

NeRF w/o \( \mathcal{L}_\text{area} \)~\cite{mildenhall2020nerf,Barron2021ICCV}&            31.20 &                    30.65 &                    26.25 &                     22.53 &  27.66 &                   0.950 &                    0.956 &                    0.930 &                    0.871 &  0.927 &                  0.055 &                    0.034 &                   0.043 &                    0.075 &                    0.052
\\
NeRF~\cite{mildenhall2020nerf}&                  29.90 &                    32.13 &                    33.40 &                     29.47 &  31.23 &                   0.938 &                    0.959 &                    0.973 &                    0.962 &  0.958 &                  0.074 &                    0.040 &                    0.024 &                    0.039 &                  0.044 
\\
MipNeRF~\cite{Barron2021ICCV}&\cellcolor{yellow}  32.63 &\cellcolor{orange}     34.34 &\cellcolor{tablered}35.47 &\cellcolor{tablered}35.60 &\cellcolor{orange} 34.51 &\cellcolor{yellow} 0.958 & 0.970 &\cellcolor{yellow} 0.979 &\cellcolor{yellow} 0.983 & 0.973&\cellcolor{yellow} 0.047 &\cellcolor{orange} 0.026 &\cellcolor{orange} 0.017 &\cellcolor{orange} 0.012 &\cellcolor{orange} 0.026 
\\
\hline
Plenoxels~\cite{yu2022plenoxels}&  31.60 & 32.85 & 30.26 & 26.63 & 30.34 &  0.956 &  0.967 & 0.961 & 0.936 & 0.955 &  0.052 &   0.032 &  0.045 & 0.077 & 0.051
\\
TensoRF~\cite{Chen2022ECCV}&  32.11 &                    33.03 & 30.45 &  26.80 &                    30.60 &  0.956 &  0.966 &  0.962 &                    0.939 &                    0.956 & 0.056 & 0.038 &0.047 & 0.076 & 0.054 
\\
Instant-NGP~\cite{muller2022instant}&     30.00 & 32.15 &                    33.31 &                    29.35 & 31.20 &   0.939 &                    0.961 &  0.974 &  0.963 &0.959 &                    0.079 & 0.043 &  0.026 & 0.040 &0.047  
\\
Tri-MipRF~\cite{Hu2023ICCV}*&\cellcolor{orange}   32.65 &\cellcolor{yellow} 34.24 &\cellcolor{yellow} 35.02 &\cellcolor{orange} 35.53 &\cellcolor{yellow} 34.36 &\cellcolor{yellow} 0.958 &\cellcolor{yellow} 0.971 &\cellcolor{orange} 0.980 &\cellcolor{orange} 0.987 &\cellcolor{orange} 0.974 &\cellcolor{yellow} 0.047 & 0.027 &\cellcolor{yellow} 0.018 &\cellcolor{orange} 0.012 &\cellcolor{orange} 0.026
\\
3DGS~\cite{kerbl3Dgaussians}&  28.79 & 30.66 & 31.64 & 27.98 & 29.77 & 0.943 & 0.962 & 0.972 & 0.960 & 0.960 & 0.065 & 0.038 & 0.025 & 0.031 & 0.040
\\
\hline
3DGS~\cite{kerbl3Dgaussians} + EWA~\cite{zwicker2001ewa}& 31.54 & 33.26 & 33.78 & 33.48 & 33.01 &\cellcolor{orange} 0.961 &\cellcolor{orange} 0.973 &\cellcolor{yellow} 0.979 &\cellcolor{yellow} 0.983 &\cellcolor{orange} 0.974 &\cellcolor{orange} 0.043 &\cellcolor{orange} 0.026 & 0.021 & 0.019 & 0.027
\\
Mip-Splatting (ours)&\cellcolor{tablered}32.81 &\cellcolor{tablered}34.49 &\cellcolor{orange} 35.45 &\cellcolor{yellow} 35.50 &\cellcolor{tablered}34.56 &\cellcolor{tablered}0.967 &\cellcolor{tablered}0.977 &\cellcolor{tablered}0.983 &\cellcolor{tablered}0.988 &\cellcolor{tablered}0.979 &\cellcolor{tablered}0.035 &\cellcolor{tablered}0.019 &\cellcolor{tablered}0.013 &\cellcolor{tablered}0.010 &\cellcolor{tablered}0.019

    \end{tabular}
    }
    \vspace{-0.1in}
    \caption{
    \textbf{Multi-scale Training and Multi-scale Testing on the Blender dataset~\cite{mildenhall2020nerf}.} Our approach achieves state-of-the-art performance in most metrics. It significantly outperforms 3DGS~\cite{kerbl3Dgaussians} and 3DGS + EWA~\cite{zwicker2001ewa}. $*$ indicates that we retrain the model.
    }
    \label{tab:avg_multiblender_results}
    \vspace{-0.1in}
\end{table*}

While our 3D smoothing filter effectively mitigates high-frequency artifacts~\cite{kerbl3Dgaussians,zwicker2001ewa}, rendering the reconstructed scene at lower sampling rates (\eg, zooming out or moving the camera further away) would still lead to aliasing. To overcome this, we replace the screen space dilation filter of 3DGS by a 2D Mip filter.

More specifically, we replicate the physical imaging process~\cite{Shirley2023RTW1,szeliski2022computer,mildenhall2022rawnerf}, where photons hitting a pixel on the camera sensor are integrated over the pixel's area. While an ideal model would use a 2D box filter in image space, we approximate it with a 2D Gaussian filter for efficiency
\begin{equation}
\cG^{2D}_k(\bx)_{\text{mip}} = \sqrt{\frac{|\bSigma^{2D}_k|}{|\bSigma^{2D}_k + s \bI|}} \, \, e^{-\frac{1}{2} (\bx-\bp_k)^T \, (\bSigma^{2D}_k + s \bI)^{-1} \, (\bx-\bp_k)}
\label{eqn:ewa_filter}
\end{equation}
where $s$ is chosen to cover a single pixel in screen space.

While our Mip filter shares similarities with the EWA filter~\cite{zwicker2001ewa}, their underlying principles are distinct. Our Mip filter is designed to replicate the box filter in the imaging process, targeting an exact approximation of a single pixel. Conversely, the EWA filter's role is to limit the frequency signal's bandwidth, and the size of the filter is chosen empirically. The EWA paper~\cite{zwicker2001ewa,heckbert1989fundamentals} even advocates for an identity covariance matrix, effectively occupying a 3x3 pixel region on the screen. However, this approach leads to overly smooth results when zooming out as we will show in our experiments.

\section{Experiments}

\newcommand{\multiwidth}{0.16\textwidth}

\begin{figure*}[t]
    \centering
    \setlength{\tabcolsep}{0.1em}
    \renewcommand{\arraystretch}{0.4}
    \hfill{}\hspace*{-0.5em}
    \scriptsize
    \begin{tabular}{ccccccc}
    \multirow{1}{*}[3ex]{\rotatebox{90}{Full}} &
    \includegraphics[width=\multiwidth]{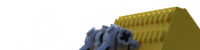}&
    \includegraphics[width=\multiwidth]{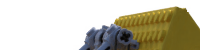}&
    \includegraphics[width=\multiwidth]{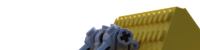}&
    \includegraphics[width=\multiwidth]{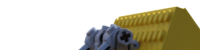}&
    \includegraphics[width=\multiwidth]{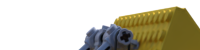}&
    \includegraphics[width=\multiwidth]{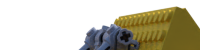}
    \\
    \multirow{1}{*}[4ex]{\rotatebox{90}{$\nicefrac{1}{2}$}} &
    \includegraphics[width=\multiwidth]{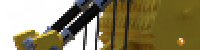}&
    \includegraphics[width=\multiwidth]{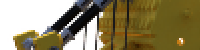}&
    \includegraphics[width=\multiwidth]{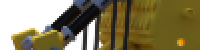}&
    \includegraphics[width=\multiwidth]{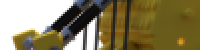}&
    \includegraphics[width=\multiwidth]{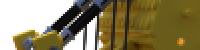}&
    \includegraphics[width=\multiwidth]{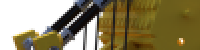}
    \\
    \multirow{1}{*}[4ex]{\rotatebox{90}{$\nicefrac{1}{4}$ }} &
    \includegraphics[width=\multiwidth]{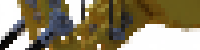}&
    \includegraphics[width=\multiwidth]{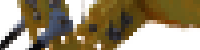}&
    \includegraphics[width=\multiwidth]{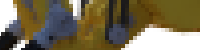}&
    \includegraphics[width=\multiwidth]{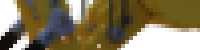}&
    \includegraphics[width=\multiwidth]{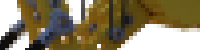}&
    \includegraphics[width=\multiwidth]{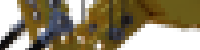}
    \\
    \multirow{1}{*}[4ex]{\rotatebox{90}{$\nicefrac{1}{8}$ }} &
    \includegraphics[width=\multiwidth]{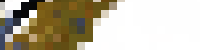}&
    \includegraphics[width=\multiwidth]{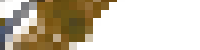}&
    \includegraphics[width=\multiwidth]{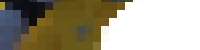}&
    \includegraphics[width=\multiwidth]{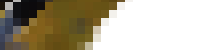}&
    \includegraphics[width=\multiwidth]{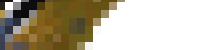}&
    \includegraphics[width=\multiwidth]{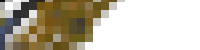}
    \\
    \multirow{1}{*}[3ex]{\rotatebox{90}{Full}} &
    \includegraphics[width=\multiwidth]{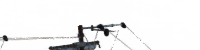}&
    \includegraphics[width=\multiwidth]{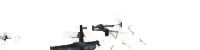}&
    \includegraphics[width=\multiwidth]{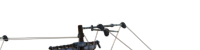}&
    \includegraphics[width=\multiwidth]{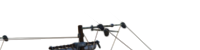}&
    \includegraphics[width=\multiwidth]{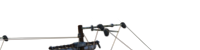}&
    \includegraphics[width=\multiwidth]{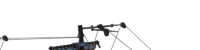}
    \\
    \multirow{1}{*}[4ex]{\rotatebox{90}{$\nicefrac{1}{2}$}} &
    \includegraphics[width=\multiwidth]{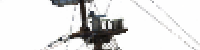}&
    \includegraphics[width=\multiwidth]{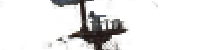}&
    \includegraphics[width=\multiwidth]{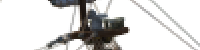}&
    \includegraphics[width=\multiwidth]{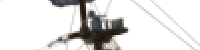}&
    \includegraphics[width=\multiwidth]{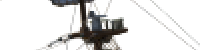}&
    \includegraphics[width=\multiwidth]{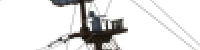}
    \\
    \multirow{1}{*}[4ex]{\rotatebox{90}{$\nicefrac{1}{4}$ }} &
    \includegraphics[width=\multiwidth]{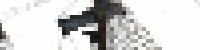}&
    \includegraphics[width=\multiwidth]{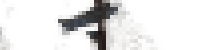}&
    \includegraphics[width=\multiwidth]{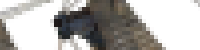}&
    \includegraphics[width=\multiwidth]{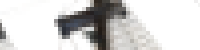}&
    \includegraphics[width=\multiwidth]{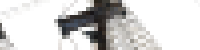}&
    \includegraphics[width=\multiwidth]{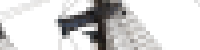}
    \\
    \multirow{1}{*}[4ex]{\rotatebox{90}{$\nicefrac{1}{8}$ }} &
    \includegraphics[width=\multiwidth]{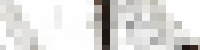}&
    \includegraphics[width=\multiwidth]{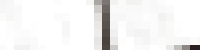}&
    \includegraphics[width=\multiwidth]{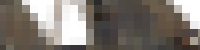}&
    \includegraphics[width=\multiwidth]{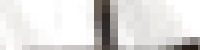}&
    \includegraphics[width=\multiwidth]{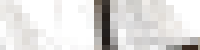}&
    \includegraphics[width=\multiwidth]{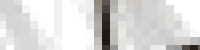}
    \\
    \multirow{1}{*}[3ex]{\rotatebox{90}{Full}} &
    \includegraphics[width=\multiwidth]{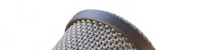}&
    \includegraphics[width=\multiwidth]{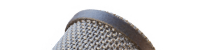}&
    \includegraphics[width=\multiwidth]{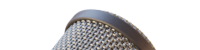}&
    \includegraphics[width=\multiwidth]{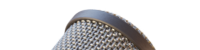}&
    \includegraphics[width=\multiwidth]{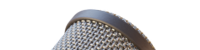}&
    \includegraphics[width=\multiwidth]{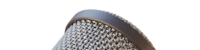}
    \\
    \multirow{1}{*}[4ex]{\rotatebox{90}{$\nicefrac{1}{2}$}} &
    \includegraphics[width=\multiwidth]{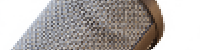}&
    \includegraphics[width=\multiwidth]{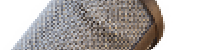}&
    \includegraphics[width=\multiwidth]{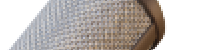}&
    \includegraphics[width=\multiwidth]{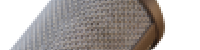}&
    \includegraphics[width=\multiwidth]{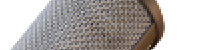}&
    \includegraphics[width=\multiwidth]{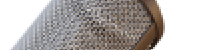}
    \\
    \multirow{1}{*}[4ex]{\rotatebox{90}{$\nicefrac{1}{4}$ }} &
    \includegraphics[width=\multiwidth]{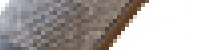}&
    \includegraphics[width=\multiwidth]{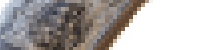}&
    \includegraphics[width=\multiwidth]{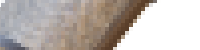}&
    \includegraphics[width=\multiwidth]{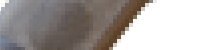}&
    \includegraphics[width=\multiwidth]{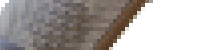}&
    \includegraphics[width=\multiwidth]{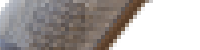}
    \\
    \multirow{1}{*}[4ex]{\rotatebox{90}{$\nicefrac{1}{8}$ }} &
    \includegraphics[width=\multiwidth]{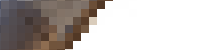}&
    \includegraphics[width=\multiwidth]{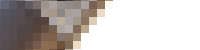}&
    \includegraphics[width=\multiwidth]{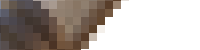}&
    \includegraphics[width=\multiwidth]{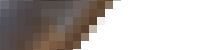}&
    \includegraphics[width=\multiwidth]{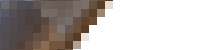}&
    \includegraphics[width=\multiwidth]{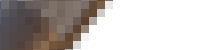}
    \\
    {}& Mip-NeRF~\cite{Barron2021ICCV} & Tri-MipRF~\cite{Hu2023ICCV} & 3DGS~\cite{kerbl3Dgaussians} & 3DGS~\cite{kerbl3Dgaussians} + EWA~\cite{zwicker2001ewa} & Mip-Splatting (ours) & GT  
    \end{tabular}
    \vspace{-0.1in}
    \caption{
    \textbf{Single-scale Training and Multi-scale Testing on the Blender Dataset~\cite{mildenhall2020nerf}.}
    All methods are trained at full resolution and evaluated at different (smaller) resolutions to mimic zoom-out. Methods based on 3DGS capture fine details better than Mip-NeRF~\cite{Barron2021ICCV} and Tri-MipRF~\cite{Hu2023ICCV} at training resolution. Mip-Splatting surpasses both 3DGS~\cite{kerbl3Dgaussians} and 3DGS + EWA~\cite{zwicker2001ewa} at lower resolutions.
    }
    \label{fig:blender_single_train_multi_test}
    \vspace{-0.1in}
\end{figure*}

\begin{table*}[]
    \renewcommand{\tabcolsep}{1pt}
    \centering
    \resizebox{0.95\linewidth}{!}{
    \begin{tabular}{@{}l@{\,\,}|ccccc|ccccc|ccccc}
    & \multicolumn{5}{c|}{PSNR $\uparrow$} & \multicolumn{5}{c|}{SSIM $\uparrow$} & \multicolumn{5}{c}{LPIPS $\downarrow$}  \\
    & Full Res. & $\nicefrac{1}{2}$ Res. & $\nicefrac{1}{4}$ Res. & $\nicefrac{1}{8}$ Res. & Avg. & Full Res. & $\nicefrac{1}{2}$ Res. & $\nicefrac{1}{4}$ Res. & $\nicefrac{1}{8}$ Res. & Avg. & Full Res. & $\nicefrac{1}{2}$ Res. & $\nicefrac{1}{4}$ Res. & $\nicefrac{1}{8}$ Res & Avg.  \\ \hline
    NeRF~\cite{mildenhall2020nerf}&     31.48 & 32.43 &\cellcolor{yellow} 30.29 &\cellcolor{yellow} 26.70 & 30.23 & 0.949 & 0.962 & 0.964 &\cellcolor{yellow} 0.951 & 0.956 & 0.061 & 0.041 & 0.044 & 0.067 & 0.053
\\
MipNeRF~\cite{Barron2021ICCV}& 33.08 &\cellcolor{orange} 33.31 &\cellcolor{orange} 30.91 &\cellcolor{orange} 27.97 &\cellcolor{orange} 31.31 & 0.961 &\cellcolor{yellow} 0.970 &\cellcolor{orange} 0.969 &\cellcolor{orange} 0.961 &\cellcolor{orange} 0.965 & 0.045 &\cellcolor{yellow} 0.031 &\cellcolor{yellow} 0.036 &\cellcolor{yellow} 0.052 &\cellcolor{yellow} 0.041
\\
\hline
TensoRF~\cite{Chen2022ECCV}&  32.53 & 32.91 & 30.01 & 26.45 & 30.48 & 0.960 & 0.969 &\cellcolor{yellow} 0.965 & 0.948 &\cellcolor{yellow} 0.961 & 0.044 &\cellcolor{yellow} 0.031 & 0.044 & 0.073 & 0.048
\\
Instant-NGP~\cite{muller2022instant}& 33.09 &\cellcolor{yellow} 33.00 & 29.84 & 26.33 &\cellcolor{yellow} 30.57 & 0.962 & 0.969 & 0.964 & 0.947 &\cellcolor{yellow} 0.961 & 0.044 & 0.033 & 0.046 & 0.075 & 0.049
\\
Tri-MipRF~\cite{Hu2023ICCV}&  32.89 & 32.84 & 28.29 & 23.87 & 29.47 & 0.958 & 0.967 & 0.951 & 0.913 & 0.947 & 0.046 & 0.033 & 0.046 & 0.075 & 0.050
\\
3DGS~\cite{kerbl3Dgaussians}&\cellcolor{yellow}   33.33 & 26.95 & 21.38 & 17.69 & 24.84 &\cellcolor{tablered} 0.969 & 0.949 & 0.875 & 0.766 & 0.890 &\cellcolor{tablered} 0.030 & 0.032 & 0.066 & 0.121 & 0.063
\\
\hline
3DGS~\cite{kerbl3Dgaussians} + EWA~\cite{zwicker2001ewa}&\cellcolor{tablered}33.51 & 31.66 & 27.82 & 24.63 & 29.40 &\cellcolor{tablered} 0.969 &\cellcolor{orange} 0.971 & 0.959 & 0.940 & 0.960 &\cellcolor{yellow} 0.032 &\cellcolor{orange} 0.024 &\cellcolor{orange} 0.033 &\cellcolor{orange} 0.047 &\cellcolor{orange} 0.034
\\
Mip-Splatting (ours)&\cellcolor{orange}   33.36 &\cellcolor{tablered} 34.00 &\cellcolor{tablered}31.85 &\cellcolor{tablered}28.67 &\cellcolor{tablered}31.97 &\cellcolor{tablered}0.969 &\cellcolor{tablered}0.977 &\cellcolor{tablered}0.978 &\cellcolor{tablered}0.973 &\cellcolor{tablered}0.974 &\cellcolor{orange} 0.031 &\cellcolor{tablered}0.019 &\cellcolor{tablered}0.019 &\cellcolor{tablered}0.026 &\cellcolor{tablered}0.024
    \end{tabular}
    }
    \vspace{-0.1in}
    \caption{
    \textbf{Single-scale Training and Multi-scale Testing on the Blender Dataset~\cite{mildenhall2020nerf}.}
    All methods are trained on full-resolution images and evaluated at four different (smaller) resolutions, with lower resolutions simulating zoom-out effects.
    While Mip-Splatting yields comparable results at training resolution, it significantly surpasses previous work at all other scales.
    }
    \label{tab:avg_blender_results_single_train_multi_test}
    \vspace{-0.1in}
\end{table*}

We first present the implementation details of Mip-Splatting. We then assess its performance on the Blender dataset~\cite{mildenhall2020nerf} and the challenging Mip-NeRF 360 dataset~\cite{barron2022mipnerf360}. Finally, we discuss the limitations of our approach.

\subsection{Implementation}
We build our method upon the popular open-source 3DGS code base~\cite{kerbl3Dgaussians}\footnote{\url{https://github.com/graphdeco-inria/gaussian-splatting}}. Following~\cite{kerbl3Dgaussians}, we train our models for 30K iterations across all scenes and use the same loss function, Gaussian density control strategy, schedule and hyperparameters. For efficiency, we recompute the sampling rate of each 3D Gaussian every $m=100$ iterations. We choose the variance of our 2D Mip filter as 0.1, approximating a single pixel, and the variance of our 3D smoothing filter as 0.2, totaling 0.3 for a fair comparison with 3DGS~\cite{kerbl3Dgaussians} and 3DGS + EWA~\cite{zwicker2001ewa} which replaces the dilation of 3DGS with the EWA filter. 

\subsection{Evaluation on the Blender Dataset}
\boldparagraph{Multi-scale Training and Multi-scale Testing}
Following previous work~\cite{Barron2021ICCV,Hu2023ICCV}, we train our model with multi-scale data and evaluate on multi-scale data. Similar to~\cite{Barron2021ICCV,Hu2023ICCV} where rays of full resolution images are sampled more frequently compared to lower resolution images, we sample 40 percent of full resolution images and 20 percent from other image resolutions each. Our quantitative evaluation is shown in~\tabref{tab:avg_multiblender_results}. Our approach attains comparable or superior performance compared to state-of-the-art methods such as Mip-NeRF~\cite{Barron2021ICCV} and Tri-MipRF~\cite{Hu2023ICCV}. Notably, our method outperforms 3DGS~\cite{kerbl3Dgaussians} and 3DGS + EWA~\cite{zwicker2001ewa} by a substantial margin, owing to its 2D Mip filter. 

\boldparagraph{Single-scale Training and Multi-scale Testing}
Contrary to prior work that evaluates models trained on single-scale data at the same scale, we consider the an important new setting that involves training on full-resolution images and rendering at various resolutions (\ie $1 \times$, $\nicefrac{1}{2}$, $\nicefrac{1}{4}$, and $\nicefrac{1}{8}$) to mimic zoom-out effects. In the absence of a public benchmark for this setting, we trained all baseline methods ourselves. We use NeRFAcc~\cite{Li2023NeRFAcc}'s implementation for NeRF~\cite{mildenhall2020nerf}, Instant-NGP~\cite{muller2022instant}, and TensoRF~\cite{Chen2022ECCV} for its efficiency. Official implementations were employed for Mip-NeRF~\cite{Barron2021ICCV}, Tri-MipRF~\cite{Hu2023ICCV}, and 3DGS~\cite{kerbl3Dgaussians}. The quantitative results, as presented in Table~\ref{tab:avg_blender_results_single_train_multi_test}, indicate that our method significantly outperforms all existing state-of-the-art methods. A qualitative comparison is provided in Figure~\ref{fig:blender_single_train_multi_test}. Methods based on 3DGS~\cite{kerbl3Dgaussians} capture fine details more effectively than Mip-NeRF~\cite{Barron2021ICCV} and Tri-MipRF~\cite{Hu2023ICCV}, but only at the original training scale. Notably, our method surpasses both 3DGS~\cite{kerbl3Dgaussians} and 3DGS + EWA~\cite{zwicker2001ewa} in rendering quality at lower resolutions. In particular, 3DGS~\cite{kerbl3Dgaussians} exhibits dilation artifacts. EWA splatting~\cite{zwicker2001ewa} uses a large low pass filter to limit the frequency of the rendered images, resulting in oversmoothed images, which becomes particularly apparent at lower resolutions.

\subsection{Evaluation on the Mip-NeRF 360 Dataset}
\newcommand{\upwidth}{0.16\textwidth}

\begin{figure*}[t]
    \centering
    \setlength{\tabcolsep}{0.1em}
    \renewcommand{\arraystretch}{0.4}
    \scriptsize
    \begin{tabular}{cccccc}
    \includegraphics[width=\upwidth]{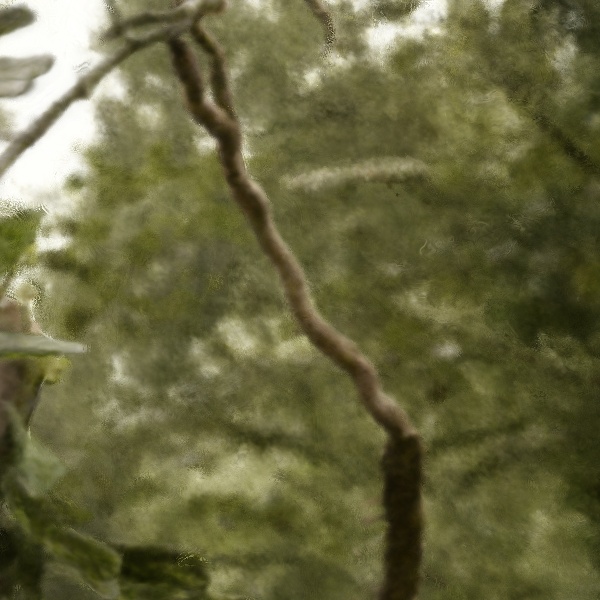}&    
    \includegraphics[width=\upwidth]{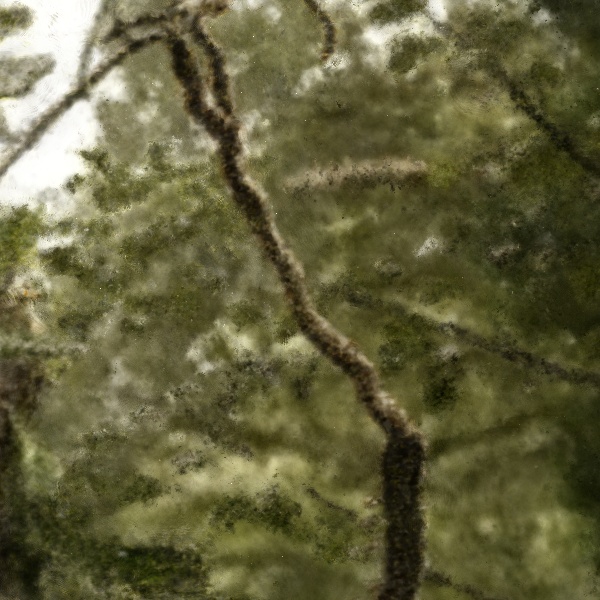}&    
    \includegraphics[width=\upwidth]{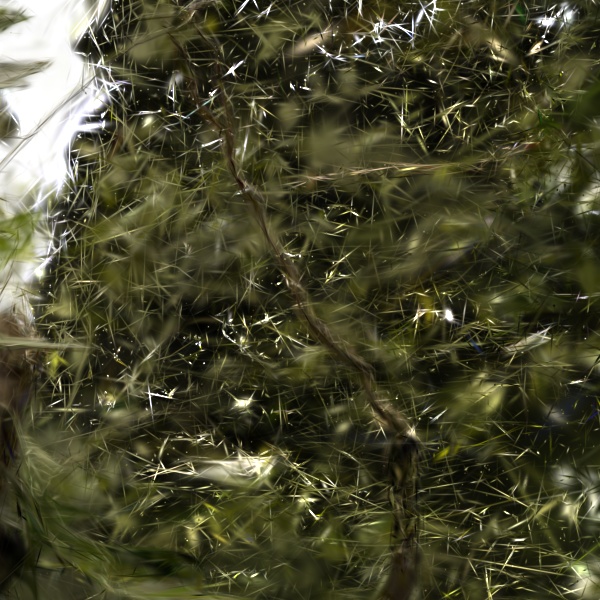}&    
    \includegraphics[width=\upwidth]{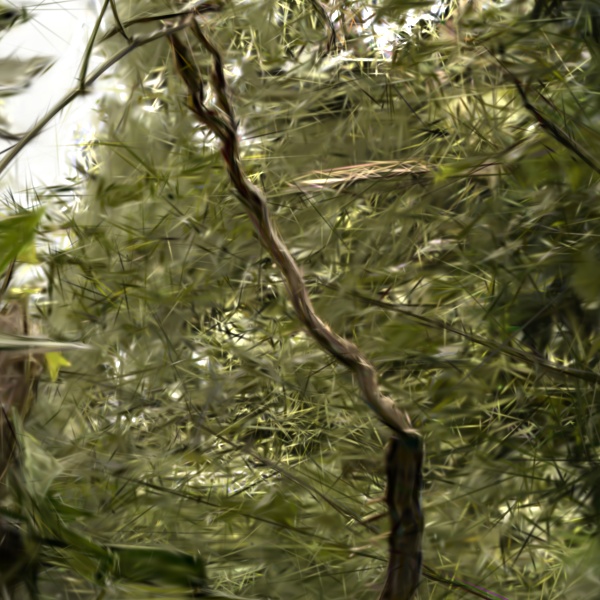}&    
    \includegraphics[width=\upwidth]{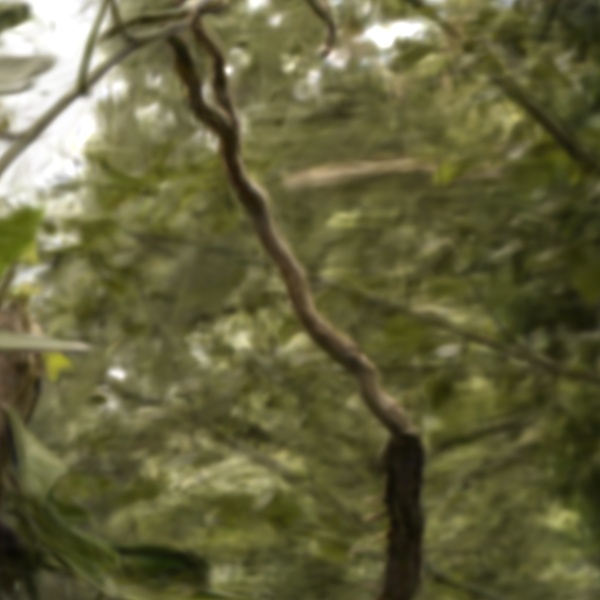}&    
    \includegraphics[width=\upwidth]{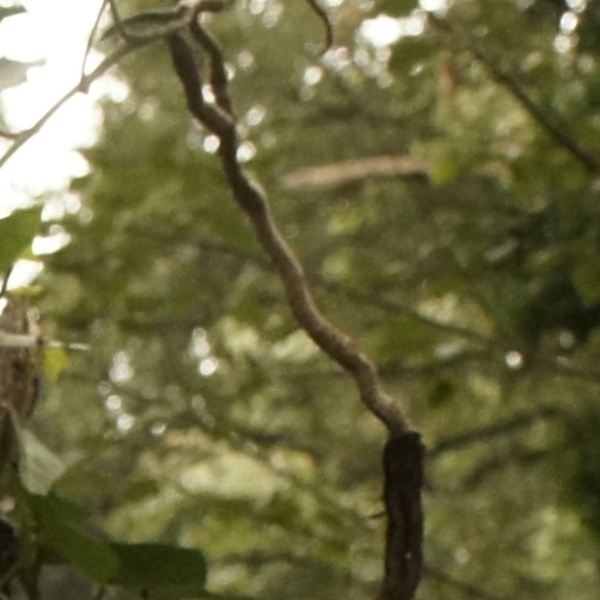}  
    \\
    \includegraphics[width=\upwidth]{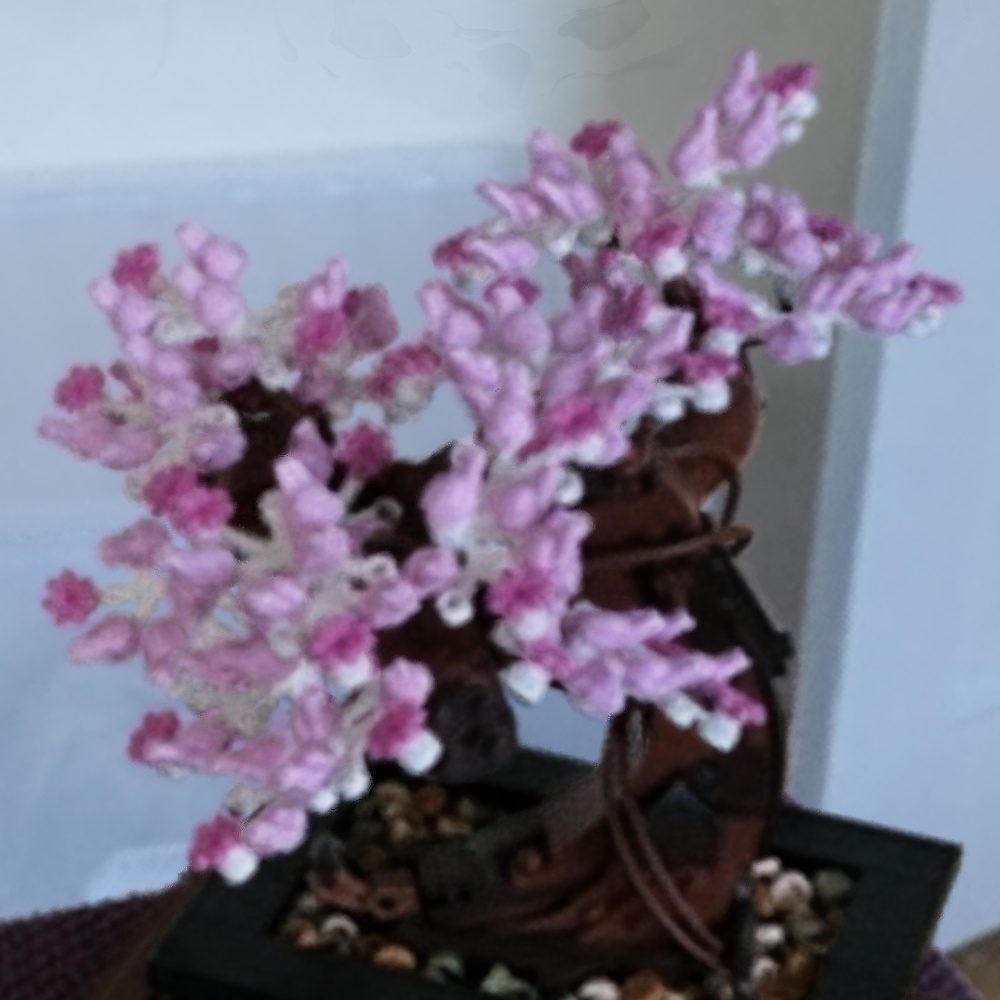}&    
    \includegraphics[width=\upwidth]{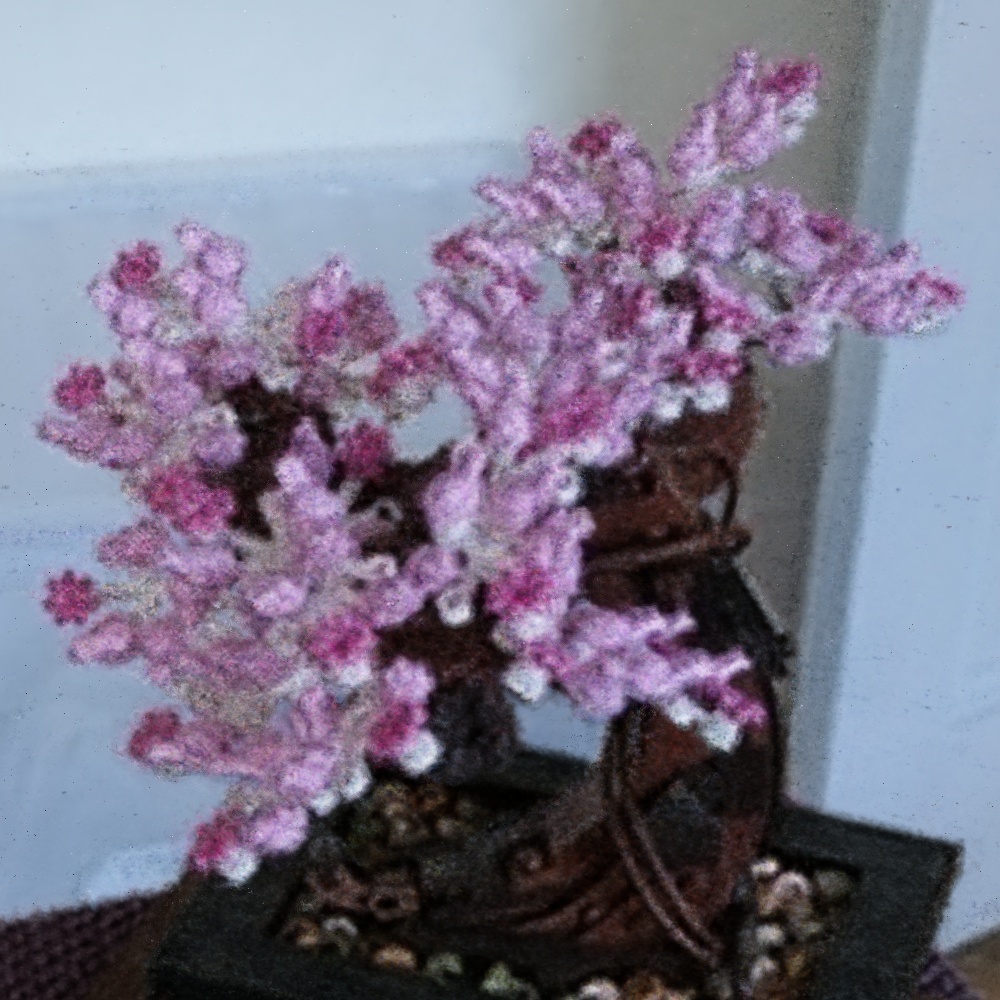}&    
    \includegraphics[width=\upwidth]{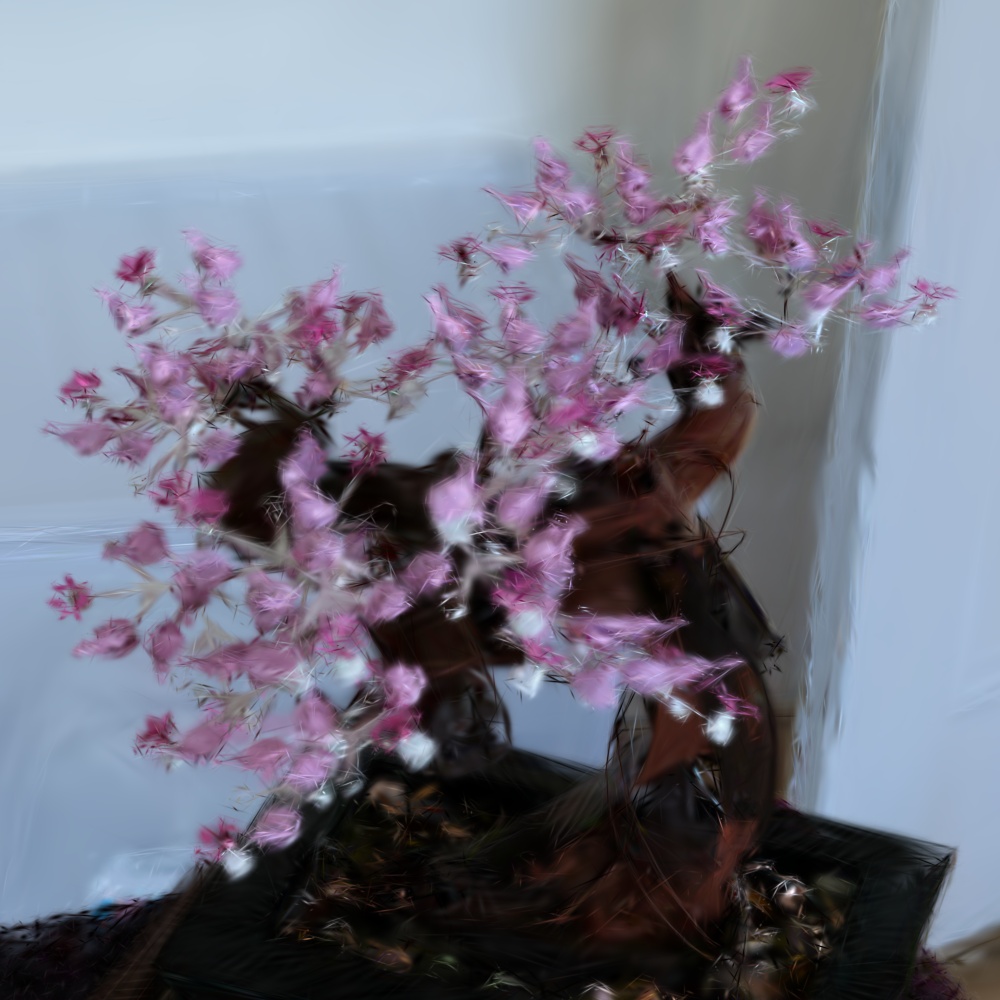}&    
    \includegraphics[width=\upwidth]{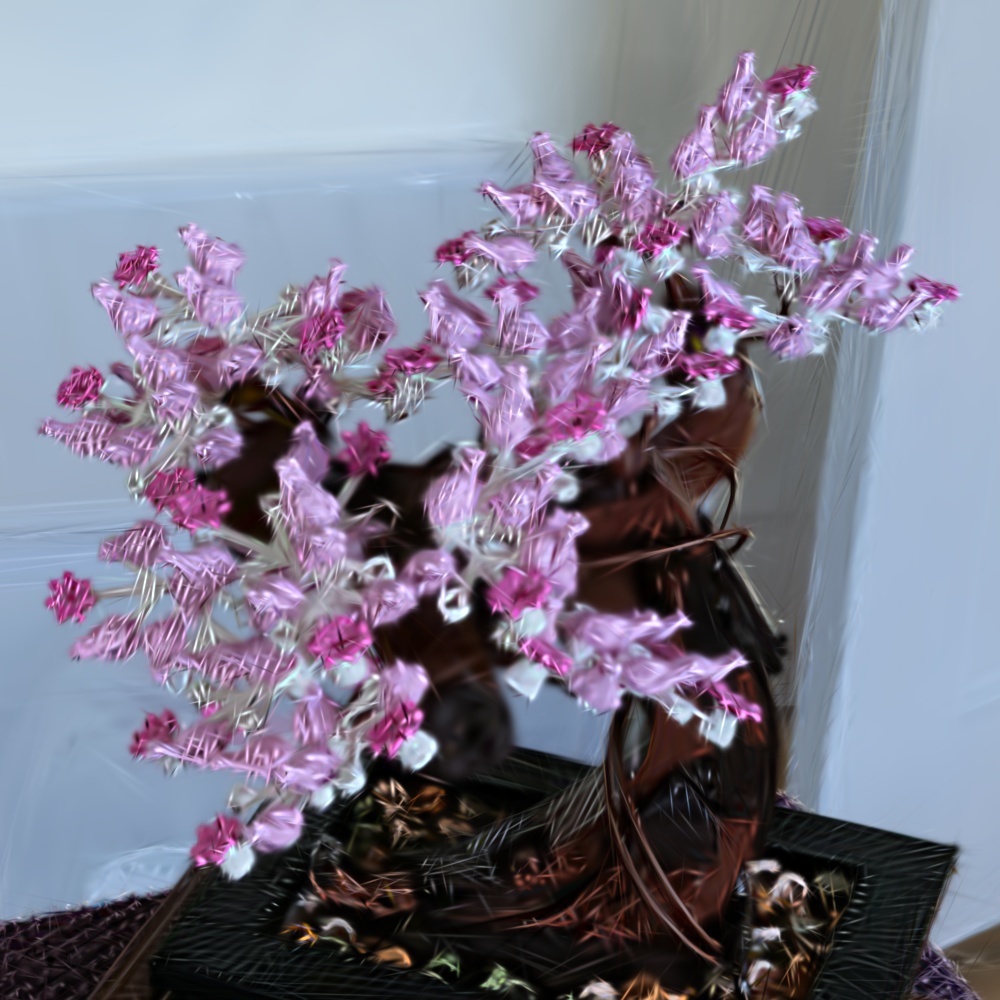}&    
    \includegraphics[width=\upwidth]{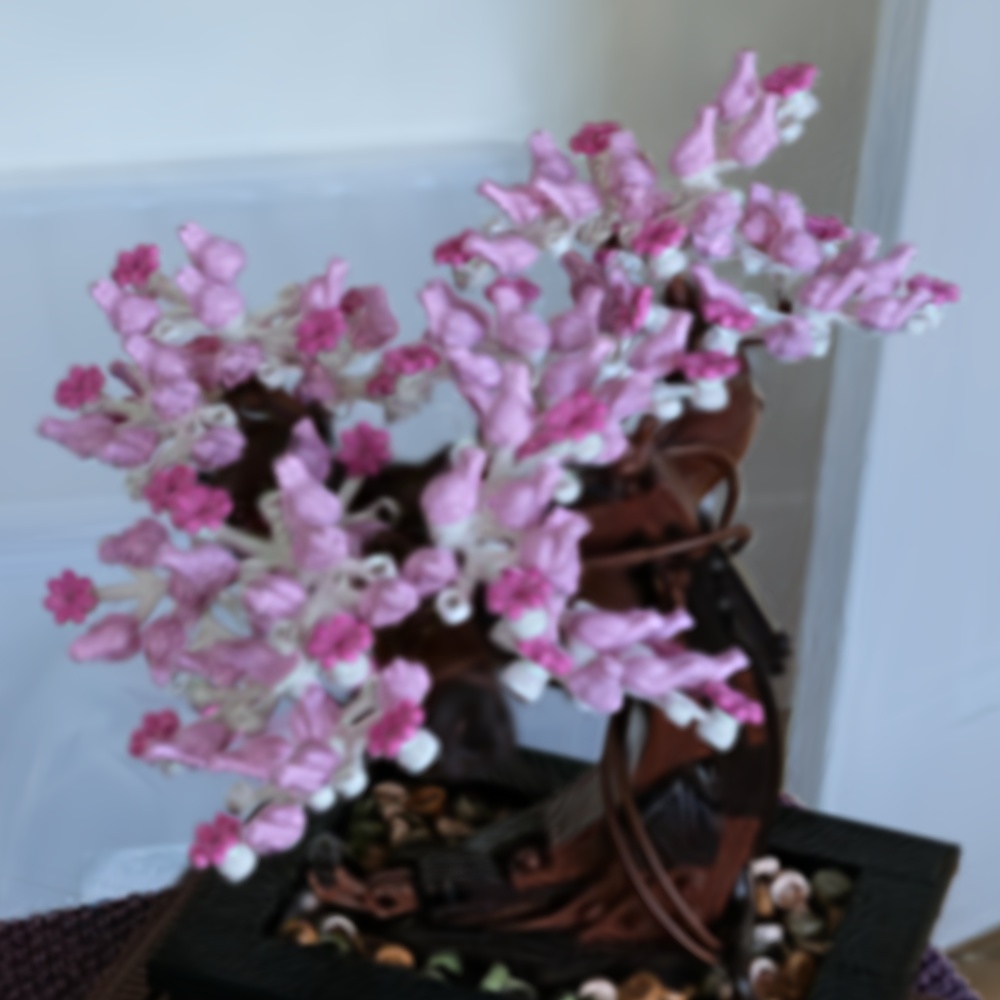}&    
    \includegraphics[width=\upwidth]{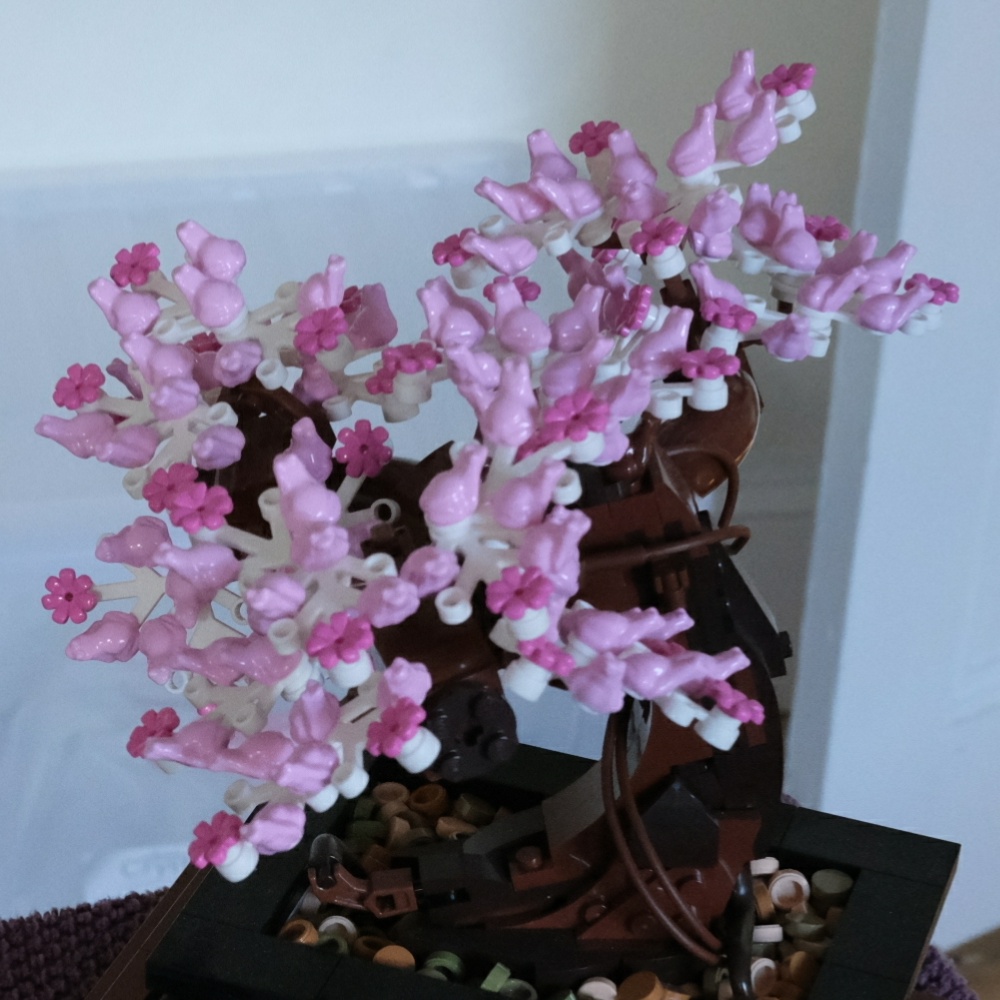}  
    \\
    \includegraphics[width=\upwidth]{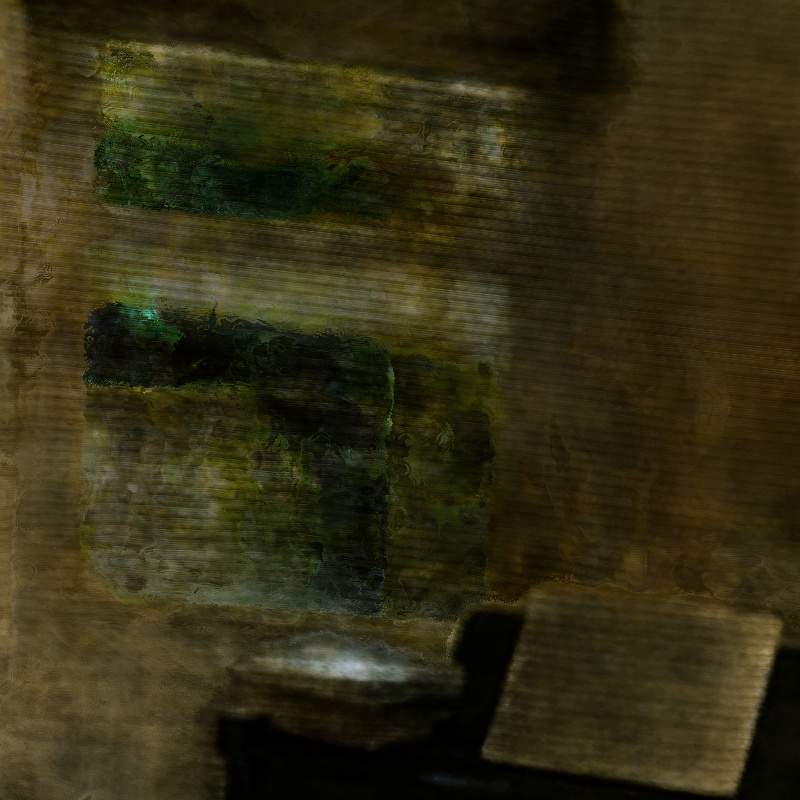}&    
    \includegraphics[width=\upwidth]{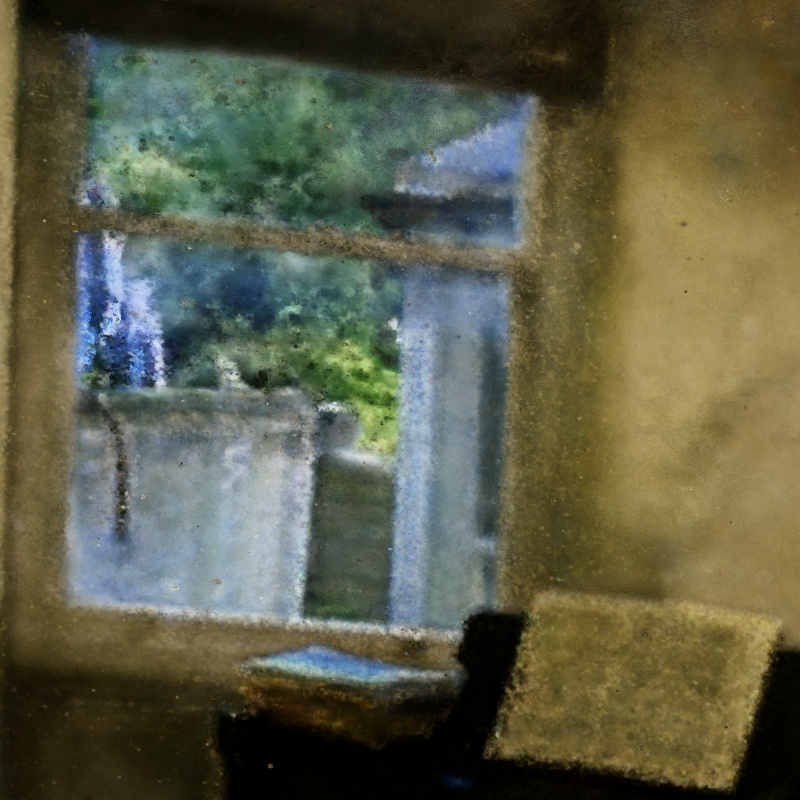}&    
    \includegraphics[width=\upwidth]{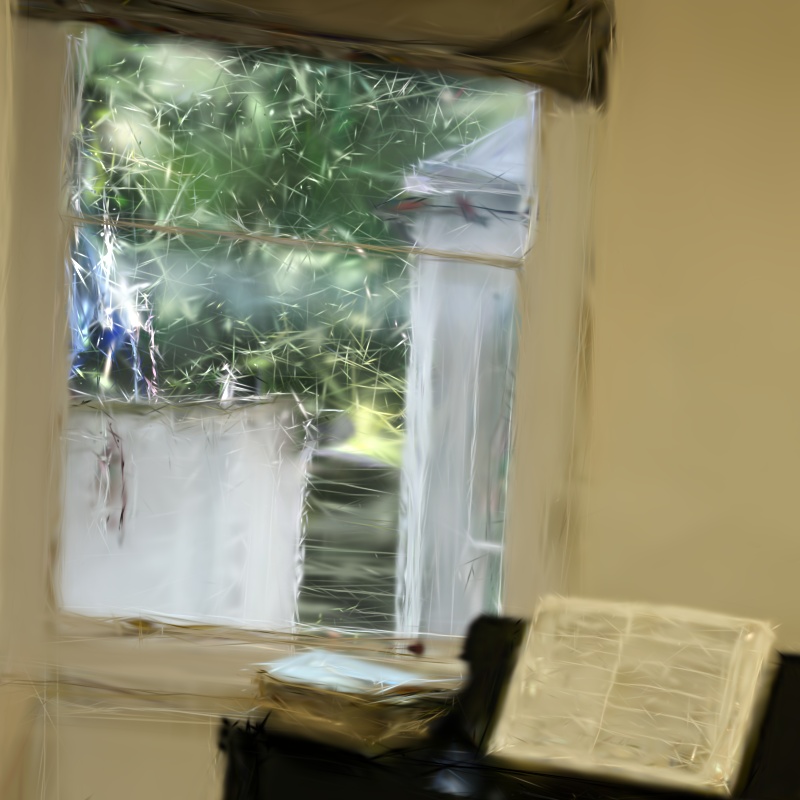}&    
    \includegraphics[width=\upwidth]{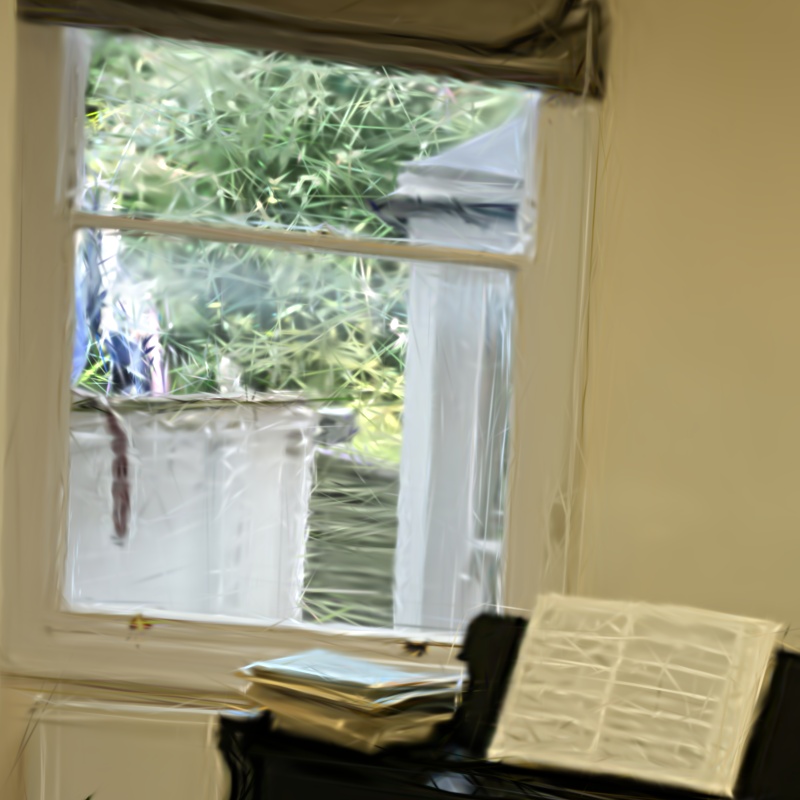}&    
    \includegraphics[width=\upwidth]{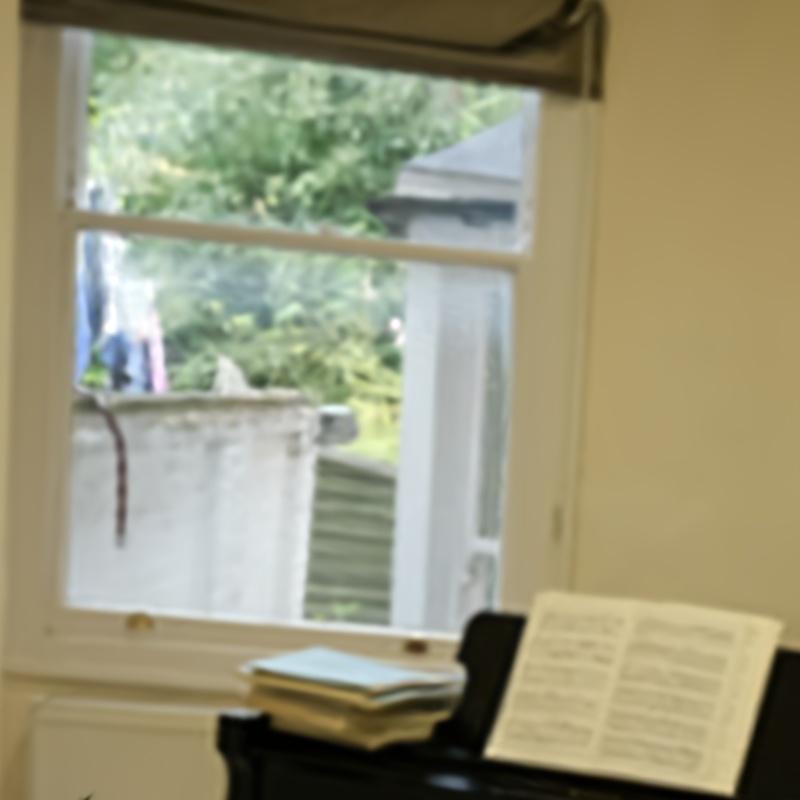}&    
    \includegraphics[width=\upwidth]{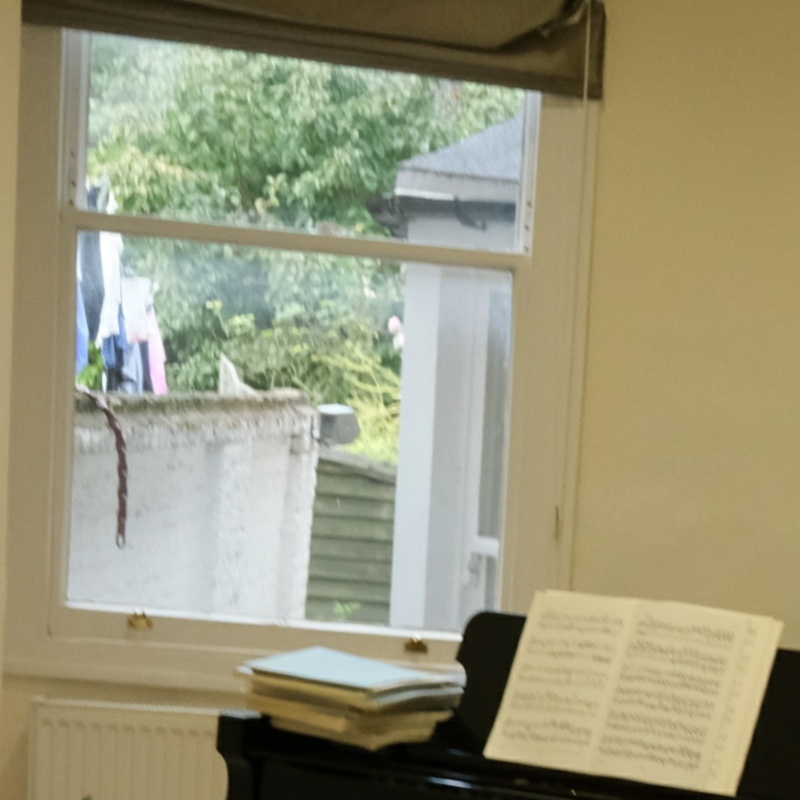}  
    \\
    Mip-NeRF 360~\cite{barron2022mipnerf360} & Zip-NeRF~\cite{Barron2023ICCV} & 3DGS~\cite{kerbl3Dgaussians} & 3DGS~\cite{kerbl3Dgaussians} + EWA~\cite{zwicker2001ewa} & Mip-Splatting (ours) & GT  
    \end{tabular}
    \vspace{-0.1in}
    \caption{\textbf{Single-scale Training and Multi-scale Testing on the Mip-NeRF 360 Dataset~\cite{barron2022mipnerf360}.} 
    All models are trained on images downsampled by a factor of eight and rendered at full resolution to demonstrate zoom-in/moving closer effects. In contrast to prior work, Mip-Splatting renders images that closely approximate ground truth. Please also note the high-frequency artifacts of 3DGS + EWA~\cite{zwicker2001ewa}.
    }
    \label{fig:360_single_train_multi_test}
    \vspace{-0.1in}
\end{figure*}
\begin{table*}[]
    \renewcommand{\tabcolsep}{1pt}
    \centering
    \resizebox{0.95\linewidth}{!}{
    \begin{tabular}{@{}l@{\,\,}|ccccc|ccccc|ccccc}
    & \multicolumn{5}{c|}{PSNR $\uparrow$} & \multicolumn{5}{c|}{SSIM $\uparrow$} & \multicolumn{5}{c}{LPIPS $\downarrow$}  \\
    & $1 \times$ Res. & $2 \times$ Res. & $4 \times$ Res. & $8 \times$ Res. & Avg. & $1 \times$ Res. & $2 \times$ Res. & $4 \times$ Res. & $8 \times$ Res. & Avg. & $1 \times$ Res. & $2 \times$ Res. & $4 \times$ Res. & $8 \times$ Res. & Avg.  \\ \hline

Instant-NGP~\cite{muller2022instant}&                            26.79  &                            24.76  &\cellcolor{orange}24.27  &\cellcolor{orange}24.27  &                            25.02  &                            0.746  &                            0.639  &                            0.626  &\cellcolor{yellow}0.698  &                            0.677  &                            0.239  &                            0.367  &                            0.445  &                            0.475  &                             0.382  
\\
mip-NeRF 360~\cite{barron2022mipnerf360}&                            29.26  &\cellcolor{yellow}25.18  &\cellcolor{yellow}24.16  &\cellcolor{yellow}24.10  &\cellcolor{orange}25.67  &                            0.860  &                            0.727  &\cellcolor{orange}0.670  &\cellcolor{orange}0.706  &\cellcolor{orange}0.741  &                            0.122  &                            0.260  &\cellcolor{yellow}0.370  &\cellcolor{orange}0.428  &\cellcolor{yellow}                             0.295 
\\
zip-NeRF~\cite{Barron2023ICCV}&\cellcolor{tablered}29.66  &                            23.27  &                            20.87  &                            20.27  &                            23.52  &                            0.875  &                            0.696  &                            0.565  &                            0.559  &                            0.674  &\cellcolor{tablered}0.097  &                            0.257  &                            0.421  &                            0.494  &                             0.318 
\\
3DGS~\cite{kerbl3Dgaussians}&                            29.19  &                            23.50  &                            20.71  &                            19.59  &                            23.25  &\cellcolor{orange}0.880  &\cellcolor{yellow}0.740  &                            0.619  &                            0.619  &                            0.715  &\cellcolor{orange}0.107  &\cellcolor{yellow}0.243  &                            0.394  &                            0.476  &0.305 
\\
\hline
3DGS~\cite{kerbl3Dgaussians} + EWA~\cite{zwicker2001ewa}&\cellcolor{yellow}29.30  &\cellcolor{orange}25.90  &                            23.70  &                            22.81  &\cellcolor{yellow}25.43  &\cellcolor{orange}0.880  &\cellcolor{orange}0.775  &\cellcolor{yellow}0.667  &                            0.643  &\cellcolor{orange}0.741  &                            0.114  &\cellcolor{orange}0.236  &\cellcolor{orange}0.369  &\cellcolor{yellow}0.449  &\cellcolor{orange}0.292
\\
Mip-Splatting (ours)&\cellcolor{orange}29.39  &\cellcolor{tablered}27.39  &\cellcolor{tablered}26.47  &\cellcolor{tablered}26.22  &\cellcolor{tablered}27.37  &\cellcolor{tablered}0.884  &\cellcolor{tablered}0.808  &\cellcolor{tablered}0.754  &\cellcolor{tablered}0.765  &\cellcolor{tablered}0.803  &\cellcolor{yellow}0.108  &\cellcolor{tablered}0.205  &\cellcolor{tablered}0.305  &\cellcolor{tablered}0.392  &\cellcolor{tablered}0.252
    \end{tabular}
    }
    \vspace{-0.1in}
    \caption{
    \textbf{Single-scale Training and Multi-scale Testing on the Mip-NeRF 360 Dataset~\cite{barron2022mipnerf360}.} All methods are trained on the smallest scale ($1 \times$) and evaluated across four scales ($1 \times$, $2 \times$, $4 \times$, and $8 \times$), with evaluations at higher sampling rates simulating zoom-in effects. While our method yields comparable results at the training resolution, it significantly surpasses all previous work at all other scales.}    \label{tab:avg_360_results_single_train_multi_test}
\end{table*}

\boldparagraph{Single-scale Training and Multi-scale Testing}
To simulate zoom-in effects, we train models on data downsampled by a factor of 8 and rendered at successively higher resolutions ($1 \times$, $2 \times$, $4 \times$, and $8 \times$). In the absence of a public benchmark for this setting, we trained all baseline methods ourselves. We use the official implementation for Mip-NeRF 360~\cite{Barron2021ICCV} and 3DGS~\cite{kerbl3Dgaussians} and use a community reimplementation for Zip-NeRF~\cite{Barron2023ICCV}\footnote{\url{https://github.com/SuLvXiangXin/zipnerf-pytorch}} as the code is not available. The results in Table~\ref{tab:avg_360_results_single_train_multi_test} show that our method performs comparable to prior work at the training scale ($1 \times$) and significantly exceeds all state-of-the-art methods at higher resolutions. As depicted in Figure~\ref{fig:360_single_train_multi_test}, our method generates high fidelity imagery without high-frequency artifacts. Notably, both Mip-NeRF 360~\cite{barron2022mipnerf360} and Zip-NeRF~\cite{Barron2023ICCV} exhibit subpar performance at increased resolutions, likely due to their MLPs' inability to extrapolate to out-of-distribution frequencies. While 3DGS~\cite{kerbl3Dgaussians} introduces notable erosion artifacts due to dilation operations, 3DGS + EWA~\cite{zwicker2001ewa} performs better while still yielding pronounced high-frequency artifacts. In contrast, our method avoids such artifacts, yielding aesthetically pleasing images that more closely resemble ground truth. It's important to remark that rendering at higher resolutions is a super-resolution task, and models should not hallucinate high-frequency details absent from the training data.

\boldparagraph{Single-scale Training and Same-scale Testing}
We further evaluate our method on the Mip-NeRF 360 dataset~\cite{barron2022mipnerf360} following the widely used setting, where models are trained and tested at the same scale, with indoor scenes downsampled by a factor of two and outdoor scenes by four. As shown in Table~\ref{tab:avg_360_results}, our method performs on par with 3DGS~\cite{kerbl3Dgaussians} and 3DGS + EWA~\cite{zwicker2001ewa} in this challenging benchmark, without any decrease in performance. This confirms our method's effectiveness to handle various settings.

\begin{table}[]
    \centering
    \resizebox{0.75\linewidth}{!}{
    \begin{tabular}{@{}l@{\,\,}|ccc}
    & \!PSNR $\uparrow$\! & \!SSIM $\uparrow$\! & \!LPIPS $\downarrow$\!\\ 
    \hline
    NeRF~\cite{mildenhall2020nerf,jaxnerf2020github}          &                   23.85 &                   0.605 &                   0.451\\
mip-NeRF~\cite{Barron2021ICCV}                            &                   24.04 &                   0.616 &                   0.441 \\
NeRF++~\cite{kaizhang2020}                                &                   25.11 &                   0.676 &                   0.375 \\
Plenoxels~\cite{yu2022plenoxels}                          &                   23.08 &                   0.626 &                   0.463 \\
Instant NGP ~\cite{muller2022instant,yariv2023bakedsdf}   &                   25.68 &                   0.705 &                   0.302 \\
mip-NeRF 360~\cite{barron2022mipnerf360, multinerf2022}   &                   27.57 &                   0.793 &                   0.234 \\
Zip-NeRF~\cite{Barron2023ICCV}                            & \cellcolor{tablered} 28.54 &   \cellcolor{tablered}   0.828 & \cellcolor{tablered}   0.189 \\
3DGS~\cite{kerbl3Dgaussians}                      &                   27.21 &                   0.815 &                   0.214 \\
3DGS~\cite{kerbl3Dgaussians}*                     &                   27.70 & \cellcolor{yellow}0.826 & \cellcolor{orange}0.202 \\
\hline
3DGS~\cite{kerbl3Dgaussians} + EWA~\cite{zwicker2001ewa}             &  \cellcolor{yellow}  27.77 & \cellcolor{yellow} 0.826 &                   0.206 \\
Mip-Splatting (ours)         &   \cellcolor{orange}27.79 & \cellcolor{orange}0.827 & \cellcolor{yellow}0.203 

  \!\!\!
    \end{tabular}
    }
    \vspace{-0.1in}
    \caption{
    \textbf{Single-scale Training and Same-scale Testing on the Mip-NeRF 360 dataset~\cite{barron2022mipnerf360}.} In the standard in-distribution setting, our approach demonstrates performance on par with many established techniques. $*$ indicates that we retrain the model.
    }
    \label{tab:avg_360_results}
    \vspace{-0.1in}
\end{table}

\subsection{Limitations}
Our method employs a Gaussian filter as an approximation to a box filter for efficiency. However, this approximation introduces errors, particularly when the Gaussian is small in screen space. This issue correlates with our experimental findings, where increased zooming out leads to larger errors, as evidenced in Table~\ref{tab:avg_blender_results_single_train_multi_test}. Additionally, there is a slight increase in training overhead as the sampling rate for each 3D Gaussian must be calculated every $m=100$ iterations. Currently, this computation is performed using PyTorch~\cite{NEURIPS2019_9015} and a more efficient CUDA implementation could potentially reduce this overhead. Designing a better data structure for precomputing and storing the sampling rate, as it depends solely on the camera poses and intrinsics, is an avenue for future work. As mentioned before, the sampling rate computation is the only prerequisite during training and the 3D smoothing filter can be fused with the Gaussian primitives per~\eqnref{eqn:filter_3d_full}, thereby eliminating any additional overhead during rendering.

\section{Conclusion}
We presented Mip-Splatting, a modification to 3D Gaussian Splatting, which introduces two novel filters, namely a 3D smoothing filter and a 2D Mip filter, to achieve alias-free rendering at arbitrary scales. Our 3D smoothing filter effectively limits the maximal frequency of Gaussian primitives to match the sampling constraints imposed by the training images, while the 2D Mip filter approximates the box filter to simulate the physical imaging process. Our experimental results demonstrate that Mip-Splatting is competitive with state-of-the-art methods in terms of performance when training and testing at the same scale / sampling rate. Importantly, it significantly outperforms state-of-the-art methods in out-of-distribution scenarios, when testing at sampling rates different from training, resulting in better generalization to out-of-distribution camera poses and zoom factors.

{
    \small
    \bibliographystyle{ieeenat_fullname}
    \bibliography{bibliography,bibliography_long,bibliography_custom}
}

\clearpage
\setcounter{page}{1}
\maketitlesupplementary

In this \textbf{supplementary document}, we first present ablation studies of Mip-Splatting in~\secref{sec:ablation}. %
Next, we report additional quantitative and quality results in~\secref{sec:addtional_results}.

\section{Ablation}
\label{sec:ablation}
In this section, we evaluate the effectiveness of our 3D smoothing filter and 2D Mip filter in~\secref{sec:ab_3d} and ~\secref{sec:ab_2d}. Then, we present an additional experiment to evaluate both zoom-in and zoom-out effects in the same dataset in~\secref{sec:ab_5scales}.

\subsection{Effectiveness of the 3D Smoothing Filter}
\label{sec:ab_3d}
\newcommand{\ablationwidth}{0.16\textwidth}

\begin{figure*}[t]
    \centering
    \setlength{\tabcolsep}{0.1em}
    \renewcommand{\arraystretch}{0.4}
    \scriptsize
    \begin{tabular}{cccccc}
    \includegraphics[width=\ablationwidth]{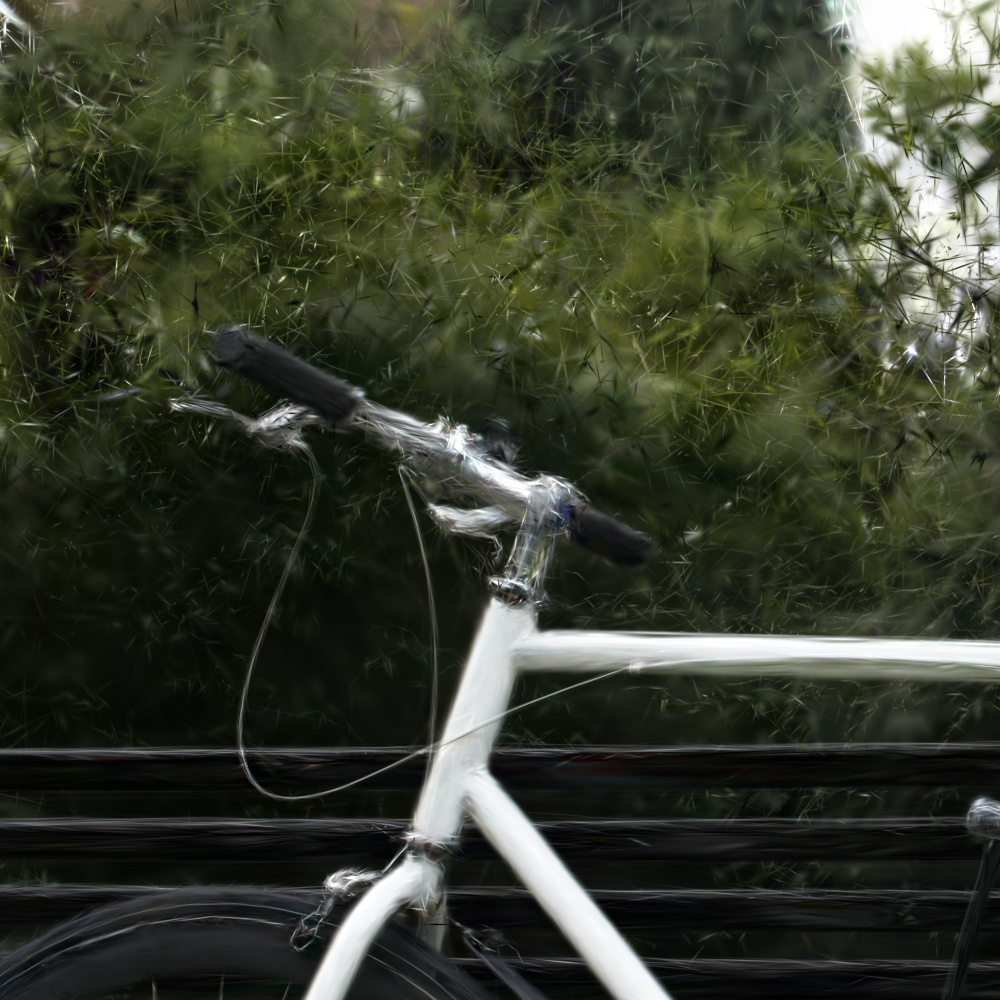}& \includegraphics[width=\ablationwidth]{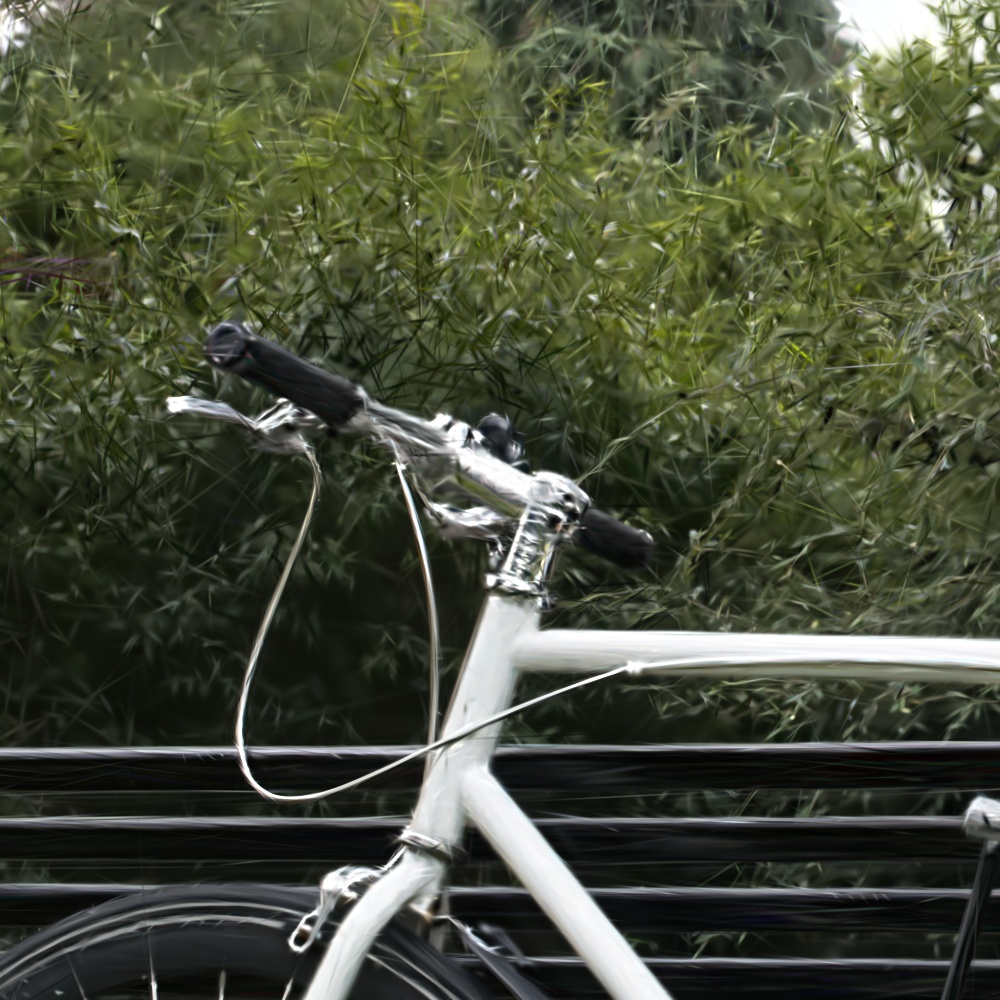}&
    \includegraphics[width=\ablationwidth]{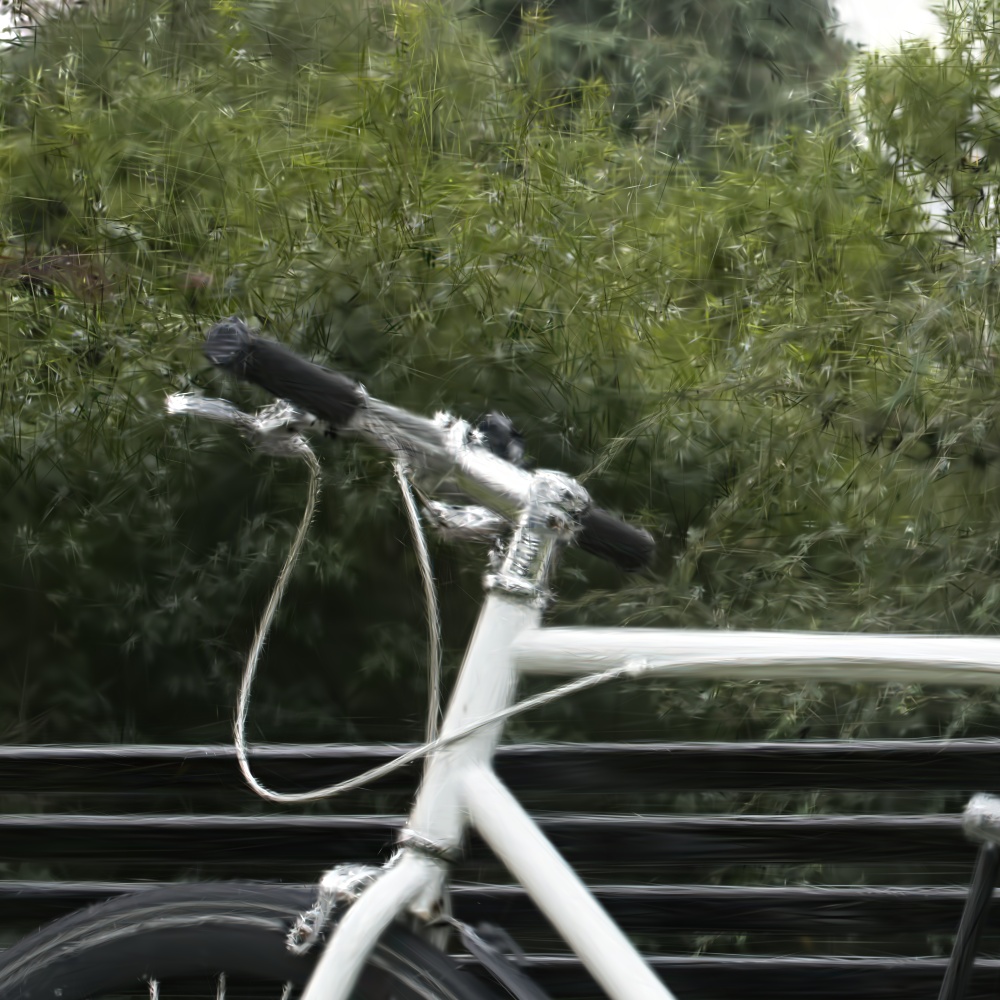}&
    \includegraphics[width=\ablationwidth]{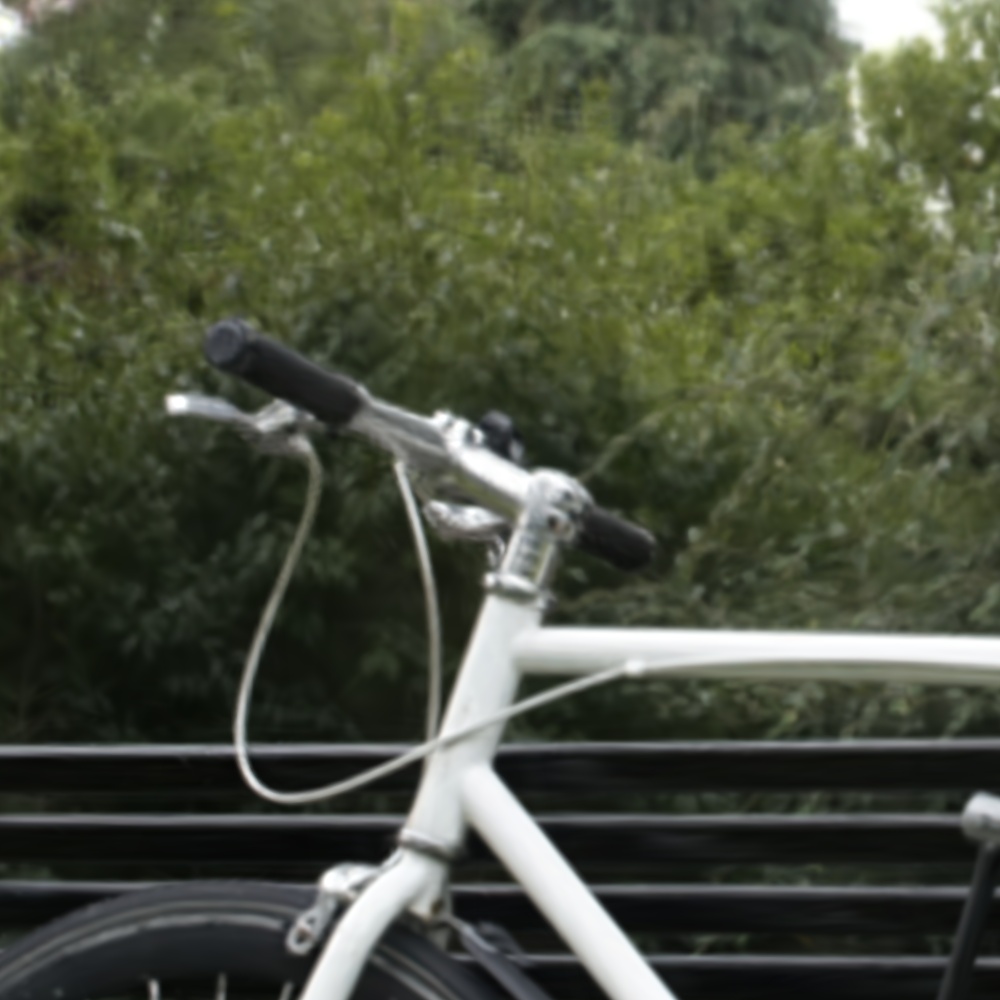}& 
    \includegraphics[width=\ablationwidth]{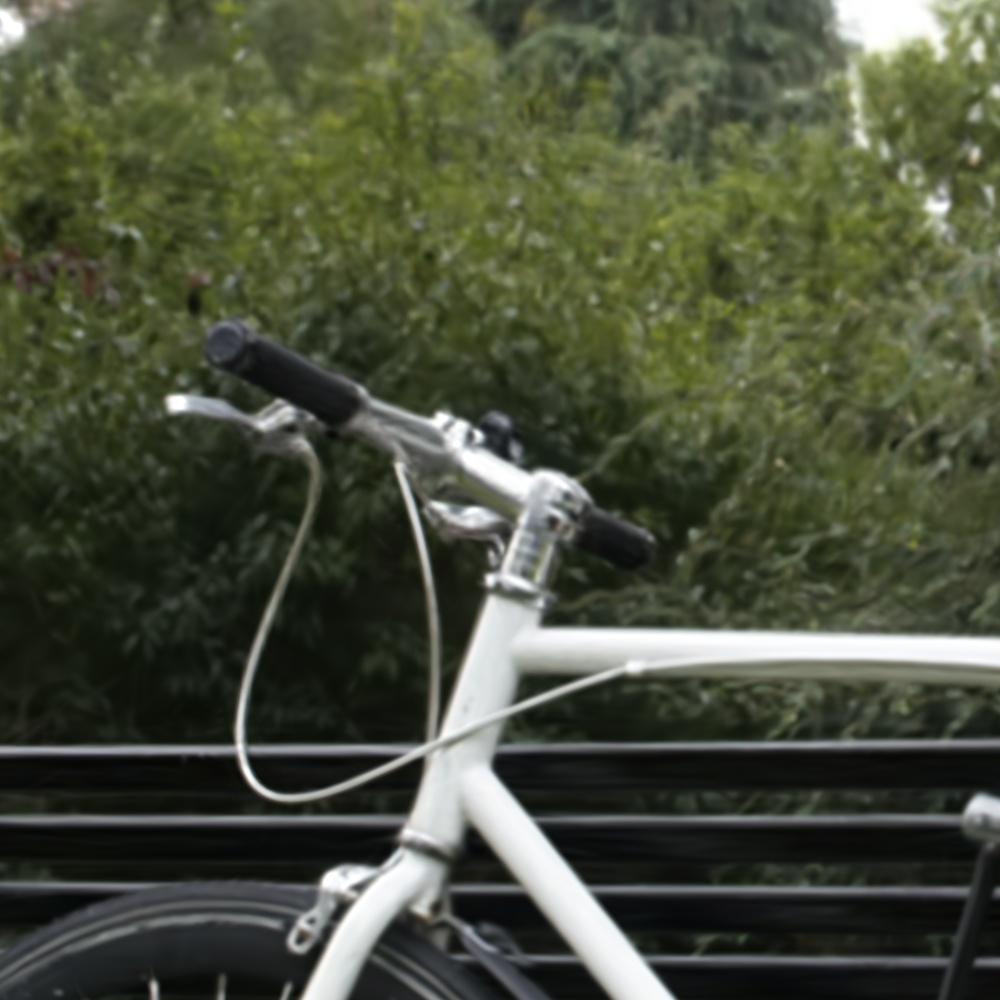}& 
    \includegraphics[width=\ablationwidth]{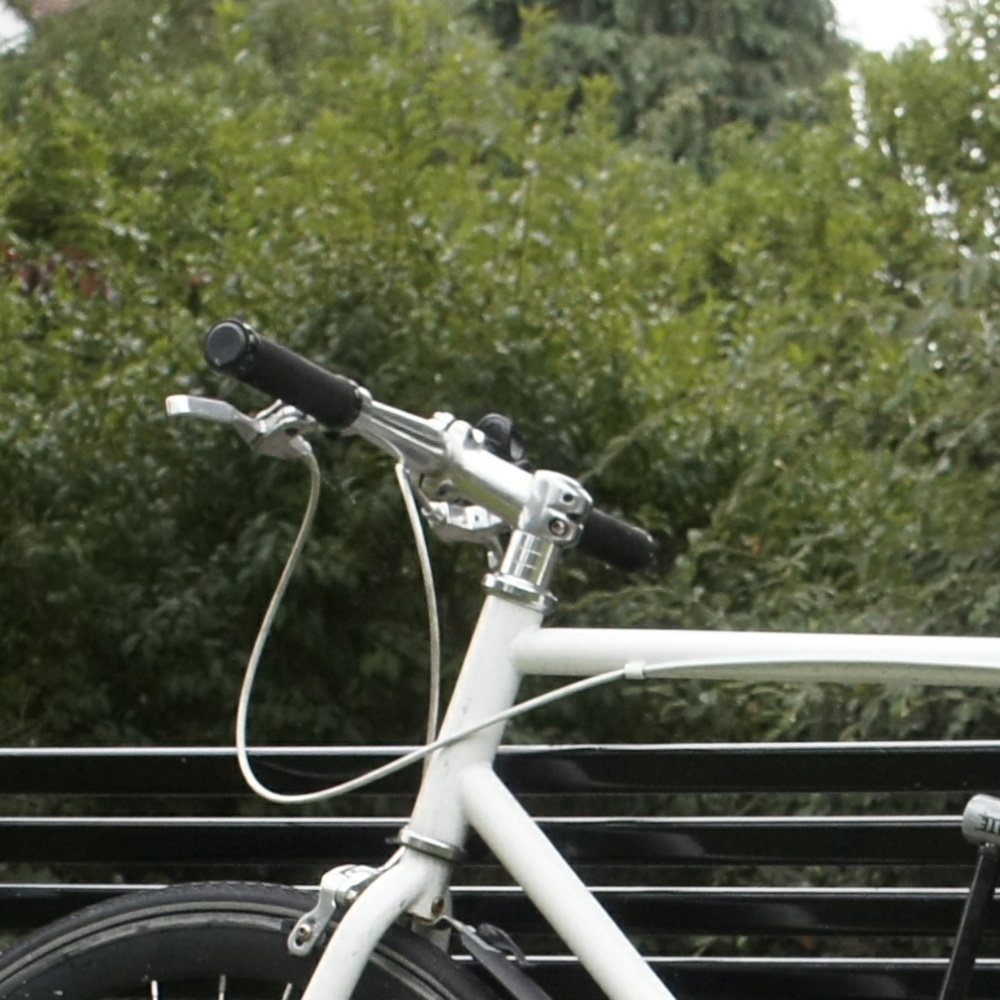}  
    \\ 
    \includegraphics[width=\ablationwidth]{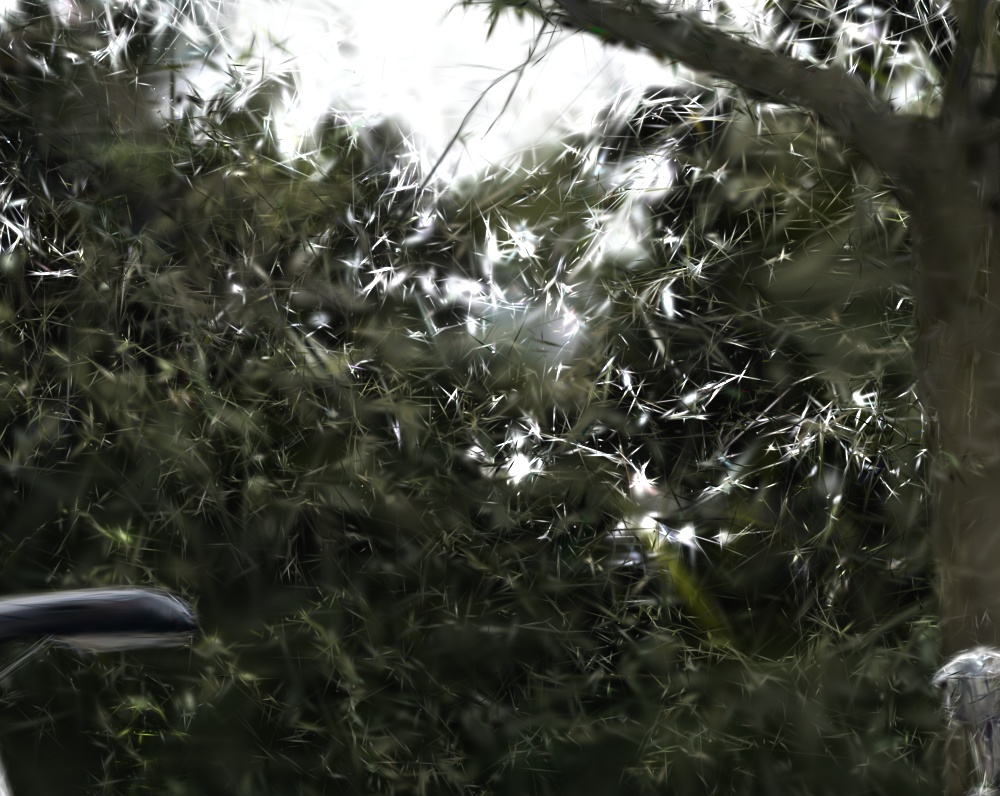}& \includegraphics[width=\ablationwidth]{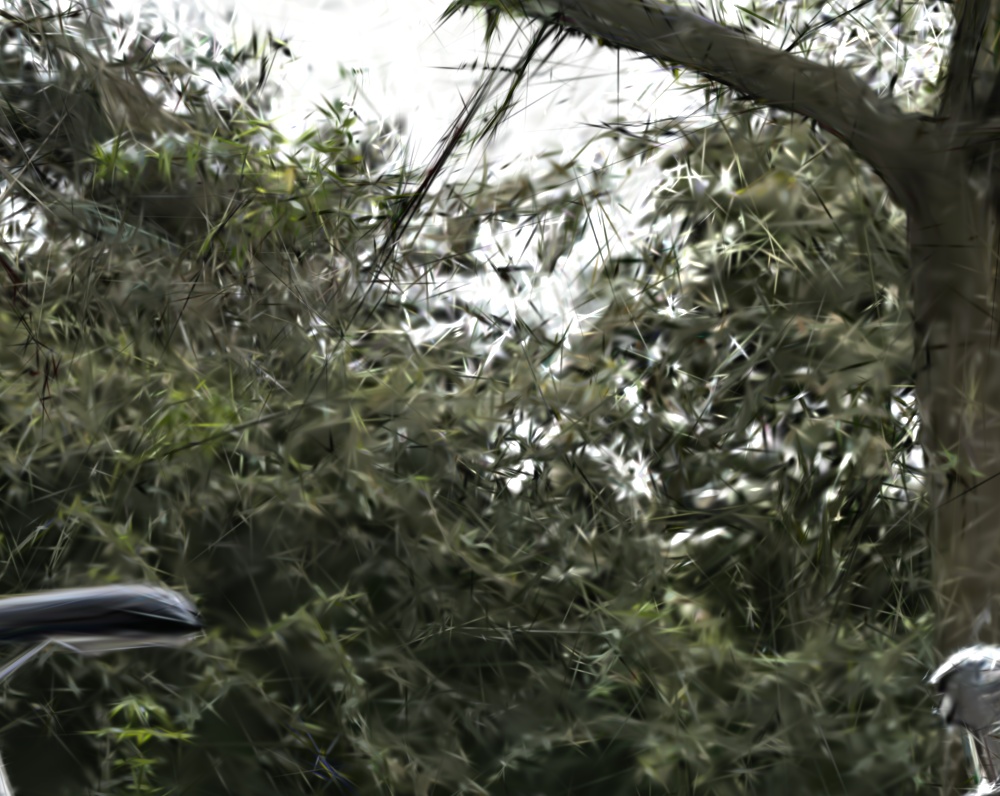}&
    \includegraphics[width=\ablationwidth]{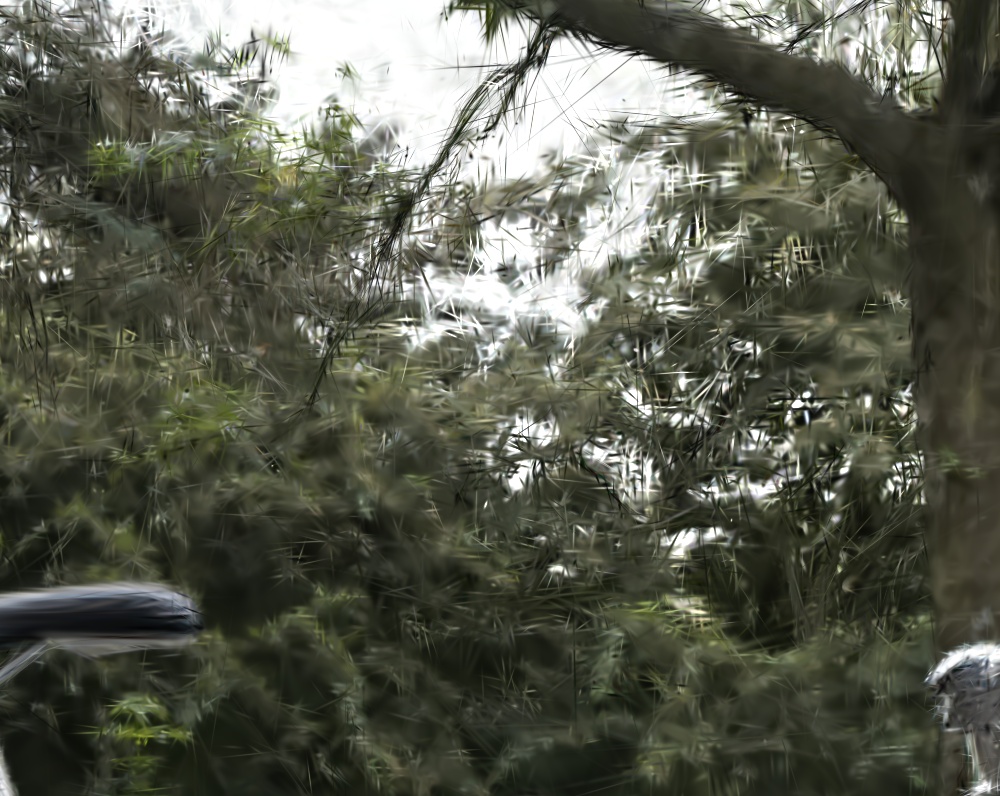}&
    \includegraphics[width=\ablationwidth]{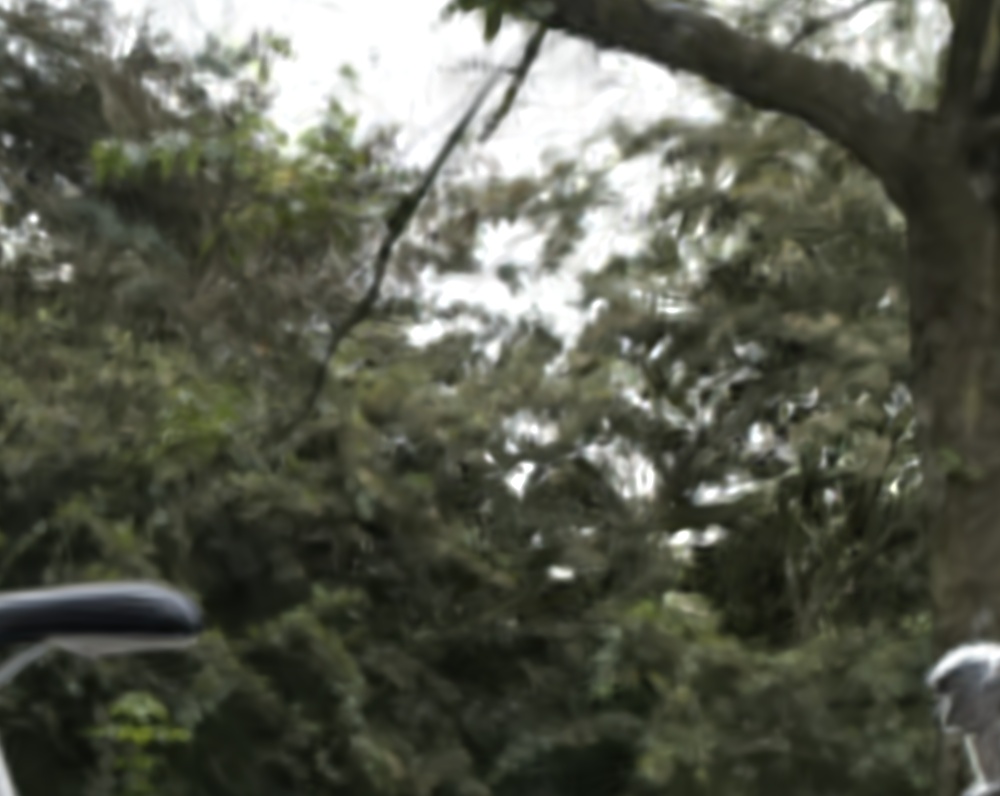}& 
    \includegraphics[width=\ablationwidth]{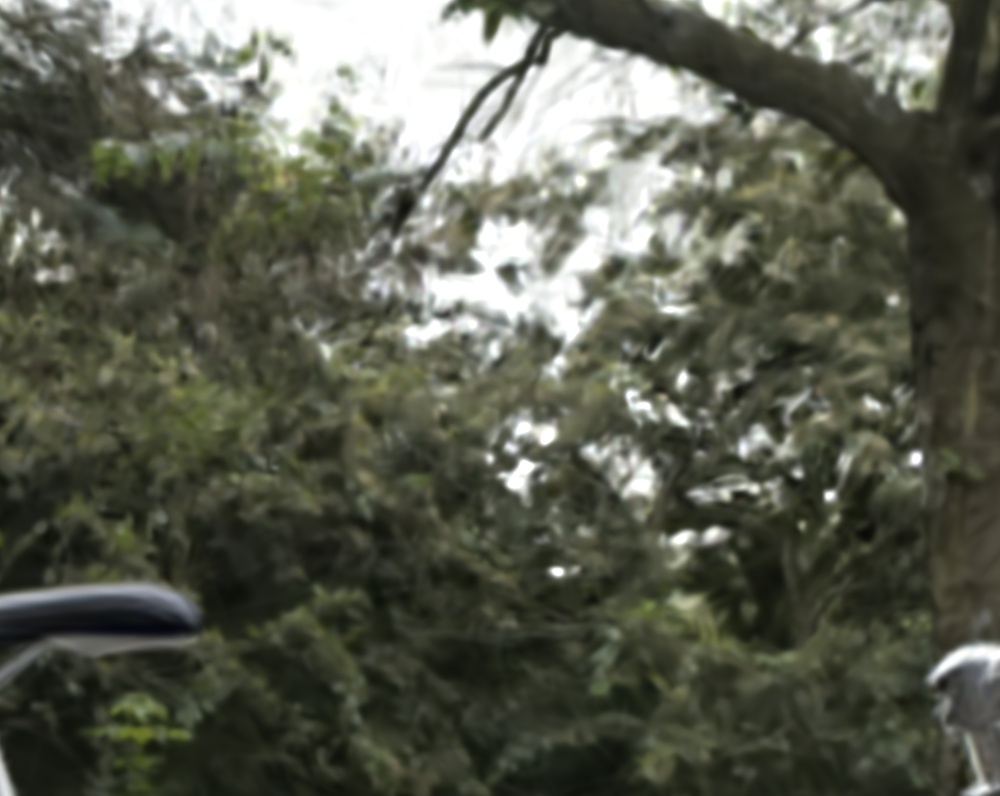}& 
    \includegraphics[width=\ablationwidth]{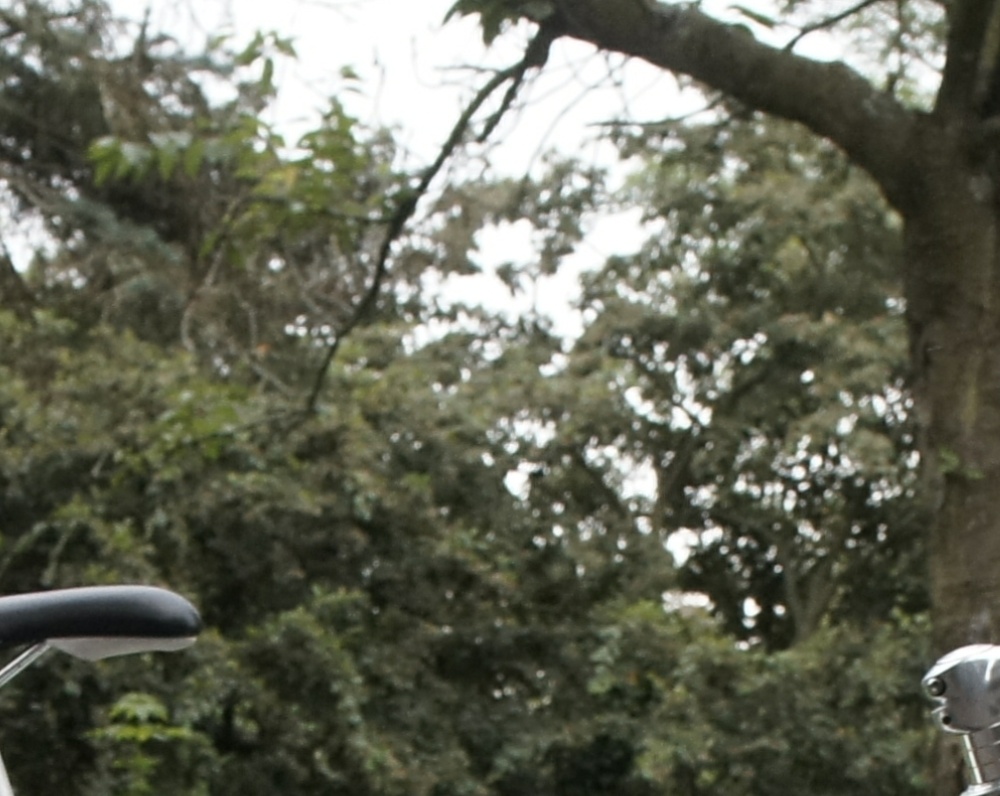}  
    \\ 
    \includegraphics[width=\ablationwidth]{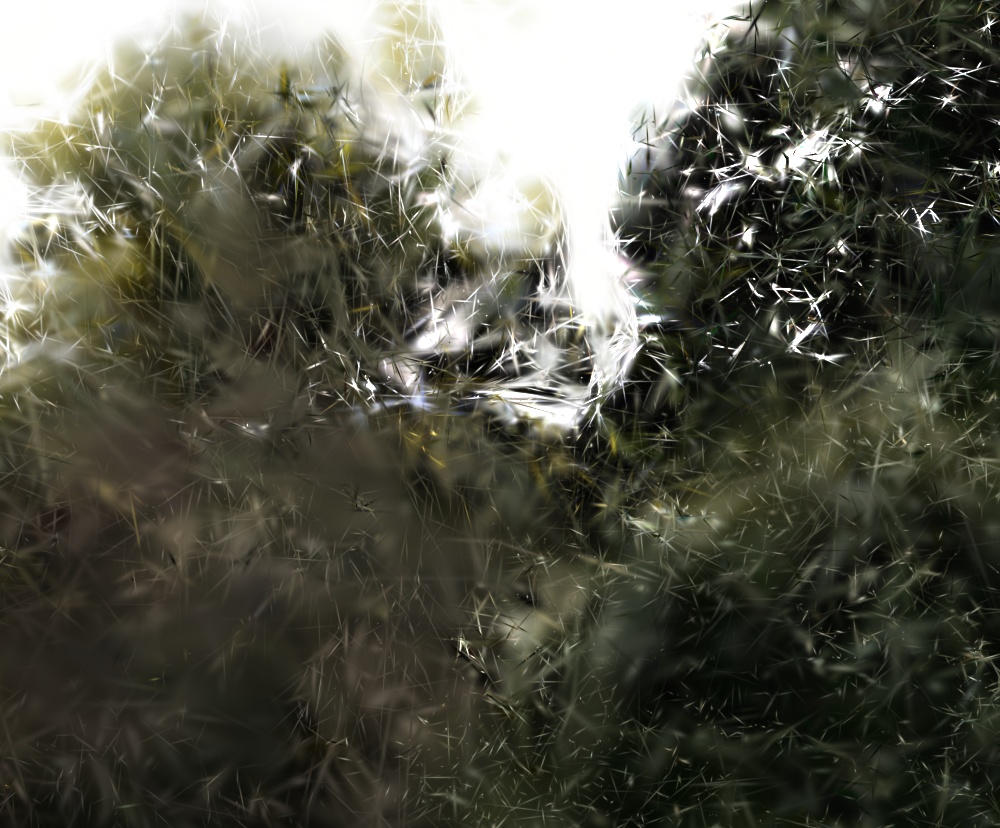}& \includegraphics[width=\ablationwidth]{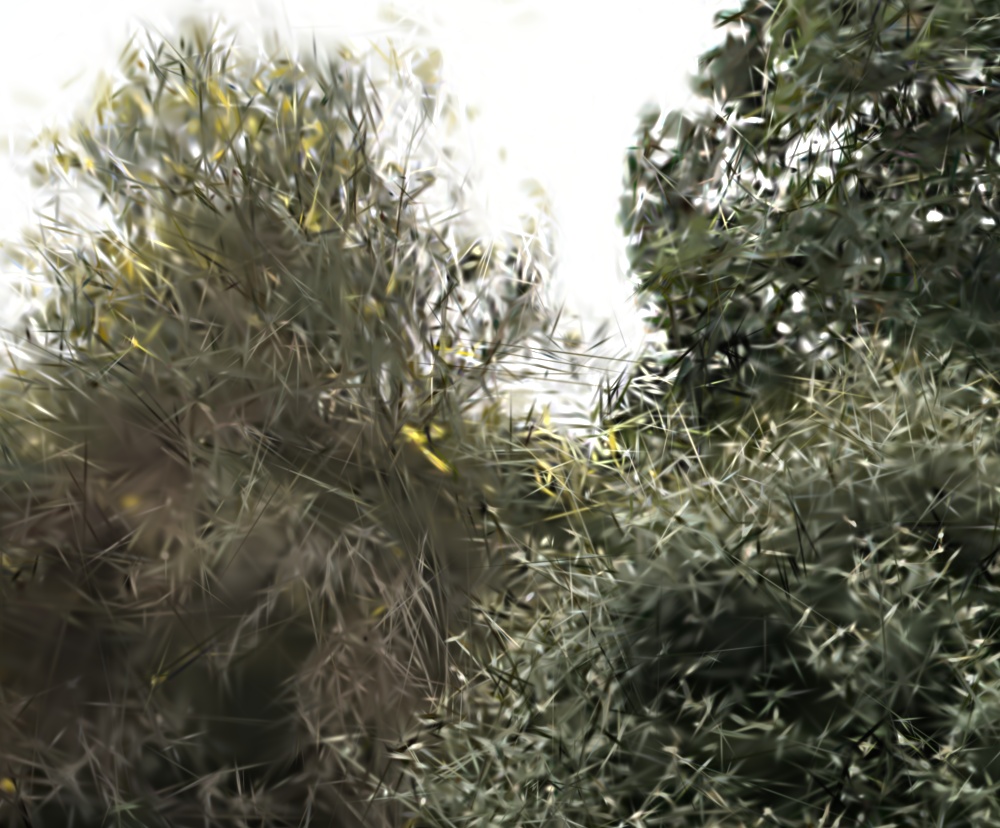}&
    \includegraphics[width=\ablationwidth]{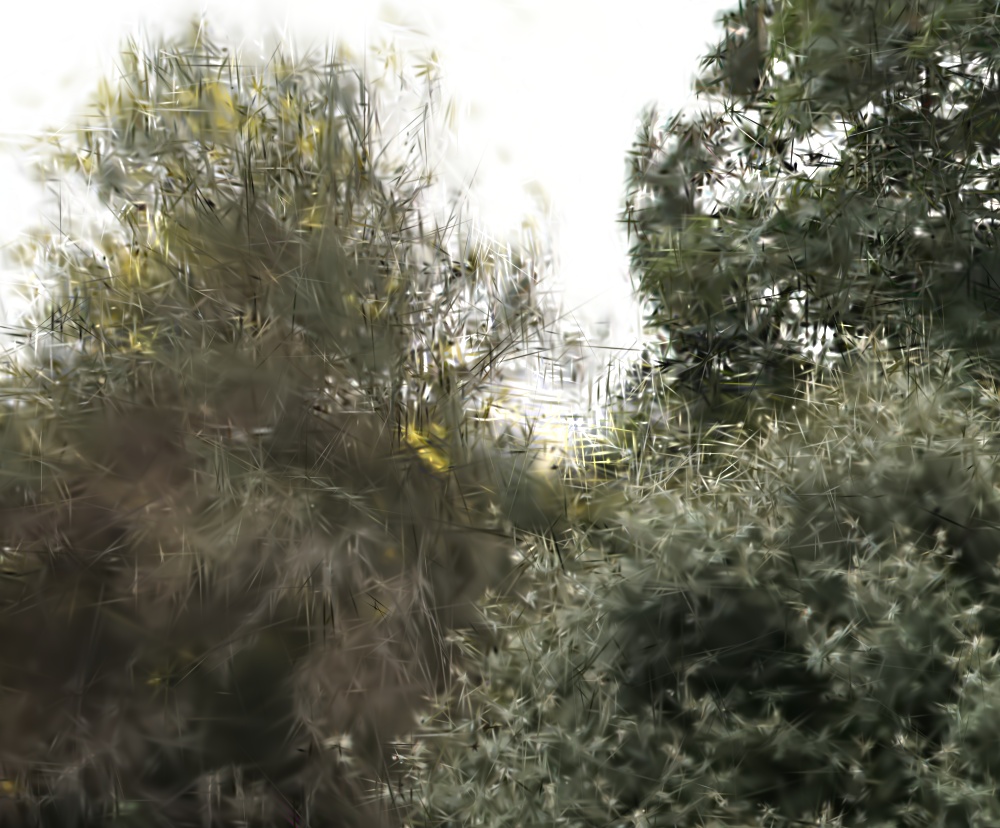}&
    \includegraphics[width=\ablationwidth]{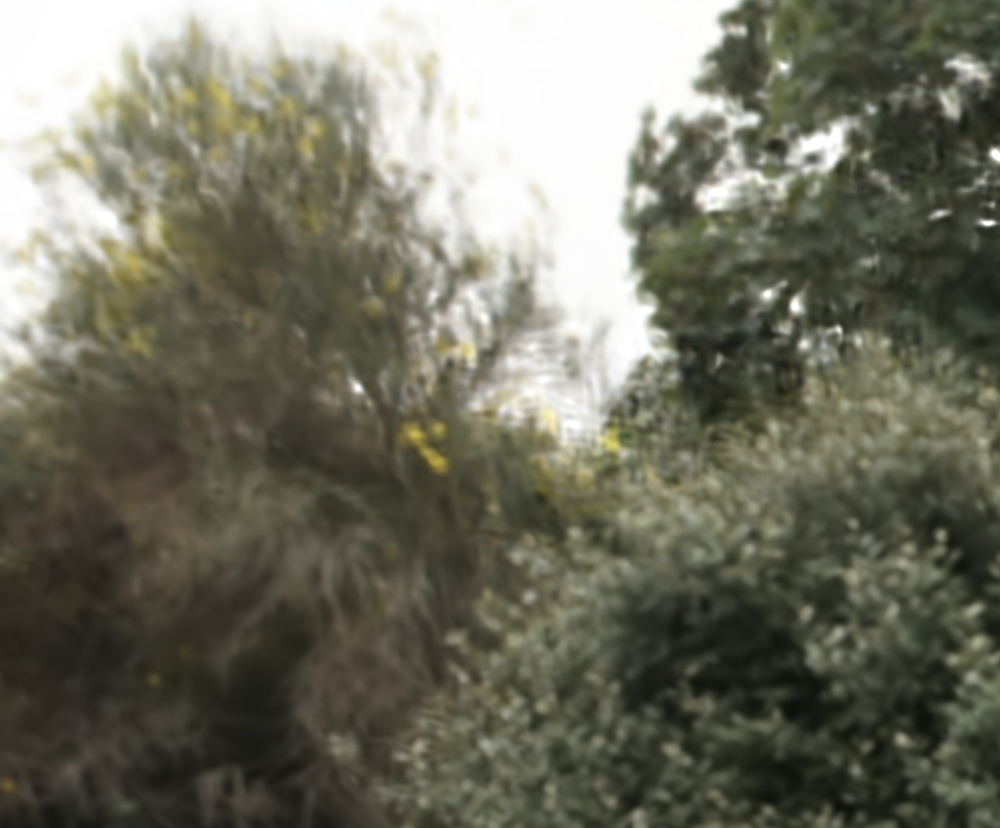}& 
    \includegraphics[width=\ablationwidth]{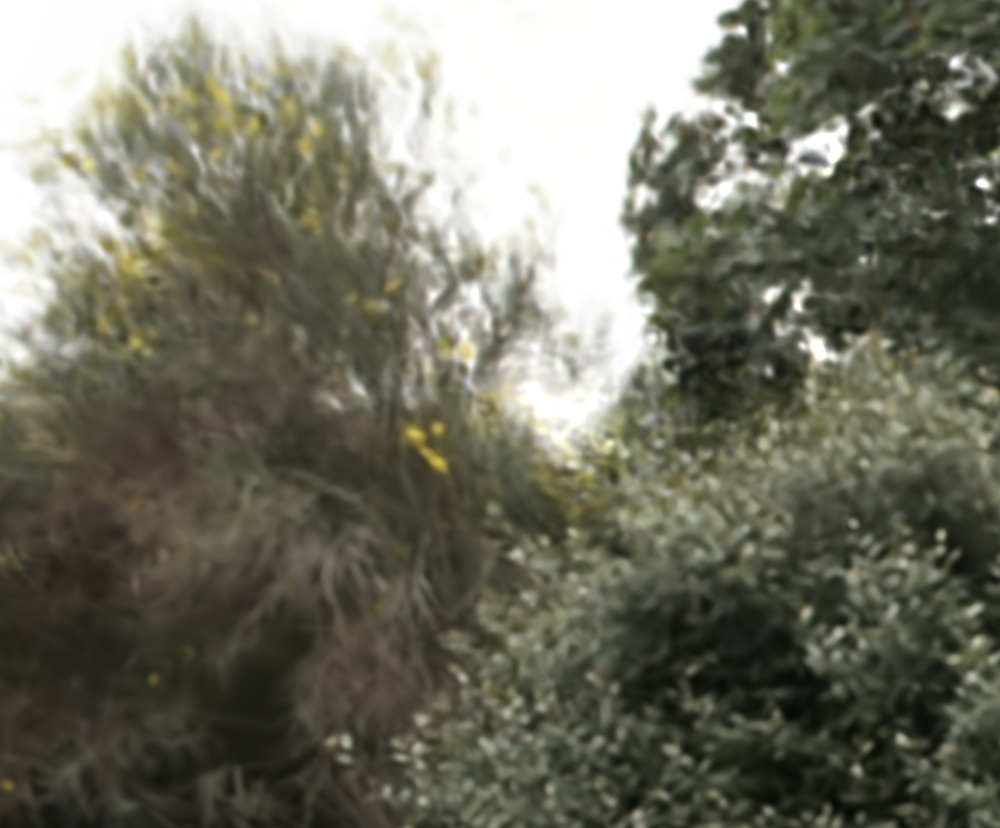}& 
    \includegraphics[width=\ablationwidth]{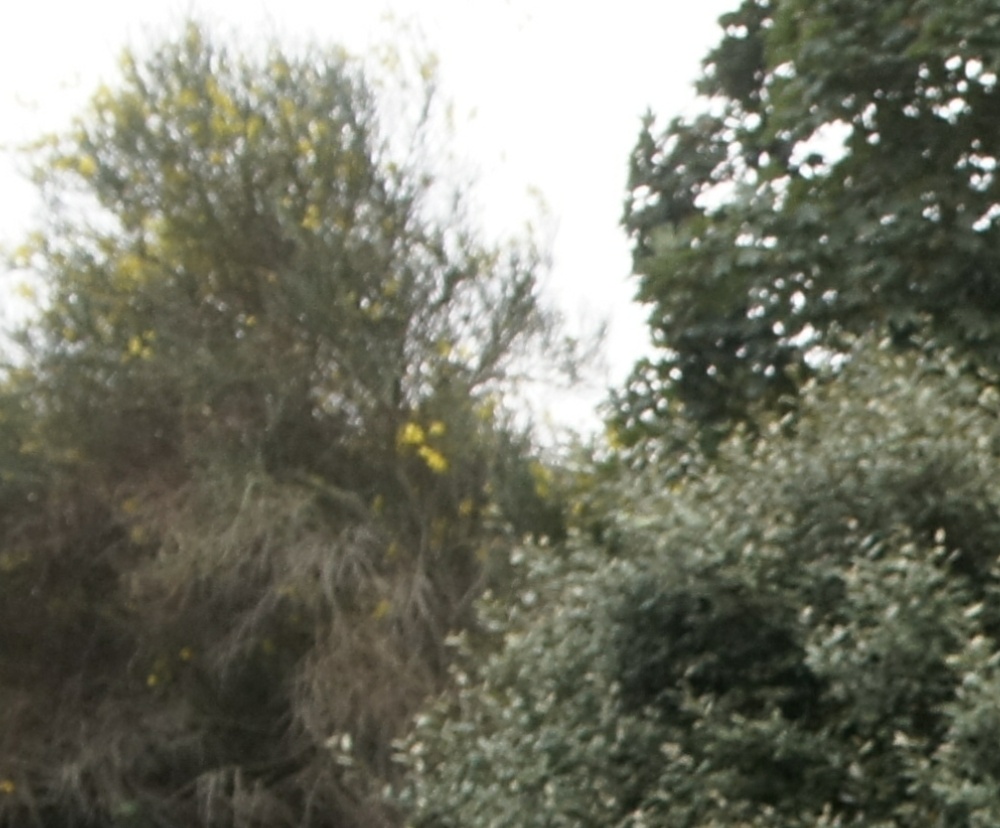}  
    \\ 
    \includegraphics[width=\ablationwidth]{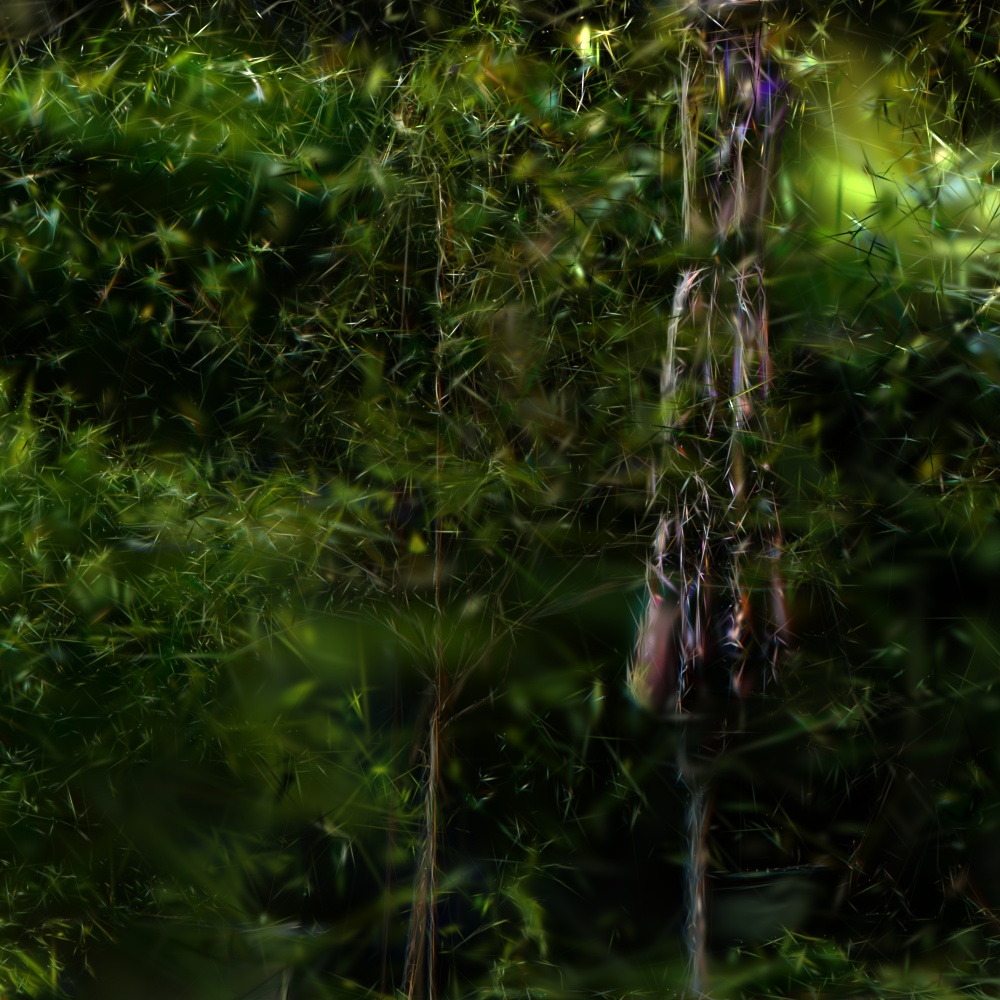}& \includegraphics[width=\ablationwidth]{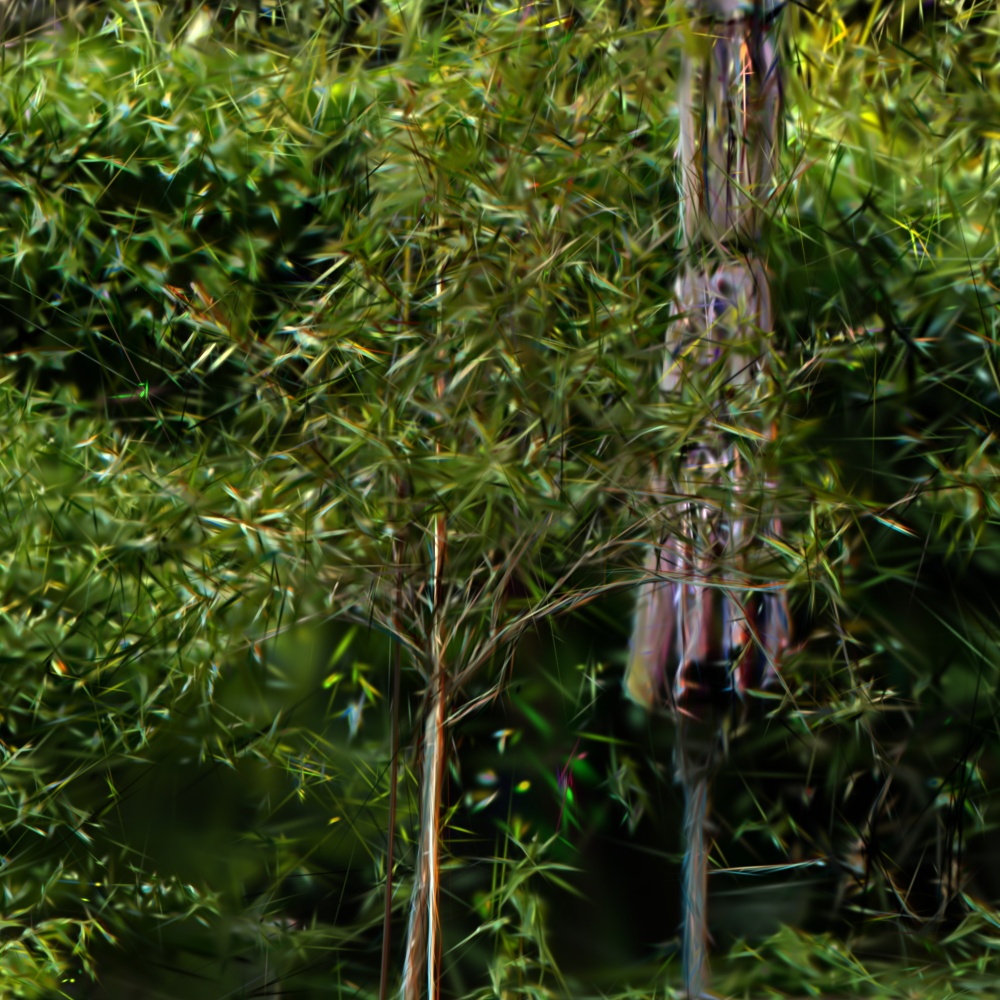}&
    \includegraphics[width=\ablationwidth]{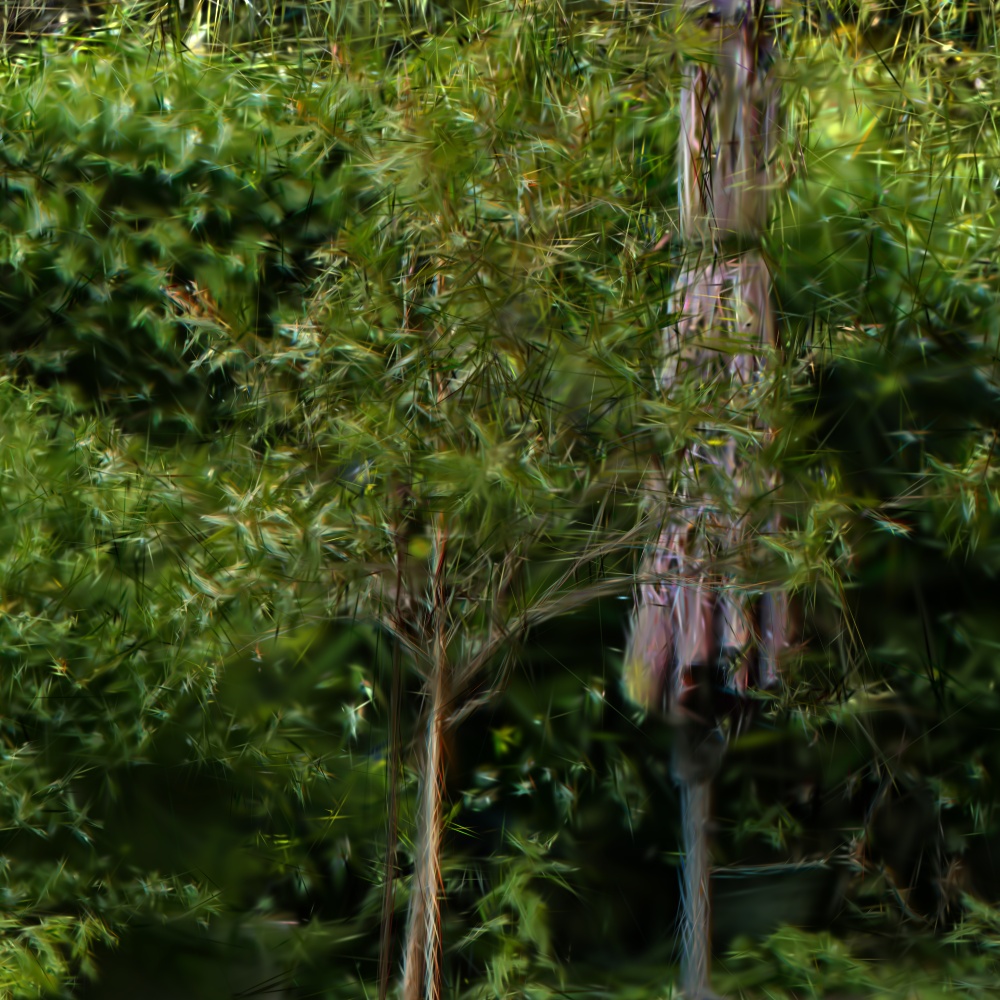}&
    \includegraphics[width=\ablationwidth]{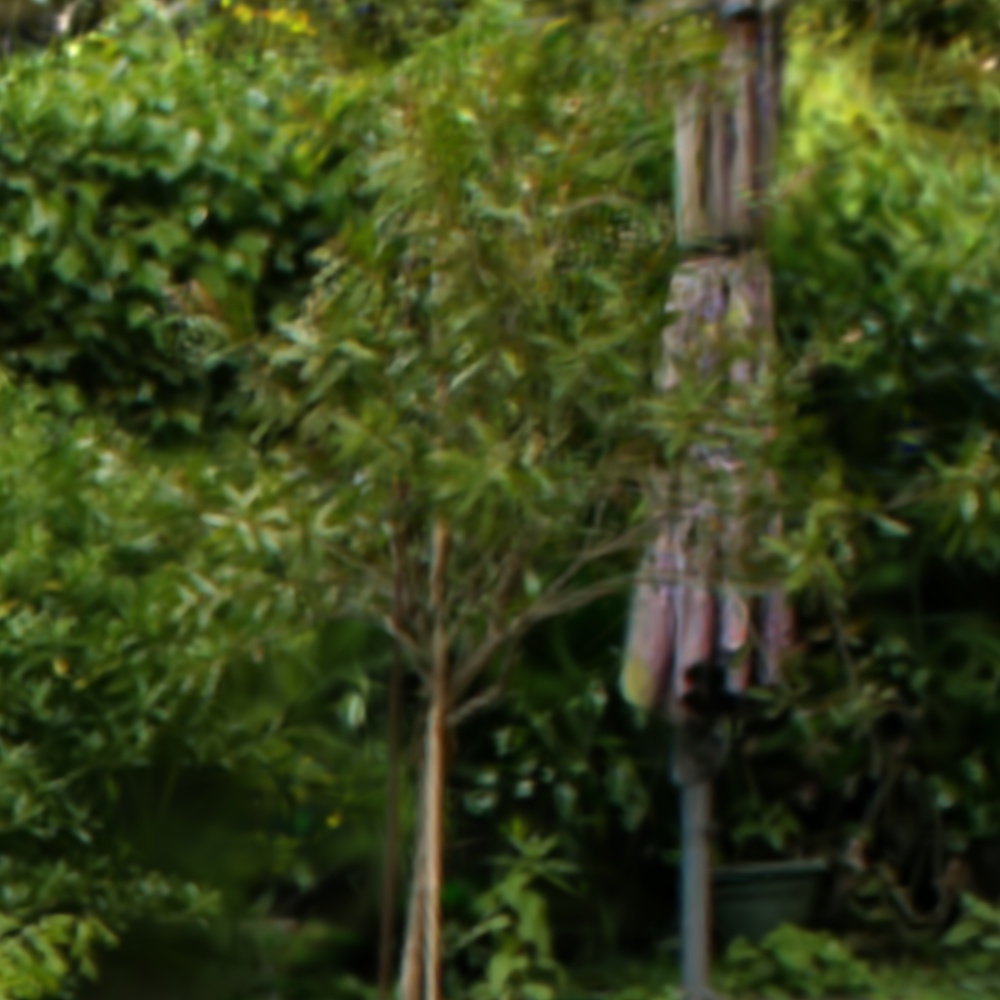}& 
    \includegraphics[width=\ablationwidth]{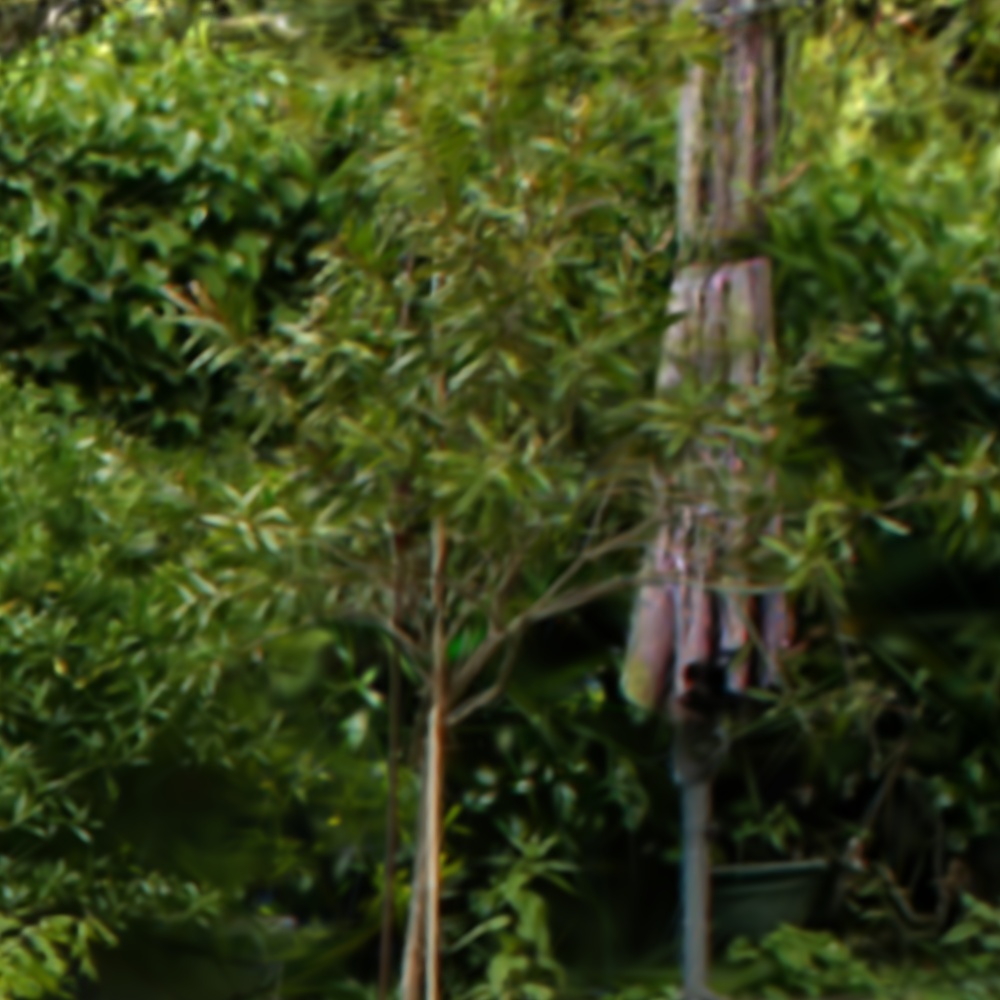}& 
    \includegraphics[width=\ablationwidth]{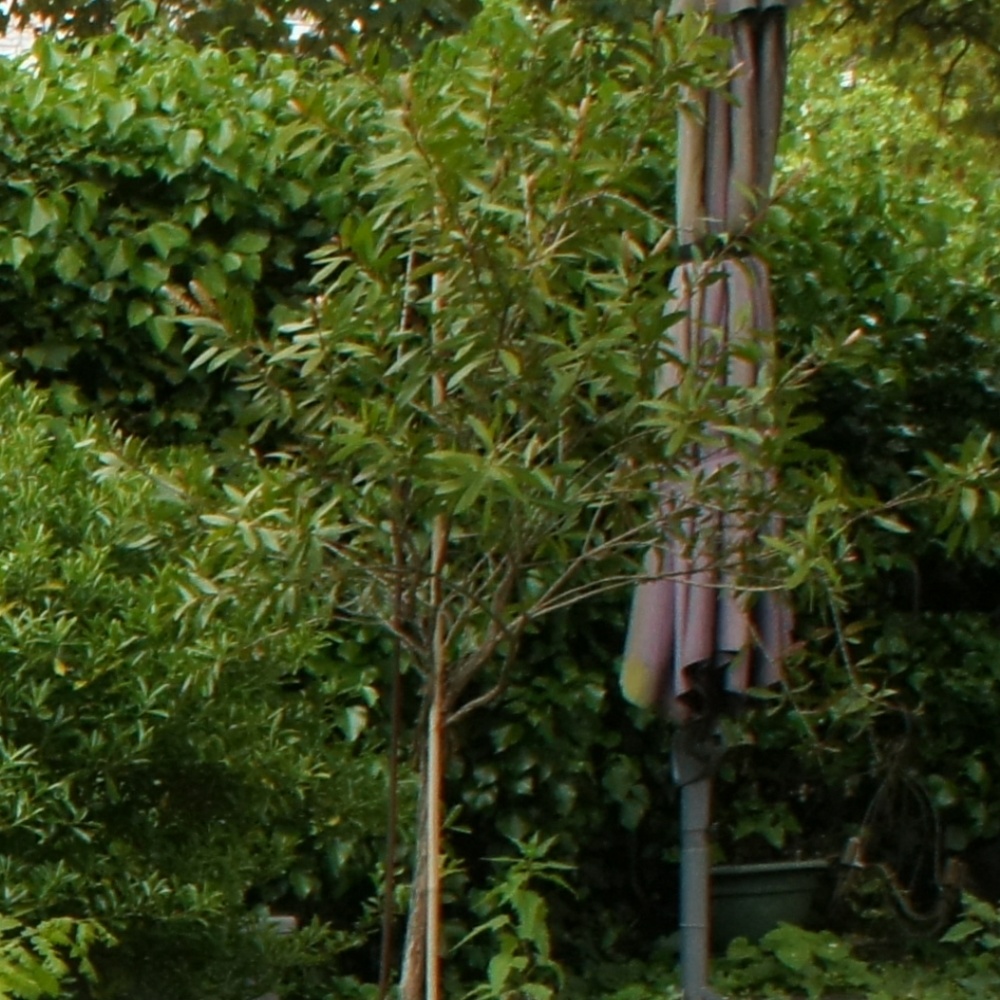}  
    \\ 
    \includegraphics[width=\ablationwidth]{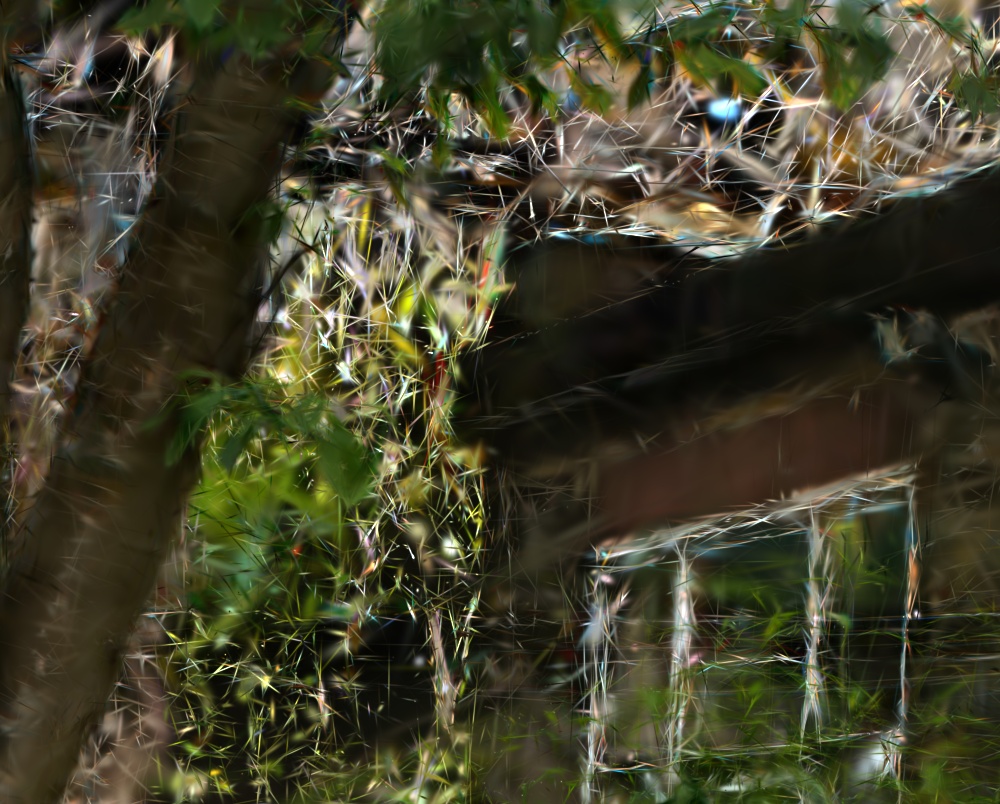}& \includegraphics[width=\ablationwidth]{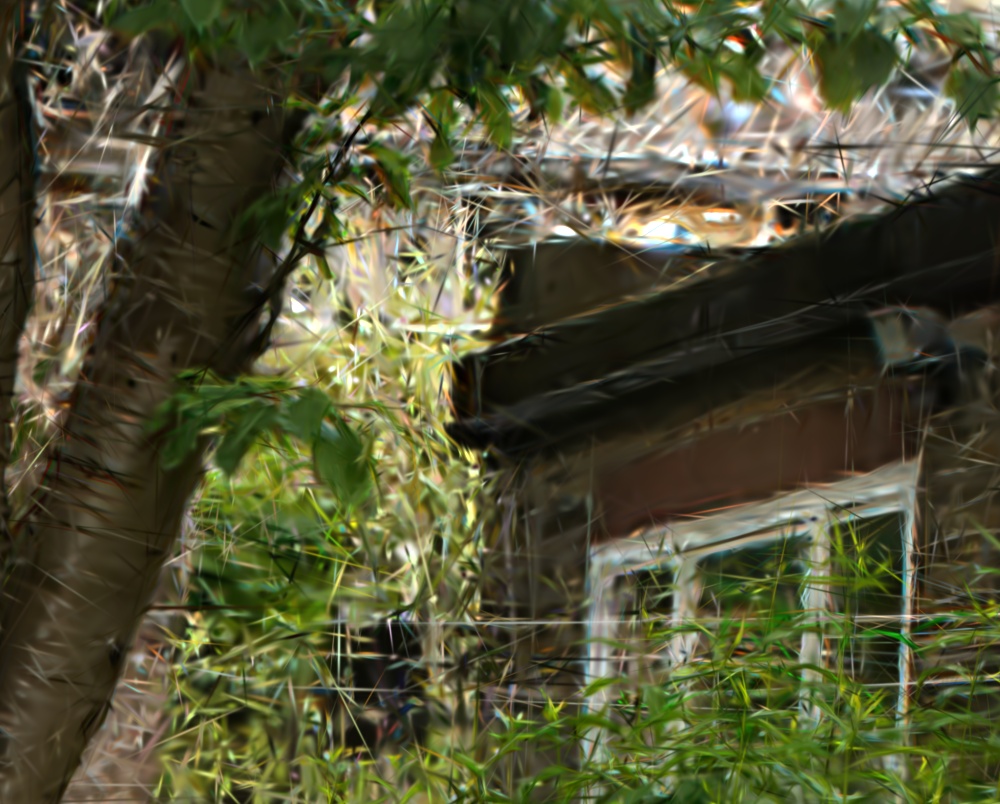}&
    \includegraphics[width=\ablationwidth]{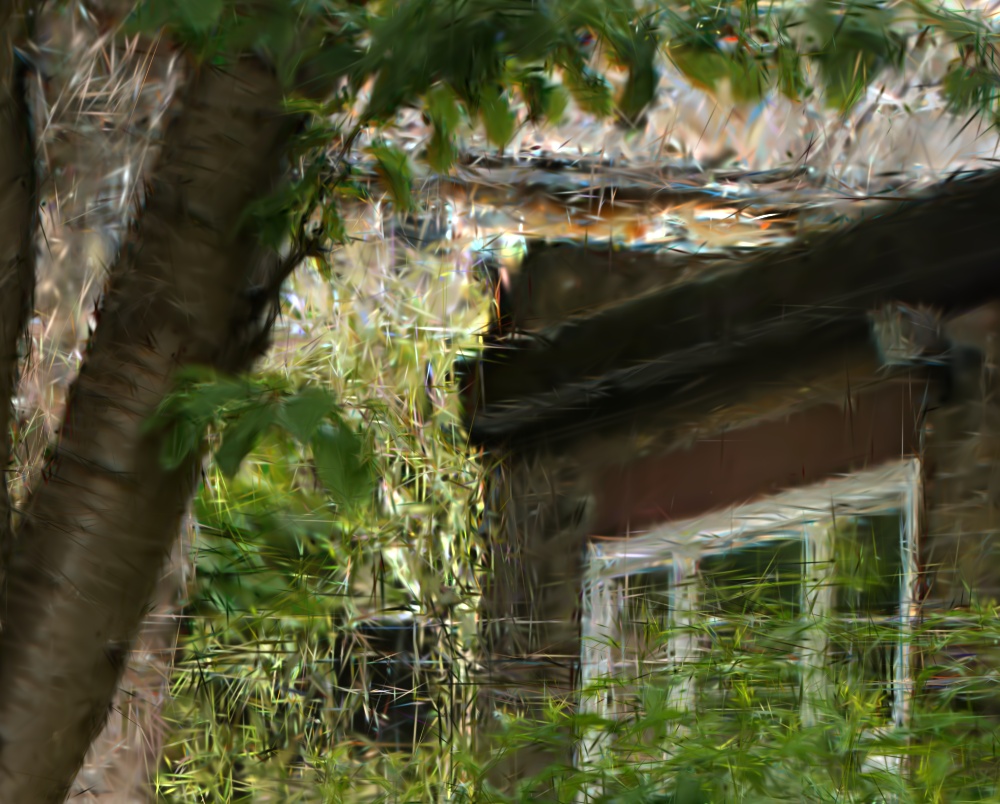}&
    \includegraphics[width=\ablationwidth]{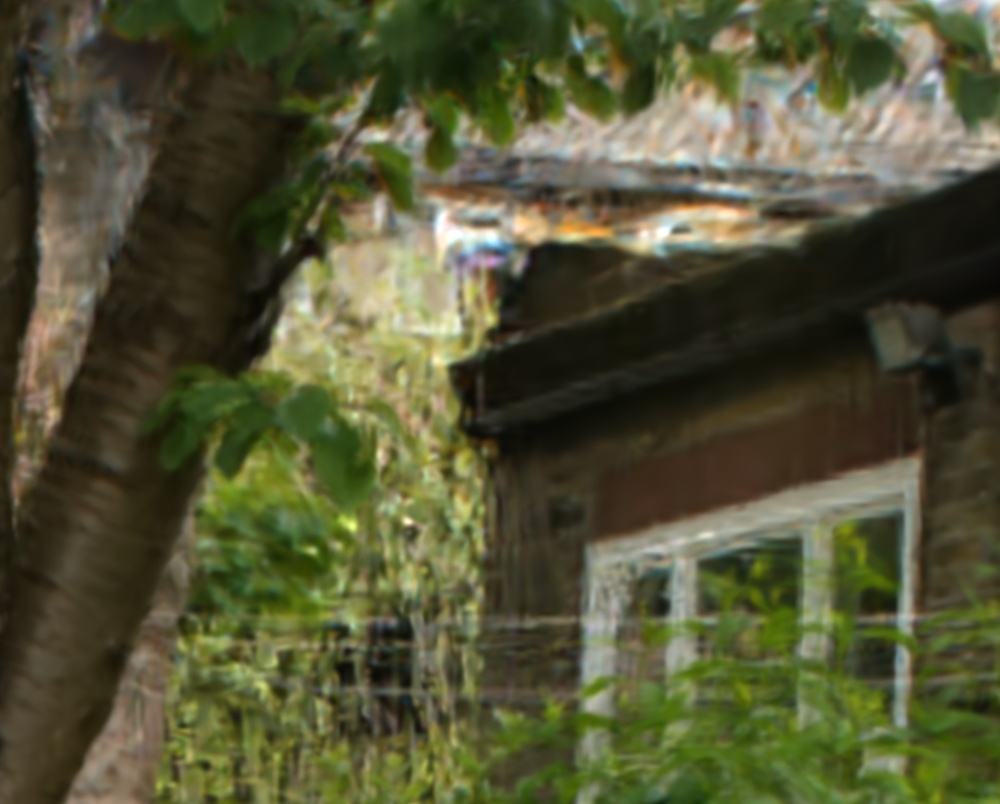}& 
    \includegraphics[width=\ablationwidth]{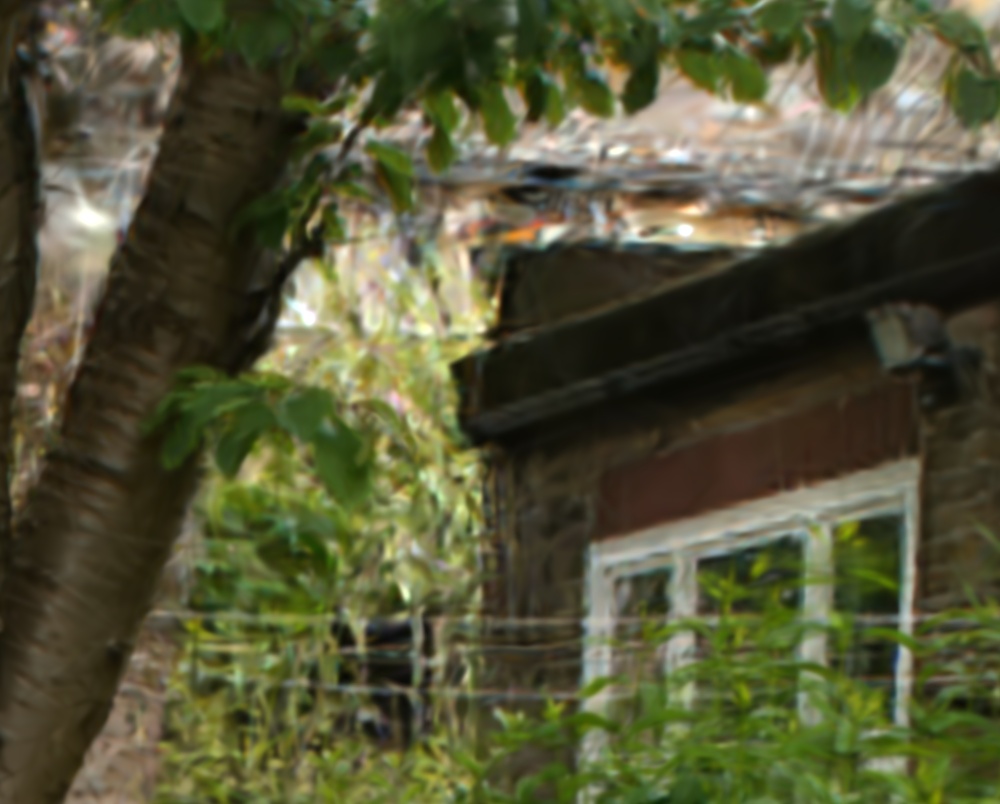}& 
    \includegraphics[width=\ablationwidth]{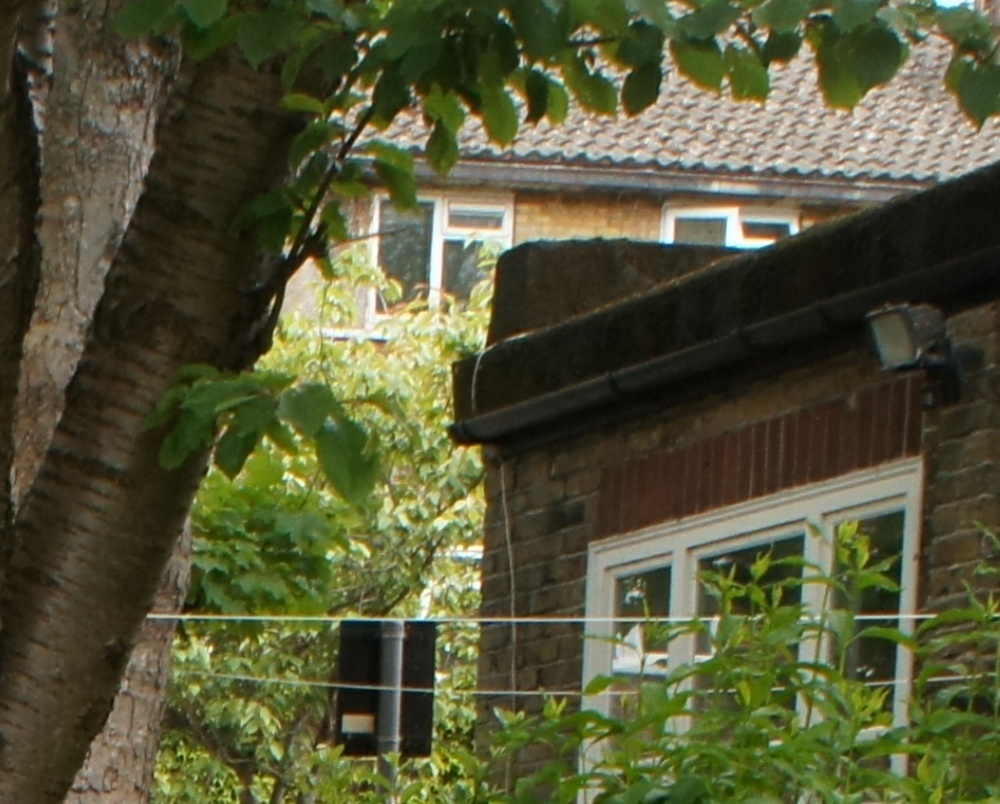}  
    \\ 
    \includegraphics[width=\ablationwidth]{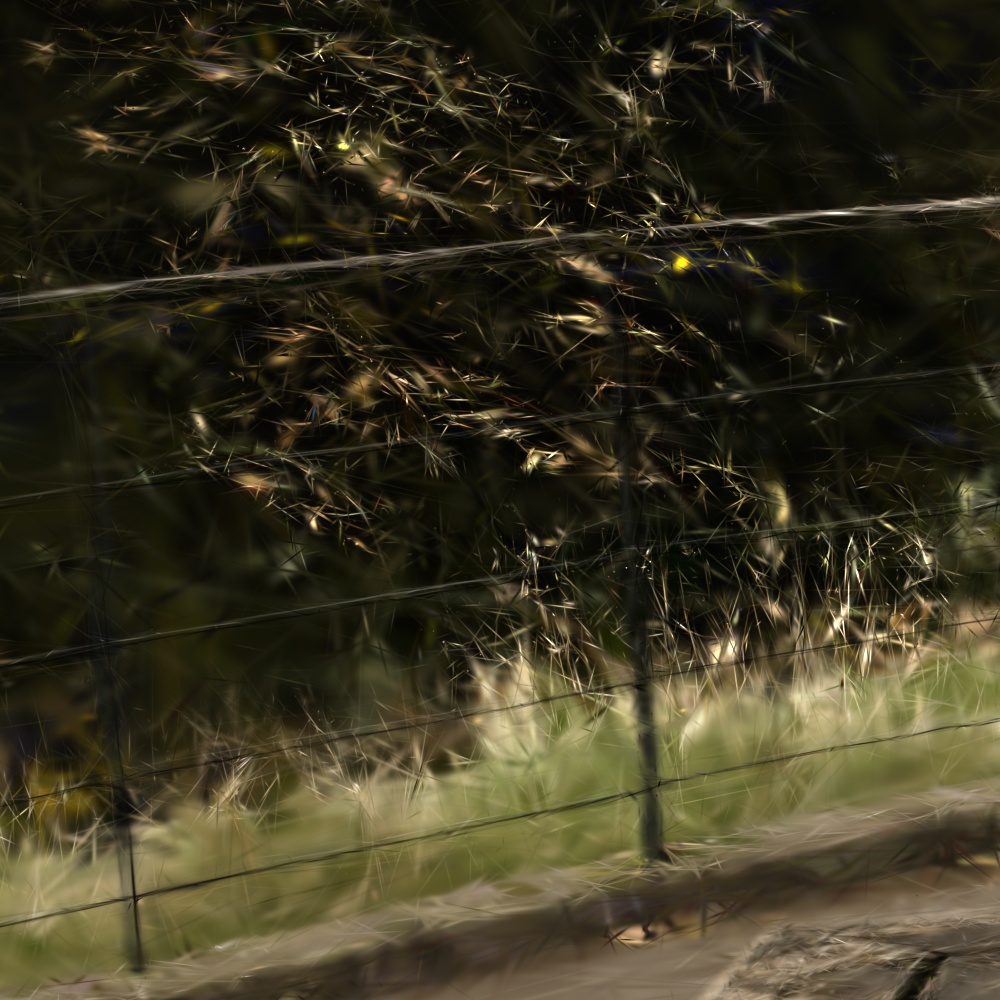}& \includegraphics[width=\ablationwidth]{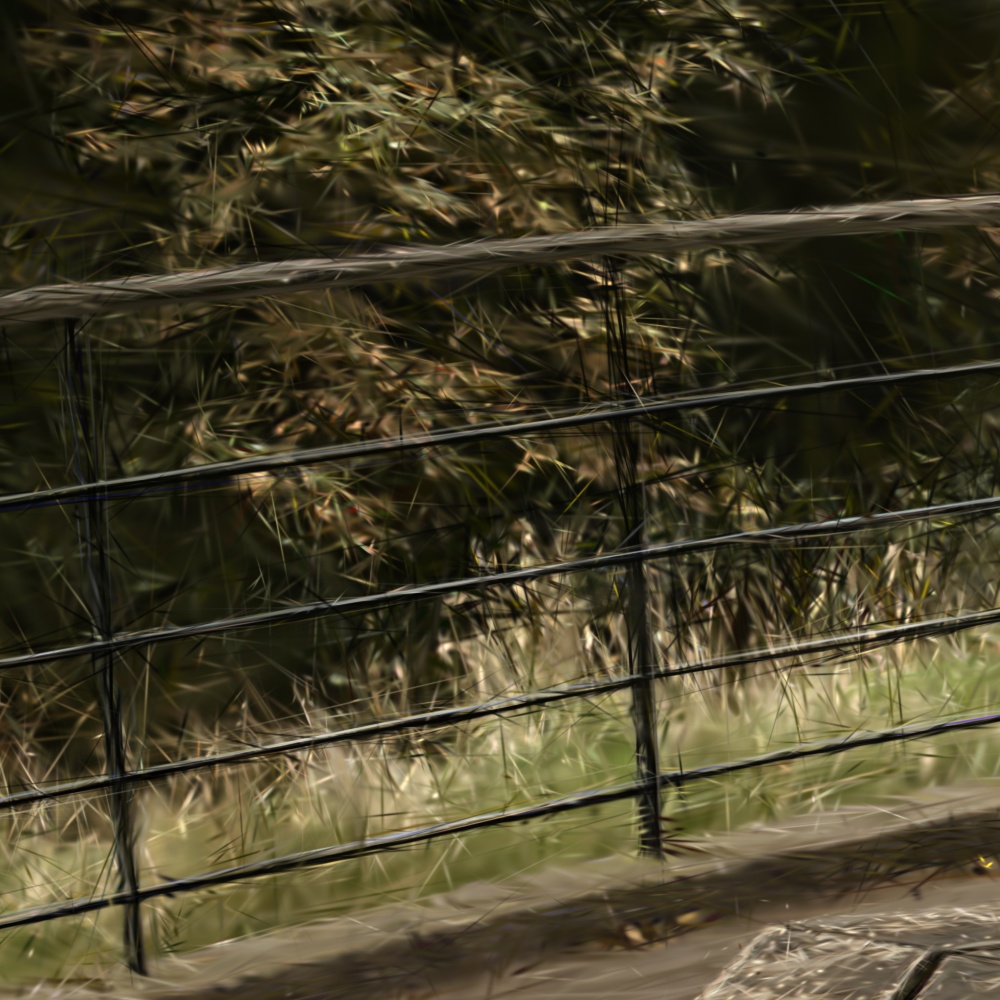}&
    \includegraphics[width=\ablationwidth]{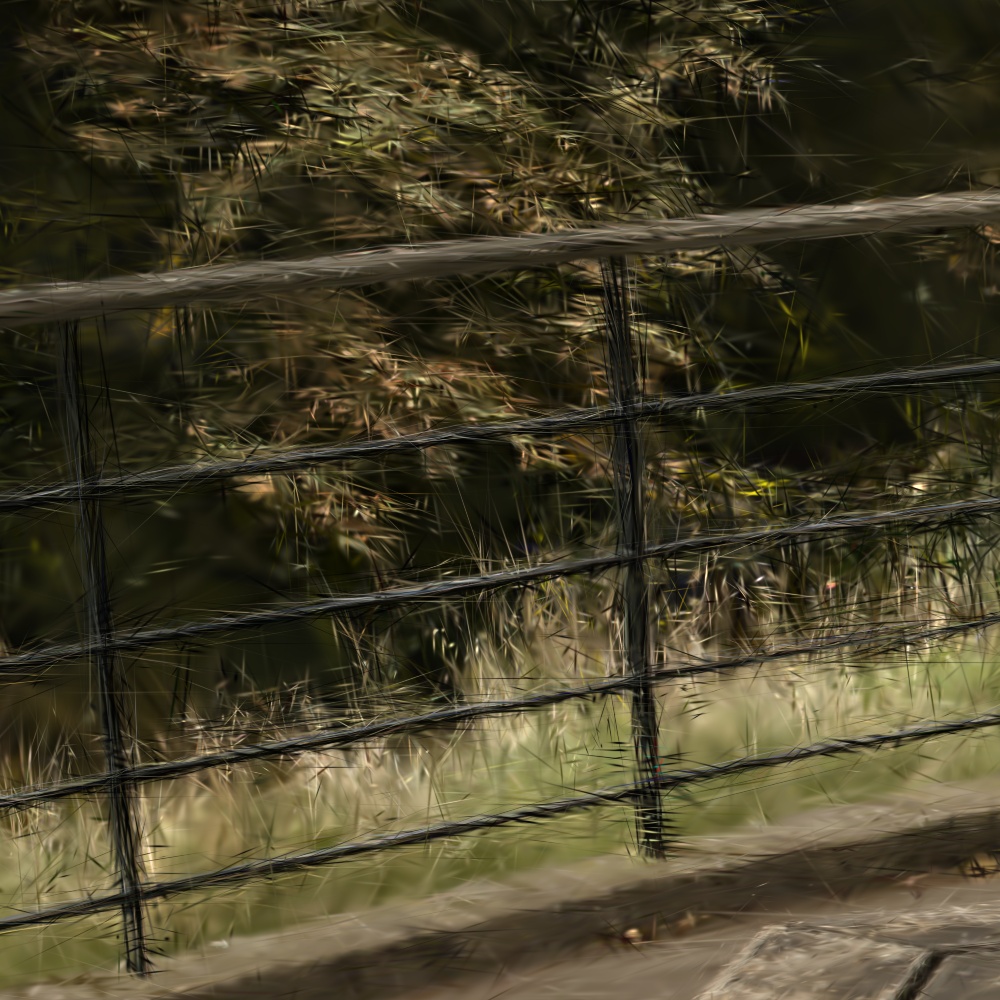}&
    \includegraphics[width=\ablationwidth]{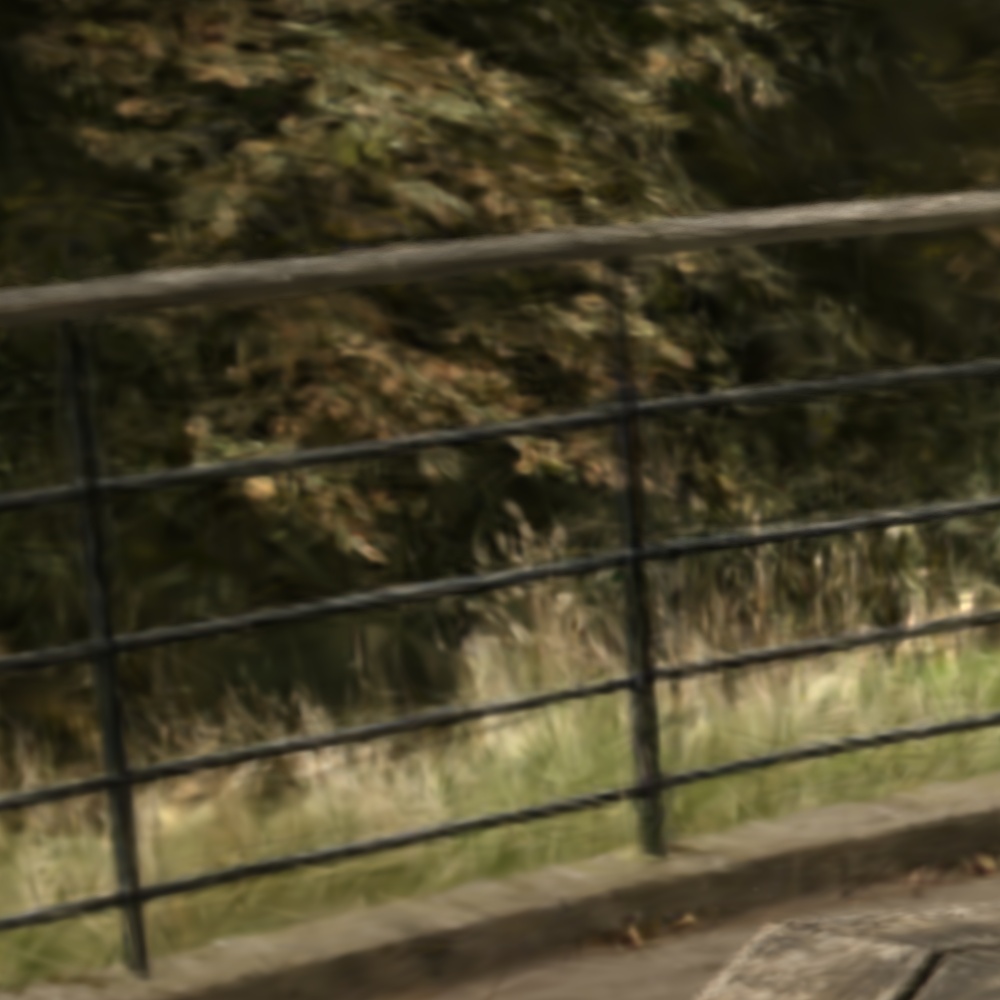}& 
    \includegraphics[width=\ablationwidth]{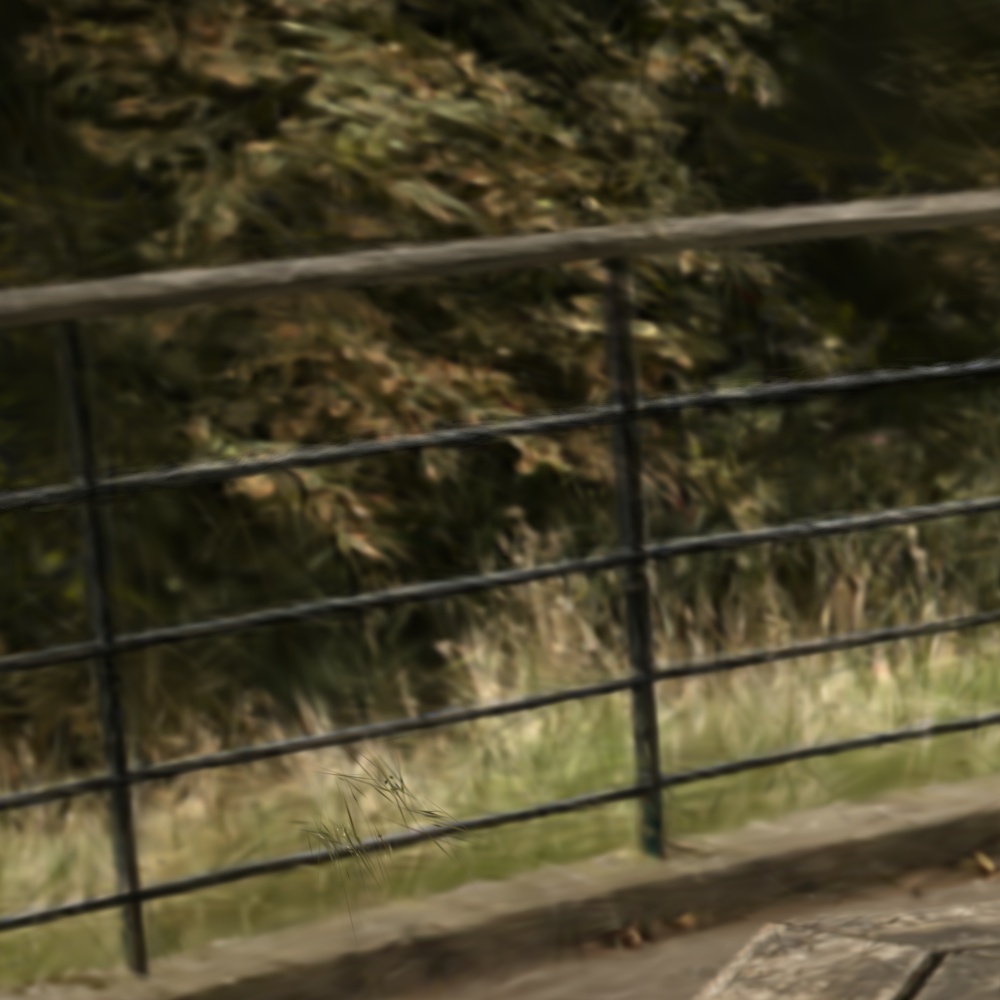}& 
    \includegraphics[width=\ablationwidth]{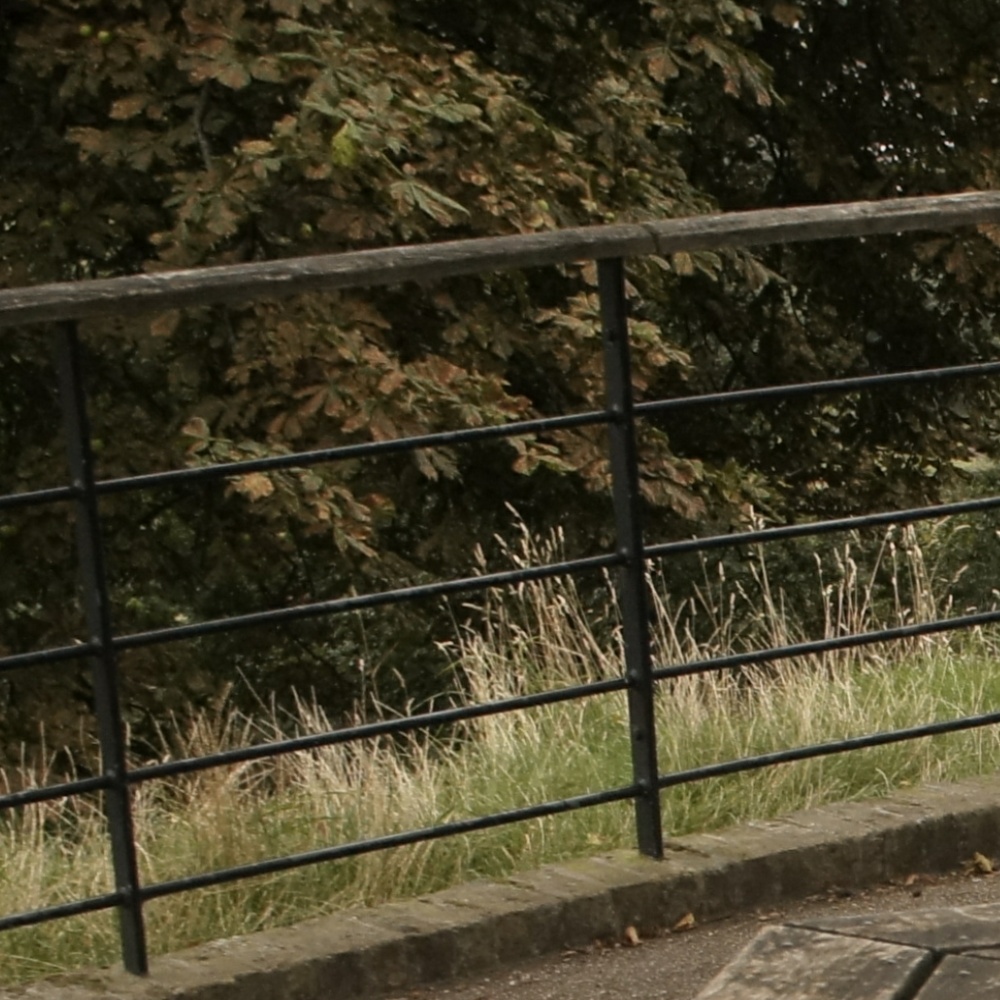}  
    \\ 
    3DGS~\cite{kerbl3Dgaussians} & 3DGS~\cite{kerbl3Dgaussians} + EWA~\cite{zwicker2001ewa} & Ours w/o 3D smoothing filter & Ours w/o 2D Mip filter & Mip-Splatting (ours) & GT  
    \end{tabular}
    \vspace{-0.1in}
    \caption{\textbf{Single-scale Training and Multi-scale Testing on the Mip-NeRF 360 Dataset~\cite{barron2022mipnerf360}.} 
    All models are trained on images downsampled by a factor of 8 and rendered at full resolution to demonstrate zoom-in/moving closer effects. Removing the 3D smoothing filter results in high-frequency artifacts. Mip-Splatting renders images that closely approximate ground truth. Zoom in for a better view. %
    }
    \label{fig:360_single_train_multi_test_ablation}
    \vspace{-0.1in}
\end{figure*}
\begin{table*}[]
    \renewcommand{\tabcolsep}{1pt}
    \centering
    \resizebox{1.0\linewidth}{!}{
    \begin{tabular}{@{}l@{\,\,}|ccccc|ccccc|ccccc}
    & \multicolumn{5}{c|}{PSNR $\uparrow$} & \multicolumn{5}{c|}{SSIM $\uparrow$} & \multicolumn{5}{c}{LPIPS $\downarrow$}  \\
    & $1 \times$ Res. & $2 \times$ Res. & $4 \times$ Res. & $8 \times$ Res. & Avg. & $1 \times$ Res. & $2 \times$ Res. & $4 \times$ Res. & $8 \times$ Res. & Avg. & $1 \times$ Res. & $2 \times$ Res. & $4 \times$ Res. & $8 \times$ Res. & Avg.  \\ \hline
    
3DGS~\cite{kerbl3Dgaussians}&                            29.19  &                            23.50  &                            20.71  &                            19.59  &                            23.25  & 0.880  & 0.740  &                            0.619  &                            0.619  &                            0.715  &\cellcolor{tablered} 0.107  & 0.243  &                            0.394  &                            0.476  &0.305 
\\
3DGS~\cite{kerbl3Dgaussians} + EWA~\cite{zwicker2001ewa}&\cellcolor{yellow} 29.30  & 25.90  &                            23.70  &                            22.81  & 25.43  & 0.880  & 0.775  & 0.667  &                            0.643  & 0.741  &                            0.114  & 0.236  & 0.369  & 0.449  & 0.292
\\
\hline
Mip-Splatting (ours)&\cellcolor{orange} 29.39  &\cellcolor{tablered} 27.39  &\cellcolor{tablered} 26.47  &\cellcolor{tablered} 26.22  &\cellcolor{tablered} 27.37  &\cellcolor{tablered} 0.884  &\cellcolor{tablered} 0.808  &\cellcolor{tablered} 0.754  &\cellcolor{tablered} 0.765  &\cellcolor{tablered} 0.803  & 0.108  &\cellcolor{tablered} 0.205  &\cellcolor{tablered} 0.305  &\cellcolor{tablered} 0.392  &\cellcolor{tablered} 0.252
\\
Mip-Splatting (ours) w/o 3D smoothing filter&\cellcolor{tablered} 29.41 &\cellcolor{yellow} 27.09 &\cellcolor{yellow} 25.83 &\cellcolor{yellow} 25.38 &\cellcolor{yellow} 26.93 &\cellcolor{yellow} 0.881 &\cellcolor{yellow} 0.795 &\cellcolor{yellow} 0.722 &\cellcolor{yellow} 0.713 &\cellcolor{yellow} 0.778 &\cellcolor{tablered} 0.107 &\cellcolor{orange} 0.214 &\cellcolor{yellow} 0.342 &\cellcolor{yellow} 0.424 &\cellcolor{yellow} 0.272
\\
Mip-Splatting (ours) w/o 2D Mip filter& 29.29 &\cellcolor{orange} 27.22 &\cellcolor{orange} 26.31 &\cellcolor{orange} 26.08 &\cellcolor{orange} 27.23 &\cellcolor{orange} 0.882 &\cellcolor{orange} 0.798 &\cellcolor{orange} 0.742 &\cellcolor{orange} 0.759 &\cellcolor{orange} 0.795 &\cellcolor{tablered} 0.107 &\cellcolor{orange} 0.214 &\cellcolor{orange} 0.319 &\cellcolor{orange} 0.407 &\cellcolor{orange} 0.262
    \end{tabular}
    }
    \vspace{-0.1in}
    \caption{
    \textbf{Single-scale Training and Multi-scale Testing on the Mip-NeRF 360 Dataset~\cite{barron2022mipnerf360}.} All methods are trained on the smallest scale ($1 \times$) and evaluated across four scales ($1 \times$, $2 \times$, $4 \times$, and $8 \times$), with evaluations at higher sampling rates simulating zoom-in effects. While our method yields comparable results at the training resolution, it significantly surpasses all previous work at all other scales. Omitting the 3D smoothing filter results in high-frequency artifacts when rendering higher resolution image as shown in~\ref{fig:360_single_train_multi_test_ablation}, while the excluding the 2D Mip filter only causes a slight decline in performance as this filter's role is mainly for mitigating zoom-out artifacts. }    \label{tab:avg_360_results_single_train_multi_test_ablation}
\end{table*}

To evaluate the effectiveness of the 3D smoothing filter, we conduct an ablation with the single-scale training and multi-scale testing setting to simulate zoom-in effects in the Mip-NeRF 360 dataset~\cite{barron2022mipnerf360}. The quantitative result is presented in ~\tabref{tab:avg_360_results_single_train_multi_test_ablation}. Omitting the 3D smoothing filter results in high-frequency artifacts when rendering higher resolution image, as depicted in~\figref{fig:360_single_train_multi_test_ablation}. Excluding the 2D Mip filter causes a slight decline in performance as this filter's role is mainly for mitigating zoom-out artifacts, as we will shown next. The absence of both the 3D smoothing filter and the 2D Mip filter leads to an excessive generation of small Gaussian primitives, due to the density control mechanism, resulting in out of memory error even on an A100 GPU with 40GB memory. Hence, we don't report the result.

\subsection{Effectiveness of the 2D Mip Filter}
\label{sec:ab_2d}
\begin{table*}[]
    \renewcommand{\tabcolsep}{1pt}
    \centering
    \resizebox{1.0\linewidth}{!}{
    \begin{tabular}{@{}l@{\,\,}|ccccc|ccccc|ccccc}
    & \multicolumn{5}{c|}{PSNR $\uparrow$} & \multicolumn{5}{c|}{SSIM $\uparrow$} & \multicolumn{5}{c}{LPIPS $\downarrow$}  \\
    & Full Res. & $\nicefrac{1}{2}$ Res. & $\nicefrac{1}{4}$ Res. & $\nicefrac{1}{8}$ Res. & Avg. & Full Res. & $\nicefrac{1}{2}$ Res. & $\nicefrac{1}{4}$ Res. & $\nicefrac{1}{8}$ Res. & Avg. & Full Res. & $\nicefrac{1}{2}$ Res. & $\nicefrac{1}{4}$ Res. & $\nicefrac{1}{8}$ Res & Avg.  \\ \hline
    
3DGS~\cite{kerbl3Dgaussians}&    33.33 & 26.95 & 21.38 & 17.69 & 24.84 &\cellcolor{yellow}  0.969 & 0.949 & 0.875 & 0.766 & 0.890 &\cellcolor{tablered}  0.030 & 0.032 & 0.066 & 0.121 & 0.063
\\
3DGS~\cite{kerbl3Dgaussians} + EWA~\cite{zwicker2001ewa}&\cellcolor{orange} 33.51 & 31.66 & 27.82 & 24.63 & 29.40 &\cellcolor{yellow}  0.969 &  0.971 & 0.959 & 0.940 & 0.960 &  0.032 &  0.024 &\cellcolor{yellow}  0.033 &\cellcolor{yellow}  0.047 &\cellcolor{yellow}  0.034
\\
3DGS~\cite{kerbl3Dgaussians} - Dilation& 33.38 & 33.06 & 29.68 & 26.19 & 30.58 &\cellcolor{yellow} 0.969 & 0.973 & 0.964 & 0.945 & 0.963 &\cellcolor{tablered} 0.030 & 0.024 & 0.041 & 0.075 & 0.042
\\
\hline
Mip-Splatting (ours)&    33.36 &\cellcolor{orange}  34.00 &\cellcolor{tablered} 31.85 &\cellcolor{tablered} 28.67 &\cellcolor{tablered} 31.97 &\cellcolor{yellow} 0.969 &\cellcolor{tablered} 0.977 &\cellcolor{tablered} 0.978 &\cellcolor{tablered} 0.973 &\cellcolor{tablered} 0.974 &  0.031 &\cellcolor{orange} 0.019 &\cellcolor{tablered} 0.019 &\cellcolor{tablered} 0.026 &\cellcolor{tablered} 0.024
\\
Mip-Splatting (ours) w/o 3D smoothing filter&\cellcolor{tablered} 33.67 &\cellcolor{tablered} 34.16 &\cellcolor{orange} 31.56 &\cellcolor{orange} 28.20 &\cellcolor{orange} 31.90 &\cellcolor{tablered} 0.970 &\cellcolor{tablered} 0.977 &\cellcolor{tablered} 0.978 &\cellcolor{orange} 0.971 &\cellcolor{tablered} 0.974 &\cellcolor{tablered} 0.030 &\cellcolor{tablered} 0.018 &\cellcolor{tablered} 0.019 &\cellcolor{orange} 0.027 &\cellcolor{tablered} 0.024
\\
Mip-Splatting (ours) w/o 2D Mip filter&\cellcolor{orange} 33.51 &\cellcolor{yellow} 33.38 &\cellcolor{yellow} 29.87 &\cellcolor{yellow} 26.28 &\cellcolor{yellow} 30.76 &\cellcolor{tablered} 0.970 &\cellcolor{yellow} 0.975 &\cellcolor{yellow} 0.966 &\cellcolor{yellow} 0.946 &\cellcolor{yellow} 0.964 & 0.031 &\cellcolor{yellow} 0.022 & 0.039 & 0.073 & 0.041
    \end{tabular}
    }
    \vspace{-0.1in}
    \caption{
    \textbf{Single-scale Training and Multi-scale Testing on the Blender Dataset~\cite{mildenhall2020nerf}.}
    All methods are trained on full-resolution images and evaluated at four different (smaller) resolutions, with lower resolutions simulating zoom-out effects.
    While Mip-Splatting yields comparable results at training resolution, it significantly surpasses previous work at all other scales. Removing the 2D Mip filter results in a notable decline in performance at lower resolutions, validating its critical role in anti-aliasing. Removing the 3D smoothing filter achieves similar performance since the 3D filter aims at addressing the high-frequency artifacts when zooming in.
    }
    \label{tab:avg_blender_results_single_train_multi_test_ablation}
    \vspace{-0.1in}
\end{table*}

To evaluate the effectiveness of the 2D Mip filter, we perform an ablation study with the single-scale training and multi-scale testing setting to simulate zoom-out effects in the Blender dataset~\cite{mildenhall2020nerf}. The quantitative results are shown in ~\tabref{tab:avg_blender_results_single_train_multi_test_ablation}. Upon removing the dilation operation from 3DGS~\cite{kerbl3Dgaussians} (\textit{3DGS - Dilation}), the dilation effects are eliminated, outperforming 3DGS in this context. However, it also results in aliasing artifacts due to a lack of anti-aliasing. Mip-Splatting outperforms all baseline methods by a large margin. Removing the 2D Mip filter results in a notable decline in performance, validating its critical role in anti-aliasing. Without the 3D smoothing filter, it still produces alias-free rendering as the 3D filter aims at addressing the high-frequency artifacts when zooming in. 

\subsection{Single-scale Training and Multi-scale Testing}
\label{sec:ab_5scales}
\newcommand{\fivewidth}{0.16\textwidth}

\begin{figure*}[t]
    \centering
    \setlength{\tabcolsep}{0.1em}
    \renewcommand{\arraystretch}{0.4}
    \hfill{}\hspace*{-0.5em}
    \scriptsize
    \begin{tabular}{ccccccc}
    \multirow{1}{*}[3ex]{\rotatebox{90}{$4 \times$}} &
    \includegraphics[width=\fivewidth]{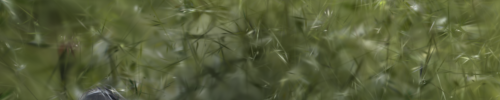}&
    \includegraphics[width=\fivewidth]{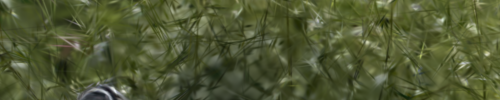}&
    \includegraphics[width=\fivewidth]{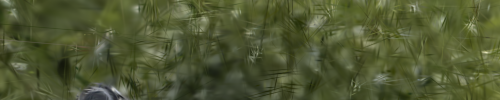}&
    \includegraphics[width=\fivewidth]{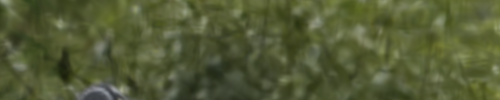}&
    \includegraphics[width=\fivewidth]{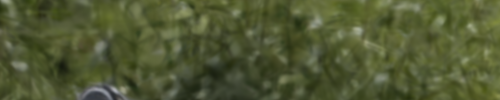}&
    \includegraphics[width=\fivewidth]{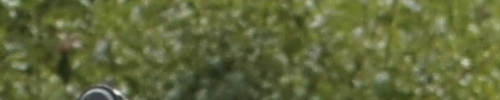}
    \\
    \multirow{1}{*}[3ex]{\rotatebox{90}{$2 \times$}} &
    \includegraphics[width=\fivewidth]{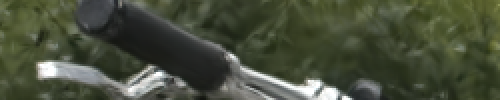}&
    \includegraphics[width=\fivewidth]{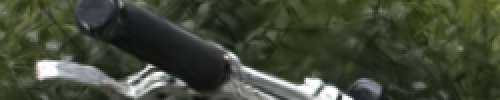}&
    \includegraphics[width=\fivewidth]{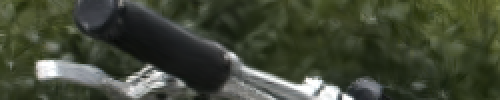}&
    \includegraphics[width=\fivewidth]{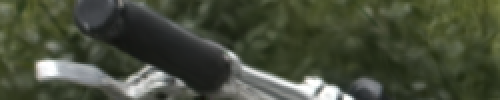}&
    \includegraphics[width=\fivewidth]{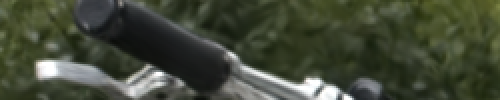}&
    \includegraphics[width=\fivewidth]{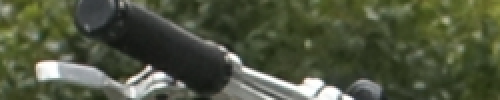}
    \\
    \multirow{1}{*}[3ex]{\rotatebox{90}{$1 \times$}} &
    \includegraphics[width=\fivewidth]{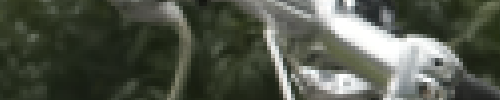}&
    \includegraphics[width=\fivewidth]{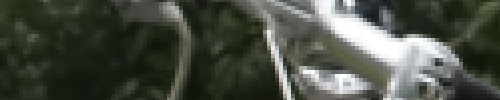}&
    \includegraphics[width=\fivewidth]{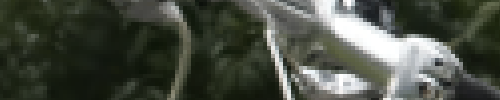}&
    \includegraphics[width=\fivewidth]{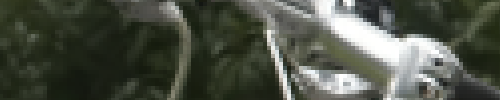}&
    \includegraphics[width=\fivewidth]{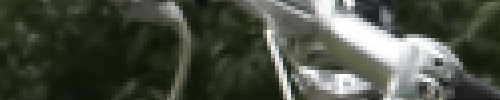}&
    \includegraphics[width=\fivewidth]{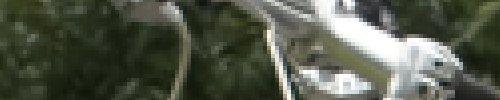}
    \\
    \multirow{1}{*}[4ex]{\rotatebox{90}{$\nicefrac{1}{2} \times$}} &
    \includegraphics[width=\fivewidth]{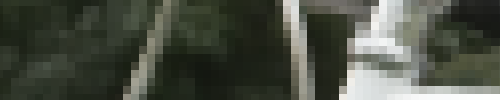}&
    \includegraphics[width=\fivewidth]{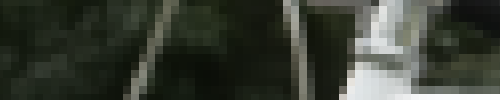}&
    \includegraphics[width=\fivewidth]{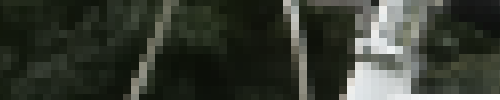}&
    \includegraphics[width=\fivewidth]{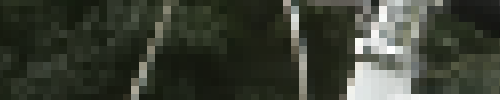}&
    \includegraphics[width=\fivewidth]{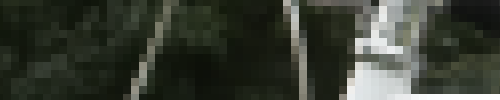}&
    \includegraphics[width=\fivewidth]{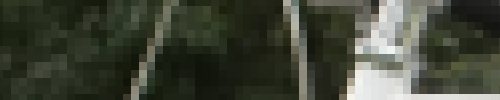}
    \\
    \multirow{1}{*}[4ex]{\rotatebox{90}{$\nicefrac{1}{4} \times$}} &
    \includegraphics[width=\fivewidth]{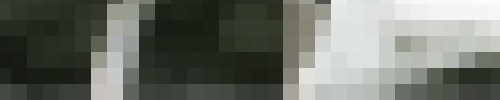}&
    \includegraphics[width=\fivewidth]{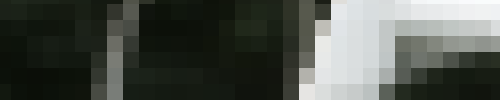}&
    \includegraphics[width=\fivewidth]{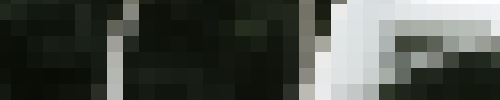}&
    \includegraphics[width=\fivewidth]{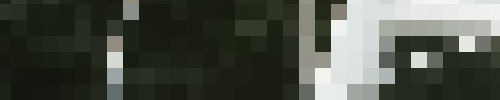}&
    \includegraphics[width=\fivewidth]{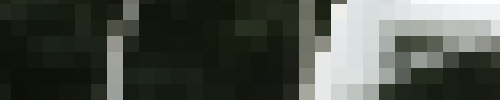}&
    \includegraphics[width=\fivewidth]{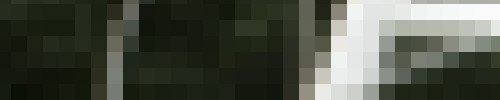}
    \\
    \\
    \multirow{1}{*}[3ex]{\rotatebox{90}{$4 \times$}} &
    \includegraphics[width=\fivewidth]{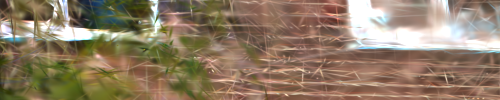}&
    \includegraphics[width=\fivewidth]{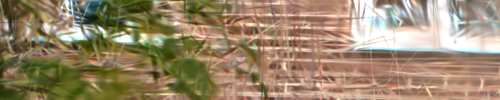}&
    \includegraphics[width=\fivewidth]{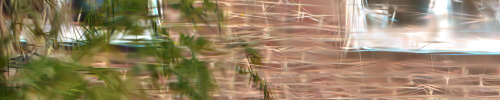}&
    \includegraphics[width=\fivewidth]{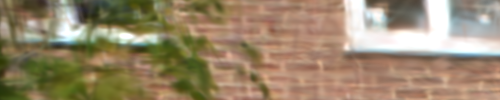}&
    \includegraphics[width=\fivewidth]{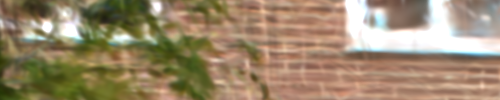}&
    \includegraphics[width=\fivewidth]{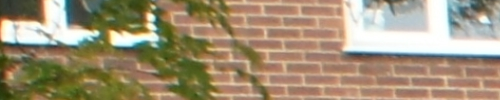}
    \\
    \multirow{1}{*}[3ex]{\rotatebox{90}{$2 \times$}} &
    \includegraphics[width=\fivewidth]{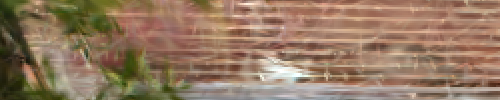}&
    \includegraphics[width=\fivewidth]{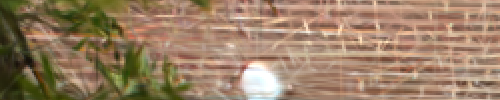}&
    \includegraphics[width=\fivewidth]{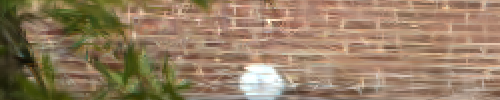}&
    \includegraphics[width=\fivewidth]{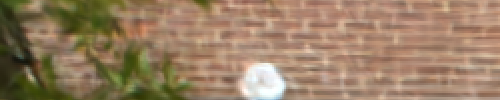}&
    \includegraphics[width=\fivewidth]{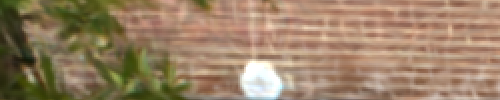}&
    \includegraphics[width=\fivewidth]{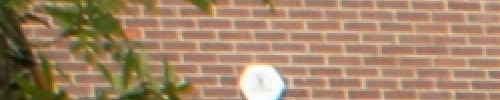}
    \\
    \multirow{1}{*}[3ex]{\rotatebox{90}{$1 \times$}} &
    \includegraphics[width=\fivewidth]{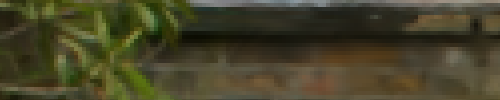}&
    \includegraphics[width=\fivewidth]{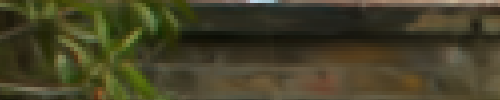}&
    \includegraphics[width=\fivewidth]{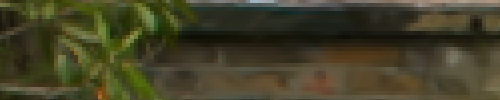}&
    \includegraphics[width=\fivewidth]{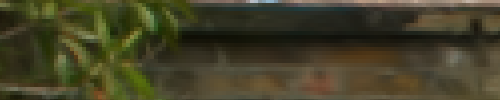}&
    \includegraphics[width=\fivewidth]{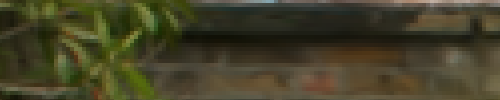}&
    \includegraphics[width=\fivewidth]{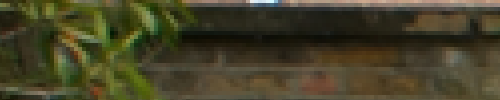}
    \\
    \multirow{1}{*}[4ex]{\rotatebox{90}{$\nicefrac{1}{2} \times$}} &
    \includegraphics[width=\fivewidth]{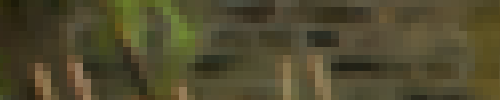}&
    \includegraphics[width=\fivewidth]{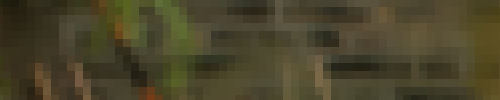}&
    \includegraphics[width=\fivewidth]{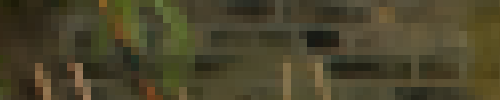}&
    \includegraphics[width=\fivewidth]{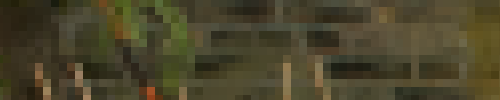}&
    \includegraphics[width=\fivewidth]{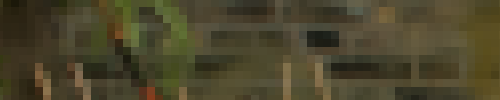}&
    \includegraphics[width=\fivewidth]{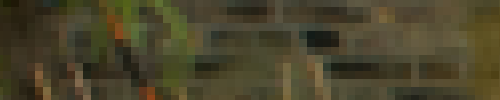}
    \\
    \multirow{1}{*}[4ex]{\rotatebox{90}{$\nicefrac{1}{4} \times$}} &
    \includegraphics[width=\fivewidth]{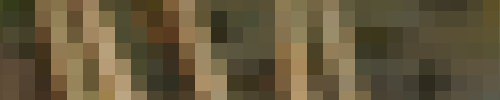}&
    \includegraphics[width=\fivewidth]{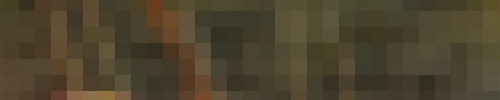}&
    \includegraphics[width=\fivewidth]{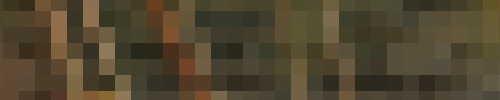}&
    \includegraphics[width=\fivewidth]{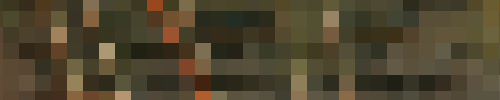}&
    \includegraphics[width=\fivewidth]{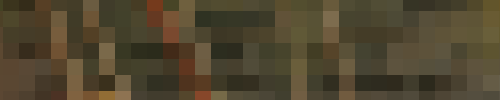}&
    \includegraphics[width=\fivewidth]{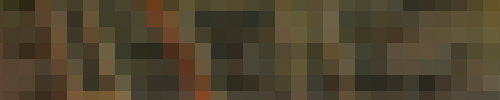}
    \\
    \\
    \multirow{1}{*}[3ex]{\rotatebox{90}{$4 \times$}} &
    \includegraphics[width=\fivewidth]{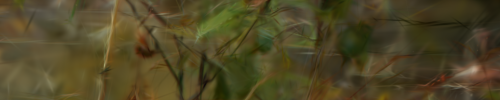}&
    \includegraphics[width=\fivewidth]{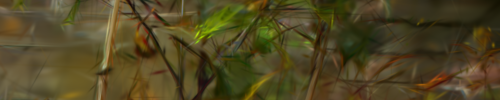}&
    \includegraphics[width=\fivewidth]{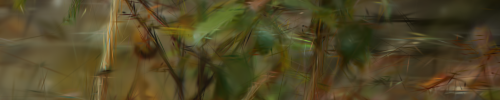}&
    \includegraphics[width=\fivewidth]{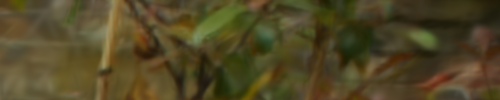}&
    \includegraphics[width=\fivewidth]{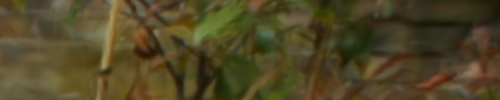}&
    \includegraphics[width=\fivewidth]{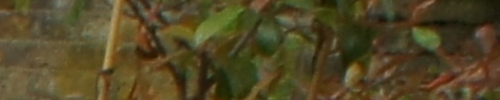}
    \\
    \multirow{1}{*}[3ex]{\rotatebox{90}{$2 \times$}} &
    \includegraphics[width=\fivewidth]{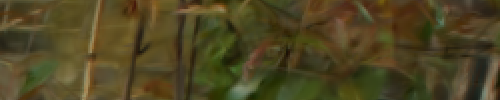}&
    \includegraphics[width=\fivewidth]{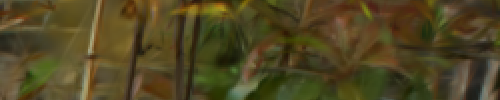}&
    \includegraphics[width=\fivewidth]{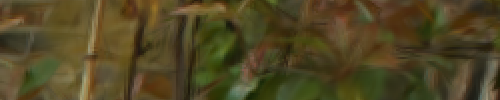}&
    \includegraphics[width=\fivewidth]{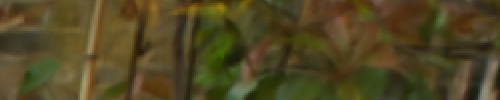}&
    \includegraphics[width=\fivewidth]{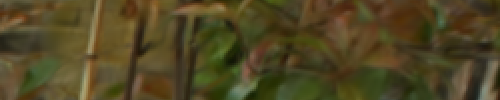}&
    \includegraphics[width=\fivewidth]{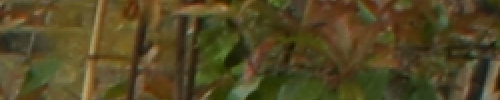}
    \\
    \multirow{1}{*}[3ex]{\rotatebox{90}{$1 \times$}} &
    \includegraphics[width=\fivewidth]{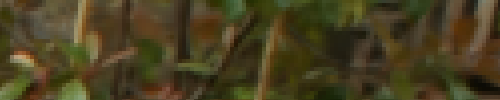}&
    \includegraphics[width=\fivewidth]{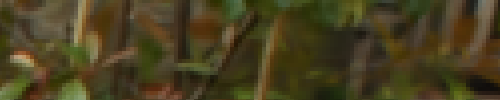}&
    \includegraphics[width=\fivewidth]{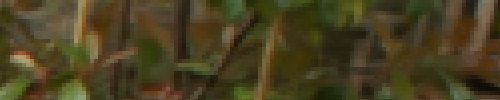}&
    \includegraphics[width=\fivewidth]{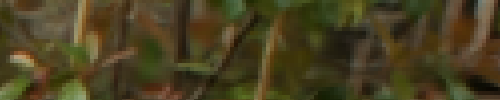}&
    \includegraphics[width=\fivewidth]{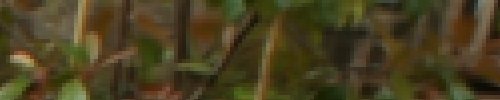}&
    \includegraphics[width=\fivewidth]{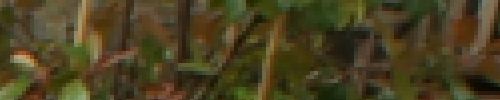}
    \\
    \multirow{1}{*}[4ex]{\rotatebox{90}{$\nicefrac{1}{2} \times$}} &
    \includegraphics[width=\fivewidth]{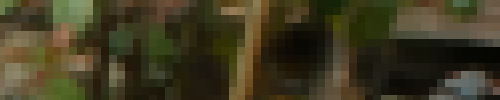}&
    \includegraphics[width=\fivewidth]{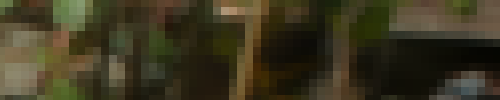}&
    \includegraphics[width=\fivewidth]{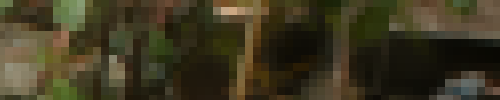}&
    \includegraphics[width=\fivewidth]{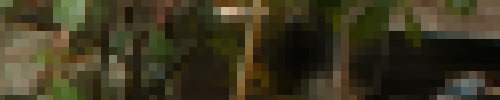}&
    \includegraphics[width=\fivewidth]{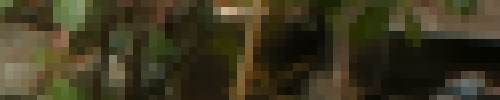}&
    \includegraphics[width=\fivewidth]{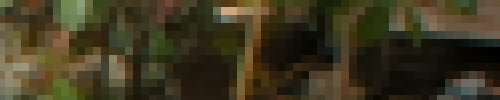}
    \\
    \multirow{1}{*}[4ex]{\rotatebox{90}{$\nicefrac{1}{4} \times$}} &
    \includegraphics[width=\fivewidth]{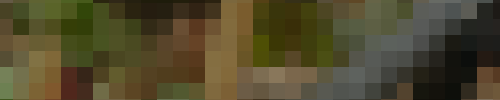}&
    \includegraphics[width=\fivewidth]{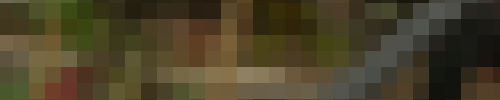}&
    \includegraphics[width=\fivewidth]{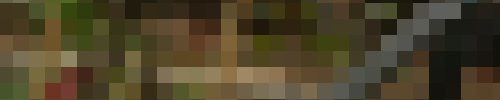}&
    \includegraphics[width=\fivewidth]{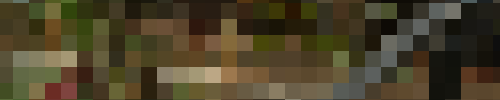}&
    \includegraphics[width=\fivewidth]{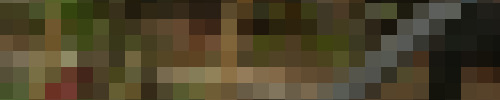}&
    \includegraphics[width=\fivewidth]{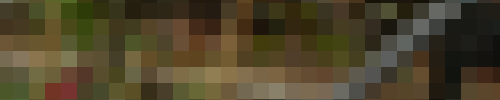}
    \\
    \\
    \multirow{1}{*}[3ex]{\rotatebox{90}{$4 \times$}} &
    \includegraphics[width=\fivewidth]{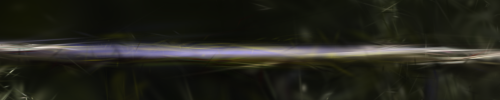}&
    \includegraphics[width=\fivewidth]{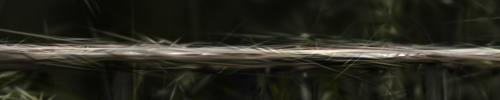}&
    \includegraphics[width=\fivewidth]{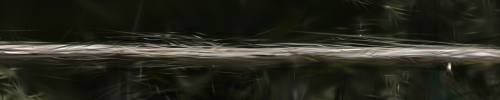}&
    \includegraphics[width=\fivewidth]{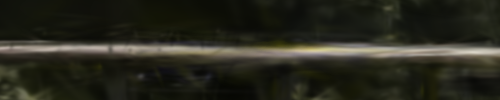}&
    \includegraphics[width=\fivewidth]{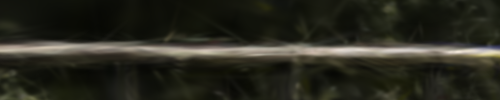}&
    \includegraphics[width=\fivewidth]{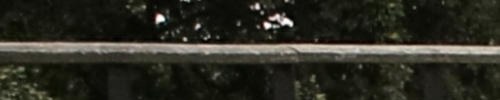}
    \\
    \multirow{1}{*}[3ex]{\rotatebox{90}{$2 \times$}} &
    \includegraphics[width=\fivewidth]{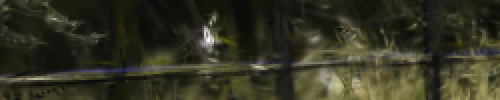}&
    \includegraphics[width=\fivewidth]{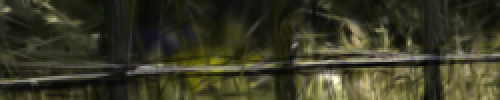}&
    \includegraphics[width=\fivewidth]{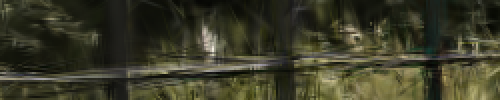}&
    \includegraphics[width=\fivewidth]{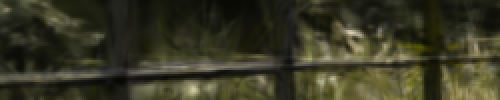}&
    \includegraphics[width=\fivewidth]{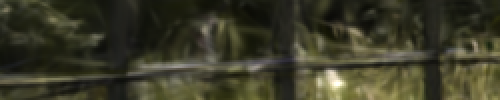}&
    \includegraphics[width=\fivewidth]{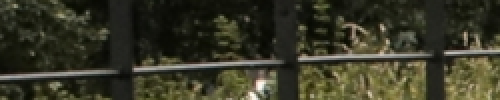}
    \\
    \multirow{1}{*}[3ex]{\rotatebox{90}{$1 \times$}} &
    \includegraphics[width=\fivewidth]{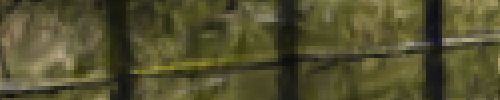}&
    \includegraphics[width=\fivewidth]{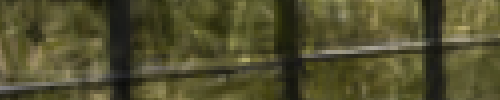}&
    \includegraphics[width=\fivewidth]{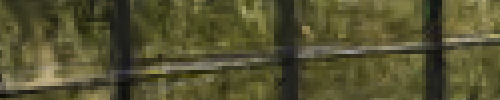}&
    \includegraphics[width=\fivewidth]{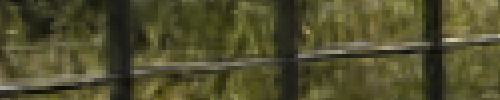}&
    \includegraphics[width=\fivewidth]{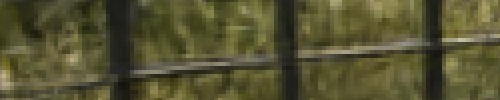}&
    \includegraphics[width=\fivewidth]{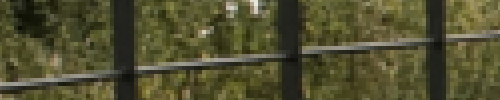}
    \\
    \multirow{1}{*}[4ex]{\rotatebox{90}{$\nicefrac{1}{2} \times$}} &
    \includegraphics[width=\fivewidth]{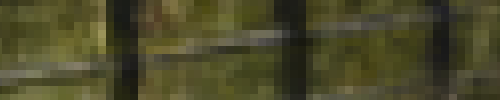}&
    \includegraphics[width=\fivewidth]{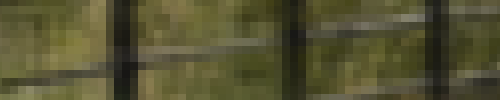}&
    \includegraphics[width=\fivewidth]{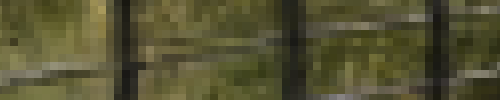}&
    \includegraphics[width=\fivewidth]{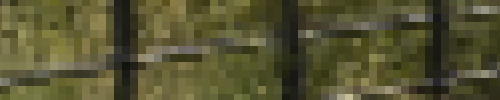}&
    \includegraphics[width=\fivewidth]{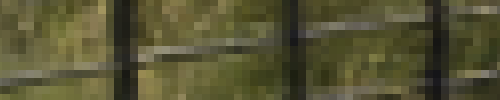}&
    \includegraphics[width=\fivewidth]{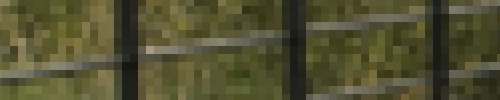}
    \\
    \multirow{1}{*}[4ex]{\rotatebox{90}{$\nicefrac{1}{4} \times$}} &
    \includegraphics[width=\fivewidth]{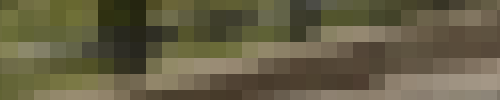}&
    \includegraphics[width=\fivewidth]{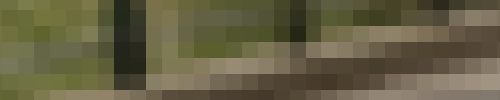}&
    \includegraphics[width=\fivewidth]{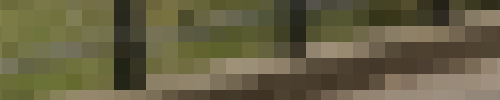}&
    \includegraphics[width=\fivewidth]{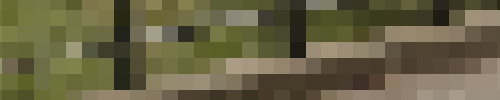}&
    \includegraphics[width=\fivewidth]{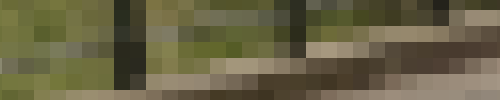}&
    \includegraphics[width=\fivewidth]{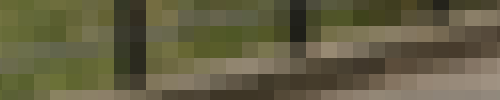}
    \\
    {}& 3DGS~\cite{kerbl3Dgaussians} & 3DGS~\cite{kerbl3Dgaussians} + EWA~\cite{zwicker2001ewa} & Ours w/o 3D smoothing filter& Ours w/o 2D Mip filter& Mip-Splatting (ours) & GT  
    \end{tabular}
    \vspace{-0.1in}
    \caption{
    \textbf{Single-scale Training and Multi-scale Testing on the Mip-NeRF 360 Dataset~\cite{barron2022mipnerf360}.}
    All methods are trained at $1\times$ resolution and evaluated at different resolutions to mimic zoom-out ($\nicefrac{1}{4}\times$ and $\nicefrac{1}{2}\times$) and zoom-in ($2\times$ and $4\times$). Mip-Splatting surpasses both 3DGS~\cite{kerbl3Dgaussians} and 3DGS + EWA~\cite{zwicker2001ewa} across different resolutions. Removing 3D smoothing filter leads to high-frquency artifacts when zooming in, while omitting 2D Mip filter results in aliasing artifacts when zooming out.
    }
    \label{fig:360_single_train_multi_test_5scales_ablation}
    \vspace{-0.1in}
\end{figure*}

\begin{table*}[]
    \renewcommand{\tabcolsep}{1pt}
    \centering
    \resizebox{1.0\linewidth}{!}{
    \begin{tabular}{@{}l@{\,\,}|cccccc|cccccc|cccccc}
    & \multicolumn{6}{c|}{PSNR $\uparrow$} & \multicolumn{6}{c|}{SSIM $\uparrow$} & \multicolumn{6}{c}{LPIPS $\downarrow$}  \\
    & $\nicefrac{1}{4} $ Res. & $\nicefrac{1}{2}$ Res. & $1 \times$ Res. & $2 \times$ Res. & $4 \times$ Res. & Avg.  
    & $\nicefrac{1}{4} $ Res. & $\nicefrac{1}{2} $ Res. & $1 \times$ Res. & $2 \times$ Res. & $4 \times$ Res. & Avg.  
    & $\nicefrac{1}{4}$ Res. & $\nicefrac{1}{2} $ Res. & $1 \times$ Res. & $2 \times$ Res. & $4 \times$ Res. & Avg.  
    \\ \hline
    
3DGS~\cite{kerbl3Dgaussians}& 20.85 & 24.66 & 28.01 & 25.08 & 23.37 & 24.39 & 0.681 & 0.812 &\cellcolor{orange} 0.834 & 0.766 & 0.735 & 0.765 & 0.203 & 0.158 &\cellcolor{tablered} 0.166 & 0.275 & 0.383 & 0.237
\\
3DGS~\cite{kerbl3Dgaussians} + EWA~\cite{zwicker2001ewa}&\cellcolor{yellow}27.40 &\cellcolor{yellow} 28.39 &\cellcolor{tablered} 28.09 & 26.43 & 25.30 & 27.12 &\cellcolor{yellow} 0.888 &\cellcolor{yellow} 0.871 & 0.833 & 0.774 & 0.738 &\cellcolor{yellow} 0.821 &\cellcolor{yellow} 0.103 &\cellcolor{yellow} 0.126 & 0.171 & 0.276 & 0.385 & 0.212
\\
\hline
Mip-Splatting (ours)&\cellcolor{tablered} 28.98 &\cellcolor{tablered} 29.02 &\cellcolor{tablered} 28.09 &\cellcolor{orange} 27.25 &\cellcolor{orange} 26.95 &\cellcolor{tablered} 28.06 &\cellcolor{tablered} 0.908 &\cellcolor{tablered} 0.880 &\cellcolor{tablered} 0.835 &\cellcolor{tablered} 0.798 &\cellcolor{orange} 0.800 &\cellcolor{tablered} 0.844 &\cellcolor{tablered} 0.086 &\cellcolor{tablered} 0.114 &\cellcolor{yellow} 0.168 &\cellcolor{tablered} 0.248 &\cellcolor{tablered} 0.331 &\cellcolor{tablered} 0.189
\\
Mip-Splatting (ours) w/o 3D smoothing filter&\cellcolor{orange} 28.69 &\cellcolor{orange} 28.94 &\cellcolor{yellow} 28.05 &\cellcolor{yellow} 27.06 &\cellcolor{yellow} 26.61 &\cellcolor{orange} 27.87 &\cellcolor{orange} 0.905 &\cellcolor{orange} 0.879 & 0.833 &\cellcolor{yellow} 0.790 &\cellcolor{yellow} 0.780 &\cellcolor{orange} 0.837 &\cellcolor{orange} 0.088 &\cellcolor{orange} 0.115 &\cellcolor{yellow} 0.168 &\cellcolor{yellow} 0.261 &\cellcolor{yellow} 0.359 &\cellcolor{orange} 0.198
\\
Mip-Splatting (ours) w/o 2D Mip filter& 26.09 & 28.04 &\cellcolor{yellow} 28.05 &\cellcolor{tablered} 27.27 &\cellcolor{tablered} 27.00 &\cellcolor{yellow} 27.29 & 0.815 & 0.856 &\cellcolor{orange} 0.834 &\cellcolor{tablered} 0.798 &\cellcolor{tablered} 0.802 &\cellcolor{yellow} 0.821 & 0.167 & 0.132 &\cellcolor{orange} 0.167 &\cellcolor{orange} 0.249 &\cellcolor{orange} 0.335 &\cellcolor{yellow} 0.210
    \end{tabular}
    }
    \vspace{-0.1in}
    \caption{
    \textbf{Single-scale Training and Multi-scale Testing on the Mip-NeRF 360 Dataset~\cite{barron2022mipnerf360}.} All methods are trained on the middle scale ($1 \times$) and evaluated across four scales ($\nicefrac{1}{4} \times$, $\nicefrac{1}{2} \times$, $1 \times$, $2 \times$, and $4 \times$), with evaluations at higher sampling rates simulating zoom-in effects. While our method yields comparable results at the training resolution, it significantly surpasses all previous work at all other scales. Omitting the 3D smoothing filter results in high-frequency artifacts when rendering higher resolution images, while removing the 2D Mip filter results in aliasing artifacts when rendering lower resolution images, as shown in ~\figref{fig:360_single_train_multi_test_5scales_ablation}.} \label{tab:avg_360_results_single_train_multi_test_5scale}
\end{table*}

In the main paper, we evaluate the zoom-out effects by rendering lower resolution images on the Blender dataset~\cite{mildenhall2020nerf} following~\cite{Barron2021ICCV,Hu2023ICCV} and simulating the zoom-in effects by rendering higher resolution images on the Mip-NeRF 360 dataset~\cite{barron2022mipnerf360}. 
Here we present an addition experiment evaluating both zoom-out and zoom-in effects on the Mip-NeRF 360 dataset~\cite{barron2022mipnerf360}. We use the images downsampled by a factor of 4 for training and evaluate it at multiple resolutions ($\nicefrac{1}{4} \times$, $\nicefrac{1}{2} \times$, $1 \times$, $2 \times$, $4 \times$). The quantitative results are presented in~\tabref{tab:avg_360_results_single_train_multi_test_5scale} and the qualitative comparison is shown in~\figref{fig:360_single_train_multi_test_5scales_ablation}. Mip-Splatting significantly outperforms 3DGS~\cite{kerbl3Dgaussians} and 3DGS + EWA~\cite{zwicker2001ewa} in rendering quality when zooming in and out, which is consistent to our main results. Further, removing our 3D smoothing filter leads to high-frequency artifacts, while removing our 2D Mip-filter results in aliasing artifacts, as evidenced in~\figref{fig:360_single_train_multi_test_5scales_ablation}.

\section{Additional Results}
\label{sec:addtional_results}
In this section, we provide more qualitative and quantitative results on the Blender dataset~\cite{mildenhall2020nerf} in~\secref{sec:additional_blender} and the Mip-NeRF 360 dataset~\cite{barron2022mipnerf360} in~\secref{sec:additional_mipnerf360}.

\subsection{Blender Dataset}
\label{sec:additional_blender}
We evaluate Mip-Splatting under two different settings in the Blender dataset~\cite{mildenhall2020nerf}. For \textit{multi-scale training and multi-scale testing}, the quantitative results are compiled in~\tabref{tab:multiblender_perscene}, where Mip-Splatting achieves state-of-the-art performance.  Additionally, per-scene metrics for \textit{single-scale training and multi-scale testing} are presented in Table~\ref{tab:singleblender_perscene}. A qualitative comparison against leading methods is shown in Figure~\ref{fig:blender_single_train_multi_test_supp}. Mip-Splatting outperforms both 3DGS~\cite{kerbl3Dgaussians} and 3DGS + EWA~\cite{zwicker2001ewa}, particularly noticeable when zooming out, \ie at lower resolutions.

\begin{table*}[h]
    \centering
    \small
    \begin{tabular}{l|cccccccc|c}
         & \multicolumn{9}{c}{\textbf{PSNR}} \\
 & \scenename{chair}  & \scenename{drums}  & \scenename{ficus}  & \scenename{hotdog}  & \scenename{lego}  & \scenename{materials}  & \scenename{mic}  & \scenename{ship} & \scenename{Average} 
 \\ 
 \hline 
NeRF w/o $\mathcal{L}_\text{area}$~\cite{mildenhall2020nerf,Barron2021ICCV}&                    29.92  &                    23.27  &                    27.15  &                    32.00  &                    27.75  &                    26.30  &                    28.40  &                    26.46  & 27.66
\\
NeRF~\cite{mildenhall2020nerf}&                    33.39  &                    25.87  &                    30.37  &                    35.64  &                    31.65  &                       30.18  &                    32.60  &                    30.09  & 31.23
\\
MipNeRF~\cite{Barron2021ICCV}&\cellcolor{yellow}                   37.14  &                    27.02  &                    33.19  &\cellcolor{tablered}                    39.31  &\cellcolor{tablered}                   35.74  &\cellcolor{tablered}                    32.56  &\cellcolor{tablered}                   38.04  &\cellcolor{tablered}                    33.08  &\cellcolor{orange} 34.51
\\
\hline
Plenoxels~\cite{yu2022plenoxels}&                    32.79  &                    25.25  &                    30.28  &                    34.65  &                    31.26  &                    28.33  &                    31.53  &                    28.59 & 30.34 
\\
TensoRF~\cite{Chen2022ECCV}&                    32.47  &                    25.37  &                    31.16  &                    34.96  &                    31.73  &                    28.53  &                    31.48  &                    29.08  & 30.60
\\
Instant-ngp~\cite{muller2022instant}&                    32.95  &                    26.43  &                    30.41  &                    35.87  &                    31.83  &                    29.31  &                    32.58  &                    30.23  & 31.20
\\
Tri-MipRF~\cite{Hu2023ICCV}*&\cellcolor{tablered} 37.67 &\cellcolor{orange} 27.35 &\cellcolor{yellow} 33.57 &\cellcolor{yellow} 38.78 &\cellcolor{orange} 35.72 &\cellcolor{yellow} 31.42 &\cellcolor{yellow} 37.63 &\cellcolor{yellow} 32.74 &\cellcolor{yellow} 34.36
\\
3DGS~\cite{kerbl3Dgaussians}& 32.73 & 25.30 & 29.00 & 35.03 & 29.44 & 27.13 & 31.17 & 28.33 & 29.77
\\
\hline
3DGS~\cite{kerbl3Dgaussians} + EWA~\cite{zwicker2001ewa}& 35.77 &\cellcolor{yellow} 27.14 &\cellcolor{orange} 33.65 & 37.74 & 32.75 & 30.21 & 35.21 & 31.63 & 33.01
\\
Mip-Splatting (ours)&\cellcolor{orange}37.48 &\cellcolor{tablered} 27.74 &\cellcolor{tablered} 34.71 &\cellcolor{orange} 39.15 &\cellcolor{yellow} 35.07 &\cellcolor{orange} 31.88 &\cellcolor{orange} 37.68 &\cellcolor{orange} 32.80 &\cellcolor{tablered} 34.56
\\

\multicolumn{9}{c}{} \\
 & \multicolumn{9}{c}{\textbf{SSIM}} \\
 & \scenename{chair}  & \scenename{drums}  & \scenename{ficus}  & \scenename{hotdog}  & \scenename{lego}  & \scenename{materials}  & \scenename{mic}  & \scenename{ship} & \scenename{Average} 
\\ 
\hline 
NeRF w/o $\mathcal{L}_\text{area}$~\cite{mildenhall2020nerf,Barron2021ICCV}&                    0.944  &                    0.891  &                    0.942  &                    0.959  &                    0.926  &                    0.934  &                    0.958  &                    0.861  &0.927
\\
NeRF~\cite{mildenhall2020nerf}&                    0.971  &                    0.932  &                    0.971  &                    0.979  &                    0.965  &                       0.967  &                    0.980  &                    0.900 &0.958  
\\
MipNeRF~\cite{Barron2021ICCV}&\cellcolor{yellow}                     0.988  &                     0.945  &                     0.984  &\cellcolor{orange}                     0.988  &\cellcolor{yellow}                     0.984  &\cellcolor{orange}                      0.977  &\cellcolor{orange}                      0.993  &                     0.922  &  0.973
\\
\hline
Plenoxels~\cite{yu2022plenoxels}&                    0.968  &                    0.929  &                    0.972  &                    0.976  &                    0.964  &                    0.959  &                    0.979  &                    0.892 & 0.955  
\\
TensoRF~\cite{Chen2022ECCV}&                    0.967  &                    0.930  &                    0.974  &                    0.977  &                    0.967  &                    0.957  &                    0.978  &                    0.895  & 0.956 
\\
Instant-ngp~\cite{muller2022instant}&                    0.971  &                    0.940  &                    0.973  &                    0.979  &                    0.966  &                    0.959  &                    0.981  &                    0.904  & 0.959
\\
Tri-MipRF~\cite{Hu2023ICCV}*&\cellcolor{orange} 0.990 &\cellcolor{yellow} 0.951 &\cellcolor{yellow} 0.985 &\cellcolor{orange} 0.988 &\cellcolor{orange} 0.986 & 0.969 &\cellcolor{yellow} 0.992 &\cellcolor{orange} 0.929 &\cellcolor{orange} 0.974
\\
3DGS~\cite{kerbl3Dgaussians}& 0.976 & 0.941 & 0.968 & 0.982 & 0.964 & 0.956 & 0.979 & 0.910 & 0.960
\\
\hline
3DGS~\cite{kerbl3Dgaussians} + EWA~\cite{zwicker2001ewa}& 0.986 &\cellcolor{orange} 0.958 &\cellcolor{orange} 0.988 &\cellcolor{orange} 0.988 & 0.979 &\cellcolor{yellow} 0.972 & 0.990 &\cellcolor{orange} 0.929 &\cellcolor{orange} 0.974
\\
Mip-Splatting (ours)&\cellcolor{tablered} 0.991 &\cellcolor{tablered} 0.963 &\cellcolor{tablered} 0.990 &\cellcolor{tablered} 0.990 &\cellcolor{tablered} 0.987 &\cellcolor{tablered} 0.978 &\cellcolor{tablered} 0.994 &\cellcolor{tablered} 0.936 &\cellcolor{tablered} 0.979
\\

\multicolumn{9}{c}{} \\
 & \multicolumn{9}{c}{\textbf{LPIPS}} \\
 & \scenename{chair}  & \scenename{drums}  & \scenename{ficus}  & \scenename{hotdog}  & \scenename{lego}  & \scenename{materials}  & \scenename{mic}  & \scenename{ship} & \scenename{Average} 
\\ 
\hline 
NeRF w/o $\mathcal{L}_\text{area}$~\cite{mildenhall2020nerf,Barron2021ICCV}&                    0.035  &                    0.069  &                    0.032  &                    0.028  &                    0.041  &                    0.045  &                    0.031  &                    0.095  & 0.052
\\
NeRF~\cite{mildenhall2020nerf}&                    0.028  &                       0.059  &                    0.026  &                       0.024  &                    0.035  &                       0.033  &                    0.025  &                       0.085  & 0.044
\\
MipNeRF~\cite{Barron2021ICCV}&\cellcolor{orange}                     0.011  &\cellcolor{yellow}                     0.044  &\cellcolor{yellow}                      0.014  &\cellcolor{orange}                      0.012  &\cellcolor{orange}                     0.013  &\cellcolor{orange}                      0.019  &\cellcolor{orange}                      0.007  &\cellcolor{orange}                      0.062  &\cellcolor{orange}  0.026
\\
\hline
Plenoxels~\cite{yu2022plenoxels}&                    0.040  &                    0.070  &                    0.032  &                    0.037  &                    0.038  &                    0.055  &                    0.036  &                    0.104  & 0.051
\\
TensoRF~\cite{Chen2022ECCV}&                    0.042  &                    0.075  &                    0.032  &                    0.035  &                    0.036  &                    0.063  &                    0.040  &                    0.112  & 0.054
\\
Instant-ngp~\cite{muller2022instant}&                    0.035  &                    0.066  &                    0.029  &                    0.028  &                    0.040  &                    0.051  &                    0.032  &                    0.095  & 0.047
\\
Tri-MipRF~\cite{Hu2023ICCV}*&\cellcolor{orange}0.011 & 0.046 & 0.016 &\cellcolor{yellow} 0.014 &\cellcolor{orange} 0.013 & 0.033 &\cellcolor{yellow} 0.008 &\cellcolor{yellow} 0.069 &\cellcolor{orange} 0.026
\\
3DGS~\cite{kerbl3Dgaussians}& 0.025 & 0.056 & 0.030 & 0.022 & 0.038 & 0.040 & 0.023 & 0.086 & 0.040
\\
\hline
3DGS~\cite{kerbl3Dgaussians} + EWA~\cite{zwicker2001ewa}& 0.017 &\cellcolor{orange} 0.039 &\cellcolor{orange} 0.013 & 0.016 & 0.024 &\cellcolor{yellow} 0.026 & 0.011 & 0.070 & 0.027
\\
Mip-Splatting (ours)&\cellcolor{tablered} 0.010 &\cellcolor{tablered} 0.031 &\cellcolor{tablered} 0.009 &\cellcolor{tablered} 0.011 &\cellcolor{tablered} 0.012 &\cellcolor{tablered} 0.018 &\cellcolor{tablered} 0.005 &\cellcolor{tablered} 0.059 &\cellcolor{tablered} 0.019
\\

\multicolumn{9}{c}{}

    \end{tabular}
    \caption{\textbf{Multi-scale Training and Multi-scale Testing on the the Blender dataset~\cite{mildenhall2020nerf}}. For each scene, we report the arithmetic mean of each metric averaged over the 4 scales used in the dataset. 
    }
    \label{tab:multiblender_perscene}
\end{table*}

\begin{table*}[h]
    \centering
    \small
    \begin{tabular}{l|cccccccc|c}
         & \multicolumn{9}{c}{\textbf{PSNR}} \\
 & \scenename{chair}  & \scenename{drums}  & \scenename{ficus}  & \scenename{hotdog}  & \scenename{lego}  & \scenename{materials}  & \scenename{mic}  & \scenename{ship} & \scenename{Average} 
 \\ 
 \hline 
NeRF~\cite{mildenhall2020nerf}& 31.99 & 25.31 & 30.74 & 34.45 & 30.69 &\cellcolor{yellow} 28.86 & 31.41 & 28.36 & 30.23 
\\
MipNeRF~\cite{Barron2021ICCV}&\cellcolor{orange} 32.89 &\cellcolor{orange} 25.58 &\cellcolor{yellow} 31.80 &\cellcolor{orange} 35.40 &\cellcolor{orange} 32.24 &\cellcolor{orange} 29.46 &\cellcolor{tablered} 33.26 &\cellcolor{orange} 29.88 &\cellcolor{orange} 31.31    
\\
\hline
TensoRF~\cite{Chen2022ECCV}&  32.17 &\cellcolor{yellow} 25.51 & 31.19 & 34.69 & 31.46 & 28.60 & 31.50 & 28.71 & 30.48
\\
Instant-ngp~\cite{muller2022instant}& 32.18 & 25.05 & 31.32 &\cellcolor{yellow} 34.85 &\cellcolor{yellow} 31.53 & 28.59 &\cellcolor{orange} 32.15 &\cellcolor{yellow} 28.84 &\cellcolor{yellow} 30.57 
\\
Tri-MipRF~\cite{Hu2023ICCV}& 32.48 & 24.01 & 28.41 & 34.45 & 30.41 & 27.82 & 31.19 & 27.02 & 29.47
\\
3DGS~\cite{kerbl3Dgaussians}& 26.81 & 21.17 & 26.02 & 28.80 & 25.36 & 23.10 & 24.39 & 23.05 & 24.84
\\
\hline
3DGS~\cite{kerbl3Dgaussians} + EWA~\cite{zwicker2001ewa}&\cellcolor{yellow} 32.85 & 24.91 &\cellcolor{orange} 31.94 & 33.33 & 29.76 & 27.36 & 27.68 & 27.41 & 29.40
\\
Mip-Splatting (ours)&\cellcolor{tablered}35.69 &\cellcolor{tablered} 26.50 &\cellcolor{tablered} 32.99 &\cellcolor{tablered} 36.18 &\cellcolor{tablered} 32.76 &\cellcolor{tablered} 30.01 &\cellcolor{yellow} 31.66 &\cellcolor{tablered} 29.98 &\cellcolor{tablered} 31.97
\\

\multicolumn{9}{c}{} \\
 & \multicolumn{9}{c}{\textbf{SSIM}} \\
 & \scenename{chair}  & \scenename{drums}  & \scenename{ficus}  & \scenename{hotdog}  & \scenename{lego}  & \scenename{materials}  & \scenename{mic}  & \scenename{ship} & \scenename{Average} 
\\ 
\hline 
NeRF~\cite{mildenhall2020nerf}& 0.968 & 0.936 & 0.976 & 0.977 & 0.963 &\cellcolor{yellow} 0.964 & 0.980 & 0.887 & 0.956 
\\
MipNeRF~\cite{Barron2021ICCV}&\cellcolor{yellow}0.974 &\cellcolor{yellow} 0.939 &\cellcolor{yellow} 0.981 &\cellcolor{orange} 0.982 &\cellcolor{orange} 0.973 &\cellcolor{orange} 0.969 &\cellcolor{tablered} 0.987 &\cellcolor{orange} 0.915 &\cellcolor{orange} 0.965    
\\
\hline
TensoRF~\cite{Chen2022ECCV}& 0.970 & 0.938 & 0.978 & 0.979 &\cellcolor{yellow} 0.970 & 0.963 & 0.981 & 0.906 &\cellcolor{yellow} 0.961       
\\
Instant-ngp~\cite{muller2022instant}& 0.970 & 0.935 & 0.977 &\cellcolor{yellow} 0.980 & 0.969 & 0.962 &\cellcolor{yellow} 0.982 & 0.909 &\cellcolor{yellow} 0.961
\\
Tri-MipRF~\cite{Hu2023ICCV}& 0.971 & 0.908 & 0.957 & 0.975 & 0.957 & 0.953 & 0.975 & 0.883 & 0.947
\\
3DGS~\cite{kerbl3Dgaussians}& 0.915 & 0.851 & 0.921 & 0.930 & 0.882 & 0.882 & 0.909 & 0.827 & 0.890
\\
\hline
3DGS~\cite{kerbl3Dgaussians} + EWA~\cite{zwicker2001ewa}&\cellcolor{orange} 0.978 &\cellcolor{orange} 0.942 &\cellcolor{orange} 0.983 & 0.977 & 0.964 & 0.958 & 0.963 &\cellcolor{yellow} 0.912 & 0.960
\\
Mip-Splatting (ours)&\cellcolor{tablered} 0.988 &\cellcolor{tablered} 0.958 &\cellcolor{tablered} 0.988 &\cellcolor{tablered} 0.987 &\cellcolor{tablered} 0.982 &\cellcolor{tablered} 0.974 &\cellcolor{orange} 0.986 &\cellcolor{tablered} 0.930 &\cellcolor{tablered} 0.974
\\

\multicolumn{9}{c}{} \\
 & \multicolumn{9}{c}{\textbf{LPIPS}} \\
 & \scenename{chair}  & \scenename{drums}  & \scenename{ficus}  & \scenename{hotdog}  & \scenename{lego}  & \scenename{materials}  & \scenename{mic}  & \scenename{ship} & \scenename{Average} 
\\ 
\hline 
NeRF~\cite{mildenhall2020nerf}& 0.040 & 0.067 & 0.027 & 0.034 & 0.043 & 0.049 & 0.035 & 0.132 & 0.053  
\\
MipNeRF~\cite{Barron2021ICCV}&  0.033 &\cellcolor{yellow} 0.062 &\cellcolor{yellow} 0.022 & 0.025 &\cellcolor{orange} 0.030 &\cellcolor{yellow} 0.041 &\cellcolor{orange} 0.023 &\cellcolor{yellow} 0.092 &\cellcolor{yellow} 0.041      
\\
\hline
TensoRF~\cite{Chen2022ECCV}&  0.036 & 0.066 & 0.027 & 0.030 & 0.035 & 0.052 & 0.034 & 0.102 & 0.048 
\\
Instant-ngp~\cite{muller2022instant}& 0.036 & 0.074 & 0.035 & 0.030 & 0.035 & 0.054 & 0.034 & 0.096 & 0.049
\\
Tri-MipRF~\cite{Hu2023ICCV}&\cellcolor{yellow} 0.026 & 0.086 & 0.041 &\cellcolor{yellow} 0.023 & 0.036 & 0.048 &\cellcolor{orange} 0.023 & 0.117 & 0.050
\\
3DGS~\cite{kerbl3Dgaussians}& 0.047 & 0.087 & 0.055 & 0.034 & 0.064 & 0.055 & 0.046 & 0.113 & 0.063
\\
\hline
3DGS~\cite{kerbl3Dgaussians} + EWA~\cite{zwicker2001ewa}&\cellcolor{orange} 0.023 &\cellcolor{orange} 0.051 &\cellcolor{orange} 0.017 &\cellcolor{orange} 0.018 &\cellcolor{yellow} 0.033 &\cellcolor{orange} 0.027 & 0.024 &\cellcolor{orange} 0.077 &\cellcolor{orange} 0.034
\\
Mip-Splatting (ours)&\cellcolor{tablered}0.014 &\cellcolor{tablered} 0.035 &\cellcolor{tablered} 0.012 &\cellcolor{tablered} 0.014 &\cellcolor{tablered} 0.016 &\cellcolor{tablered} 0.019 &\cellcolor{tablered} 0.015 &\cellcolor{tablered} 0.066 &\cellcolor{tablered} 0.024
\\

\multicolumn{9}{c}{}

    \end{tabular}
    \caption{\textbf{Single-scale Training and Multi-scale Testing on the the Blender dataset~\cite{mildenhall2020nerf}}. For each scene, we report the arithmetic mean of each metric averaged over the four scales used in the dataset. 
    }
    \label{tab:singleblender_perscene}
\end{table*}

\subsection{Mip-NeRF 360 Dataset}
\label{sec:additional_mipnerf360}
We further evaluate Mip-Splatting on the Mip-NeRF 360 dataset~\cite{barron2022mipnerf360} across two experimental setups. In the first setup, we follow the standard approach where models are trained and evaluated at the same scale, with indoor scenes downsampled by a factor of two and outdoor scenes by four. Quantitative results with per-scene metrics are shown in~\tabref{tab:normal_360_perscene}, our method performs on par with 3DGS~\cite{kerbl3Dgaussians} and 3DGS + EWA~\cite{zwicker2001ewa} in this challenging benchmark, without any decrease in performance. 

In the second setup, models are trained on data downsampled by a factor of 8 and rendered at successively higher resolutions ($1 \times$, $2 \times$, $4 \times$, and $8 \times$) to simulate zoom-in effects. The quantitative results with per-scene metrics can be found in~\tabref{tab:multi_360_perscene}. Qualitative comparison with state-of-the-art methods are provided in~\figref{fig:360_single_train_multi_test_supp}. Mip-Splatting effectively eliminates high-frequency artifacts, yielding high quality renderings that more closely resemble ground truth.

\begin{table*}[h]
    \centering
    \small
    \begin{tabular}{l|ccccc|cccc}
         & \multicolumn{9}{c}{\textbf{PSNR}} \\
 & \scenename{bicycle}  & \scenename{flowers}  & \scenename{garden}  & \scenename{stump}  & \scenename{treehill}  & \scenename{room}  & \scenename{counter}  & \scenename{kitchen} & \scenename{bonsai} 
 \\ 
 \hline 
NeRF~\cite{mildenhall2020nerf,jaxnerf2020github}&          21.76 &                   19.40 &                   23.11 &                   21.73 &                   21.28 &                   28.56 &                   25.67 &                   26.31 &                   26.81
\\
mip-NeRF~\cite{Barron2021ICCV}&        21.69 &                   19.31 &                   23.16 &                   23.10 &                   21.21 &                   28.73 &                   25.59 &                   26.47 &                   27.13 
\\
NeRF++~\cite{kaizhang2020}&         22.64 &                   20.31 &                   24.32 &                   24.34 &                   22.20 &                   28.87 &    26.38 &                   27.80 &                   29.15
\\
Plenoxels~\cite{yu2022plenoxels}& 21.91 & 20.10 & 23.49 & 20.661 & 22.25&27.59& 23.62& 23.42& 24.67
\\
Instant NGP ~\cite{muller2022instant,yariv2023bakedsdf}&     22.79 &                   19.19 &                   25.26 &                   24.80 &                   22.46 &                   30.31 &                   26.21 &                   29.00 &                   31.08 
\\
mip-NeRF 360~\cite{barron2022mipnerf360, multinerf2022}&24.40 &    21.64 &    26.94 &    26.36 &    22.81 &    31.40 &\cellcolor{tablered}       29.44 &\cellcolor{orange}    32.02 &\cellcolor{orange}    33.11
\\
Zip-NeRF~\cite{Barron2023ICCV}&\cellcolor{tablered} 25.80 &\cellcolor{tablered}       22.40 &\cellcolor{tablered}       28.20 &\cellcolor{tablered}       27.55 &\cellcolor{tablered}       23.89 &\cellcolor{tablered}       32.65 &\cellcolor{orange}    29.38 &\cellcolor{tablered}       32.50 &\cellcolor{tablered}       34.46
\\
3DGS~\cite{kerbl3Dgaussians}& 25.25 & 21.52& 27.41& 26.55& 22.49& 30.63& 28.70& 30.32& 31.98
\\
3DGS~\cite{kerbl3Dgaussians}*&25.63 & 21.77 &\cellcolor{yellow} 27.70 &\cellcolor{yellow} 26.87 & 22.75 &\cellcolor{yellow} 31.69 & 29.08 & 31.56 & 32.29
\\
\hline
3DGS~\cite{kerbl3Dgaussians} + EWA~\cite{zwicker2001ewa}&\cellcolor{yellow} 25.64 &\cellcolor{yellow} 21.86 & 27.65 &\cellcolor{yellow} 26.87 &\cellcolor{yellow} 22.91 & 31.68 &\cellcolor{yellow} 29.21 &\cellcolor{yellow} 31.59 &\cellcolor{yellow} 32.51
\\
Mip-Splatting (ours)&\cellcolor{orange}   25.72 &\cellcolor{orange} 21.93 &\cellcolor{orange} 27.76 &\cellcolor{orange} 26.94 &\cellcolor{orange} 22.98 &\cellcolor{orange} 31.74 & 29.16 & 31.55 & 32.31
\\

\multicolumn{9}{c}{} \\
 & \multicolumn{9}{c}{\textbf{SSIM}} \\
 & \scenename{bicycle}  & \scenename{flowers}  & \scenename{garden}  & \scenename{stump}  & \scenename{treehill}  & \scenename{room}  & \scenename{counter}  & \scenename{kitchen} & \scenename{bonsai} 
\\ 
\hline 
NeRF~\cite{mildenhall2020nerf,jaxnerf2020github}&         0.455 &                   0.376 &                   0.546 &                   0.453 &                   0.459 &                   0.843 &                   0.775 &                   0.749 &                   0.792
\\
mip-NeRF~\cite{Barron2021ICCV}&          0.454 &                   0.373 &                   0.543 &                   0.517 &                   0.466 &                   0.851 &                   0.779 &                   0.745 &                   0.818
\\
NeRF++~\cite{kaizhang2020}&             0.526 &                   0.453 &                   0.635 &                   0.594 &                   0.530 &                   0.852 &                   0.802 &                   0.816 &                   0.876
\\
Plenoxels~\cite{yu2022plenoxels}& 0.496& 0.431& 0.606& 0.523& 0.509& 0.842& 0.759& 0.648& 0.814
\\
Instant NGP ~\cite{muller2022instant,yariv2023bakedsdf}&  0.540 &                   0.378 &                   0.709 &                   0.654 &                   0.547 &                   0.893 &    0.845 &                   0.857 &                   0.924 
\\
mip-NeRF 360~\cite{barron2022mipnerf360, multinerf2022}&0.693 &    0.583 &    0.816 &    0.746 &    0.632 &    0.913 &    0.895 &    0.920 &    0.939
\\
Zip-NeRF~\cite{Barron2023ICCV}&0.769 &\cellcolor{tablered}       0.642 &       0.860 &\cellcolor{tablered}       0.800 &\cellcolor{tablered}       0.681 &       0.925 &       0.902 &       0.928 &\cellcolor{tablered}       0.949
\\
3DGS~\cite{kerbl3Dgaussians}&0.771& 0.605& 0.868& 0.775& 0.638& 0.914& 0.905& 0.922& 0.938
\\
3DGS~\cite{kerbl3Dgaussians}*&\cellcolor{orange} 0.777 &\cellcolor{yellow} 0.622 &\cellcolor{orange} 0.873 & 0.783 & 0.652 &\cellcolor{tablered} 0.928 &\cellcolor{tablered} 0.916 &\cellcolor{tablered} 0.933 &\cellcolor{orange} 0.948
\\
\hline
3DGS~\cite{kerbl3Dgaussians} + EWA~\cite{zwicker2001ewa}&\cellcolor{orange} 0.777 & 0.620 &\cellcolor{yellow} 0.871 &\cellcolor{yellow} 0.784 &\cellcolor{orange} 0.655 &\cellcolor{yellow} 0.927 &\cellcolor{tablered} 0.916 &\cellcolor{tablered} 0.933 &\cellcolor{orange} 0.948
\\
Mip-Splatting (ours)&\cellcolor{tablered} 0.780 &\cellcolor{orange} 0.623 &\cellcolor{tablered} 0.875 &\cellcolor{orange} 0.786 &\cellcolor{orange} 0.655 &\cellcolor{tablered} 0.928 &\cellcolor{tablered} 0.916 &\cellcolor{tablered} 0.933 &\cellcolor{orange} 0.948 
\\

\multicolumn{9}{c}{} \\
 & \multicolumn{9}{c}{\textbf{LPIPS}} \\
 & \scenename{bicycle}  & \scenename{flowers}  & \scenename{garden}  & \scenename{stump}  & \scenename{treehill}  & \scenename{room}  & \scenename{counter}  & \scenename{kitchen} & \scenename{bonsai} 
\\ 
\hline 
NeRF~\cite{mildenhall2020nerf,jaxnerf2020github}&      0.536 &                   0.529 &                   0.415 &                   0.551 &                   0.546 &                   0.353 &                   0.394 &                   0.335 &                   0.398
\\
mip-NeRF~\cite{Barron2021ICCV}&       0.541 &                   0.535 &                   0.422 &                   0.490 &                   0.538 &                   0.346 &                   0.390 &                   0.336 &                   0.370 
\\
NeRF++~\cite{kaizhang2020}&   0.455 &                   0.466 &                   0.331 &                   0.416 &                   0.466 &                   0.335 &                   0.351 &                   0.260 &                   0.291
\\
Plenoxels~\cite{yu2022plenoxels}& 0.506& 0.521& 0.3864& 0.503& 0.540& 0.419& 0.441& 0.447& 0.398
\\
Instant NGP ~\cite{muller2022instant,yariv2023bakedsdf}&     0.398 &                   0.441 &                   0.255 &                   0.339 &                   0.420 &                   0.242 &    0.255 &                   0.170 &                   0.198
\\
mip-NeRF 360~\cite{barron2022mipnerf360, multinerf2022}&0.289 &    0.345 &    0.164 &    0.254 &    0.338 &    0.211 &    0.203 &    0.126 &    0.177
\\
Zip-NeRF~\cite{Barron2023ICCV}& 0.208 &\cellcolor{tablered}       0.273 &       0.118 &\cellcolor{tablered}       0.193 &\cellcolor{tablered}       0.242 &       0.196 &       0.185 &       0.116 &\cellcolor{tablered}    0.173
\\
3DGS~\cite{kerbl3Dgaussians}&\cellcolor{tablered} 0.205& 0.336&\cellcolor{tablered} 0.103& 0.210&\cellcolor{orange} 0.317& 0.220& 0.204& 0.129& 0.205
\\
3DGS~\cite{kerbl3Dgaussians}*&\cellcolor{tablered}0.205 &\cellcolor{orange} 0.329 &\cellcolor{tablered} 0.103 &\cellcolor{orange} 0.208 &\cellcolor{yellow} 0.318 &\cellcolor{tablered} 0.192 &\cellcolor{tablered} 0.178 &\cellcolor{tablered} 0.113 & 0.174
\\
\hline
3DGS~\cite{kerbl3Dgaussians} + EWA~\cite{zwicker2001ewa}& 0.213 & 0.335 & 0.111 & 0.210 & 0.325 &\cellcolor{tablered} 0.192 &\cellcolor{orange} 0.179 &\cellcolor{tablered} 0.113 &\cellcolor{tablered} 0.173
\\
Mip-Splatting (ours)&\cellcolor{yellow} 0.206 &\cellcolor{yellow} 0.331 &\cellcolor{tablered} 0.103 &\cellcolor{yellow} 0.209 & 0.320 &\cellcolor{tablered} 0.192 &\cellcolor{orange} 0.179 &\cellcolor{tablered} 0.113 &\cellcolor{tablered} 0.173  
\\

\multicolumn{9}{c}{}
    \end{tabular}
    \caption{\textbf{Single-scale Training and Single-scale Testing on the Mip-NeRF 360 dataset~\cite{barron2022mipnerf360}}. Indoor scenes are downsampled by a factor of 2 and outdoor scenes by 4. 
    }
    \label{tab:normal_360_perscene}
\end{table*}

\begin{table*}[h]
    \centering
    \small
    \begin{tabular}{l|ccccc|cccc}
         & \multicolumn{9}{c}{\textbf{PSNR}} \\
 & \scenename{bicycle}  & \scenename{flowers}  & \scenename{garden}  & \scenename{stump}  & \scenename{treehill}  & \scenename{room}  & \scenename{counter}  & \scenename{kitchen} & \scenename{bonsai} 
 \\ 
 \hline 
Instant-NGP~\cite{muller2022instant}&  22.51 & 20.25 & 24.65 & 23.15 & 22.24 &\cellcolor{yellow} 29.48 & 26.18 & 27.10 & 29.66
\\
mip-NeRF 360~\cite{barron2022mipnerf360}&\cellcolor{orange}  24.21 &\cellcolor{orange} 21.60 &\cellcolor{orange} 25.82 &\cellcolor{orange} 25.59 &\cellcolor{orange} 22.78 & 22.95 &\cellcolor{orange} 27.72 &\cellcolor{orange} 28.78 &\cellcolor{tablered} 31.63   
\\
zip-NeRF~\cite{Barron2023ICCV}&23.05 & 20.05 & 18.07 & 23.94 &\cellcolor{yellow} 22.53 & 20.51 & 26.08 &\cellcolor{yellow} 27.37 &\cellcolor{yellow} 30.05
\\
3DGS~\cite{kerbl3Dgaussians}& 21.34 & 19.43 & 21.94 & 22.63 & 20.91 & 28.10 & 25.33 & 23.68 & 25.89
\\
\hline
3DGS~\cite{kerbl3Dgaussians} + EWA~\cite{zwicker2001ewa}&\cellcolor{yellow}23.74 &\cellcolor{yellow} 20.94 &\cellcolor{yellow} 24.69 &\cellcolor{yellow} 24.81 & 21.93 &\cellcolor{orange} 29.80 &\cellcolor{yellow} 27.23 & 27.07 & 28.63
\\
Mip-Splatting (ours)&\cellcolor{tablered}25.26 &\cellcolor{tablered} 22.02 &\cellcolor{tablered} 26.78 &\cellcolor{tablered} 26.65 &\cellcolor{tablered} 22.92 &\cellcolor{tablered} 31.56 &\cellcolor{tablered} 28.87 &\cellcolor{tablered} 30.73 &\cellcolor{orange} 31.49
\\

\multicolumn{9}{c}{} \\
 & \multicolumn{9}{c}{\textbf{SSIM}} \\
 & \scenename{bicycle}  & \scenename{flowers}  & \scenename{garden}  & \scenename{stump}  & \scenename{treehill}  & \scenename{room}  & \scenename{counter}  & \scenename{kitchen} & \scenename{bonsai} 
\\ 
\hline 
Instant-NGP~\cite{muller2022instant}& 0.538 & 0.473 & 0.647 & 0.590 & 0.544 & 0.868 & 0.795 & 0.764 &\cellcolor{yellow} 0.877  
\\
mip-NeRF 360~\cite{barron2022mipnerf360}&\cellcolor{yellow}  0.662 &\cellcolor{orange} 0.567 &\cellcolor{yellow} 0.716 &\cellcolor{orange} 0.715 &\cellcolor{orange} 0.628 & 0.795 &\cellcolor{orange} 0.845 &\cellcolor{orange} 0.828 &\cellcolor{orange} 0.910   
\\
zip-NeRF~\cite{Barron2023ICCV}& 0.640 & 0.521 & 0.548 & 0.661 & 0.590 & 0.655 & 0.784 & 0.800 & 0.865
\\
3DGS~\cite{kerbl3Dgaussians}& 0.638 & 0.536 & 0.675 & 0.662 & 0.591 &\cellcolor{yellow} 0.878 & 0.826 & 0.789 & 0.838
\\
\hline
3DGS~\cite{kerbl3Dgaussians} + EWA~\cite{zwicker2001ewa}&\cellcolor{orange} 0.671 &\cellcolor{yellow} 0.563 &\cellcolor{orange} 0.718 &\cellcolor{yellow} 0.693 &\cellcolor{yellow} 0.608 &\cellcolor{orange} 0.889 &\cellcolor{yellow} 0.843 &\cellcolor{yellow} 0.813 & 0.874
\\
Mip-Splatting (ours)&\cellcolor{tablered}0.738 &\cellcolor{tablered} 0.613 &\cellcolor{tablered} 0.786 &\cellcolor{tablered} 0.776 &\cellcolor{tablered} 0.659 &\cellcolor{tablered} 0.921 &\cellcolor{tablered} 0.897 &\cellcolor{tablered} 0.903 &\cellcolor{tablered} 0.933
\\

\multicolumn{9}{c}{} \\
 & \multicolumn{9}{c}{\textbf{LPIPS}} \\
 & \scenename{bicycle}  & \scenename{flowers}  & \scenename{garden}  & \scenename{stump}  & \scenename{treehill}  & \scenename{room}  & \scenename{counter}  & \scenename{kitchen} & \scenename{bonsai} 
\\ 
\hline 
Instant-NGP~\cite{muller2022instant}& 0.500 & 0.486 & 0.372 & 0.469 & 0.511 & 0.270 & 0.310 & 0.286 & 0.229 
\\
mip-NeRF 360~\cite{barron2022mipnerf360}& 0.358 & 0.400 & 0.296 &\cellcolor{orange} 0.333 &\cellcolor{yellow} 0.391 & 0.256 &\cellcolor{orange} 0.228 &\cellcolor{orange} 0.210 &\cellcolor{orange} 0.182  
\\
zip-NeRF~\cite{Barron2023ICCV}& 0.353 &\cellcolor{yellow} 0.397 & 0.346 & 0.349 &\cellcolor{tablered} 0.366 & 0.302 & 0.277 & 0.232 & 0.236
\\
3DGS~\cite{kerbl3Dgaussians}&\cellcolor{yellow} 0.336 & 0.406 &\cellcolor{yellow} 0.295 & 0.353 & 0.406 &\cellcolor{yellow} 0.223 & 0.239 & 0.245 & 0.242
\\
\hline
3DGS~\cite{kerbl3Dgaussians} + EWA~\cite{zwicker2001ewa}&\cellcolor{orange}0.322 &\cellcolor{orange} 0.395 &\cellcolor{orange} 0.281 &\cellcolor{yellow} 0.334 & 0.405 &\cellcolor{orange} 0.217 &\cellcolor{yellow} 0.231 &\cellcolor{yellow} 0.216 &\cellcolor{yellow} 0.227
\\
Mip-Splatting (ours)&\cellcolor{tablered}0.281 &\cellcolor{tablered} 0.373 &\cellcolor{tablered} 0.233 &\cellcolor{tablered} 0.281 &\cellcolor{orange} 0.369 &\cellcolor{tablered} 0.193 &\cellcolor{tablered} 0.199 &\cellcolor{tablered} 0.165 &\cellcolor{tablered} 0.176
\\

\multicolumn{9}{c}{}

    \end{tabular}
    \caption{\textbf{Single-scale Training and Multi-scale Testing on the Mip-NeRF 360 dataset~\cite{barron2022mipnerf360}}. All models are trained on images downsampled by a factor of 8 and rendered at higher resolutions to simulates zoom-in effects.
    }
    \label{tab:multi_360_perscene}
\end{table*}

\newcommand{\sixwidth}{0.16\textwidth}

\begin{figure*}[t]
    \centering
    \setlength{\tabcolsep}{0.1em}
    \renewcommand{\arraystretch}{0.4}
    \hfill{}\hspace*{-0.5em}
    \scriptsize
    \begin{tabular}{ccccccc}
    \multirow{1}{*}[3ex]{\rotatebox{90}{Full}} &
    \includegraphics[width=\sixwidth]{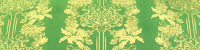}&
    \includegraphics[width=\sixwidth]{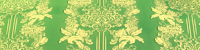}&
    \includegraphics[width=\sixwidth]{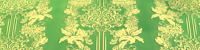}&
    \includegraphics[width=\sixwidth]{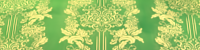}&
    \includegraphics[width=\sixwidth]{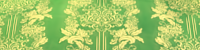}&
    \includegraphics[width=\sixwidth]{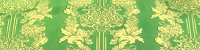}
    \\
    \multirow{1}{*}[4ex]{\rotatebox{90}{$\nicefrac{1}{2}$}} &
    \includegraphics[width=\sixwidth]{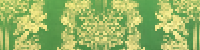}&
    \includegraphics[width=\sixwidth]{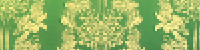}&
    \includegraphics[width=\sixwidth]{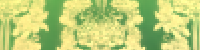}&
    \includegraphics[width=\sixwidth]{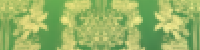}&
    \includegraphics[width=\sixwidth]{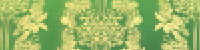}&
    \includegraphics[width=\sixwidth]{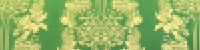}
    \\
    \multirow{1}{*}[4ex]{\rotatebox{90}{$\nicefrac{1}{4}$ }} &
    \includegraphics[width=\sixwidth]{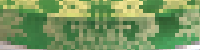}&
    \includegraphics[width=\sixwidth]{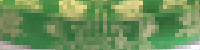}&
    \includegraphics[width=\sixwidth]{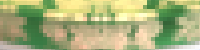}&
    \includegraphics[width=\sixwidth]{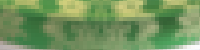}&
    \includegraphics[width=\sixwidth]{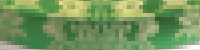}&
    \includegraphics[width=\sixwidth]{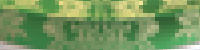}
    \\
    \multirow{1}{*}[4ex]{\rotatebox{90}{$\nicefrac{1}{8}$ }} &
    \includegraphics[width=\sixwidth]{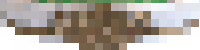}&
    \includegraphics[width=\sixwidth]{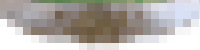}&
    \includegraphics[width=\sixwidth]{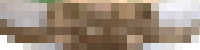}&
    \includegraphics[width=\sixwidth]{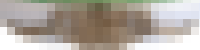}&
    \includegraphics[width=\sixwidth]{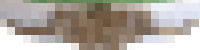}&
    \includegraphics[width=\sixwidth]{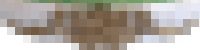}
    \\
    \\
    \multirow{1}{*}[3ex]{\rotatebox{90}{Full}} &
    \includegraphics[width=\sixwidth]{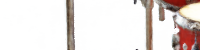}&
    \includegraphics[width=\sixwidth]{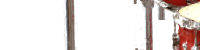}&
    \includegraphics[width=\sixwidth]{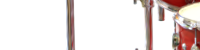}&
    \includegraphics[width=\sixwidth]{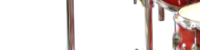}&
    \includegraphics[width=\sixwidth]{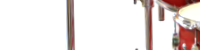}&
    \includegraphics[width=\sixwidth]{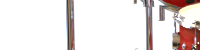}
    \\
    \multirow{1}{*}[4ex]{\rotatebox{90}{$\nicefrac{1}{2}$}} &
    \includegraphics[width=\sixwidth]{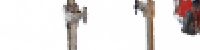}&
    \includegraphics[width=\sixwidth]{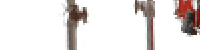}&
    \includegraphics[width=\sixwidth]{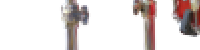}&
    \includegraphics[width=\sixwidth]{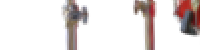}&
    \includegraphics[width=\sixwidth]{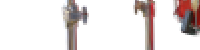}&
    \includegraphics[width=\sixwidth]{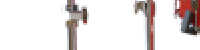}
    \\
    \multirow{1}{*}[4ex]{\rotatebox{90}{$\nicefrac{1}{4}$ }} &
    \includegraphics[width=\sixwidth]{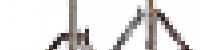}&
    \includegraphics[width=\sixwidth]{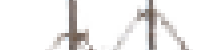}&
    \includegraphics[width=\sixwidth]{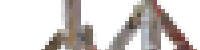}&
    \includegraphics[width=\sixwidth]{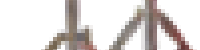}&
    \includegraphics[width=\sixwidth]{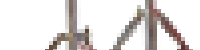}&
    \includegraphics[width=\sixwidth]{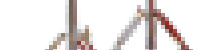}
    \\
    \multirow{1}{*}[4ex]{\rotatebox{90}{$\nicefrac{1}{8}$ }} &
    \includegraphics[width=\sixwidth]{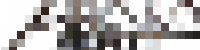}&
    \includegraphics[width=\sixwidth]{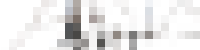}&
    \includegraphics[width=\sixwidth]{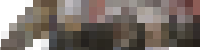}&
    \includegraphics[width=\sixwidth]{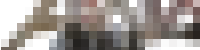}&
    \includegraphics[width=\sixwidth]{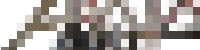}&
    \includegraphics[width=\sixwidth]{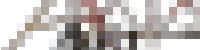}
    \\
    \\
    \multirow{1}{*}[3ex]{\rotatebox{90}{Full}} &
    \includegraphics[width=\sixwidth]{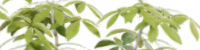}&
    \includegraphics[width=\sixwidth]{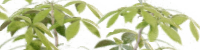}&
    \includegraphics[width=\sixwidth]{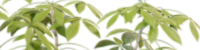}&
    \includegraphics[width=\sixwidth]{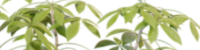}&
    \includegraphics[width=\sixwidth]{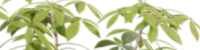}&
    \includegraphics[width=\sixwidth]{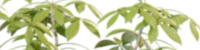}
    \\
    \multirow{1}{*}[4ex]{\rotatebox{90}{$\nicefrac{1}{2}$}} &
    \includegraphics[width=\sixwidth]{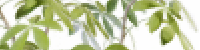}&
    \includegraphics[width=\sixwidth]{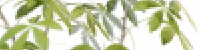}&
    \includegraphics[width=\sixwidth]{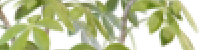}&
    \includegraphics[width=\sixwidth]{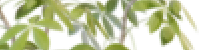}&
    \includegraphics[width=\sixwidth]{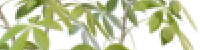}&
    \includegraphics[width=\sixwidth]{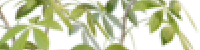}
    \\
    \multirow{1}{*}[4ex]{\rotatebox{90}{$\nicefrac{1}{4}$ }} &
    \includegraphics[width=\sixwidth]{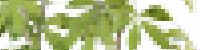}&
    \includegraphics[width=\sixwidth]{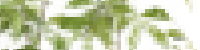}&
    \includegraphics[width=\sixwidth]{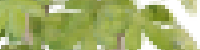}&
    \includegraphics[width=\sixwidth]{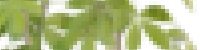}&
    \includegraphics[width=\sixwidth]{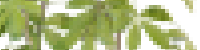}&
    \includegraphics[width=\sixwidth]{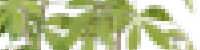}
    \\
    \multirow{1}{*}[4ex]{\rotatebox{90}{$\nicefrac{1}{8}$ }} &
    \includegraphics[width=\sixwidth]{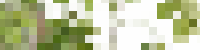}&
    \includegraphics[width=\sixwidth]{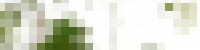}&
    \includegraphics[width=\sixwidth]{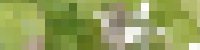}&
    \includegraphics[width=\sixwidth]{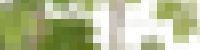}&
    \includegraphics[width=\sixwidth]{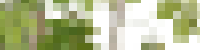}&
    \includegraphics[width=\sixwidth]{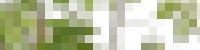}
    \\
    \\
    \multirow{1}{*}[3ex]{\rotatebox{90}{Full}} &
    \includegraphics[width=\sixwidth]{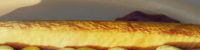}&
    \includegraphics[width=\sixwidth]{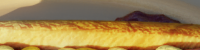}&
    \includegraphics[width=\sixwidth]{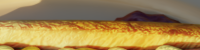}&
    \includegraphics[width=\sixwidth]{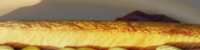}&
    \includegraphics[width=\sixwidth]{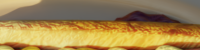}&
    \includegraphics[width=\sixwidth]{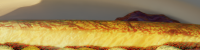}
    \\
    \multirow{1}{*}[4ex]{\rotatebox{90}{$\nicefrac{1}{2}$}} &
    \includegraphics[width=\sixwidth]{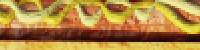}&
    \includegraphics[width=\sixwidth]{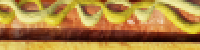}&
    \includegraphics[width=\sixwidth]{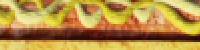}&
    \includegraphics[width=\sixwidth]{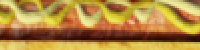}&
    \includegraphics[width=\sixwidth]{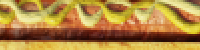}&
    \includegraphics[width=\sixwidth]{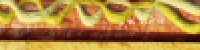}
    \\
    \multirow{1}{*}[4ex]{\rotatebox{90}{$\nicefrac{1}{4}$ }} &
    \includegraphics[width=\sixwidth]{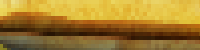}&
    \includegraphics[width=\sixwidth]{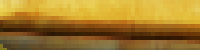}&
    \includegraphics[width=\sixwidth]{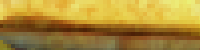}&
    \includegraphics[width=\sixwidth]{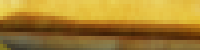}&
    \includegraphics[width=\sixwidth]{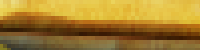}&
    \includegraphics[width=\sixwidth]{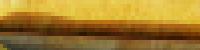}
    \\
    \multirow{1}{*}[4ex]{\rotatebox{90}{$\nicefrac{1}{8}$ }} &
    \includegraphics[width=\sixwidth]{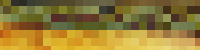}&
    \includegraphics[width=\sixwidth]{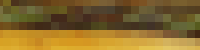}&
    \includegraphics[width=\sixwidth]{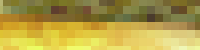}&
    \includegraphics[width=\sixwidth]{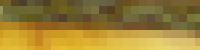}&
    \includegraphics[width=\sixwidth]{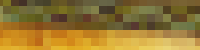}&
    \includegraphics[width=\sixwidth]{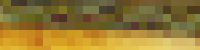}
    \\
    \\
    \multirow{1}{*}[3ex]{\rotatebox{90}{Full}} &
    \includegraphics[width=\sixwidth]{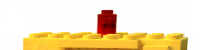}&
    \includegraphics[width=\sixwidth]{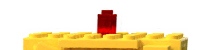}&
    \includegraphics[width=\sixwidth]{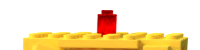}&
    \includegraphics[width=\sixwidth]{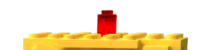}&
    \includegraphics[width=\sixwidth]{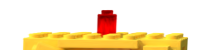}&
    \includegraphics[width=\sixwidth]{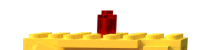}
    \\
    \multirow{1}{*}[4ex]{\rotatebox{90}{$\nicefrac{1}{2}$}} &
    \includegraphics[width=\sixwidth]{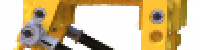}&
    \includegraphics[width=\sixwidth]{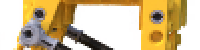}&
    \includegraphics[width=\sixwidth]{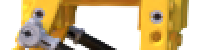}&
    \includegraphics[width=\sixwidth]{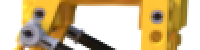}&
    \includegraphics[width=\sixwidth]{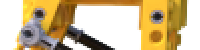}&
    \includegraphics[width=\sixwidth]{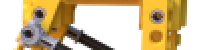}
    \\
    \multirow{1}{*}[4ex]{\rotatebox{90}{$\nicefrac{1}{4}$ }} &
    \includegraphics[width=\sixwidth]{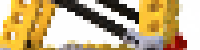}&
    \includegraphics[width=\sixwidth]{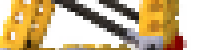}&
    \includegraphics[width=\sixwidth]{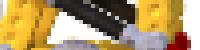}&
    \includegraphics[width=\sixwidth]{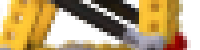}&
    \includegraphics[width=\sixwidth]{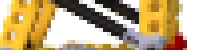}&
    \includegraphics[width=\sixwidth]{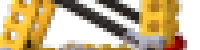}
    \\
    \multirow{1}{*}[4ex]{\rotatebox{90}{$\nicefrac{1}{8}$ }} &
    \includegraphics[width=\sixwidth]{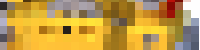}&
    \includegraphics[width=\sixwidth]{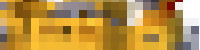}&
    \includegraphics[width=\sixwidth]{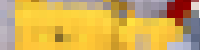}&
    \includegraphics[width=\sixwidth]{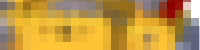}&
    \includegraphics[width=\sixwidth]{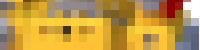}&
    \includegraphics[width=\sixwidth]{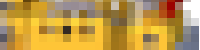}
    \\
    \\
    \multirow{1}{*}[3ex]{\rotatebox{90}{Full}} &
    \includegraphics[width=\sixwidth]{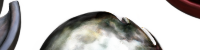}&
    \includegraphics[width=\sixwidth]{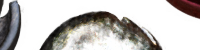}&
    \includegraphics[width=\sixwidth]{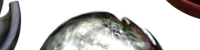}&
    \includegraphics[width=\sixwidth]{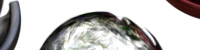}&
    \includegraphics[width=\sixwidth]{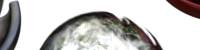}&
    \includegraphics[width=\sixwidth]{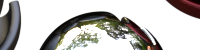}
    \\
    \multirow{1}{*}[4ex]{\rotatebox{90}{$\nicefrac{1}{2}$}} &
    \includegraphics[width=\sixwidth]{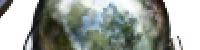}&
    \includegraphics[width=\sixwidth]{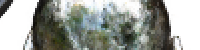}&
    \includegraphics[width=\sixwidth]{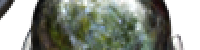}&
    \includegraphics[width=\sixwidth]{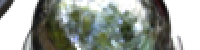}&
    \includegraphics[width=\sixwidth]{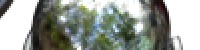}&
    \includegraphics[width=\sixwidth]{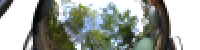}
    \\
    \multirow{1}{*}[4ex]{\rotatebox{90}{$\nicefrac{1}{4}$ }} &
    \includegraphics[width=\sixwidth]{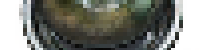}&
    \includegraphics[width=\sixwidth]{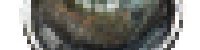}&
    \includegraphics[width=\sixwidth]{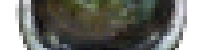}&
    \includegraphics[width=\sixwidth]{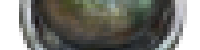}&
    \includegraphics[width=\sixwidth]{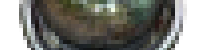}&
    \includegraphics[width=\sixwidth]{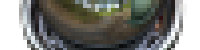}
    \\
    \multirow{1}{*}[4ex]{\rotatebox{90}{$\nicefrac{1}{8}$ }} &
    \includegraphics[width=\sixwidth]{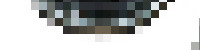}&
    \includegraphics[width=\sixwidth]{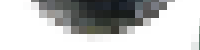}&
    \includegraphics[width=\sixwidth]{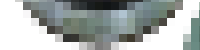}&
    \includegraphics[width=\sixwidth]{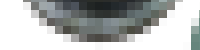}&
    \includegraphics[width=\sixwidth]{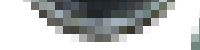}&
    \includegraphics[width=\sixwidth]{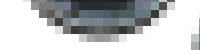}
    \\
    \\
    \multirow{1}{*}[3ex]{\rotatebox{90}{Full}} &
    \includegraphics[width=\sixwidth]{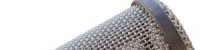}&
    \includegraphics[width=\sixwidth]{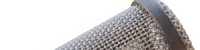}&
    \includegraphics[width=\sixwidth]{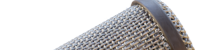}&
    \includegraphics[width=\sixwidth]{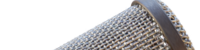}&
    \includegraphics[width=\sixwidth]{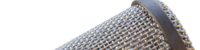}&
    \includegraphics[width=\sixwidth]{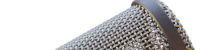}
    \\
    \multirow{1}{*}[4ex]{\rotatebox{90}{$\nicefrac{1}{2}$}} &
    \includegraphics[width=\sixwidth]{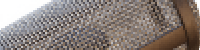}&
    \includegraphics[width=\sixwidth]{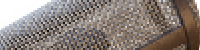}&
    \includegraphics[width=\sixwidth]{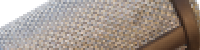}&
    \includegraphics[width=\sixwidth]{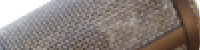}&
    \includegraphics[width=\sixwidth]{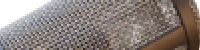}&
    \includegraphics[width=\sixwidth]{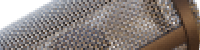}
    \\
    \multirow{1}{*}[4ex]{\rotatebox{90}{$\nicefrac{1}{4}$ }} &
    \includegraphics[width=\sixwidth]{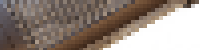}&
    \includegraphics[width=\sixwidth]{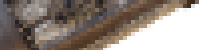}&
    \includegraphics[width=\sixwidth]{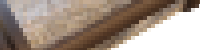}&
    \includegraphics[width=\sixwidth]{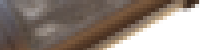}&
    \includegraphics[width=\sixwidth]{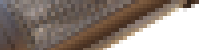}&
    \includegraphics[width=\sixwidth]{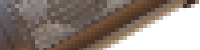}
    \\
    \multirow{1}{*}[4ex]{\rotatebox{90}{$\nicefrac{1}{8}$ }} &
    \includegraphics[width=\sixwidth]{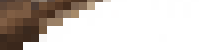}&
    \includegraphics[width=\sixwidth]{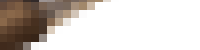}&
    \includegraphics[width=\sixwidth]{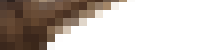}&
    \includegraphics[width=\sixwidth]{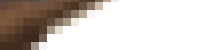}&
    \includegraphics[width=\sixwidth]{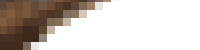}&
    \includegraphics[width=\sixwidth]{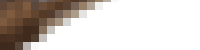}
    \\
    {}& Mip-NeRF~\cite{Barron2021ICCV} & Tri-MipRF~\cite{Hu2023ICCV} & 3DGS~\cite{kerbl3Dgaussians} & 3DGS~\cite{kerbl3Dgaussians} + EWA~\cite{zwicker2001ewa} & Mip-Splatting (ours) & GT  
    \end{tabular}
    \vspace{-0.1in}
    \caption{
    \textbf{Single-scale Training and Multi-scale Testing on the Blender Dataset~\cite{mildenhall2020nerf}.}
    All methods are trained at full resolution and evaluated at different (smaller) resolutions to mimic zoom-out. Methods based on 3DGS capture fine details better than Mip-NeRF~\cite{Barron2021ICCV} and Tri-MipRF~\cite{Hu2023ICCV} at training resolution. Mip-Splatting surpasses both 3DGS~\cite{kerbl3Dgaussians} and 3DGS + EWA~\cite{zwicker2001ewa} at lower resolutions.
    }
    \label{fig:blender_single_train_multi_test_supp}
    \vspace{-0.1in}
\end{figure*}

\newcommand{\stmtwidth}{0.16\textwidth}

\begin{figure*}[t]
    \centering
    \setlength{\tabcolsep}{0.1em}
    \renewcommand{\arraystretch}{0.4}
    \scriptsize
    \begin{tabular}{cccccc}
    \includegraphics[width=\stmtwidth]{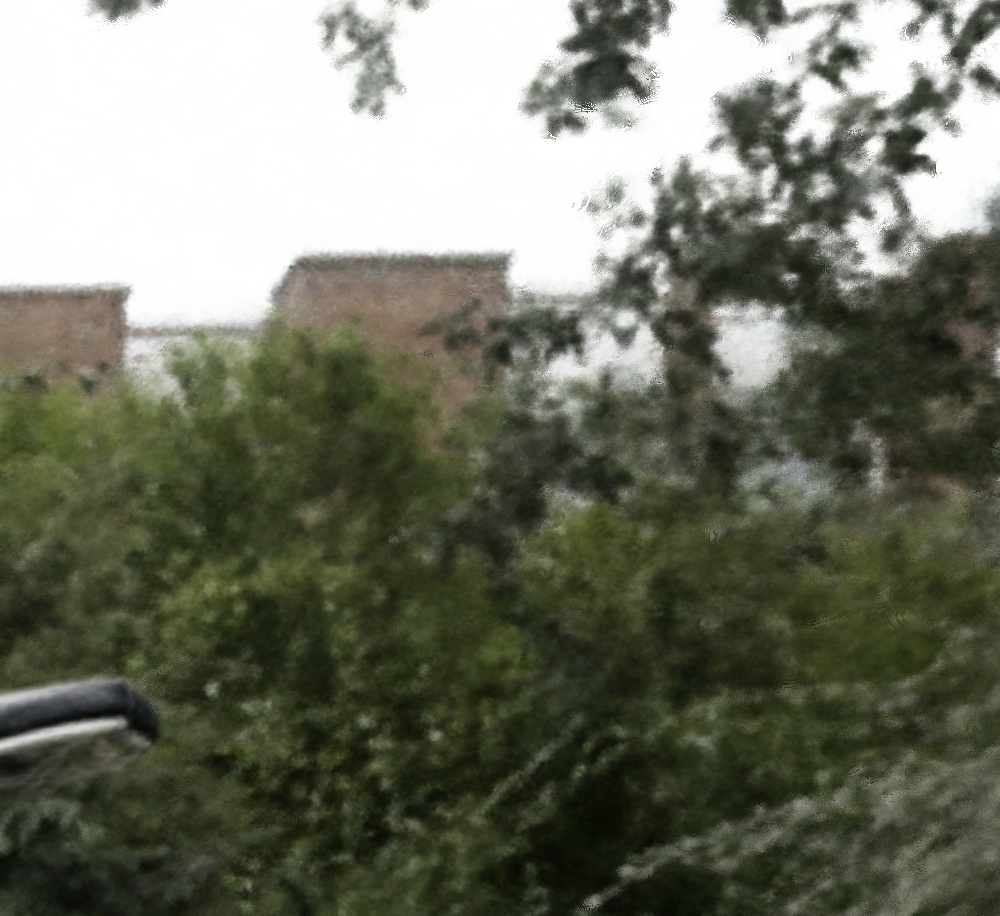}&    
    \includegraphics[width=\stmtwidth]{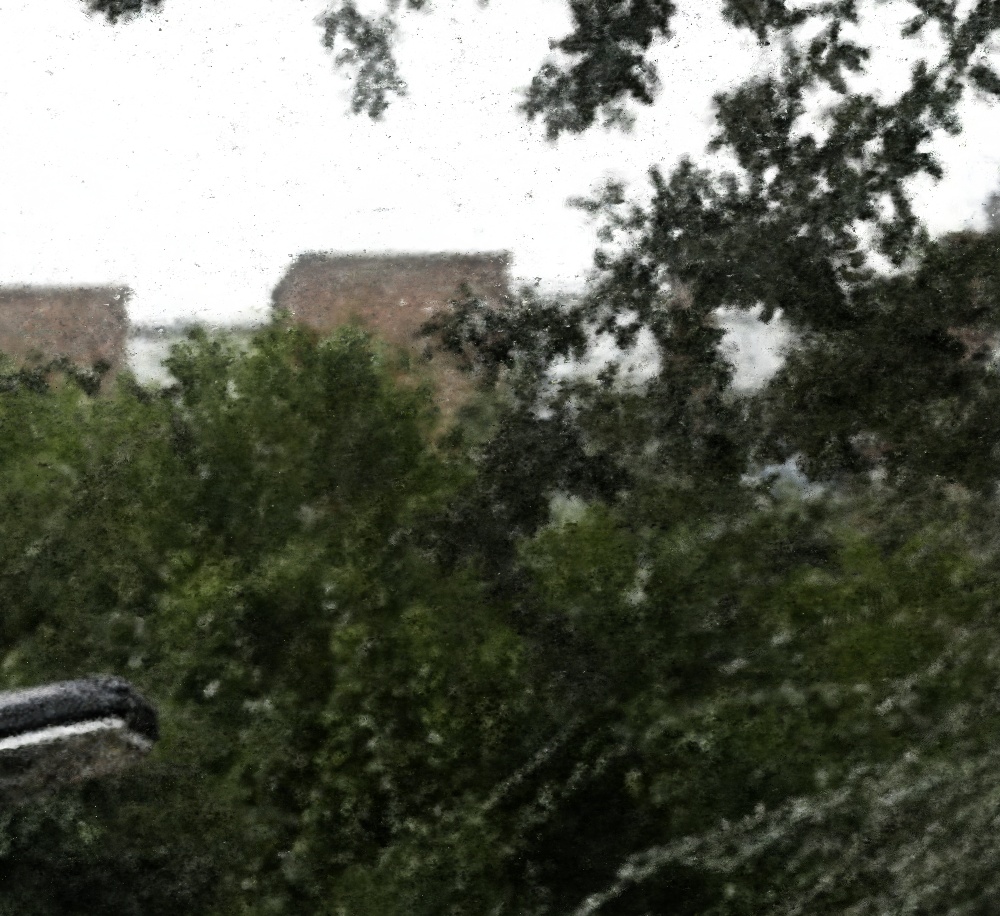}&    
    \includegraphics[width=\stmtwidth]{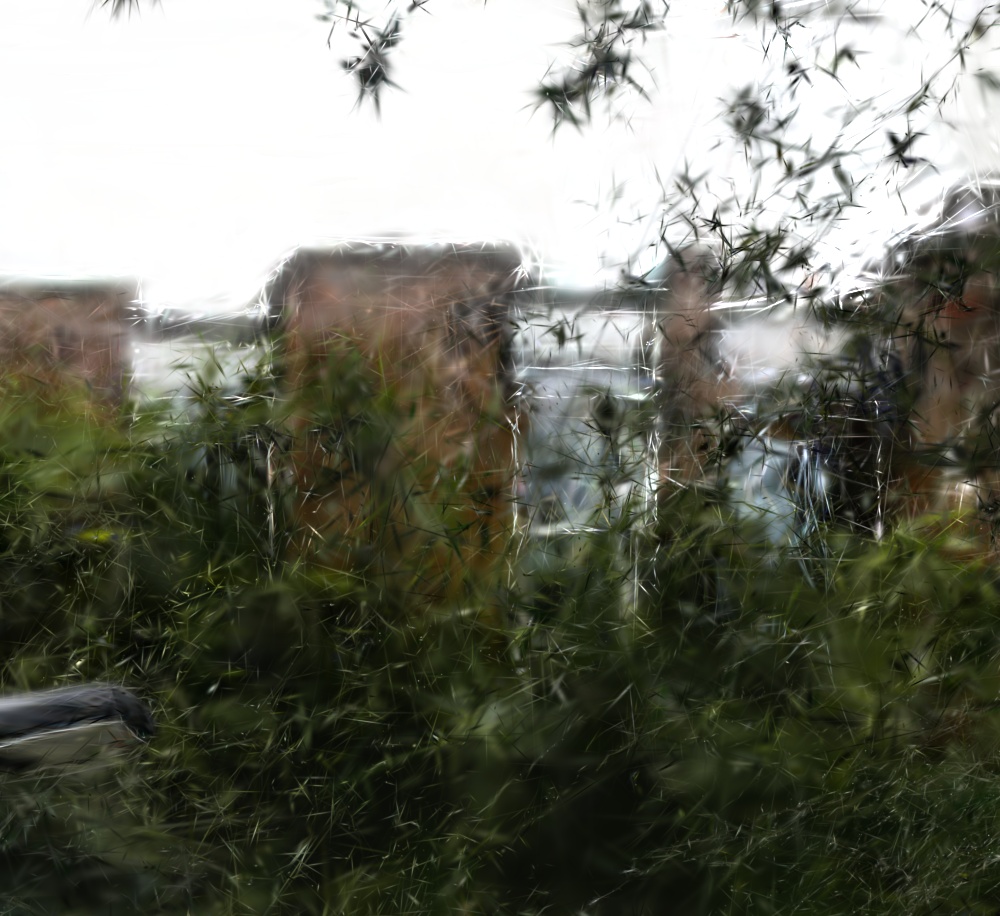}&    
    \includegraphics[width=\stmtwidth]{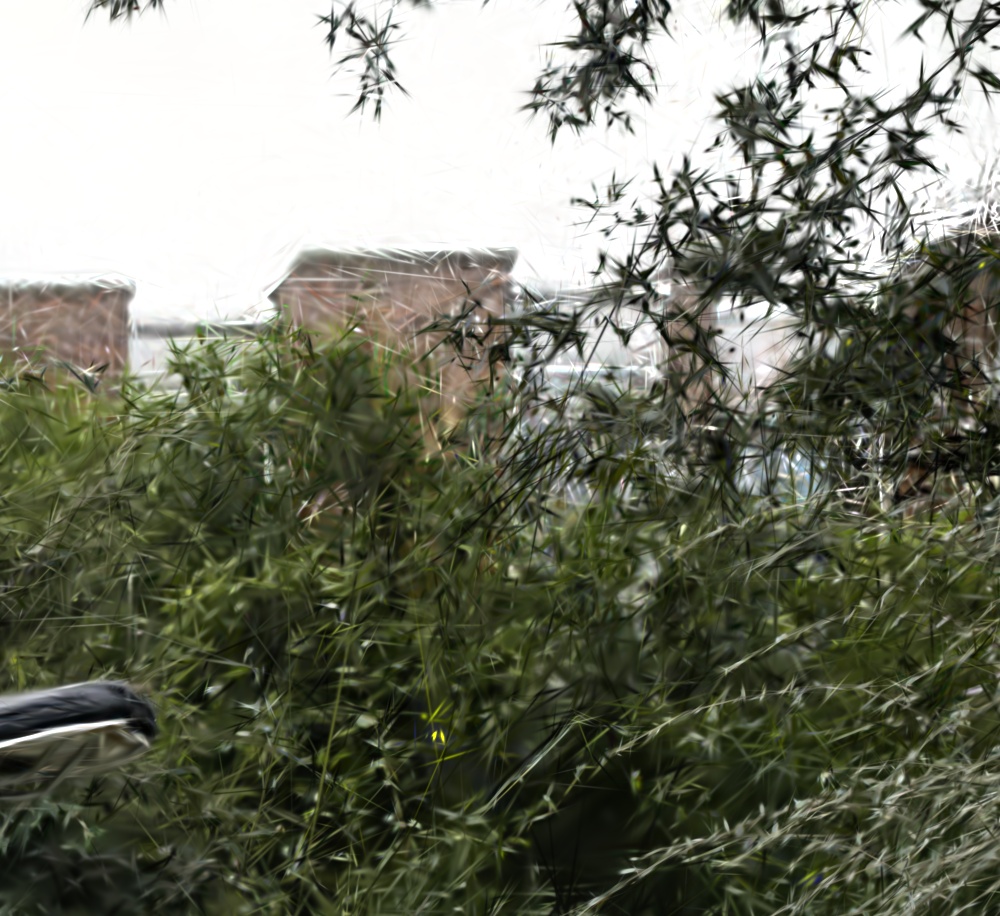}&    
    \includegraphics[width=\stmtwidth]{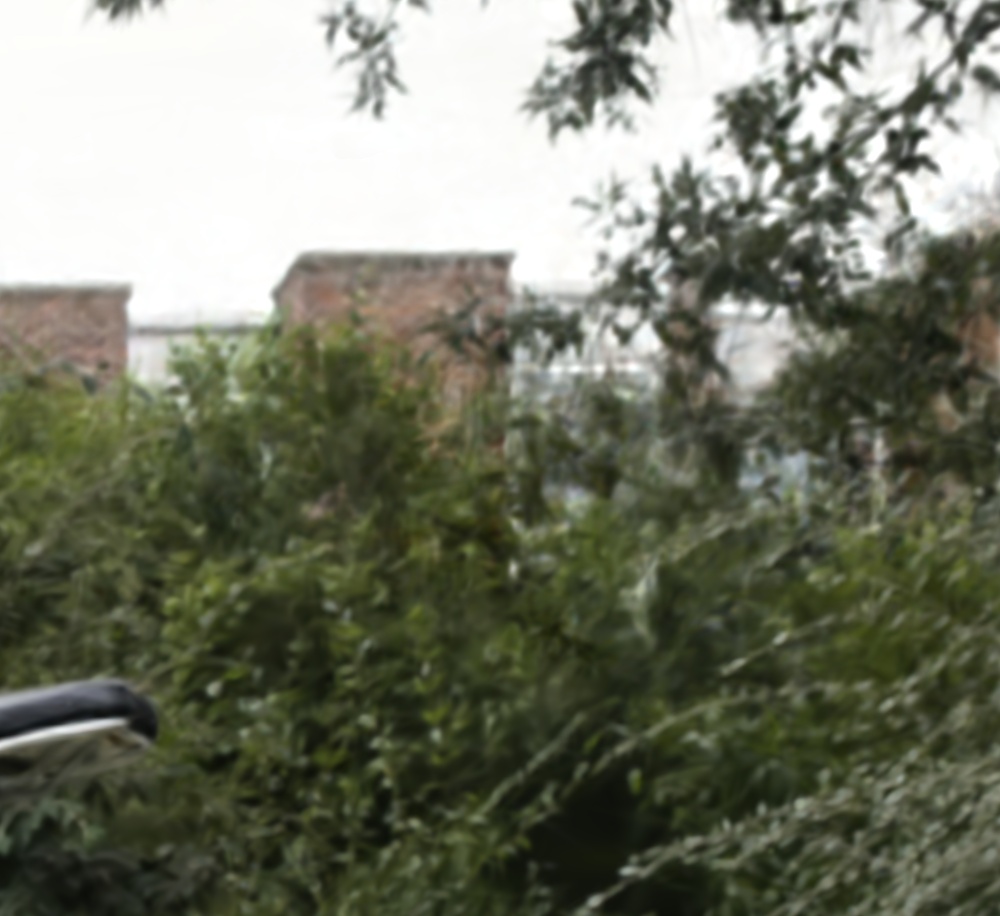}&    
    \includegraphics[width=\stmtwidth]{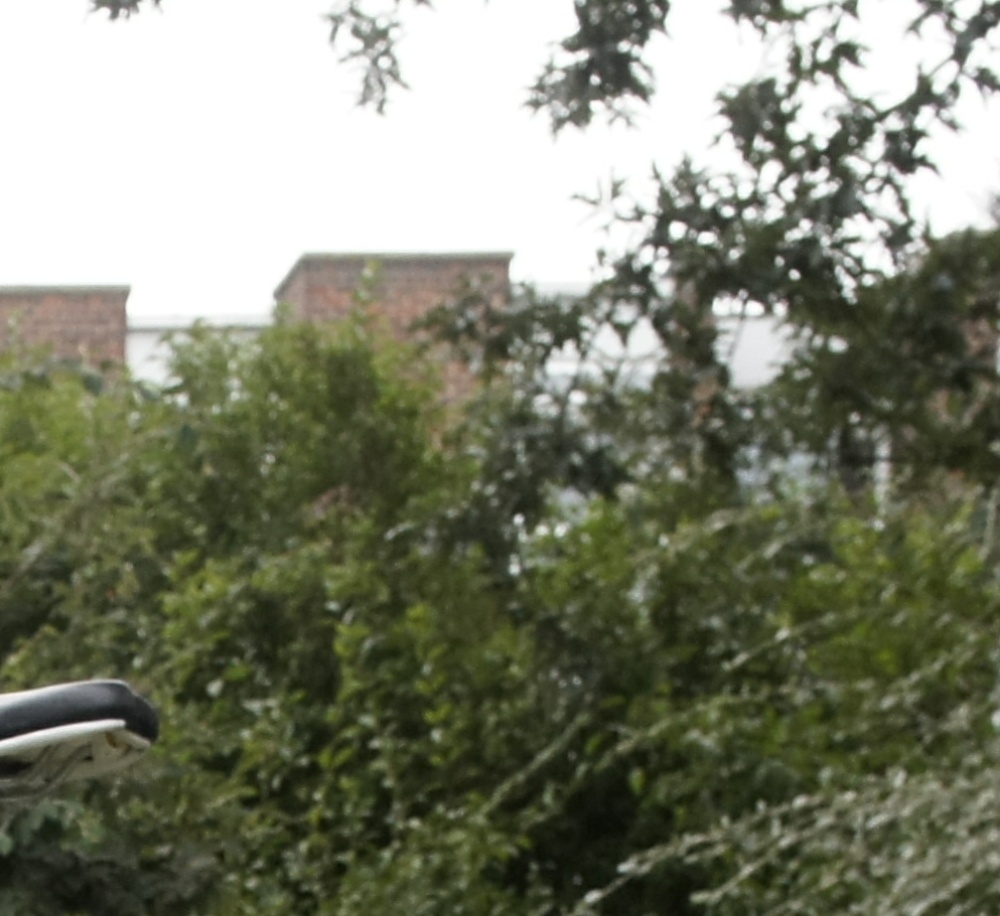}  
    \\
    \includegraphics[width=\stmtwidth]{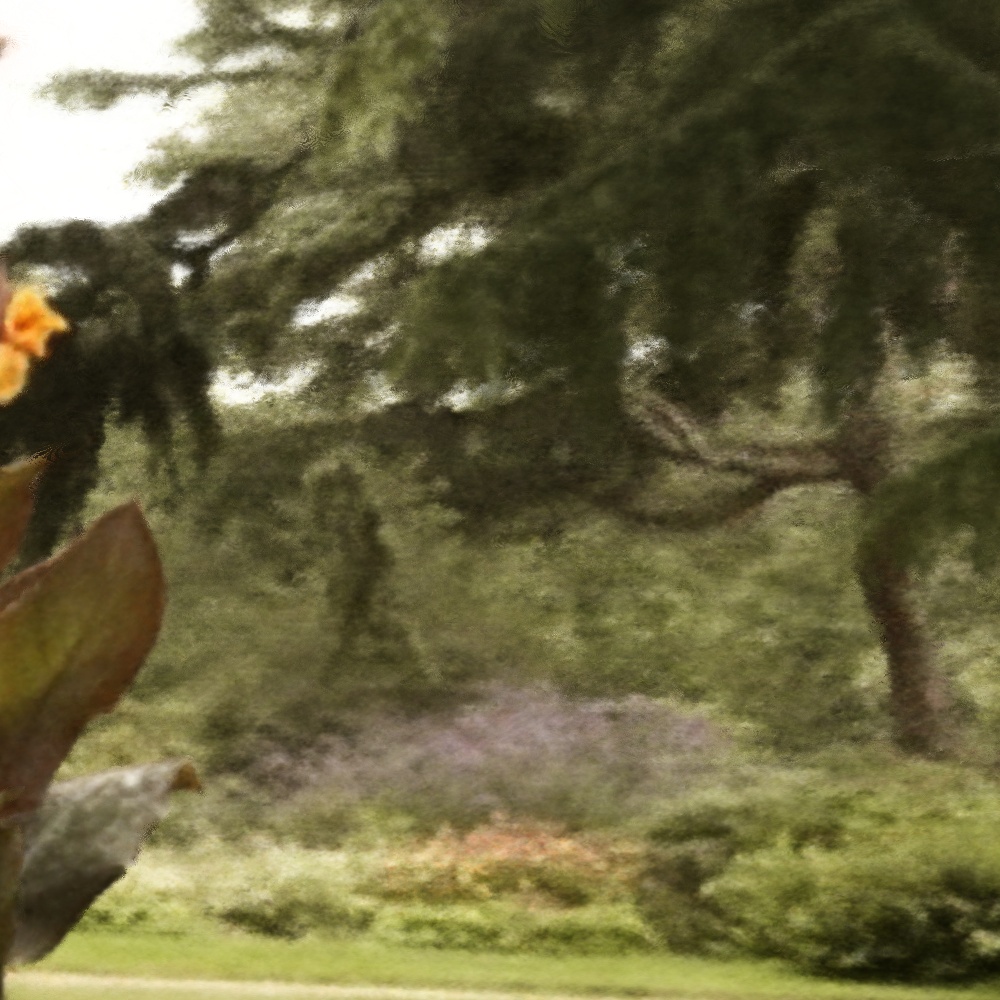}&    
    \includegraphics[width=\stmtwidth]{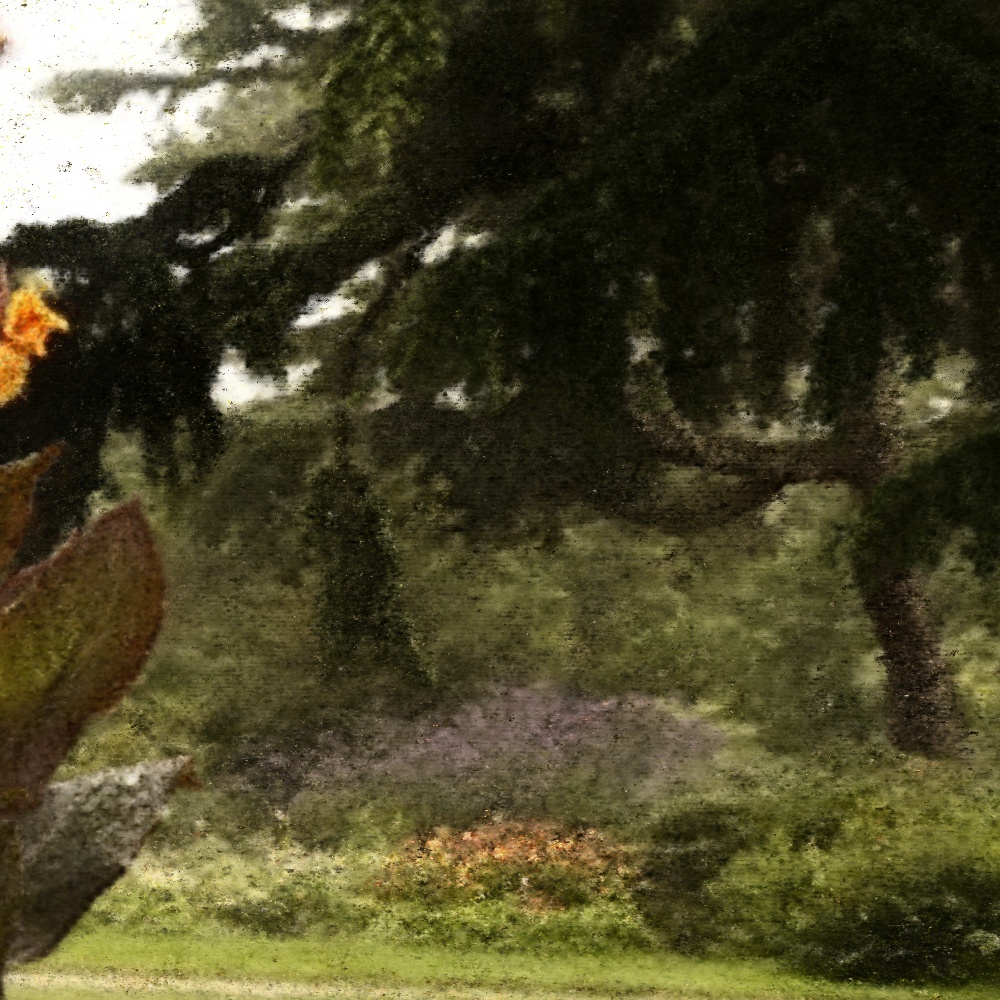}&    
    \includegraphics[width=\stmtwidth]{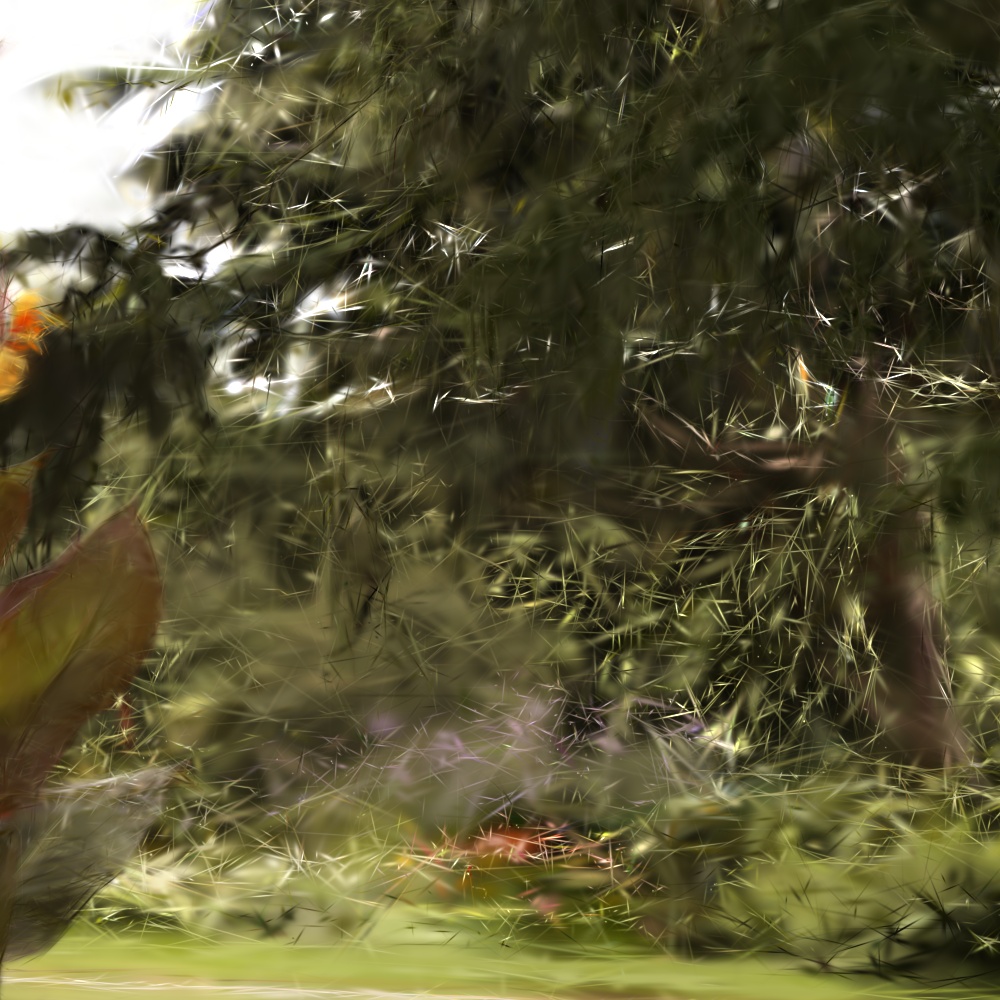}&    
    \includegraphics[width=\stmtwidth]{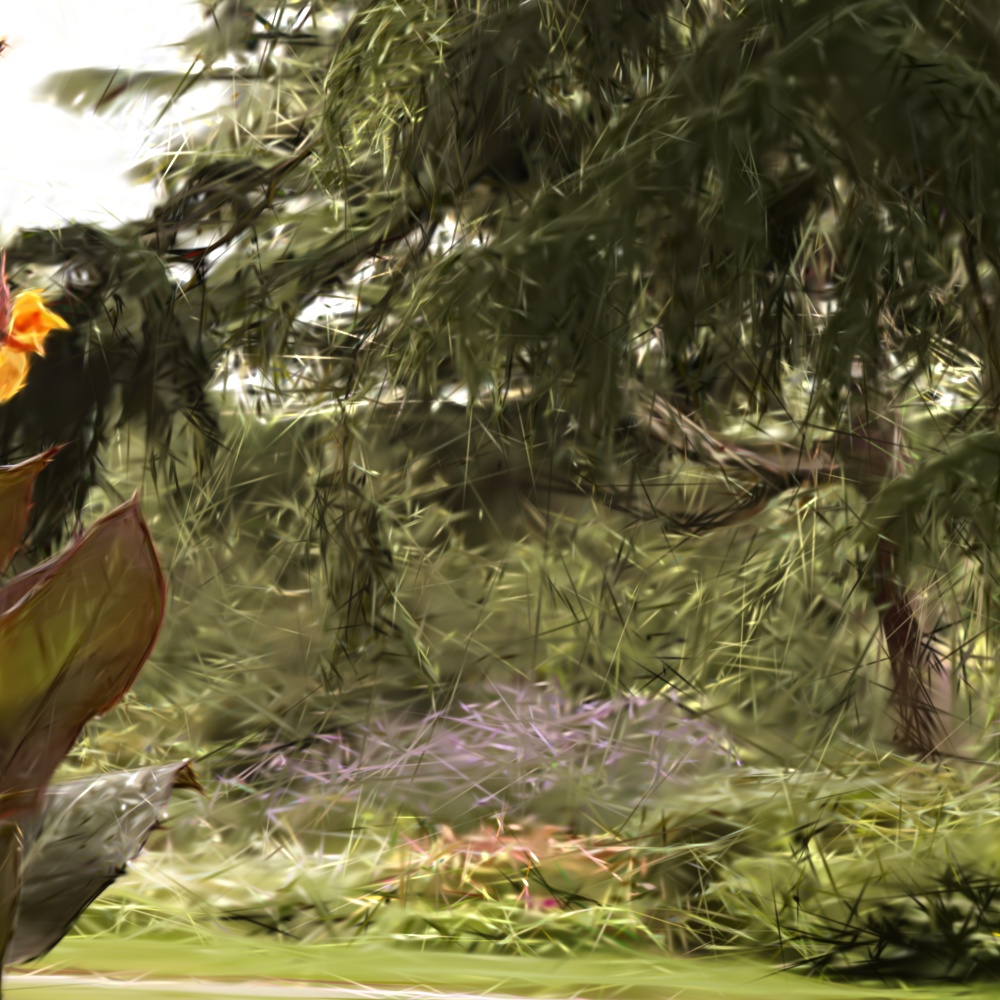}&    
    \includegraphics[width=\stmtwidth]{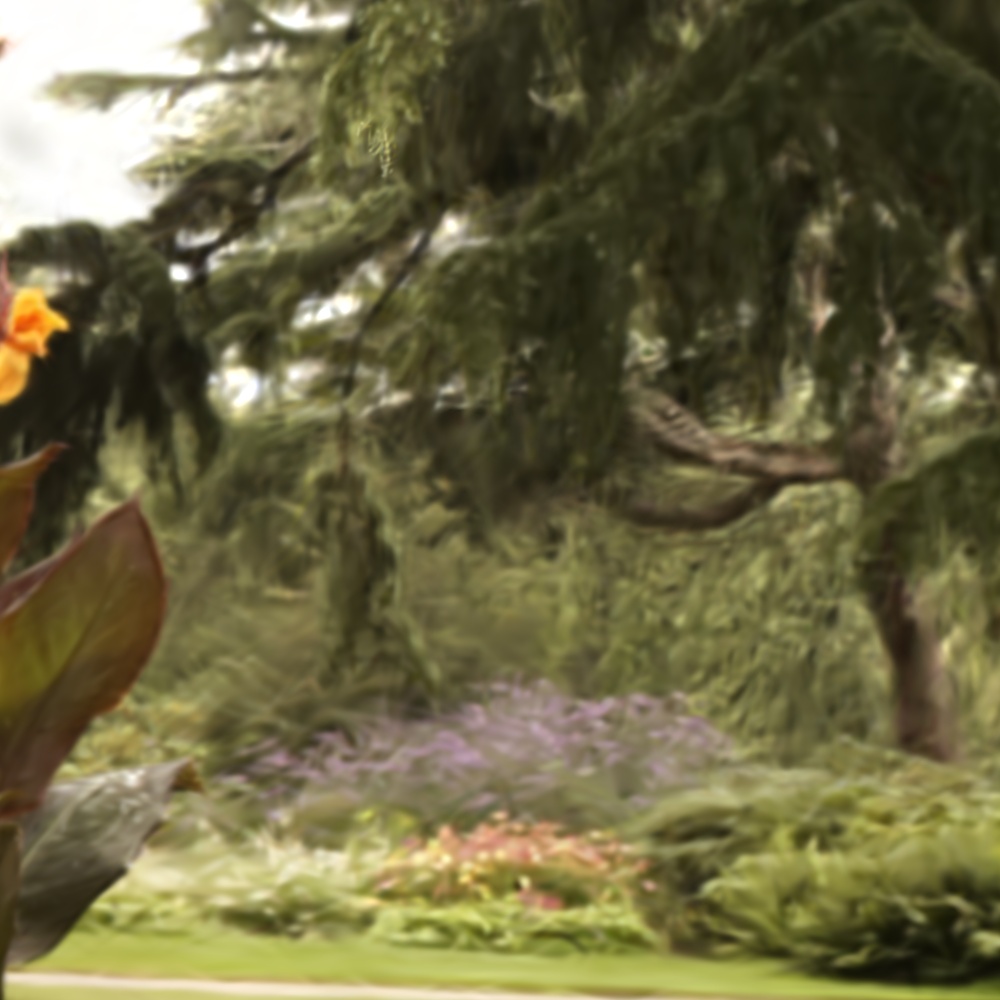}&    
    \includegraphics[width=\stmtwidth]{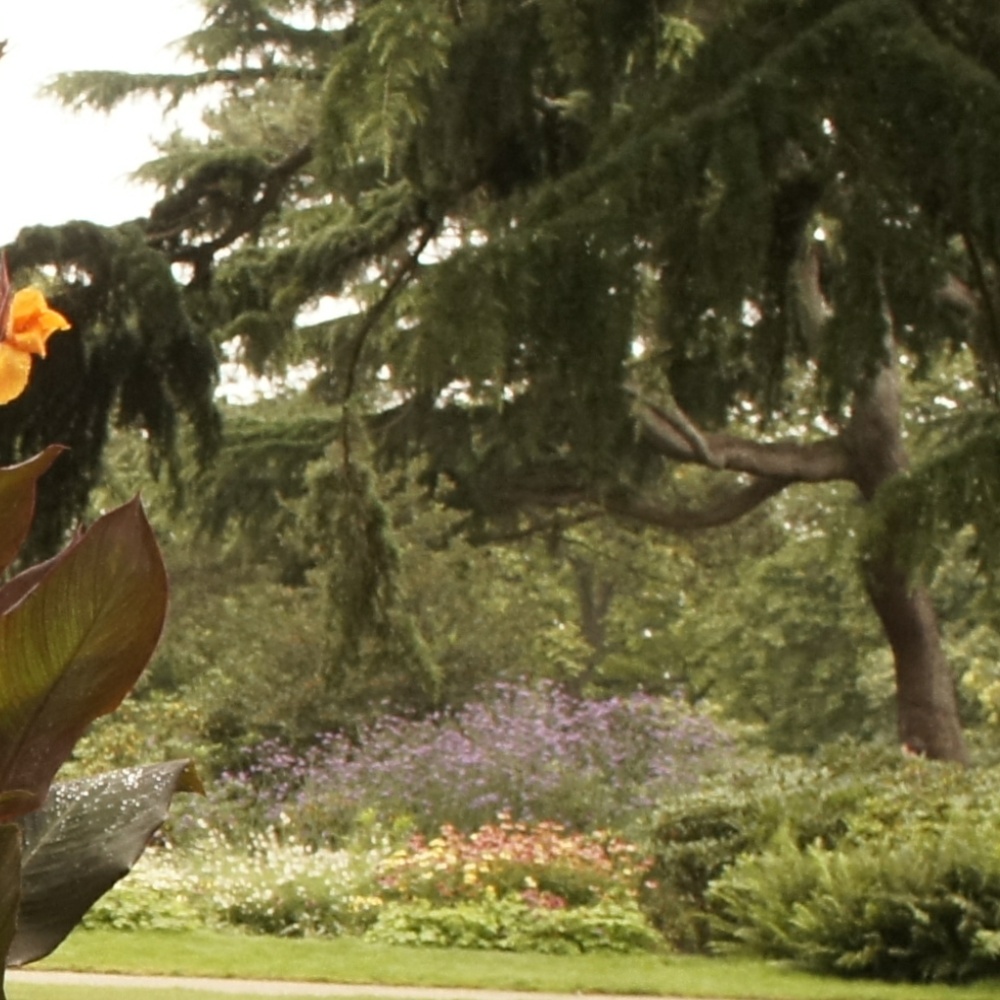}  
    \\
    \includegraphics[width=\stmtwidth]{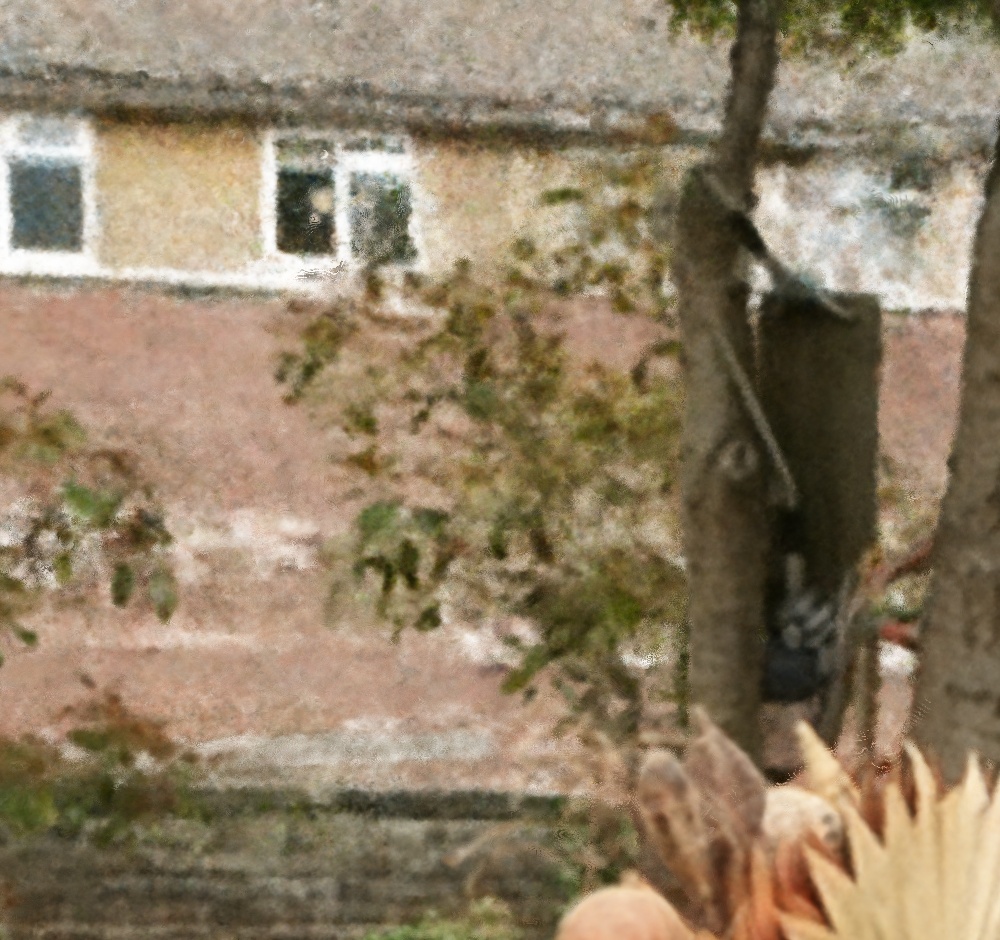}&    
    \includegraphics[width=\stmtwidth]{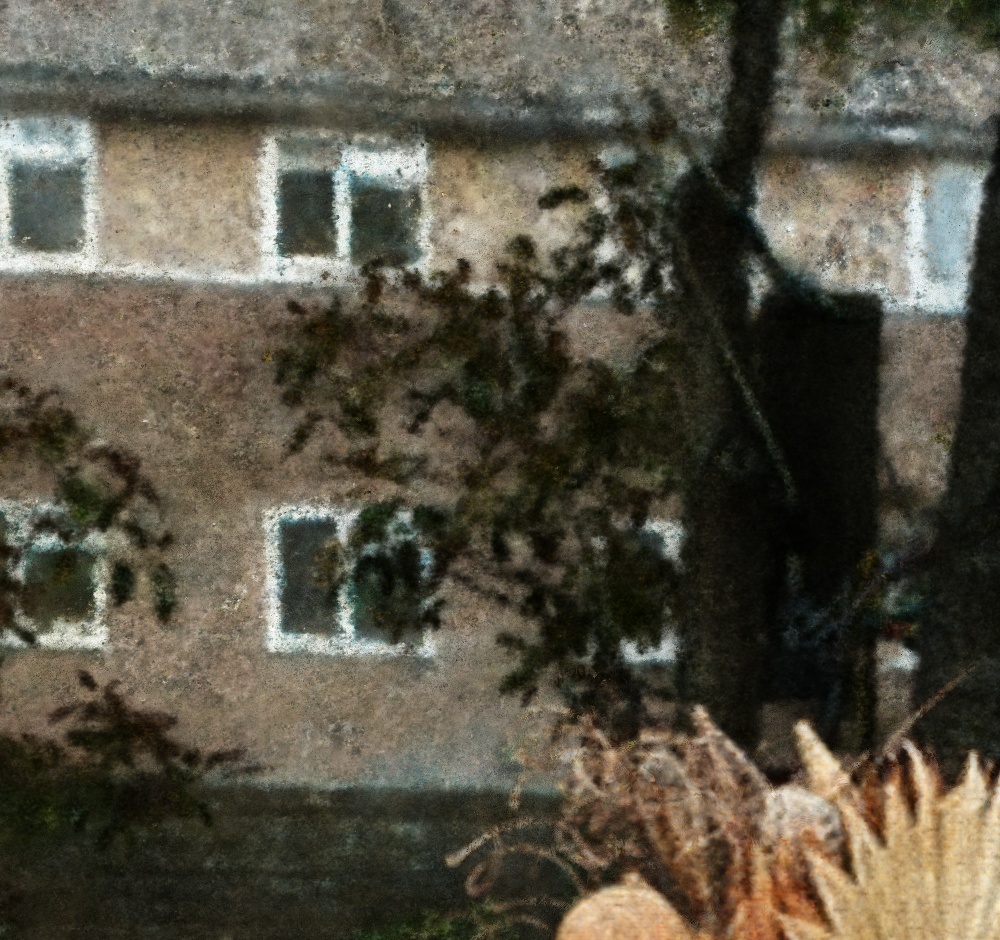}&    
    \includegraphics[width=\stmtwidth]{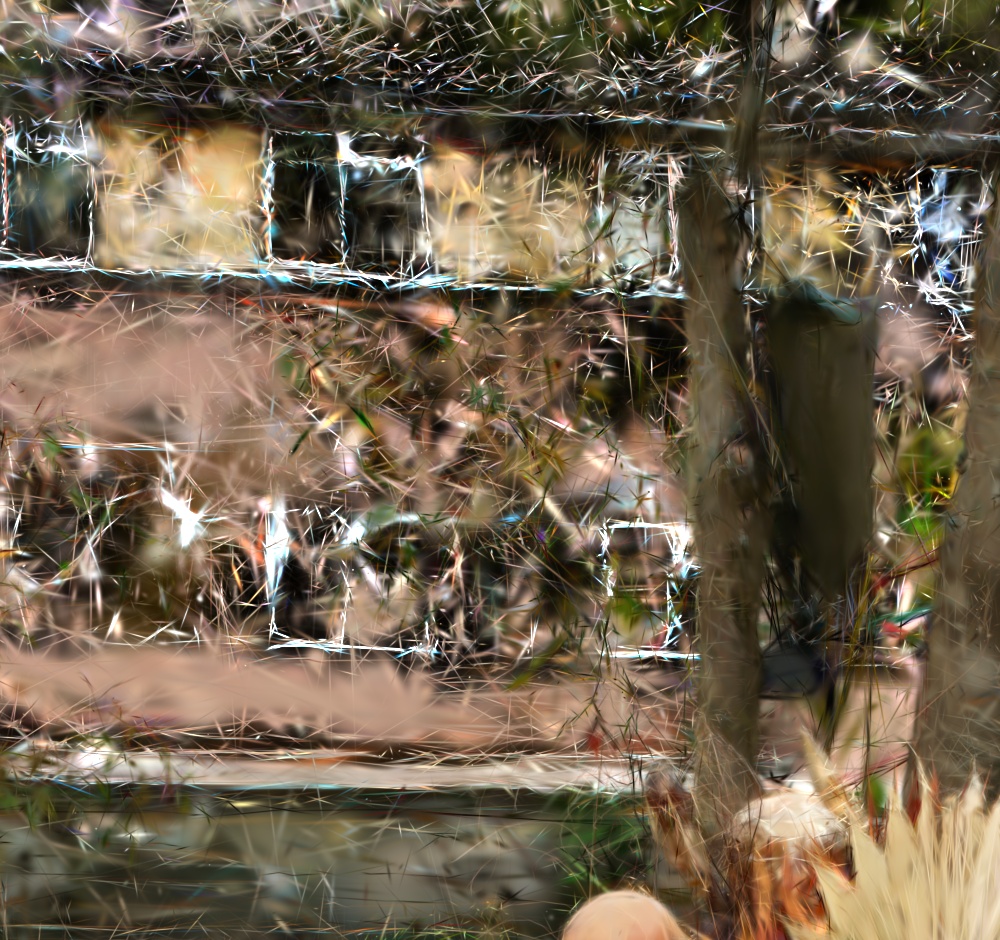}&    
    \includegraphics[width=\stmtwidth]{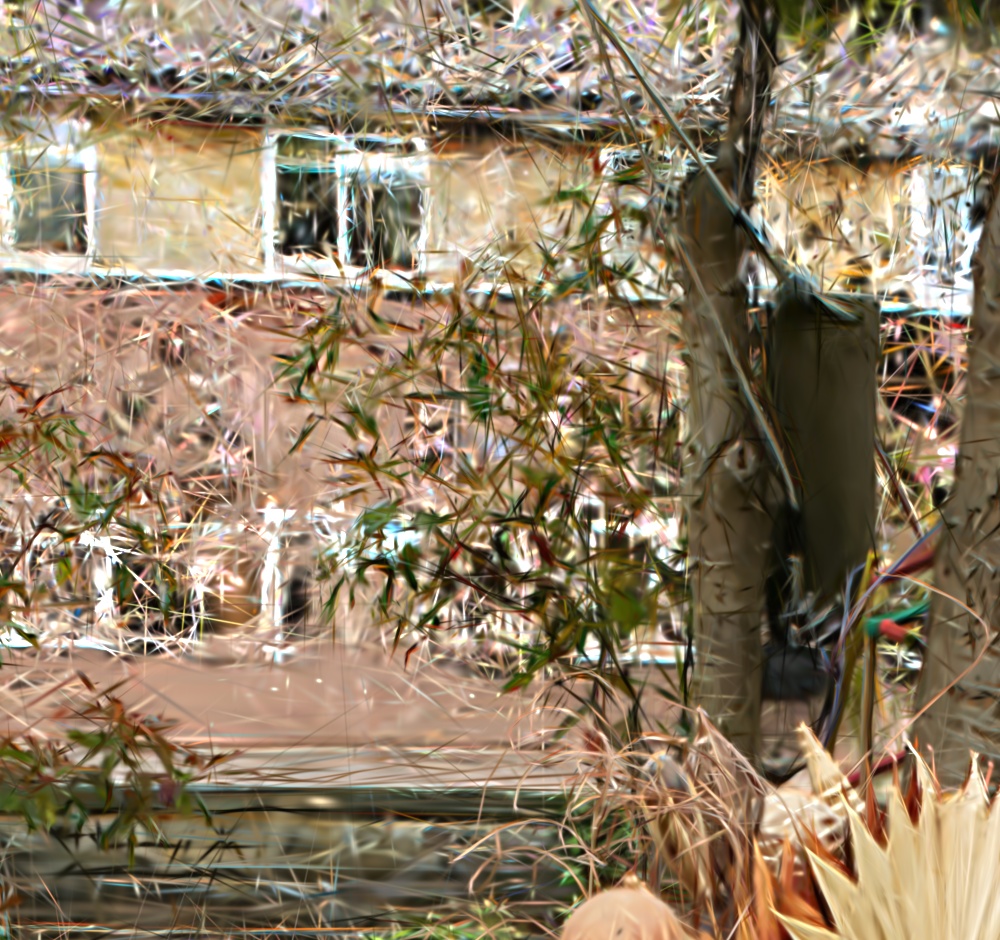}&    
    \includegraphics[width=\stmtwidth]{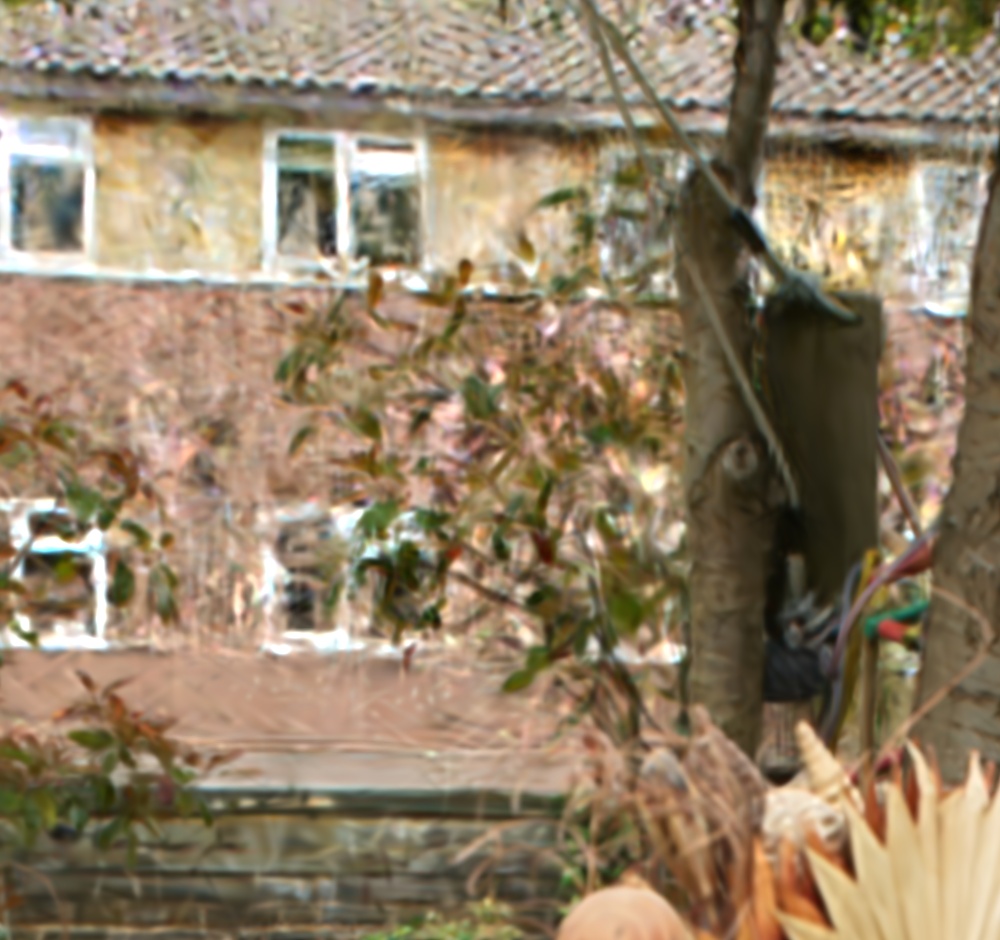}&    
    \includegraphics[width=\stmtwidth]{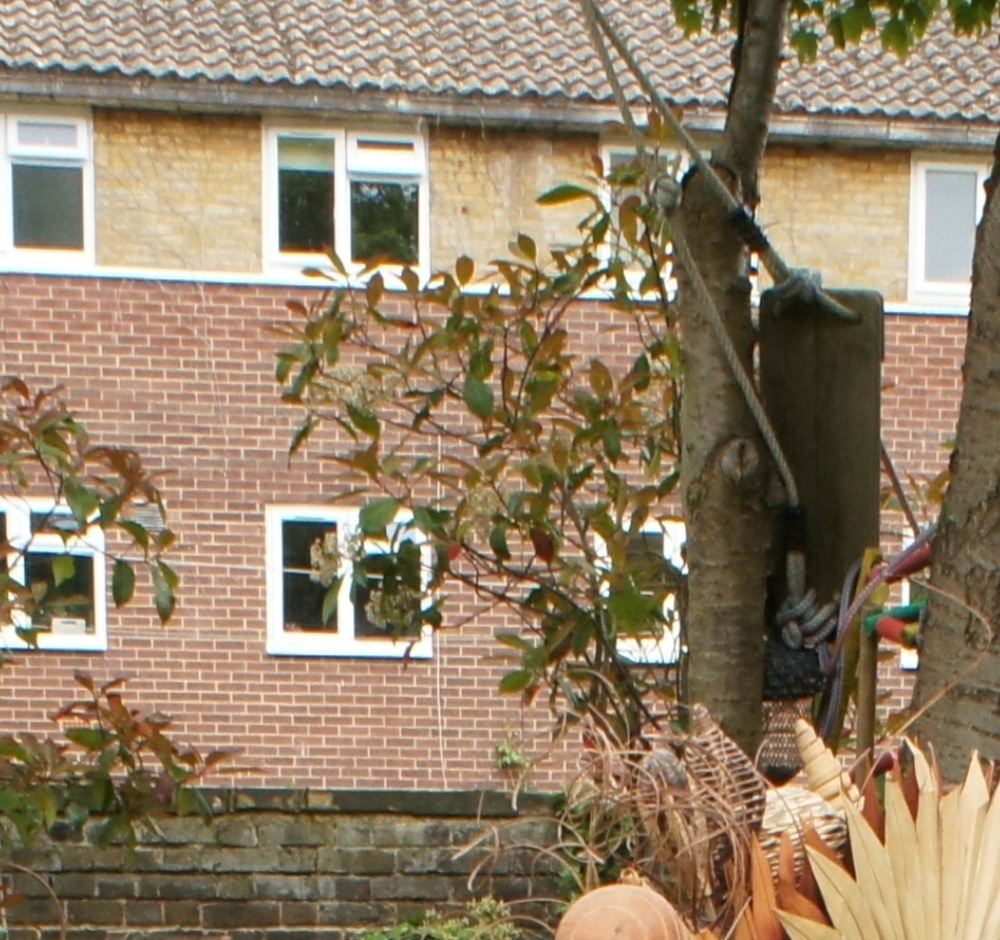}  
    \\
    \includegraphics[width=\stmtwidth]{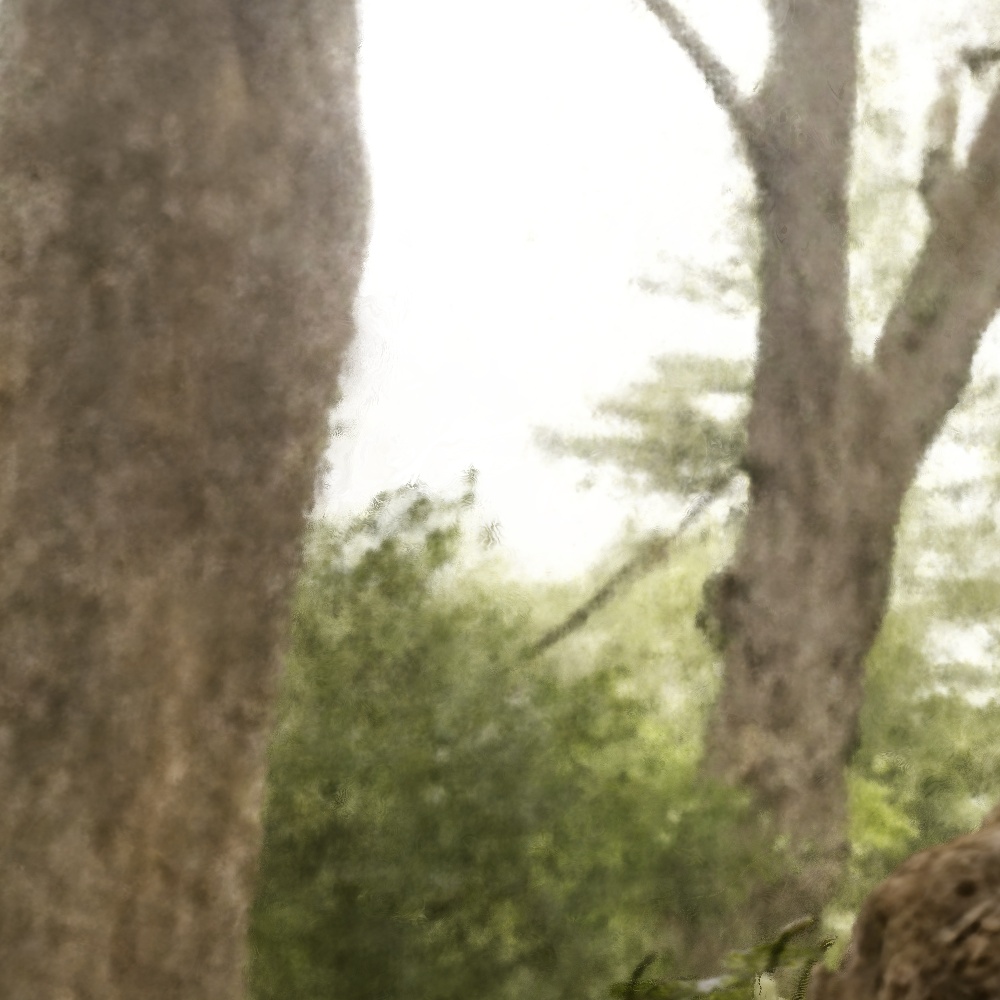}&    
    \includegraphics[width=\stmtwidth]{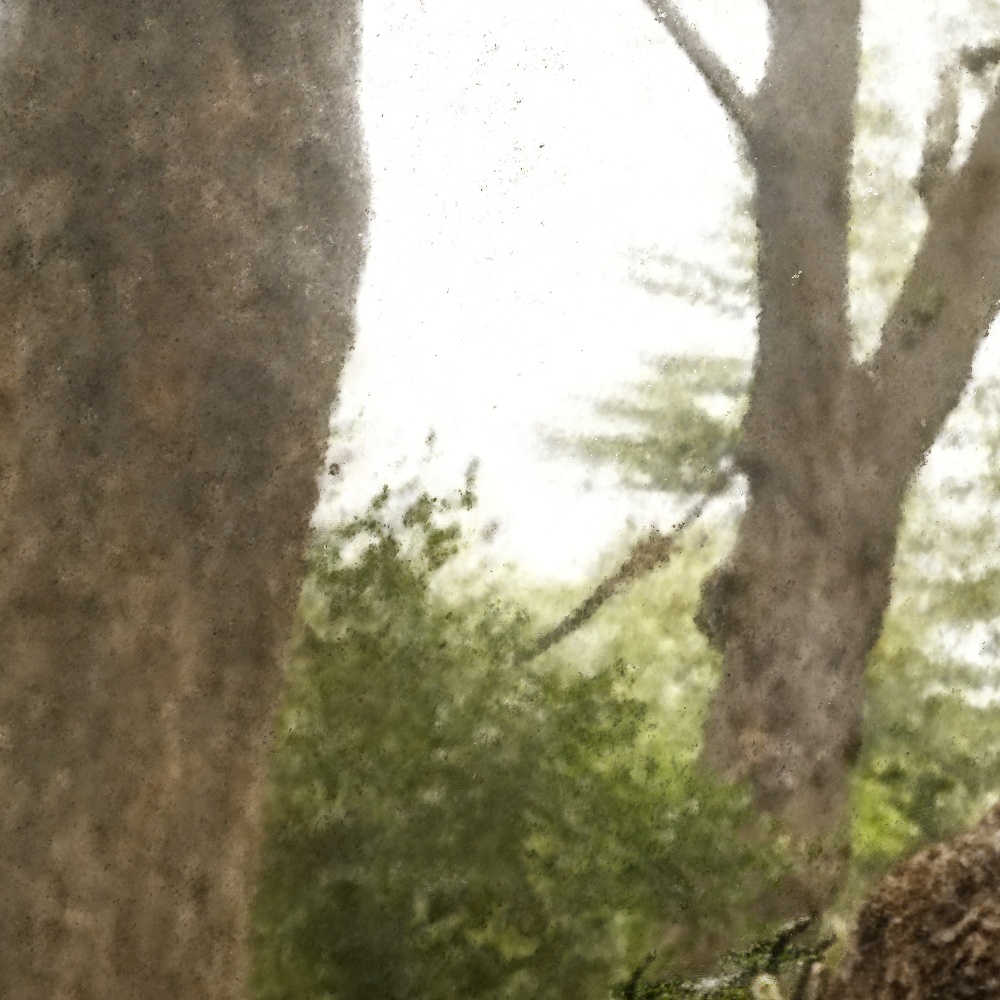}&    
    \includegraphics[width=\stmtwidth]{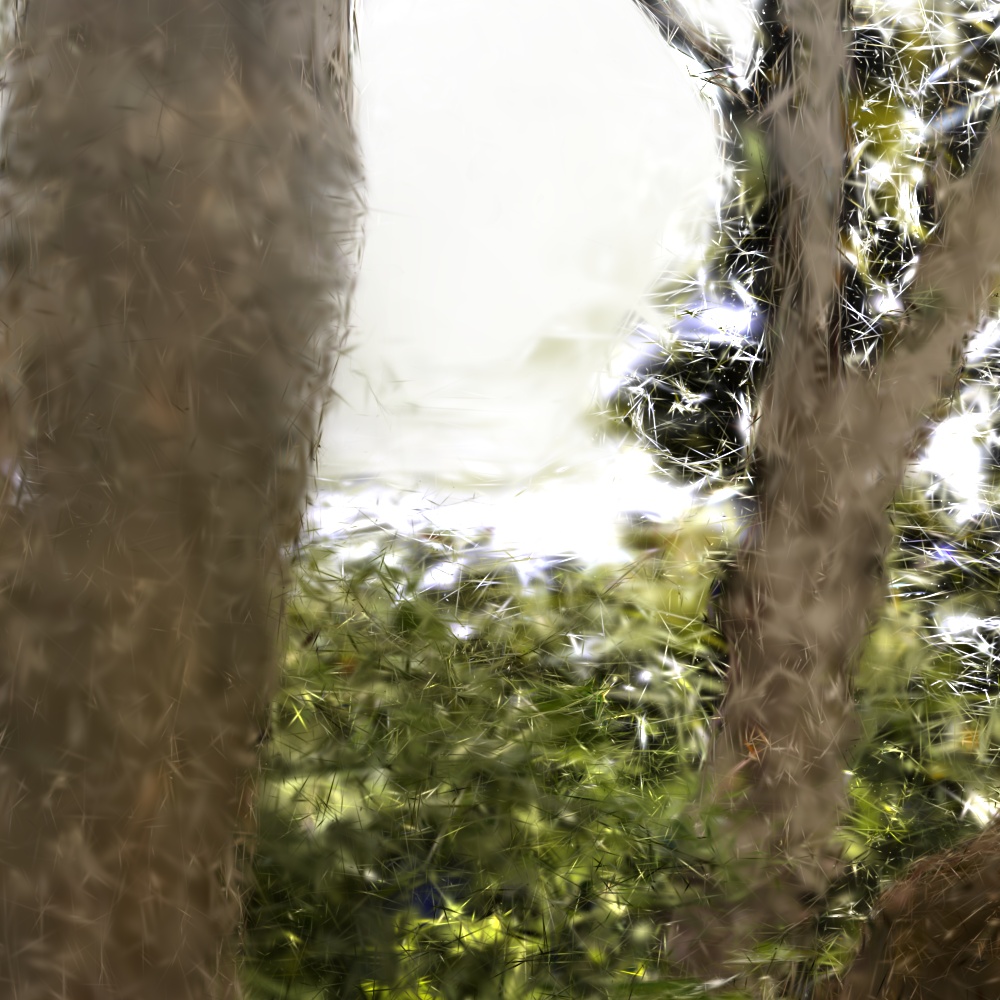}&    
    \includegraphics[width=\stmtwidth]{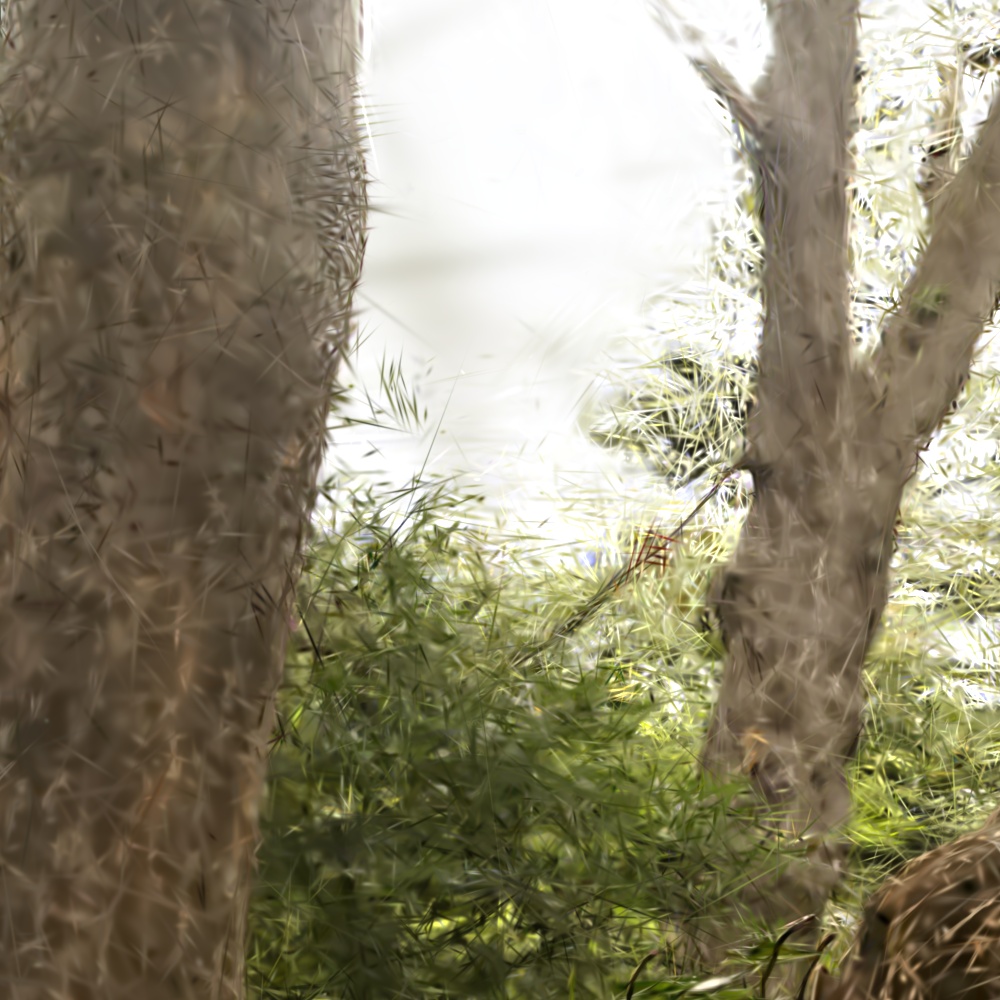}&    
    \includegraphics[width=\stmtwidth]{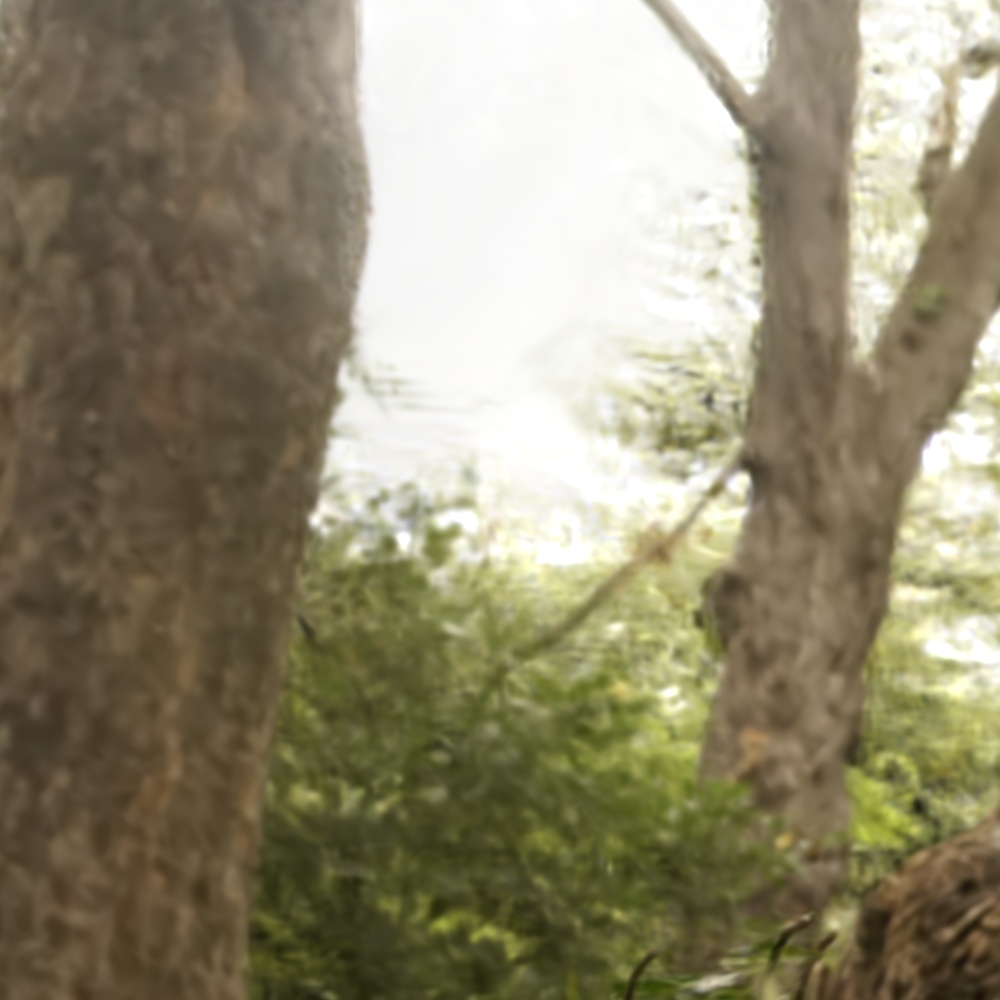}&    
    \includegraphics[width=\stmtwidth]{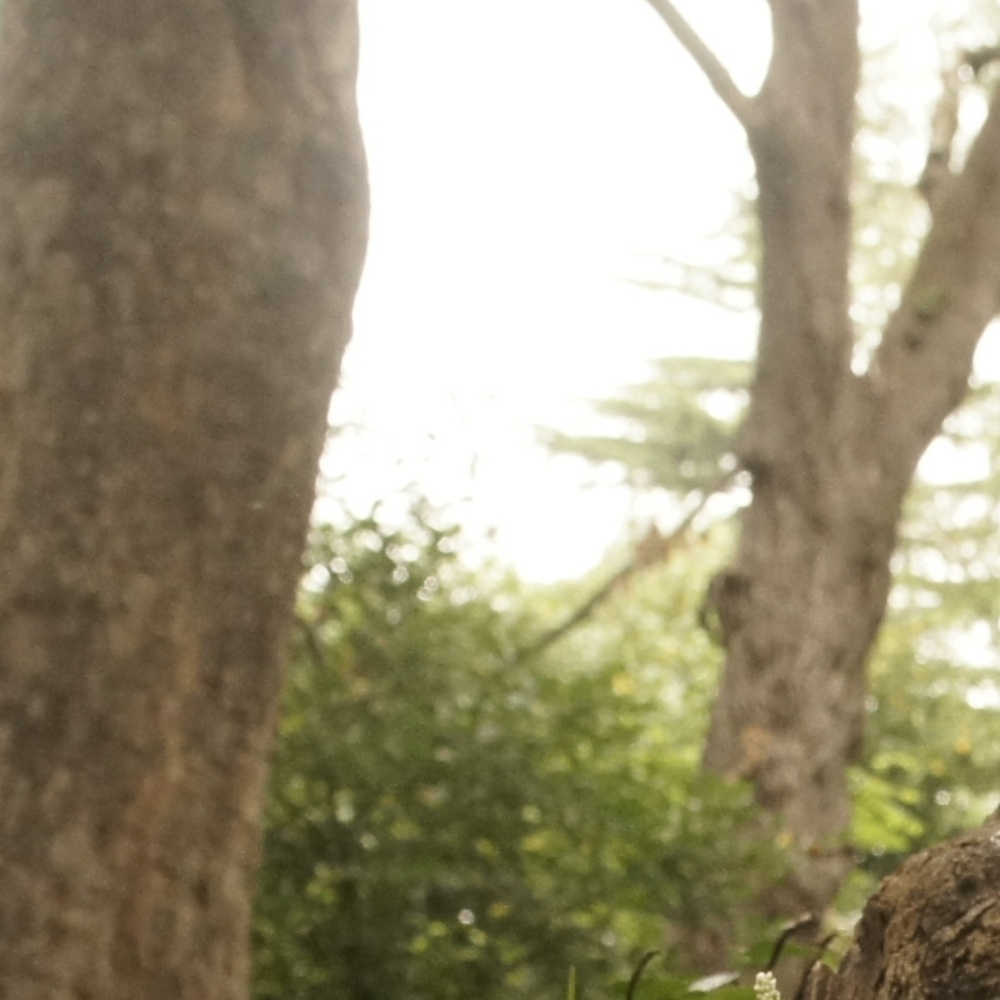}  
    \\
    \includegraphics[width=\stmtwidth]{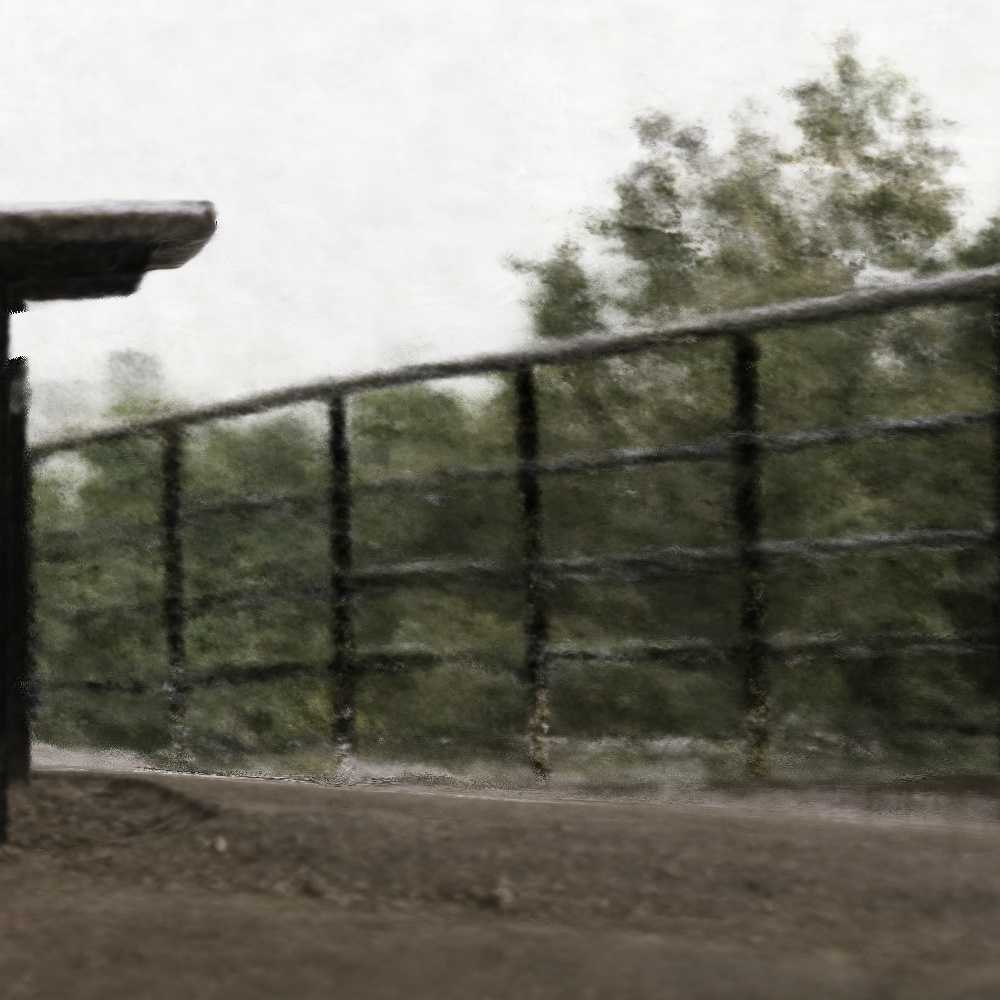}&    
    \includegraphics[width=\stmtwidth]{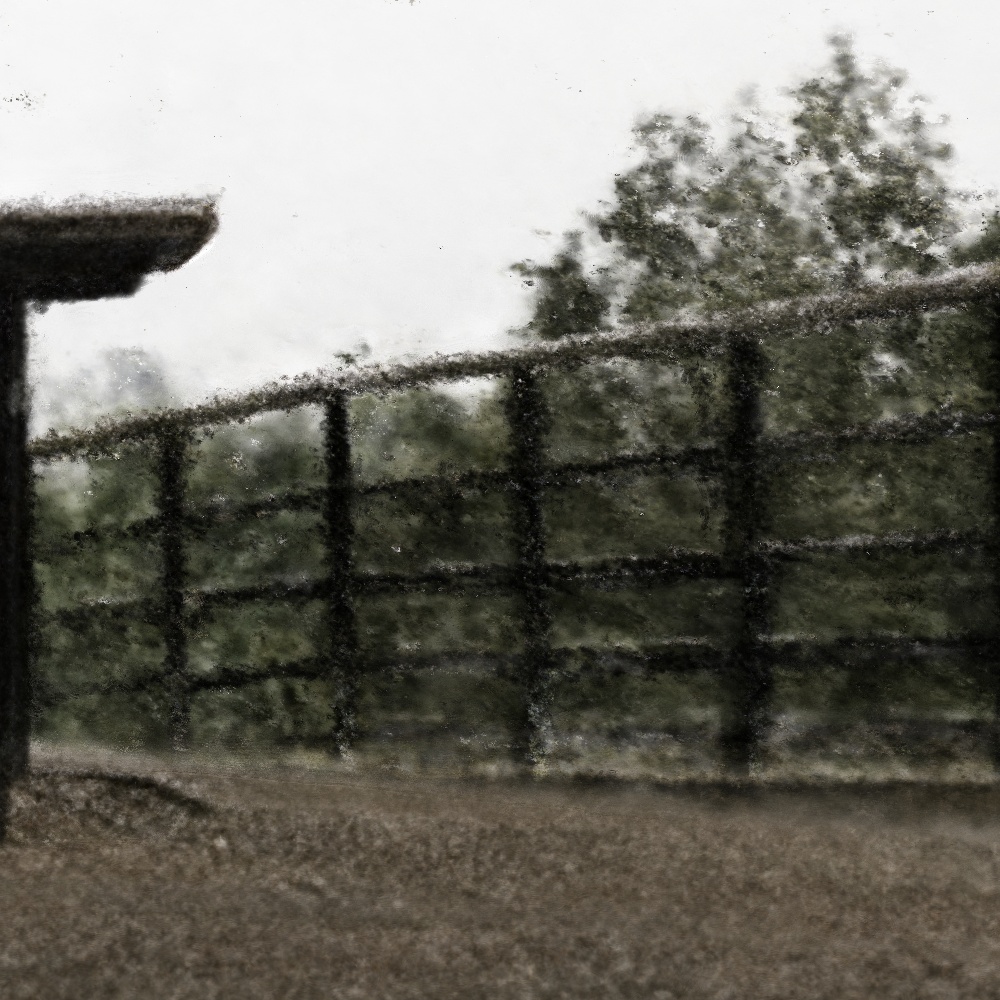}&    
    \includegraphics[width=\stmtwidth]{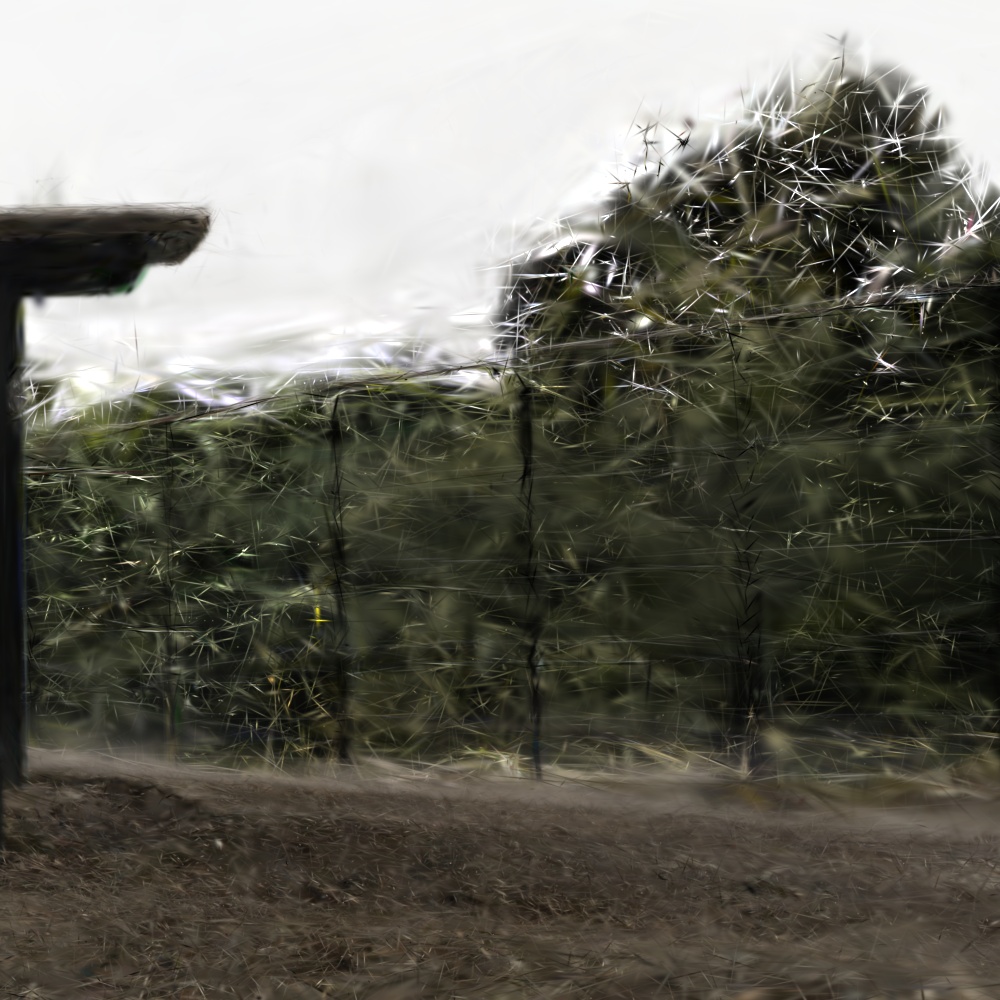}&    
    \includegraphics[width=\stmtwidth]{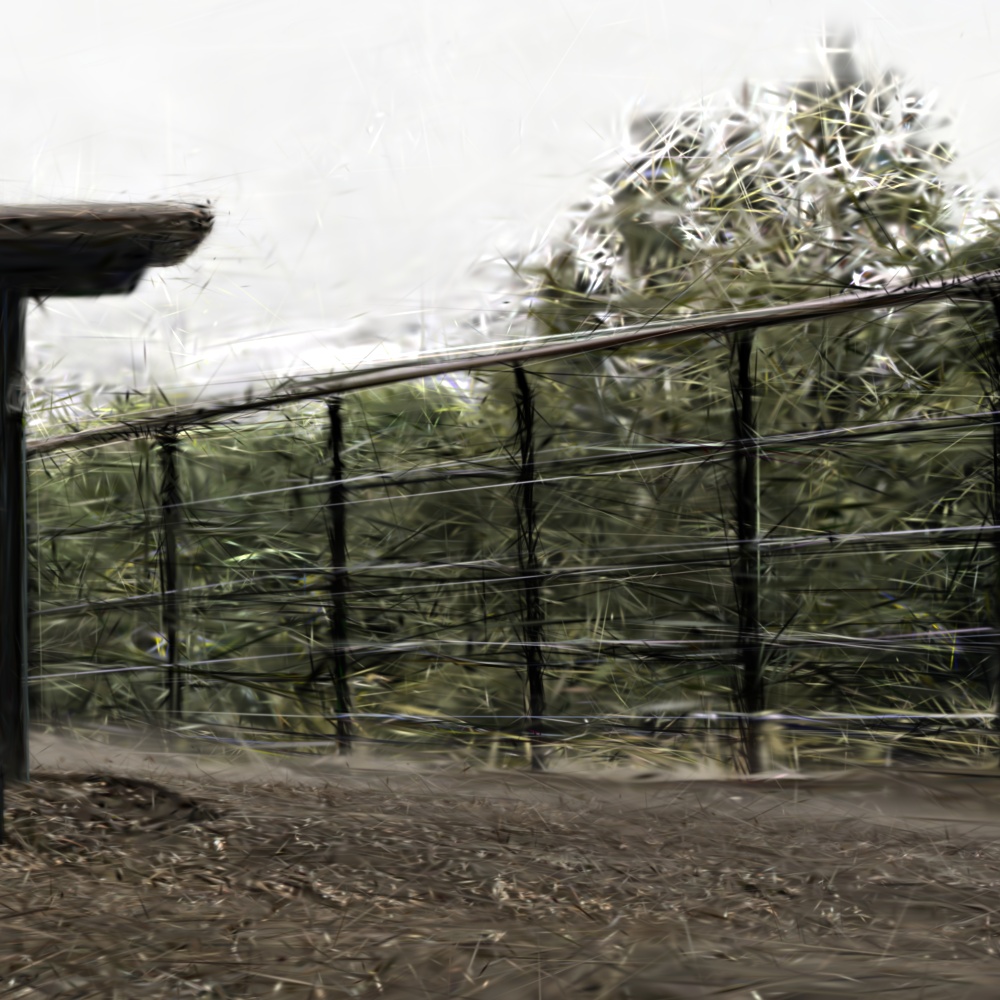}&    
    \includegraphics[width=\stmtwidth]{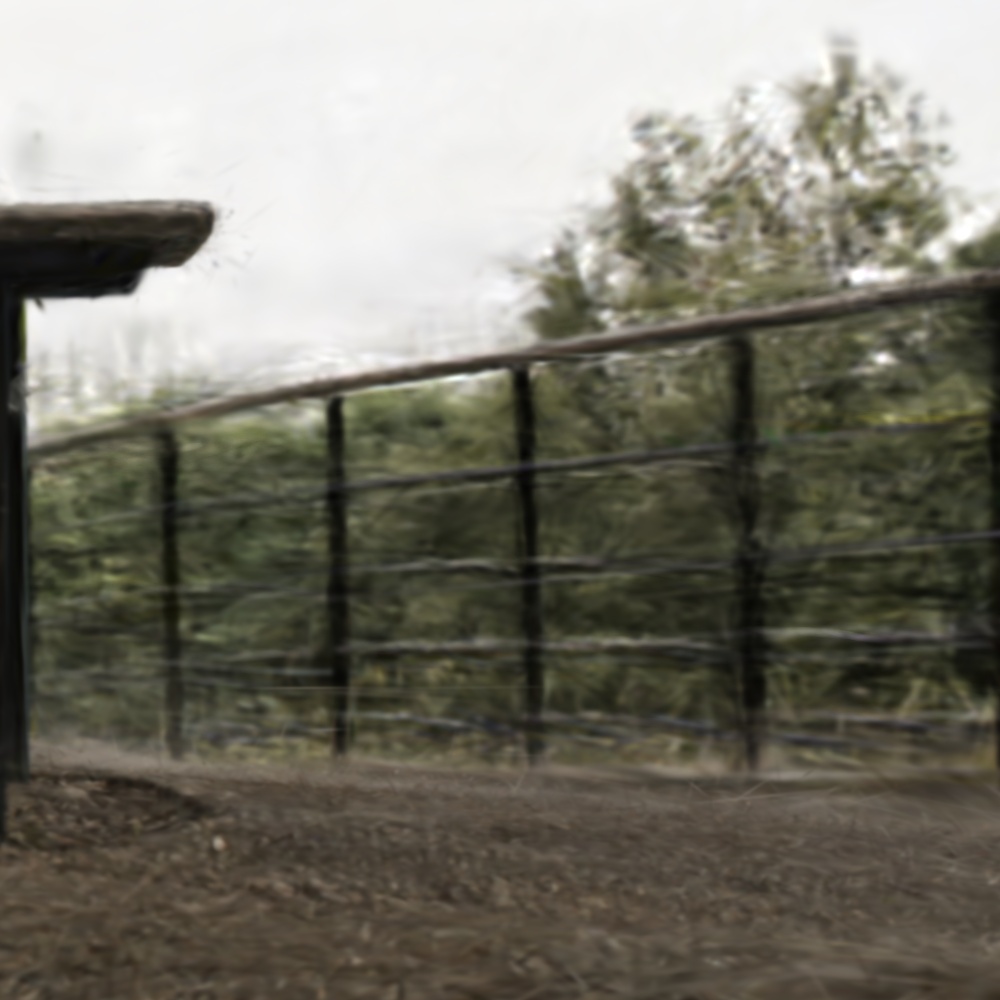}&    
    \includegraphics[width=\stmtwidth]{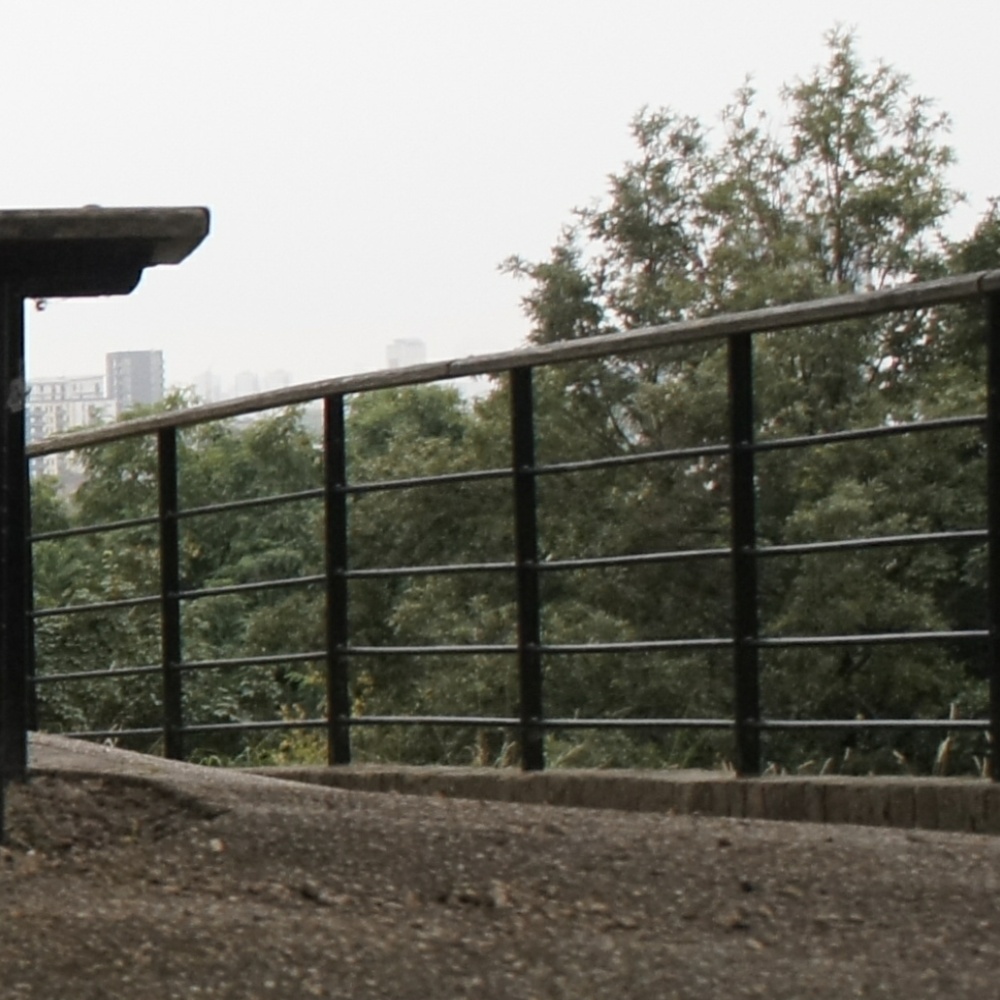}  
    \\
    \includegraphics[width=\stmtwidth]{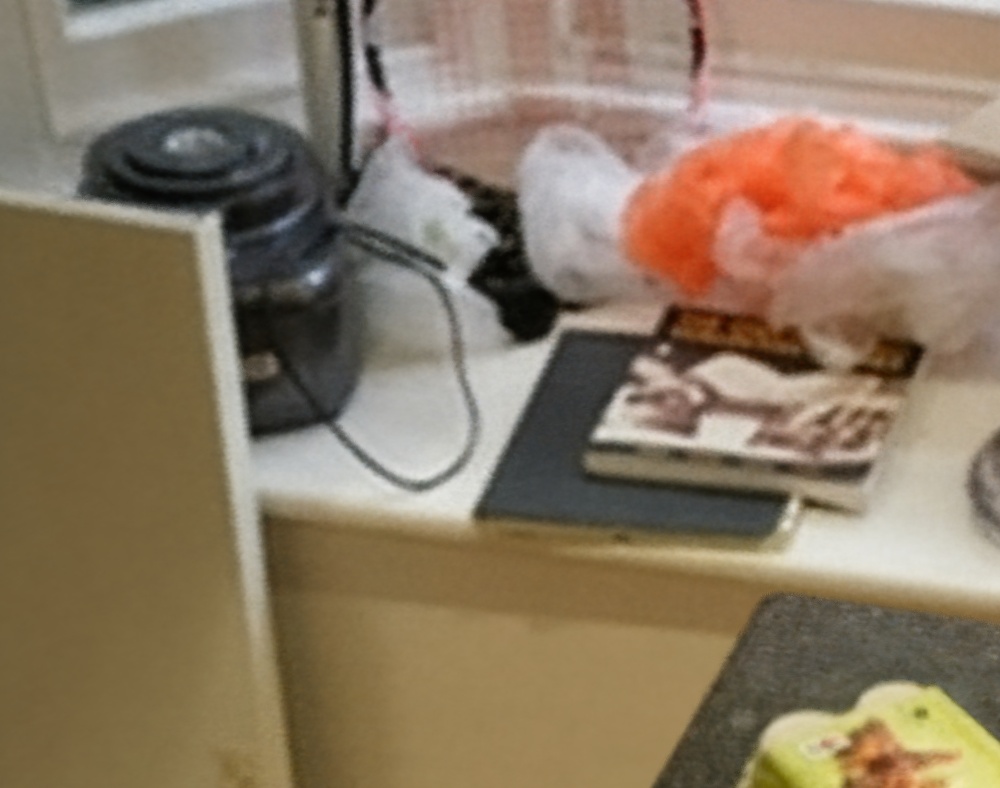}&    
    \includegraphics[width=\stmtwidth]{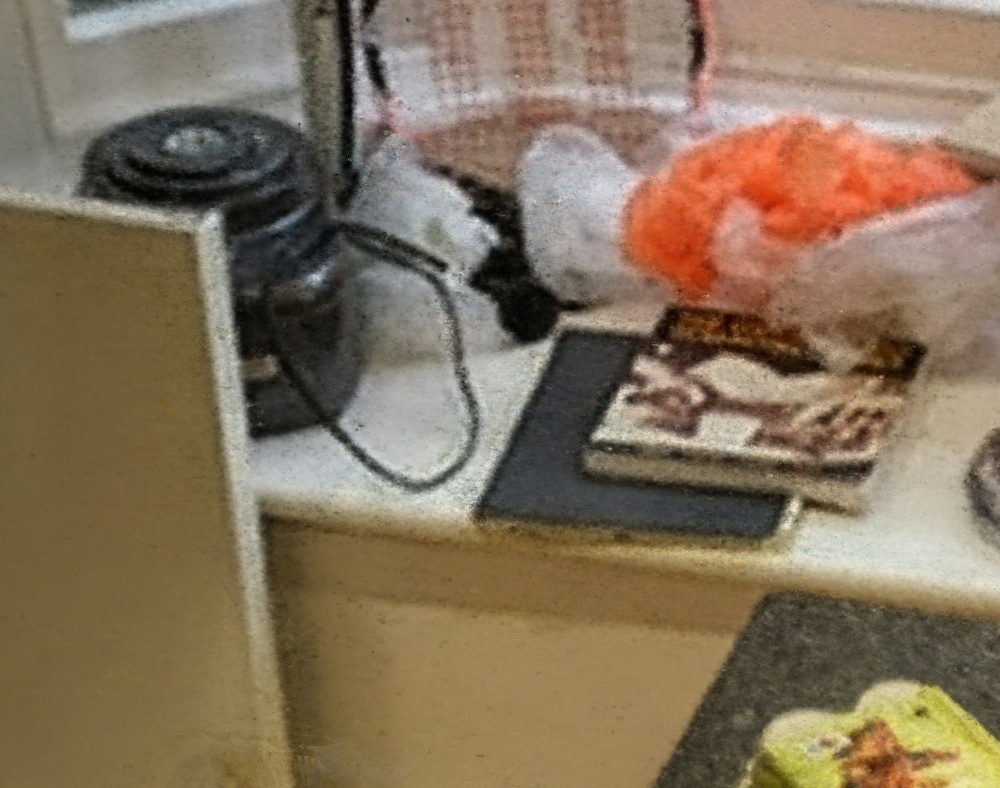}&    
    \includegraphics[width=\stmtwidth]{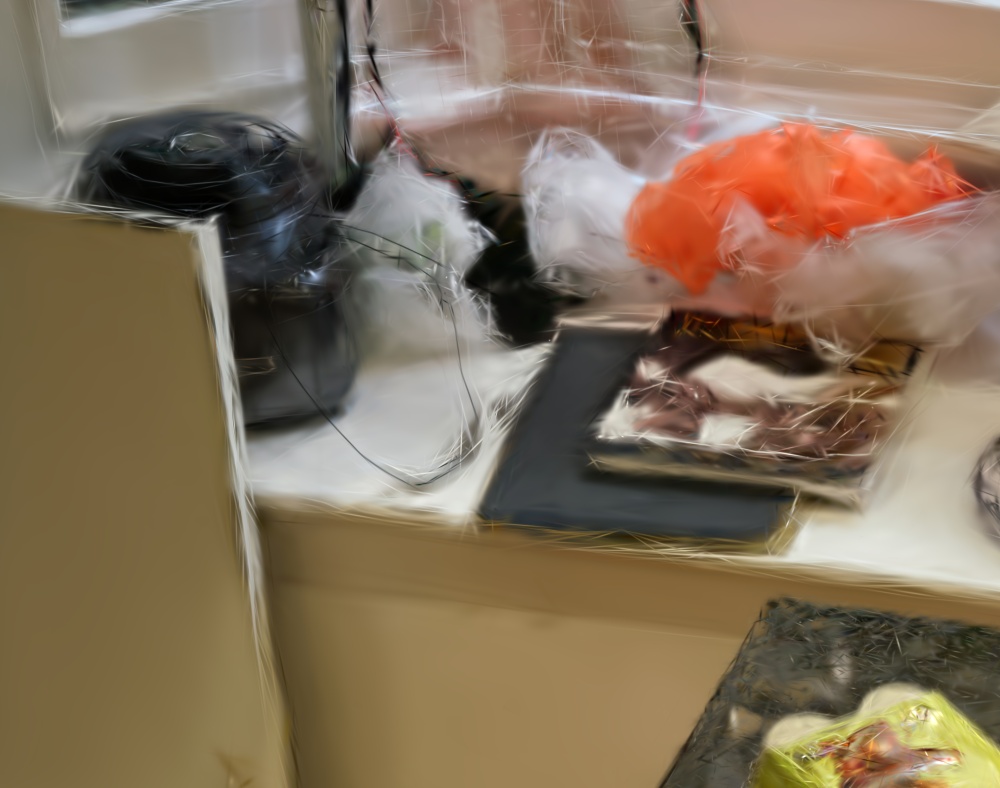}&    
    \includegraphics[width=\stmtwidth]{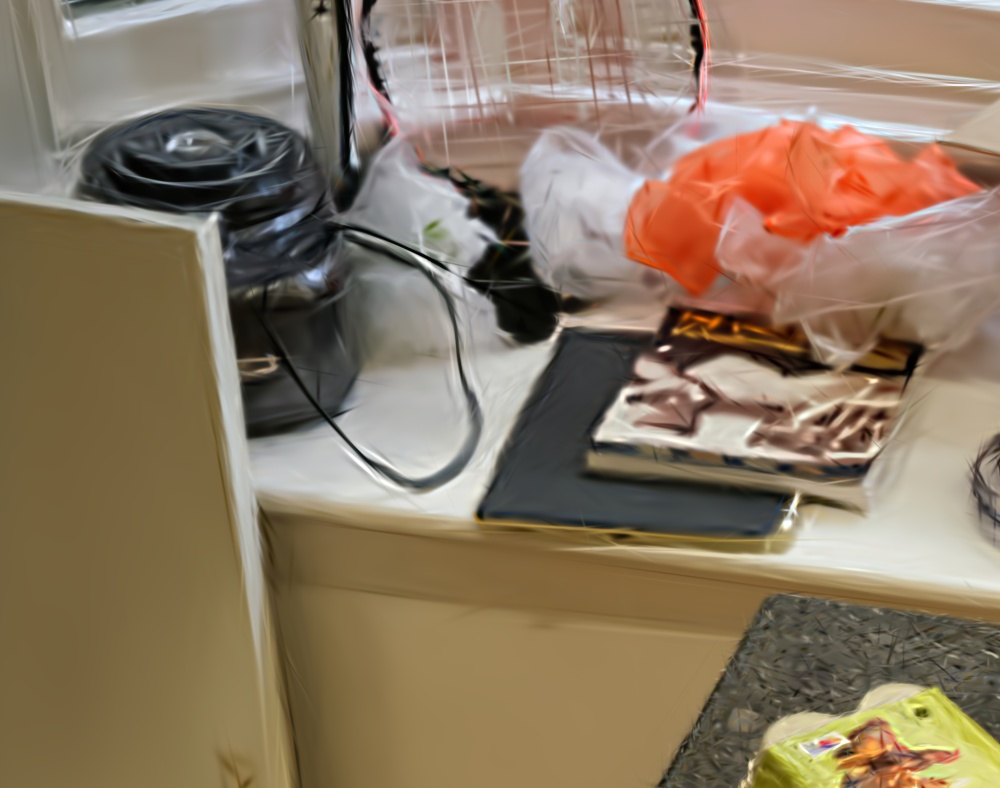}&    
    \includegraphics[width=\stmtwidth]{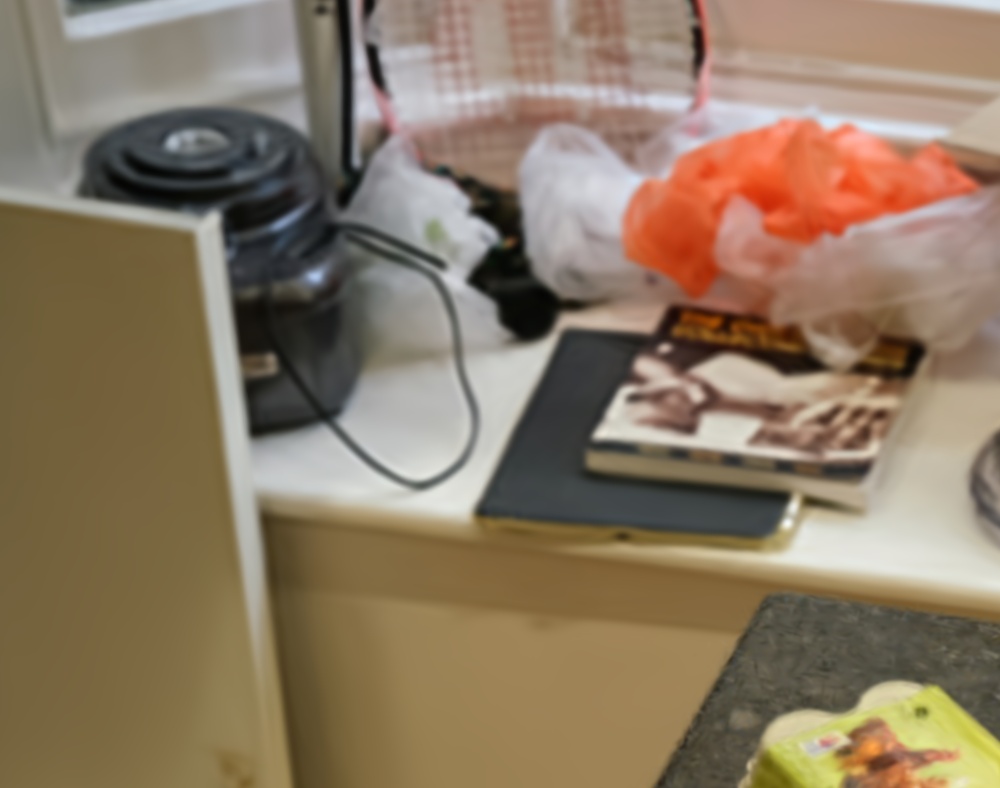}&    
    \includegraphics[width=\stmtwidth]{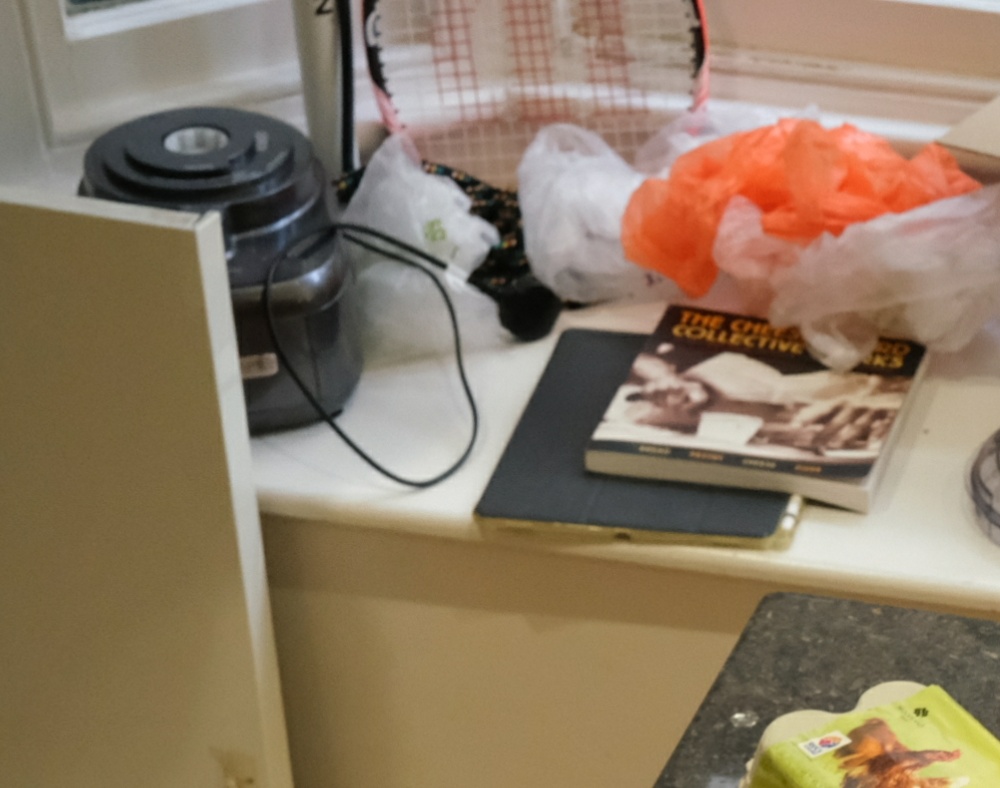}  
    \\
    \includegraphics[width=\stmtwidth]{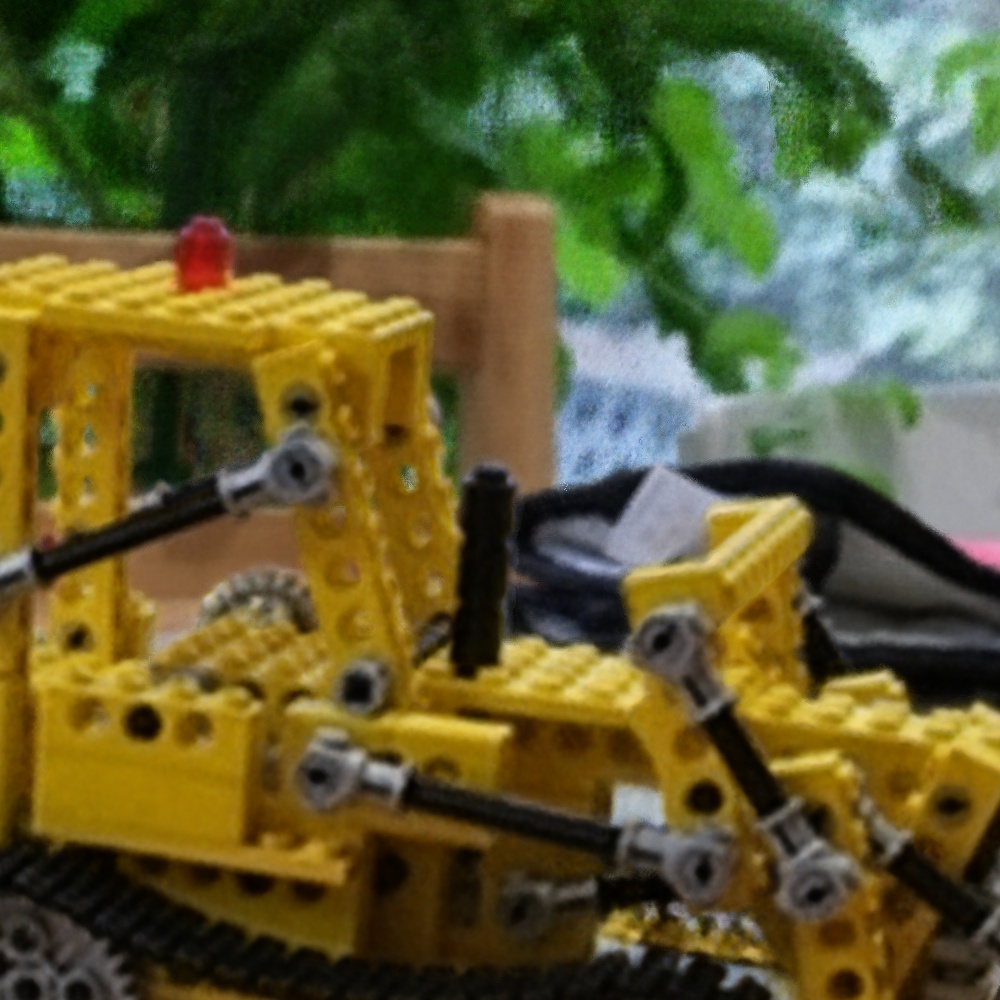}&    
    \includegraphics[width=\stmtwidth]{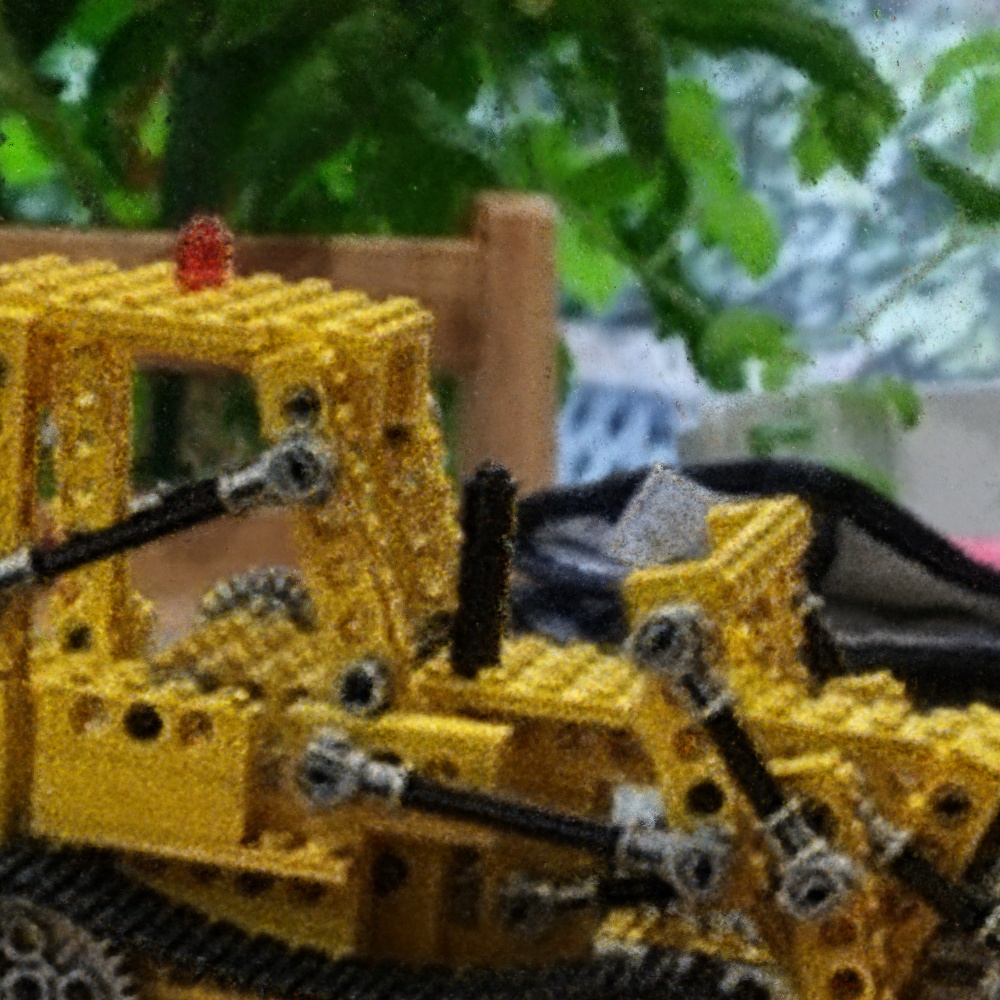}&    
    \includegraphics[width=\stmtwidth]{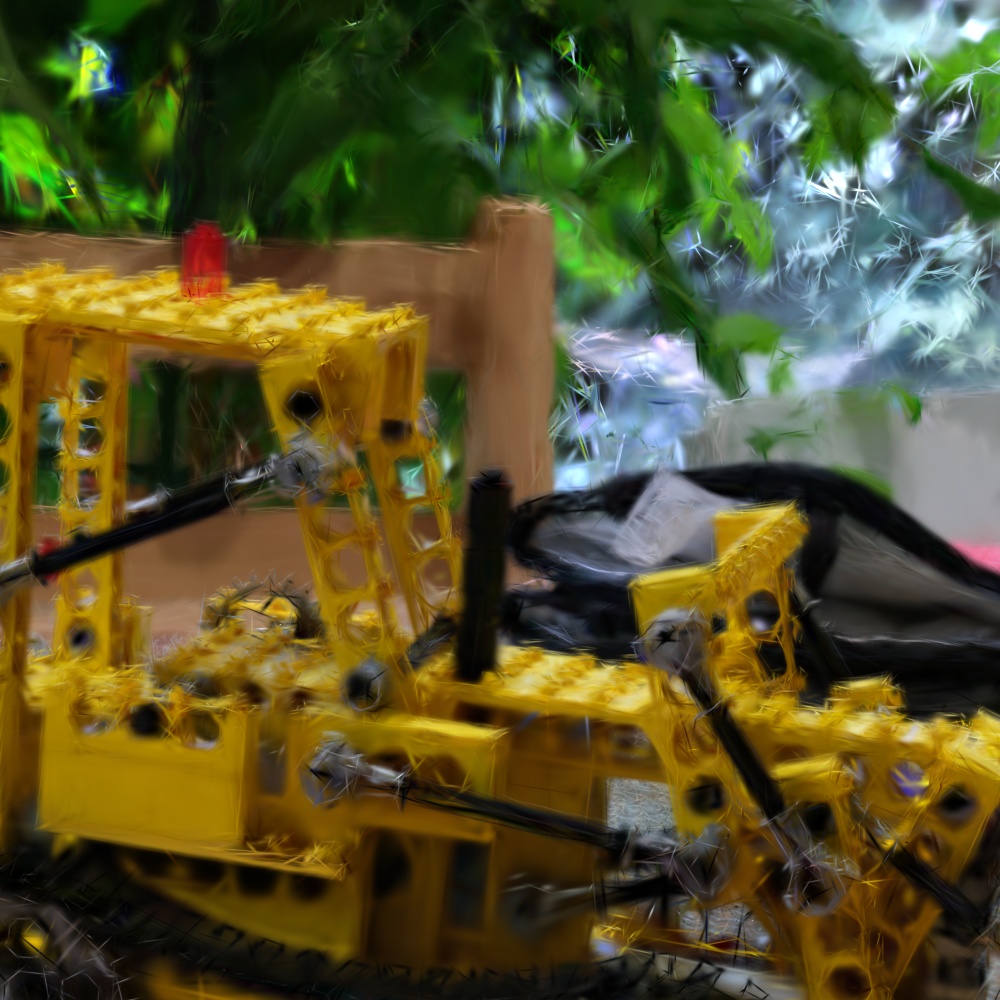}&    
    \includegraphics[width=\stmtwidth]{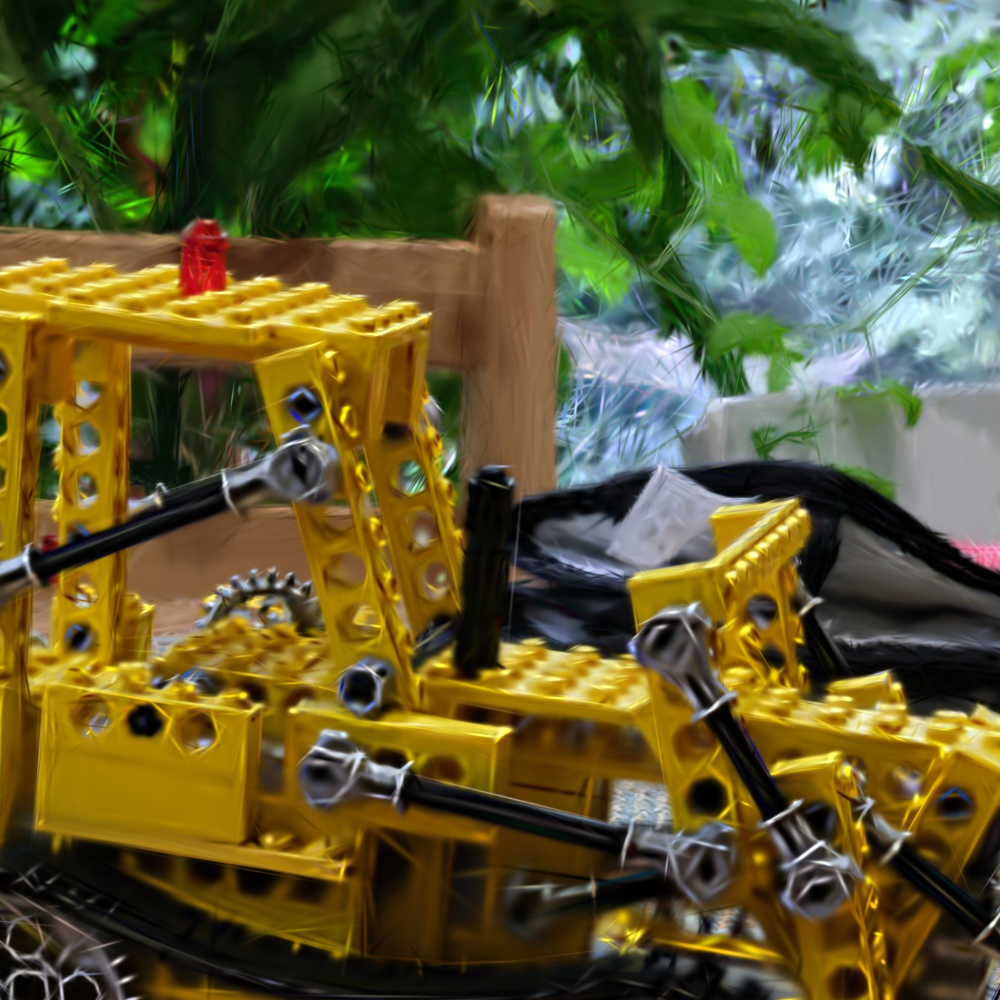}&    
    \includegraphics[width=\stmtwidth]{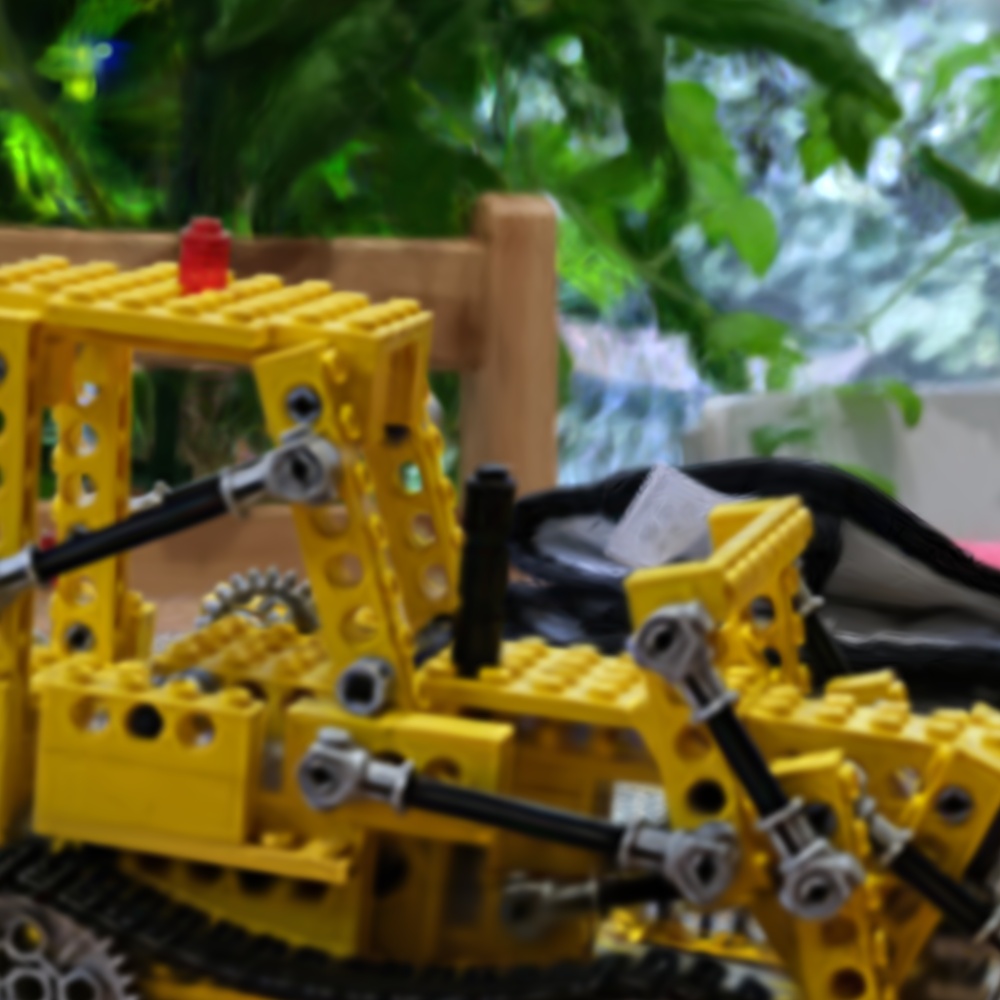}&    
    \includegraphics[width=\stmtwidth]{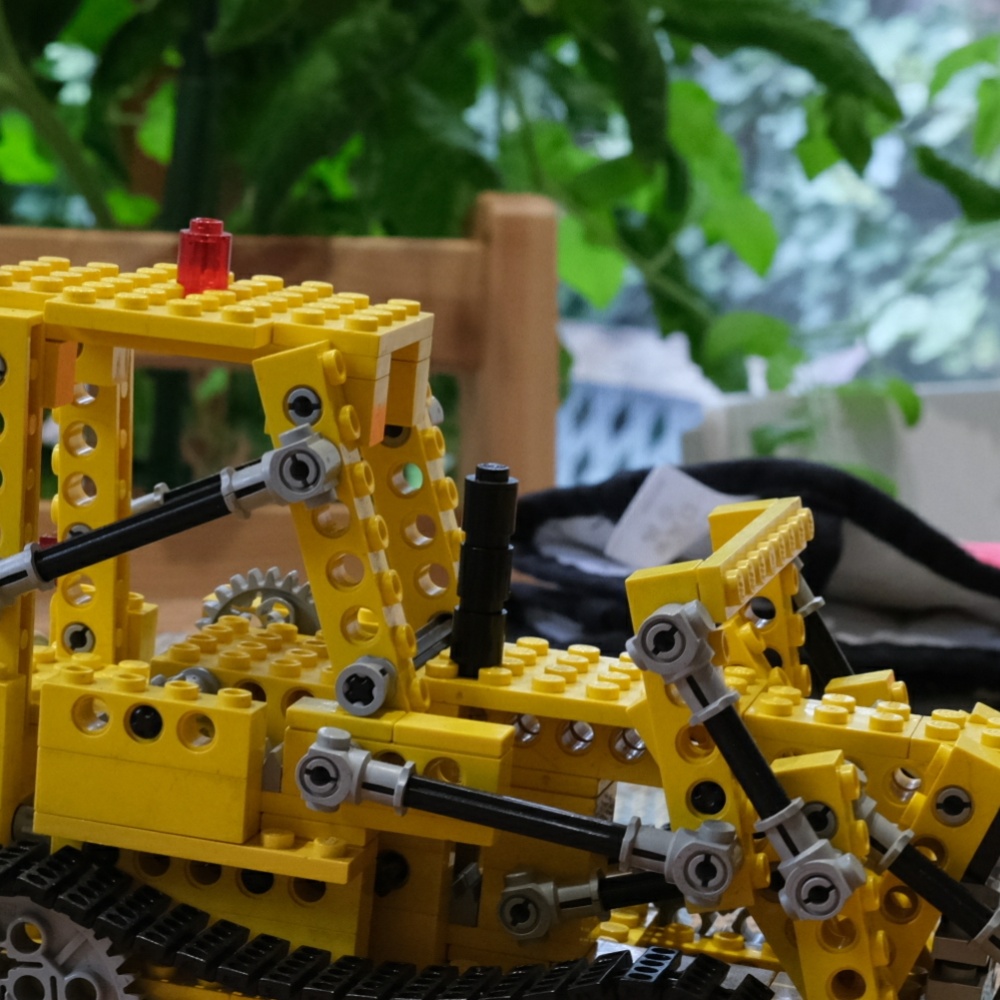}  
    \\
    \includegraphics[width=\stmtwidth]{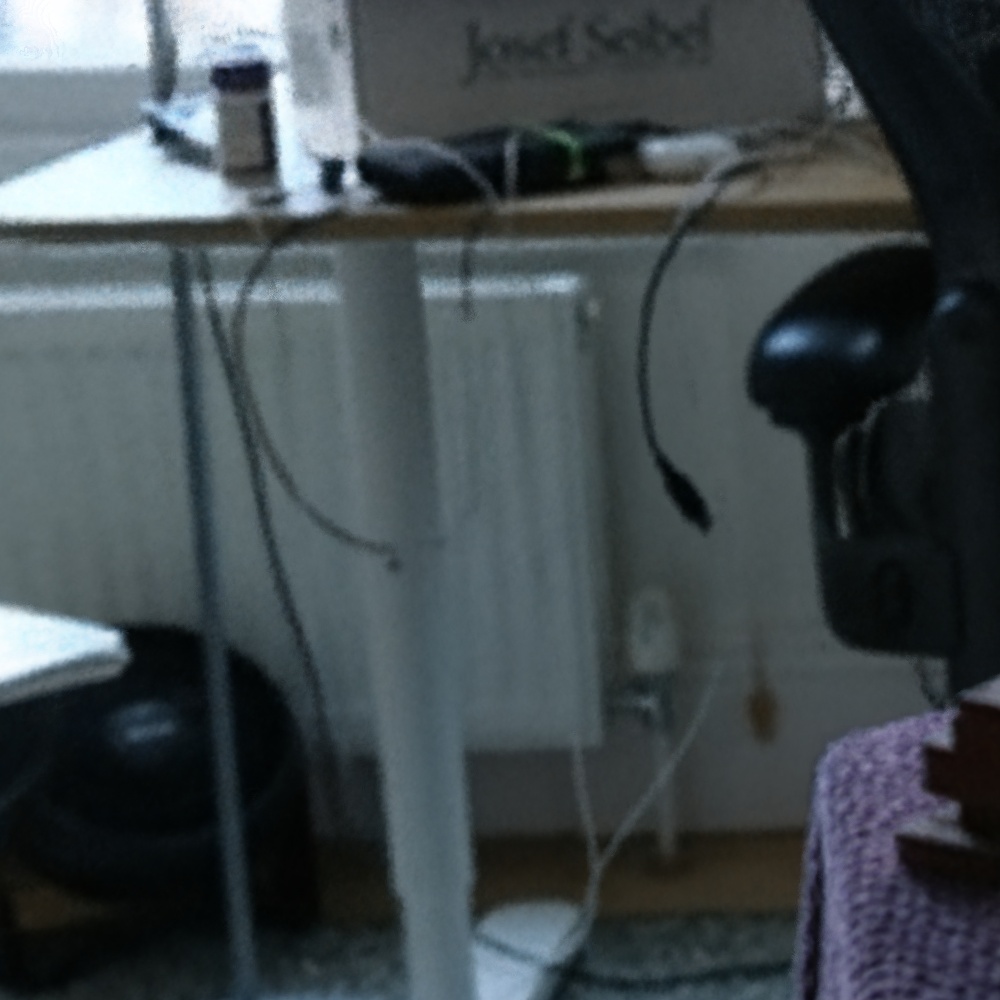}&    
    \includegraphics[width=\stmtwidth]{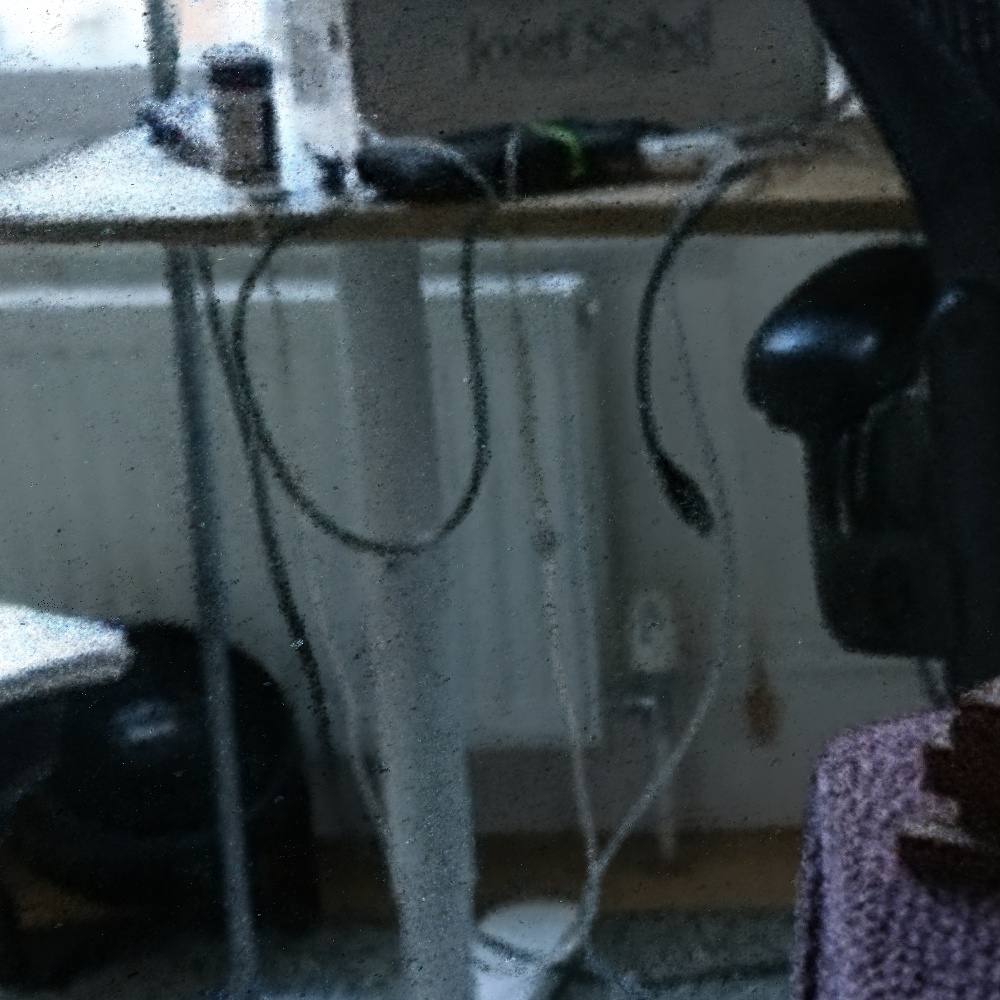}&    
    \includegraphics[width=\stmtwidth]{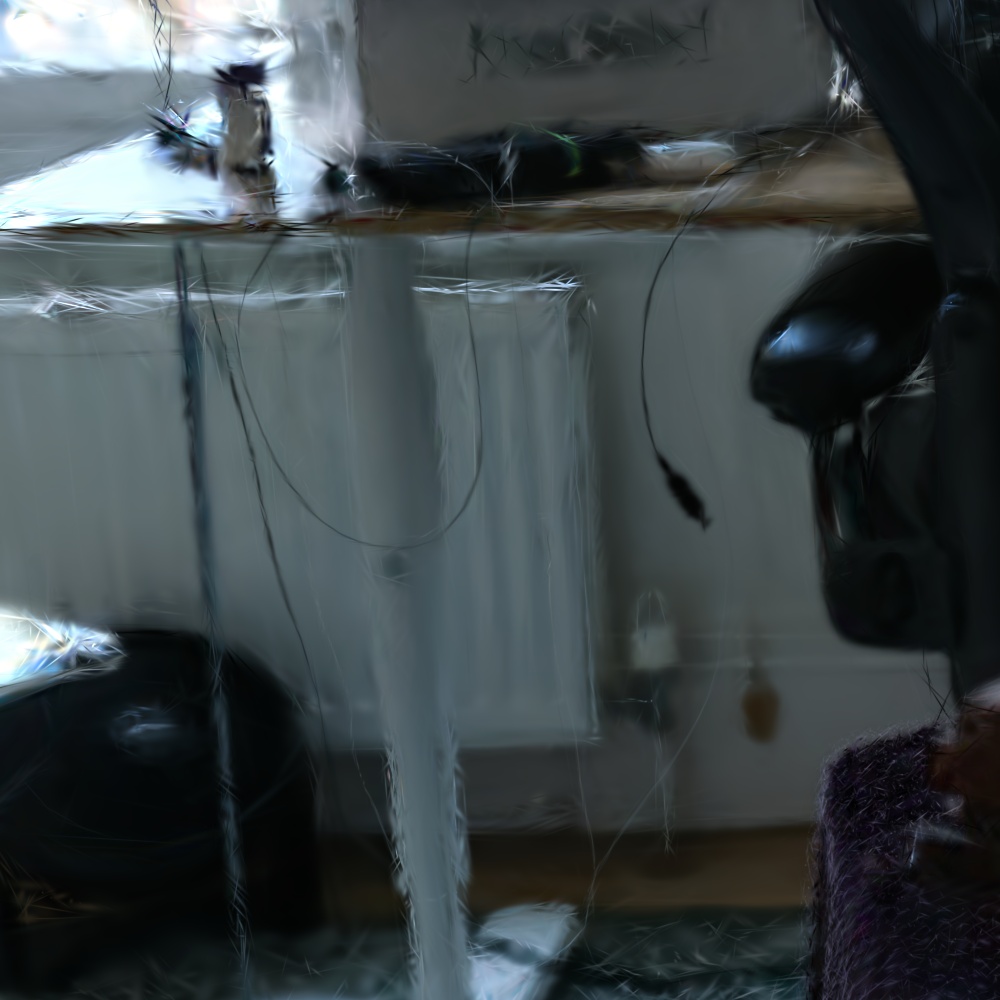}&    
    \includegraphics[width=\stmtwidth]{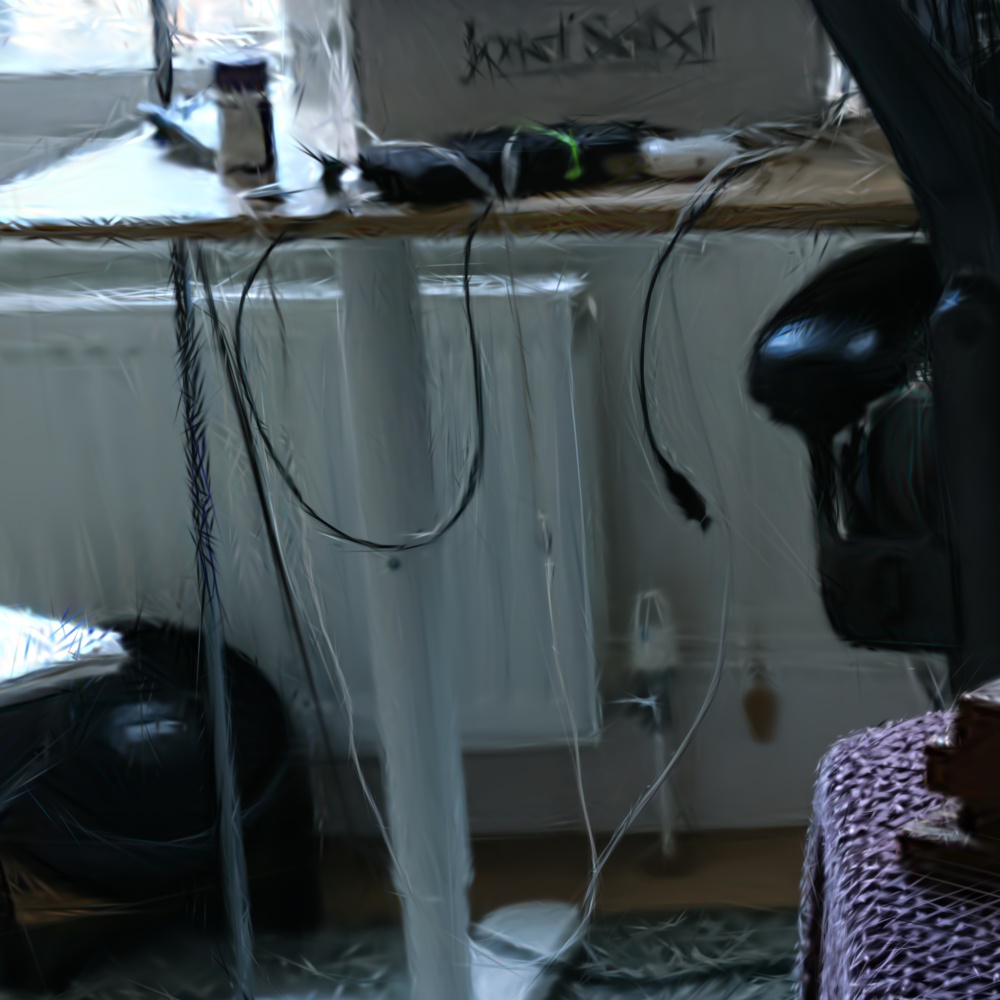}&    
    \includegraphics[width=\stmtwidth]{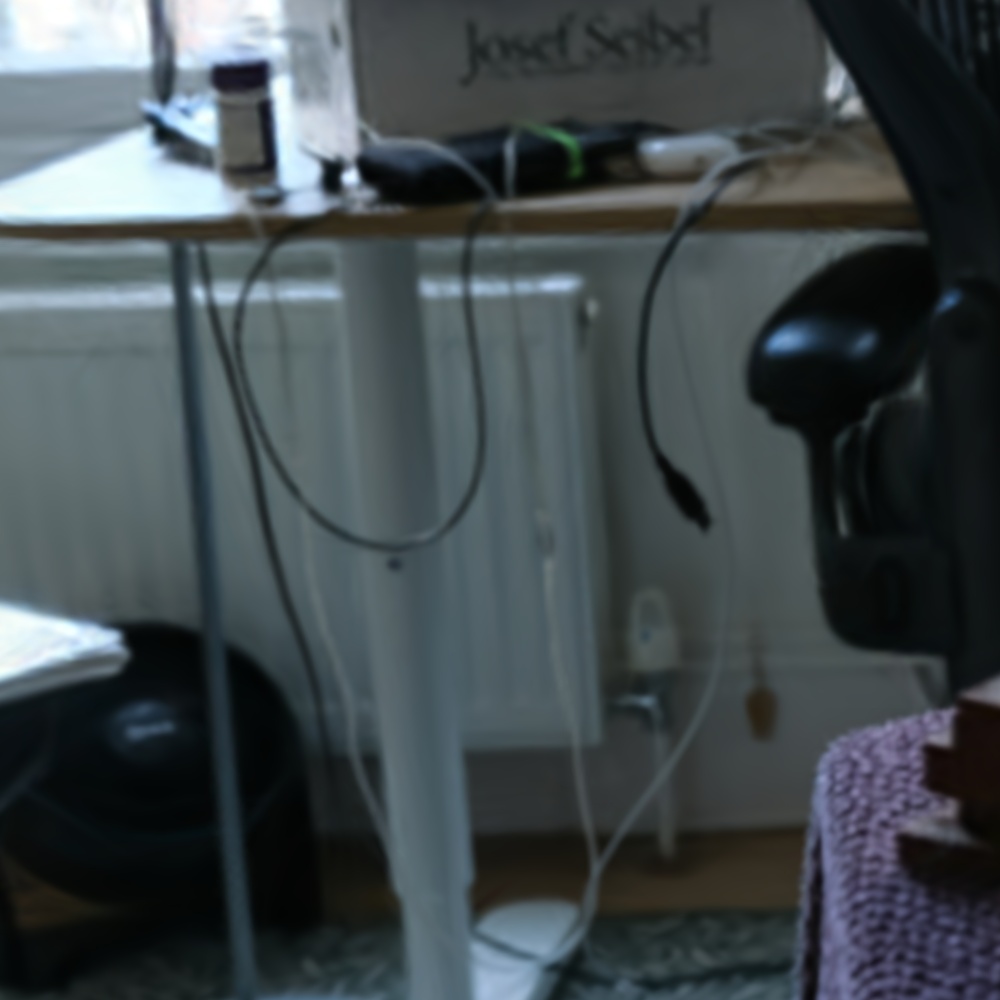}&    
    \includegraphics[width=\stmtwidth]{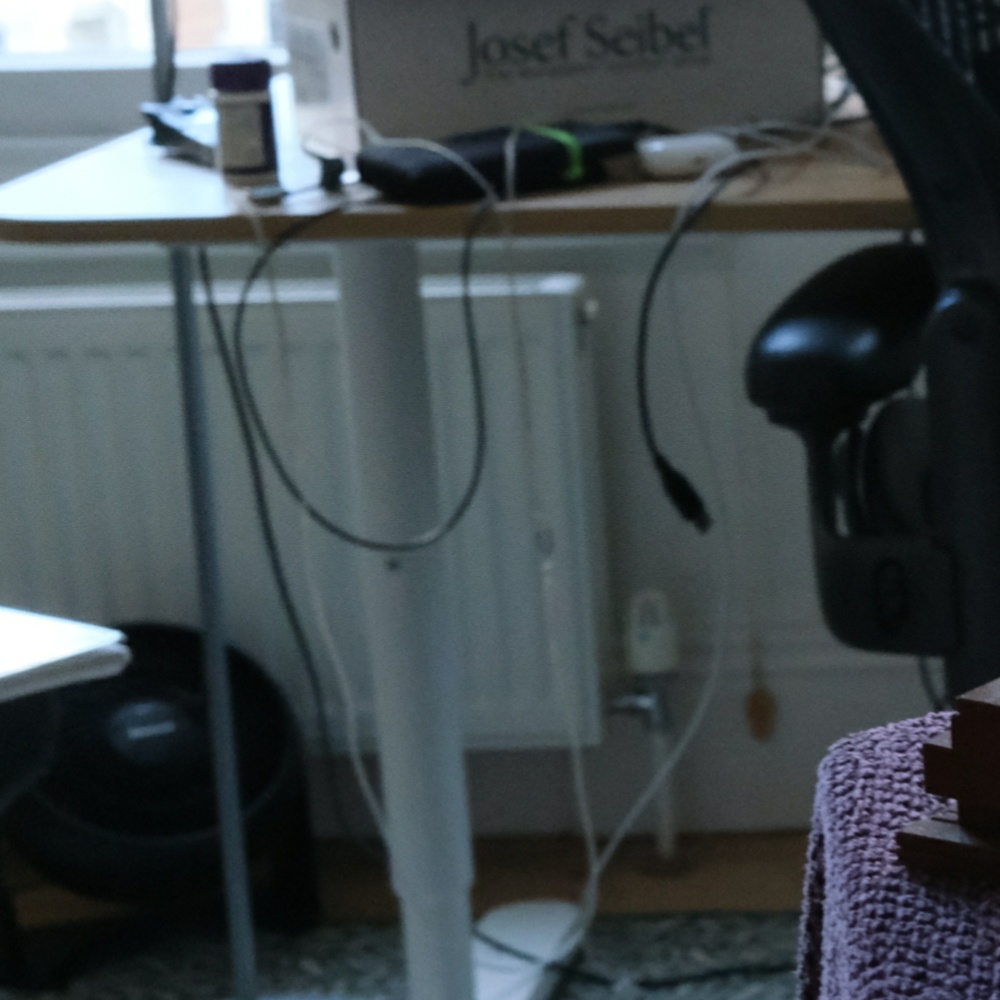}  
    \\
    Mip-NeRF 360~\cite{barron2022mipnerf360} & Zip-NeRF~\cite{Barron2023ICCV} & 3DGS~\cite{kerbl3Dgaussians} & 3DGS~\cite{kerbl3Dgaussians} + EWA~\cite{zwicker2001ewa} & Mip-Splatting (ours) & GT  
    \end{tabular}
    \vspace{-0.1in}
    \caption{\textbf{Single-scale Training and Multi-scale Testing on the Mip-NeRF 360 Dataset~\cite{barron2022mipnerf360}.} 
    All models are trained on images downsampled by a factor of eight and rendered at full resolution to demonstrate zoom-in/moving closer effects. In contrast to prior work, Mip-Splatting renders images that closely approximate ground truth. Please also note the high-frequency artifacts of 3DGS + EWA~\cite{zwicker2001ewa}.
    }
    \label{fig:360_single_train_multi_test_supp}
    \vspace{-0.1in}
\end{figure*}

\end{document}